%% file: main.tex
\documentclass[10pt,twocolumn,letterpaper]{article}

\usepackage{cvpr}              % To produce the CAMERA-READY version

\input{macros}

\definecolor{cvprblue}{rgb}{0.21,0.49,0.74}
\usepackage[pagebackref,breaklinks,colorlinks,allcolors=cvprblue]{hyperref}

\usepackage{multicol, multirow}

\def\paperID{41809} % *** Enter the Paper ID here
\def\confName{CVPR}
\def\confYear{2026}

\title{Mining Attribute Subspaces for Efficient Fine-tuning of 3D Foundation Models}

\author{
Yu Jiang\textsuperscript{1,2}\quad
Hanwen Jiang\textsuperscript{3}\quad
Ahmed Abdelkader\textsuperscript{4}\quad
Wen-Sheng Chu\textsuperscript{4}\quad
Brandon Y. Feng\textsuperscript{1}\quad \\
Zhangyang Wang\textsuperscript{1}\quad
Qixing Huang\textsuperscript{1}
\\
\textsuperscript{1}The University of Texas at Austin \quad
\textsuperscript{2}Shanghai Jiao Tong University \quad
\textsuperscript{3}Adobe Research \quad \\
\textsuperscript{4}Google Research \quad
}

\begin{document}
\maketitle
\input{sec/00_abstract}
\input{sec/01_intro}

\input{sec/02_related}

\input{sec/03_preliminary}
\input{sec/03_subspace_extration}
\input{sec/04_vggt_analysis}

\input{sec/06_evaluation}

\input{sec/07_con}
{
    \small
    \bibliographystyle{ieeenat_fullname}
    \bibliography{main_ref, main}
}


% \input{sec/supp_results}

\end{document}

% \maketitle

\maketitlesupplementary
% \beginsupplement

% \clearpage

\input{sec/supp_results}

\clearpage
\clearpage

{
    \small
    \bibliographystyle{ieeenat_fullname}
    \bibliography{main_ref}
}


%% file: macros.tex
\newcommand{\R}{\mathbb{R}}
\let \bs=\mathbf
\let \set=\mathcal

\def \path {\mathit{path}}

\let \set = \mathcal
\let \bs = \boldsymbol

%% file: sec/00_abstract.tex
\begin{abstract}
With the emergence of 3D foundation models, there is growing interest in fine-tuning them for downstream tasks, where LoRA is the dominant fine-tuning paradigm. 
As 3D datasets exhibit distinct variations in texture, geometry, camera motion, and lighting, there are interesting fundamental questions: 1) Are there LoRA subspaces associated with each type of variation? 2) Are these subspaces disentangled (i.e., orthogonal to each other)? 3) How do we compute them effectively?
This paper provides answers to all these questions. 
We introduce a robust approach that generates synthetic datasets with controlled variations, fine-tunes a LoRA adapter on each dataset, and extracts a LoRA subspace associated with each type of variation. 
We show that these subspaces are approximately disentangled.
Integrating them leads to a reduced LoRA subspace that enables efficient LoRA fine-tuning with improved prediction accuracy for downstream tasks. 
In particular, we show that such a reduced LoRA subspace, despite being derived entirely from synthetic data, generalizes to real datasets. 
An ablation study validates the effectiveness of the choices in our approach. 

\end{abstract}

%% file: sec/01_intro.tex
\section{Introduction}

Foundation Models~\cite{DBLP:journals/corr/abs-2108-07258}, pretrained on large-scale datasets with large compute, serve as powerful foundations for solving various downstream tasks via suitable fine-tuning. 
A popular fine-tuning approach is LoRA~\cite{DBLP:conf/iclr/HuSWALWWC22} and its variants, which constrain the number of trainable parameters to mitigate the problems of limited labeled data and overfitting. 
In this paper, we study efficient fine-tuning strategies for 3D foundation models. 

\begin{figure}
\includegraphics[width=1.0\linewidth]{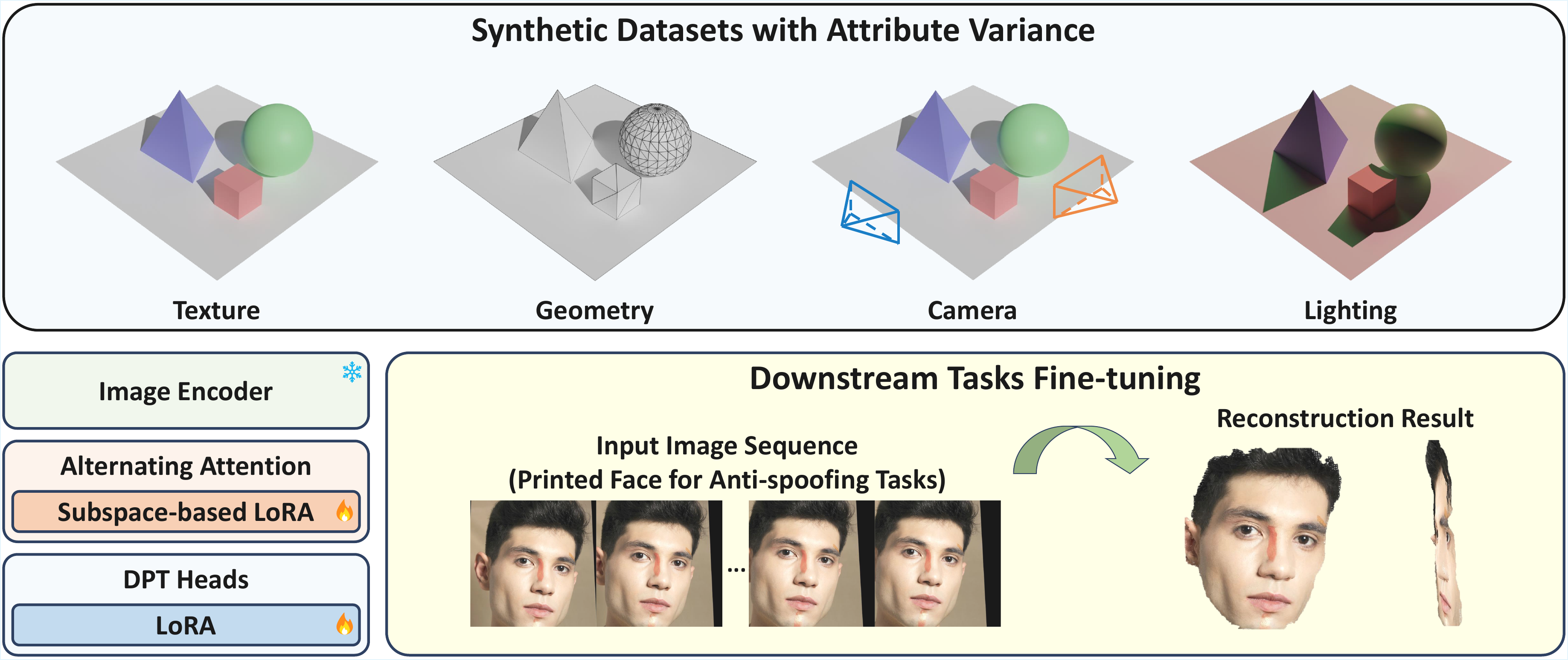}
\caption{Our approach pre-computes LoRA subspaces associated with each type of 3D dataset variation in geometry, texture, camera, and lighting via curated synthetic datasets. These subspaces are integrated into a reduced LoRA basis for efficient fine-tuning.}
\vspace{-0.2in}
\label{Fig:Teaser}   
\end{figure}

To adopt LoRA for 3D vision tasks, we argue that we need to understand how 3D data differ from other domains, and we provide two perspectives.
First, in many cases, it remains costly to obtain even a small amount of real-wold 3D data. 
One such example is to differentiate real 3D face images and 2D face (printed) images from micro-baseline multi-view images. 
This task has important forensics applications such as anti-spoofing, and yet collecting micro-baseline video data is laborious with privacy issues. 
Second, 3D data supports 3D vision tasks that usually focus on low-level visual attributes, such as
texture, geometry, camera motions, and lighting. 
Therefore, we ask a fundamental question: can we create large-scale synthetic data that has \textbf{a different data distribution} from real 3D data, while leveraging LoRA components to discover the underlying patterns of different visual properties that are transferable and can be used for improving fine-tuning performance?

In this paper, we provide positive answers to both questions.
We show how to craft synthetic datasets with controlled variations of visual attributes to fine-tune VGGT~\cite{DBLP:conf/cvpr/WangCKV0N25}. 
We then develop an algorithm that, for each type of variation, extracts a shared subspace from the resulting LoRA adapters.  
Concatenating these shared spaces together yields a concise LoRA basis.
We show the effectiveness of this basis across various downstream tasks, in which we improve both in-distribution tasks and out-of-distribution tasks. To compute the shared space of each attribute, we create synthetic datasets that hold all other attributes relatively fixed while varying the attribute of interest. Specifically, we create multiple data splits and apply LoRA-based fine-tuning on each of them. Each resulting LoRA displacement includes 1) a component shared across all LoRAs tuned for the specific attribute, which is also the component that we aim to extract to represent the attribute, and 2) the components represent data-specific features, which shall be discarded. We formulate the extraction of the shared component as a generalized least squares optimization problem. Moreover, we assess the orthogonality between LoRA subspaces computed for different attributes and find that they are disentangled. Finally, we extract the principal component from the subspaces computed for all attributes, serving as the basis for fine-tuning on new data.
% the resulting displacements include a shared component that we aim to extract, and the components orthogonally represent variance-specific features. 
% We introduce a generalized eigen-decomposition formulation to extract this shared component, addressing ambiguities of factors in low-rank decompositions.
% Based on the hypothesis that the eigenvalues of LoRA adapters are sparsely distributed, the average of these LoRAs can reflect the model's optimization direction in response to the target variance present in these datasets.
% We formulate the extraction of the shared component as a generalized least squares optimization problem.
% Furthermore, the orthogonality between the two resulting subspaces is assessed by calculating the angle between their corresponding matrices.
% In addition, we allocate layer-wise subspace dimensions based on the effective rank of the pretrained parameter matrices.

% A naive approach is to directly compute the shared subspace by randomly varying all attributes to generate the synthetic datasets for LoRA fine-tuning. However, when extracting shared LoRA subspaces, the results tend to focus on strong variations (e.g., geometry and camera motion) among synthetic datasets and can discard critical variations (e.g., lighting) for real datasets due to real-synthetic domain gap. In contrast, extracting the shared subspace for each type of variations nicely addresses this issue.

We evaluate our approach on the task of 3D face reconstruction from micro-baseline videos, 3D human reconstruction from a wide-baseline image setup, and transparent object reconstruction. Experimental results show that the subspaces discovered from our generated synthetic data are transferrable to real data, improving both efficiency and quality of fine-tuning results.
% For example, we observe that our method achieved comparable performance with LoRA when using \textbf{5x less} trainable parameters.

%We evaluate our approach on a variety of tasks, in which subspaces extraction is purely driven by synthetic datasets.
%\todo{We show that our approach offers fine-tuned models with improved performance that is competitive against fine-tuned models with labeled in-distribution data. 
%An ablation study validates the effectiveness of various design choices of our approach. }

%% file: sec/02_related.tex
\section{Related Work}

\noindent\textbf{3D foundation models.}
The 3D foundation models, e.g.,  DUSt3R~\cite{DBLP:conf/cvpr/Wang0CCR24}, VGGT~\cite{DBLP:conf/cvpr/WangCKV0N25}, and RayZer~\cite{jiang2025rayzer, jin2024lvsm, zhao2025erayzer}, have recently achieved strong performance in various 3D vision tasks.
This has created two lines of follow-up work. 
The first line develops variants~\cite{DBLP:conf/cvpr/CabonSACCR025,wang2025pi3permutationequivariantvisualgeometry,DBLP:conf/iclr/0004HHJDCS025,chen2025ttt3r3dreconstructiontesttime,zhang2025advancesfeedforward3dreconstruction,zhao2025erayzer} with expanded capabilities. 
Another line focuses on fine-tuning 3D foundation models (VGGT in particular) for various tasks~\cite{DBLP:conf/iclr/LuYXLI0025,qian2025gp33dgeometryawarepolicy,yao20253d,chen2025hart,zhao2025fastvidar}, in which LoRA~\cite{DBLP:conf/iclr/HuSWALWWC22} is a widely used strategy. 
Due to the effectiveness of LoRAs for fine-tuning VGGT and the fact that 3D datasets exhibit disentangled variations in geometry, texture, camera, and lighting, this paper studies the connection between these variations and LoRA. 
%Specifically, we design synthetic datasets to answer whether there exist subspaces of LoRAs that correspond to each type of variation and whether these subspaces are orthogonal to each other. 

\noindent\textbf{LoRA merging in generative models.}
Learning LoRA subspaces for distinct 3D variation factors and integrating them is closely related to rich prior work in image and video generation, which seeks to combine style LoRAs and content LoRAs. 
Early work, including BLoRA~\cite{blora}, ZipLoRA~\cite{DBLP:conf/eccv/ShahRCLLLJ24}, and LoRA.rar~\cite{shenaj2024lora}, develops efficient training strategies to combine LoRAs. 
Specifically, BLoRA~\cite{blora} identifies the transformer blocks responsible for style and content by curating the input conditions and learns to combine LoRAs from a single image.
ZipLoRA~\cite{DBLP:conf/eccv/ShahRCLLLJ24} introduces column-specific weights to combine two separately trained LoRAs.  
LoRA.rar~\cite{shenaj2024lora} trains a hypernetwork on LoRA corpus to predict column-wise coefficients, which are then used to fuse content and style LoRAs. 
More recent methods explore training-free approaches. 
K-LoRA~\cite{DBLP:conf/cvpr/OuyangLH25} adaptively selects LoRAs based on the analysis of layer-wise LoRA elements. 
%It shows that the initial diffusion steps are responsible for reconstructing the object and capturing coarser texture details, while subsequent steps focus on enhancing finer details of the object and the texture in style. 
EST-LoRAs~\cite{zhang2025subjectstyleadaptivetrainingfree} presents a training-free approach to combine style and content LoRAs driven by a matrix-energy criterion. 
LiONLoRA~\cite{lionlora} introduces a parameter-efficient LoRA fusion framework for video diffusion models, using three key insights: the orthogonality of camera control LoRAs, normalization of LoRA outputs, and the integration of scaling tokens into the attention mechanism for linear control over camera movement and motion strength. 

Our approach differs from this line of work in two ways.
First, rather than developing algorithms to combine two LoRAs, we investigate whether datasets for each type of variation can be encoded using a shared LoRA subspace and how to compute each subspace from synthetic data.
Second, we show that the resulting subspaces corresponding to different variation types are approximately orthogonal and that they can be integrated into a shared LoRA basis for efficient training. Although this shared LoRA subspace is derived from synthetic data, it generalizes well to real data. 

\noindent\textbf{LoRA training strategies.}
Prior work has proposed several LoRA training strategies that explicitly or implicitly control the low-rank subspace in which updates reside.
AdaLoRA~\cite{zhang2023adalora} introduces trainable incremental matrices with dynamic ranks and replaces computationally expensive SVD with a penalty orthogonality loss. However, they did not explore in depth whether dynamically adjusting the rank is task-dependent or influenced by the specific attributes of the task. 
PiSSA~\cite{meng2024pissa} opts to use principal singular values and vectors to initialize LoRA matrices for faster convergence, rather than the usual random initialization while keeping the original weights frozen.
LoRA-GA~\cite{wang2024lora} initializes the LoRA matrices by applying SVD to the gradient matrices, thus approximating the direction of fully fine-tuning. 
GaLore~\cite{zhao2024galore} projects gradients into low-rank approximations, thereby updating within a subspace for memory-efficient optimization, while effectively mimicking the trajectory of fully fine-tuning. 
In contrast to these methods, we introduce precomputed LoRA subspaces for efficient training, with precomputation aligned to distinct variations in 3D attributes. In particular, we show how to derive these LoRA subspaces from synthetic data. 

%Compared with the vast amount of papers on developing new 3D models, research on analyzing what is learned in these models has received less attention, and there are just a handful of relevant prior work. 

%% file: sec/03_preliminary.tex
%\section{Preliminary}
% \label{Sec:Preliminary}

%

%% file: sec/03_subspace_extration.tex
\section{Shared LoRA Subspaces}
\label{Sec:Shared:LoRA:Subspaces}

This section presents our algorithm for extracting the shared subspace from multiple LoRAs adapters, which are obtained by fine-tuning a 3D foundation model on controlled synthetic data. In this work, we focus on VGGT~\cite{DBLP:conf/cvpr/WangCKV0N25}, a representative 3D foundation model, which incorporates 48 sets of self-attention and linear layers. Specifically, each self-attention layer has two matrix parameters (QKV and attention projection), and each linear layer also has two matrix parameters. 

\noindent \textbf{Preliminary}. When applying vanilla LoRA to fine-tune a transformer-based foundation model, it enforces the displacement $dW \in \R^{n\times m}$ of each weight matrix $W$ to be of low-rank $dW = AB^T$, where $A\in \R^{n\times r}$, $B\in \R^{m\times r}$, and the rank $r$ satisfies $r\ll \min(m,n)$.

% In this paper, we are interested in a LoRA subspace which is defined as a pair of matrix $\overline{A}\in \R^{m\times d}$ and $\overline{B}\in \R^{n\times d}$ where $d \geq r$. When performing LoRA fine-tuning using this subspace, we  parameterize $A = \overline{A}X, X\in \R^{d\times r}$ and $B = \overline{B}Y, Y\in \R^{d\times r}$. This leads to a much fewer number of variables to optimize during fine-tuning. 

\noindent \textbf{LoRA Subspace}. We further introduce a specific LoRA subspace parameterization defined by a pair of matrices $\overline{A}\in \R^{n\times d}$ and $\overline{B}\in \R^{m\times d}$. 
When performing LoRA fine-tuning within this subspace, the weight update $dW$ is parameterized as $dW=\overline{A}M\overline{B}^T$, where the matrix $M\in\R^{d\times d}$ is the only trainable parameter. 
This formulation reduces the total number of variables optimized during fine-tuning. 

% In this section, we study how to compute a shared LoRA subspace among multiple relevant LoRAs. Specifically, suppose that we have $k$ LoRAs $\{A_i\in R^{n\times r}\}_{1\leq i \leq k}$ and $\{B_i \in \R^{m \times r}\}_{1\leq i \leq k}$. Our goal is to find a pair of $A\in R^{n\times d'}$ and $B\in \R^{m\times d'}$ for some pre-defined $d'$, so that $AB^T$ approximates all $A_i B_i^T$. We propose to solve the following optimization problem:

\noindent \textbf{Shared LoRA Subspace}. Finally, we address the problem of computing a shared LoRA subspace from an ensemble of $k$ pairs of LoRA weight matrices $\{A_i\in R^{n\times r}\}_{1\leq i \leq k}$ and $\{B_i \in \R^{m \times r}\}_{1\leq i \leq k}$. Our goal is to find a pair of $A\in R^{n\times d'}$ and $B\in \R^{m\times d'}$ for some pre-defined $d'$, such that $AB^T$ optimally approximates all individual updates $A_i B_i^T$. This objective is formalized through the following optimization problem:
\begin{equation}
\min\limits_{A,B} \sum\limits_{i=1}^{k}\|AB^T -A_i B_i^T\|_{\set{F}}^{\alpha}. 
\label{Eq:AB}
\end{equation}
Here, $\|\cdot\|_{\set{F}}$ denotes the matrix Frobenius norm, and the parameter $\alpha$ is introduced to mitigate the influence of potential LoRA outliers in $A_i B_i^T$, which may result from the construction of the datasets.

\begin{figure}
\includegraphics[width=0.46\linewidth]{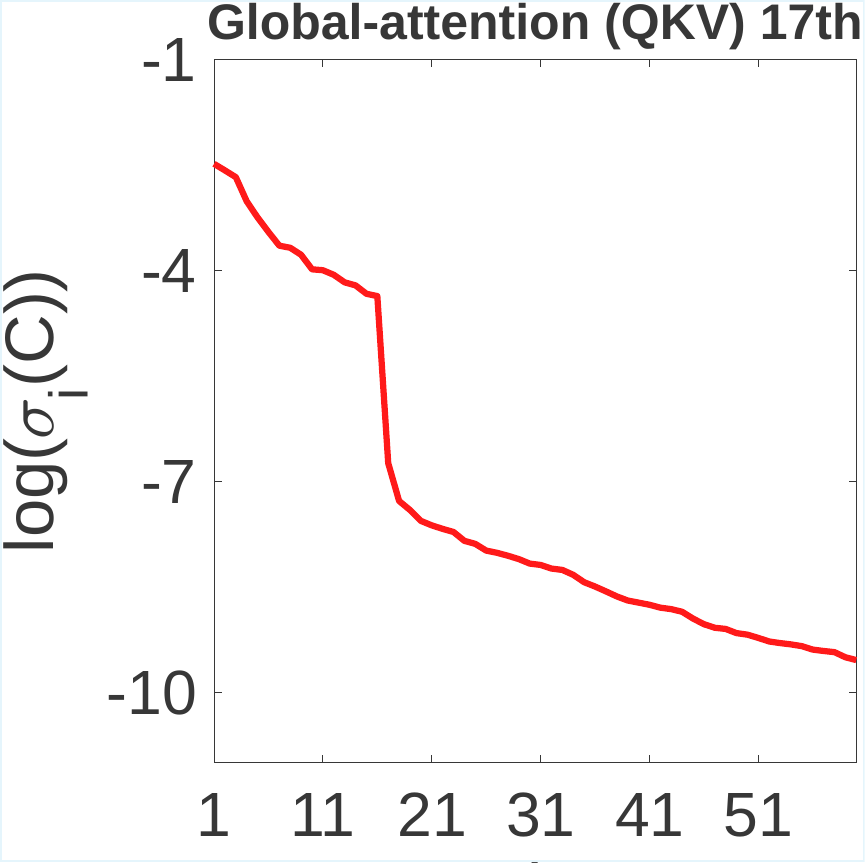}
\hfill
\includegraphics[width=0.46\linewidth]{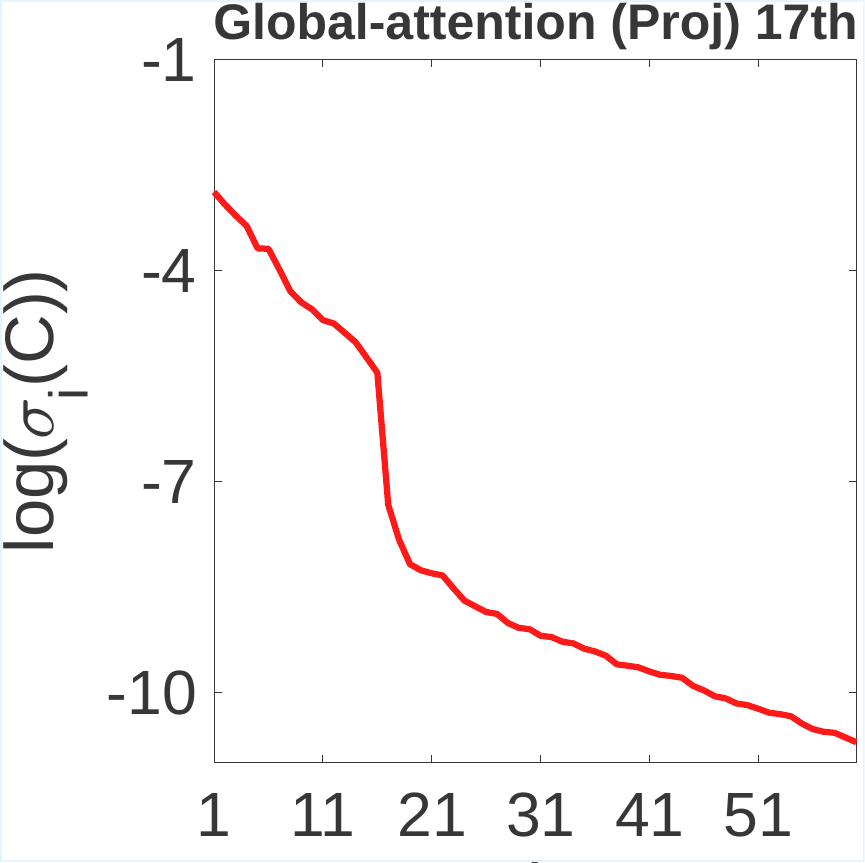}
% \vspace{-0.1in}
\caption{Singular values of $C$ computed from QKV and projection matrices of the 17-th global self-attention layer of 10 LoRAs with respect to geometry variations. Note that the singular values are reported in log-scale. }
\vspace{-0.1in}
\label{Figure:SVD:Spectrum}
\end{figure}

\begin{figure}
\setlength\tabcolsep{4pt}

\begin{tabular}{cccc}
    \includegraphics[width=0.235\linewidth]{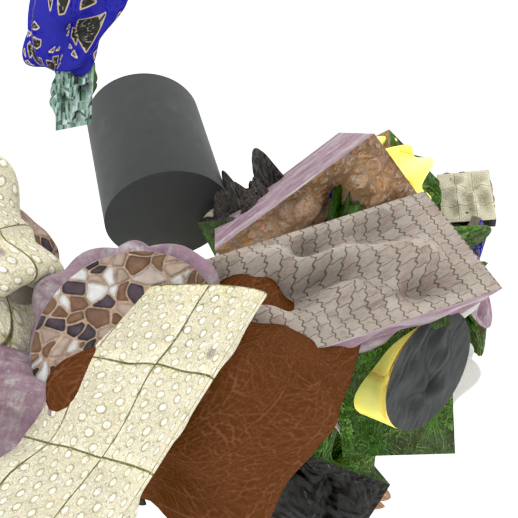} &   
    \includegraphics[width=0.235\linewidth]{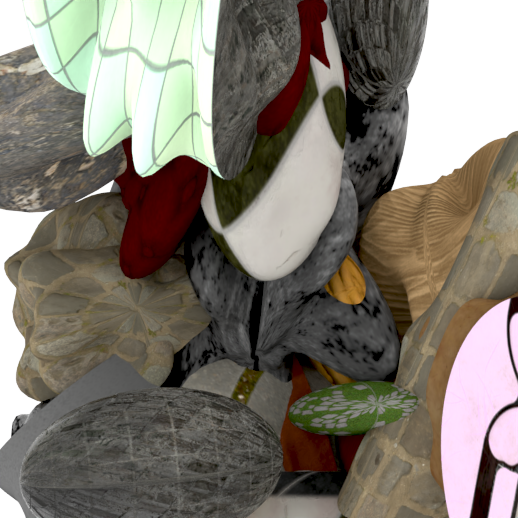}&
    \includegraphics[width=0.235\linewidth]{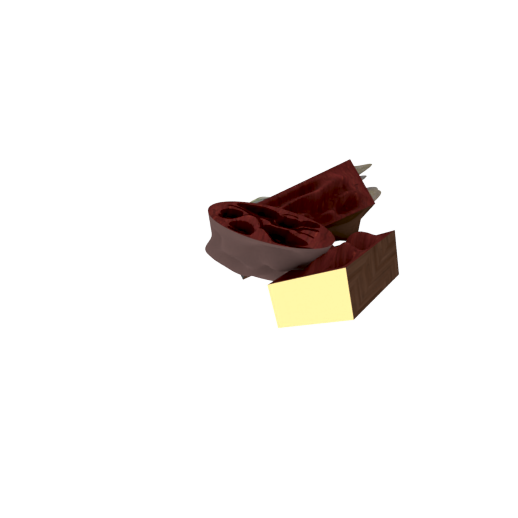} &
    \includegraphics[width=0.235\linewidth]{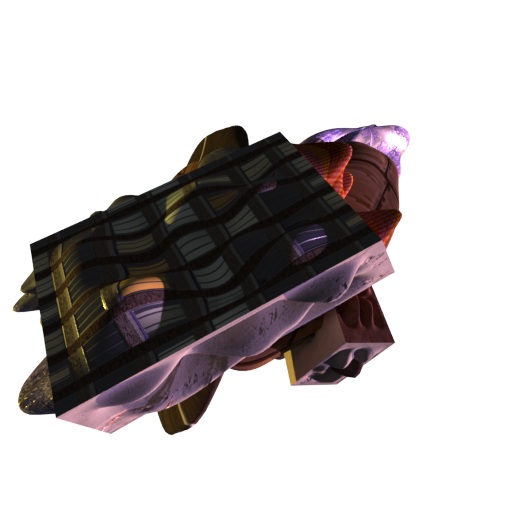}  \\
    \includegraphics[width=0.235\linewidth]{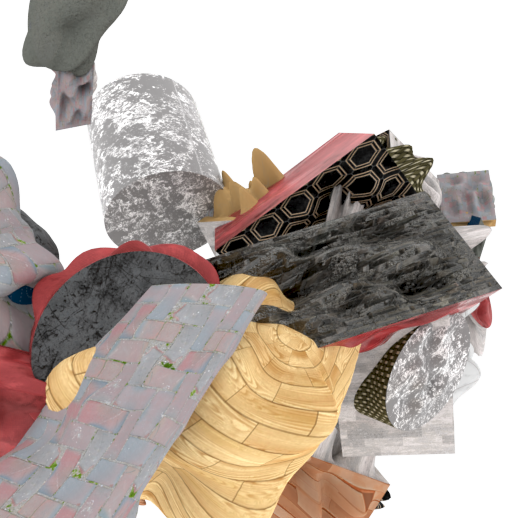} &   
    \includegraphics[width=0.235\linewidth]{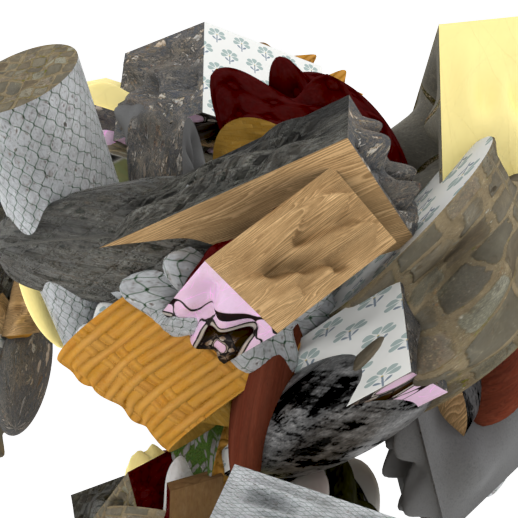}&
    \includegraphics[width=0.235\linewidth]{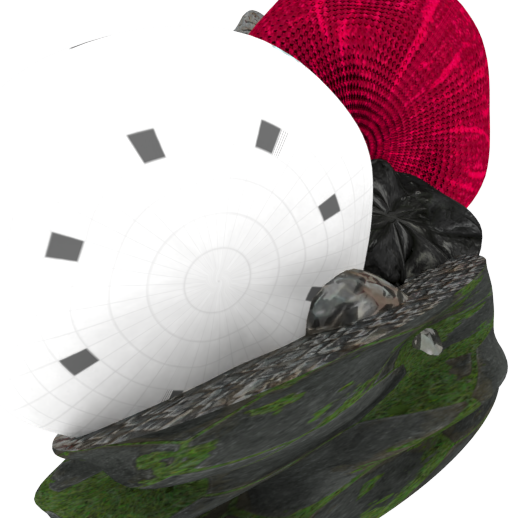} &
    \includegraphics[width=0.235\linewidth]{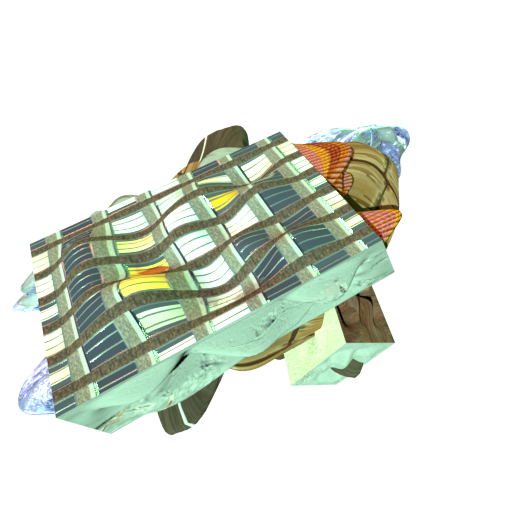} \\
  Texture & Geometry & Camera & Lighting 
\end{tabular}
\vspace{-0.1in}
\caption{Examples of synthetic datasets for each type of variations. We empirically push the variation in each type to the extreme while maintaining small variations among other types. Note that these images are very different from real-world images. }
\vspace{-0.2in}
\label{Fig:Synthetic:Datasets}    
\end{figure}

\begin{figure*}
\setlength\tabcolsep{2pt}
\begin{tabular}{cccc}
\includegraphics[width=0.24\textwidth]{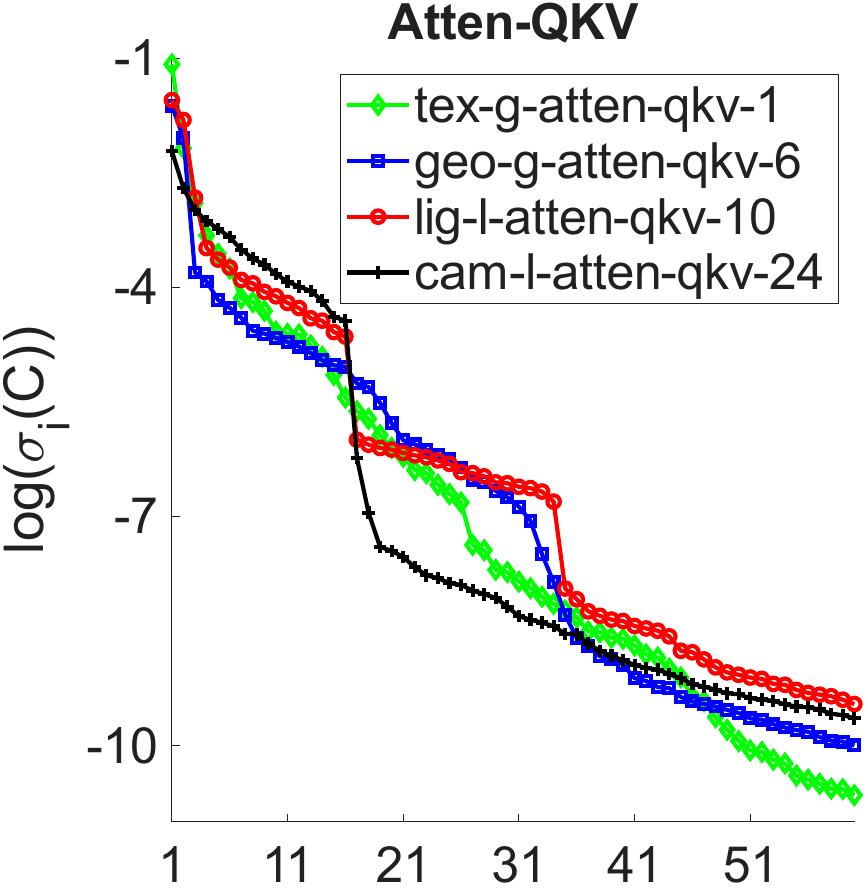}     & 
\includegraphics[width=0.24\textwidth]{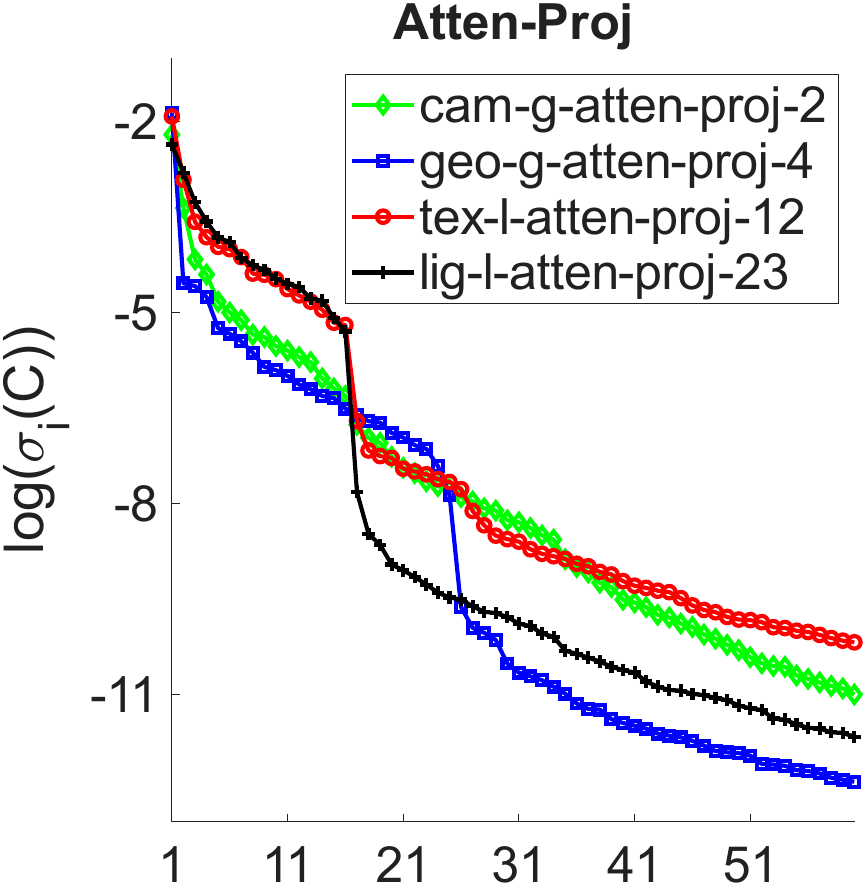}
&
\includegraphics[width=0.24\textwidth]{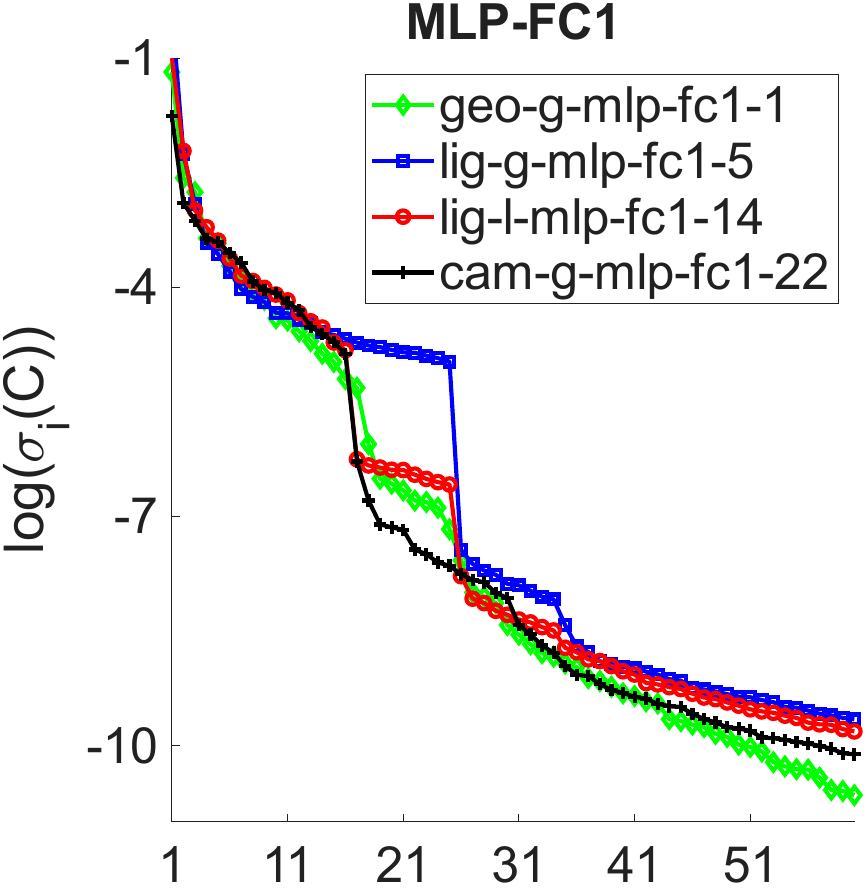}
&
\includegraphics[width=0.24\textwidth]{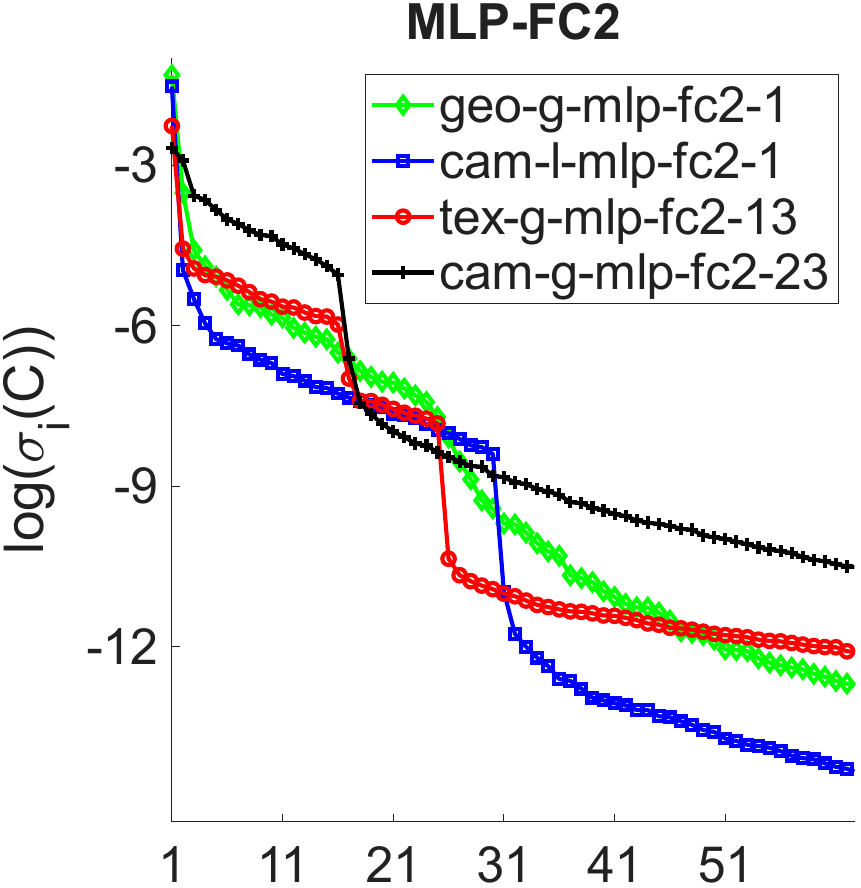}
\end{tabular}
\vspace{-0.15in}
\caption{There are four spectral patterns of matrix $C$ across some layers with respect to different dataset variations. They are colored in red, green, blue, and black. 'tex-g-atten-qkv-1' indicates the QKV matrix in the first global attention layer fine-tuned on texture variations. Note that we have applied a log scale to the singular values. It is clear that there is a significant drop in these curves, which indicates the existence of shared subspaces.}
\vspace{-0.2in}
\label{Fig:Subspace:Analysis}    
\end{figure*}
Note that Eq.~(\ref{Eq:AB}) does not admit closed-form expressions when $\alpha \neq 2$. To solve it effectively, we employ an iteratively reweighted least squares formulation by introducing a weight in front of each term:
\begin{equation}
\min\limits_{A,B} \sum\limits_{i=1}^{k}w_i\|AB^T -A_i B_i^T\|_{\set{F}}^{2}.    
\label{Eq:AB2}
\end{equation}
Starting from $w_i=1,1\leq i\leq k$, we solve Eq.~(\ref{Eq:AB}) by alternating between solving Eq.~(\ref{Eq:AB2}) with fixed $w_i$ and fixing $A$ and $B$ to update $w_i$. When $w_i$ are fixed, it is easy to see that Eq.~(\ref{Eq:AB2}) is equivalent to 
\begin{equation}
\min\limits_{A,B} \|AB^T -C\|_{\set{F}}^2, \qquad C = \sum\limits_{i=1}^k w_iA_iB_i^T/\sum\limits_{i=1}^k w_i.    
\label{Eq:AB3}    
\end{equation}
Let $C = U\Sigma V^T$ be the singular value decomposition of $C$ where the diagonal of $\Sigma = \textup{diag}(\sigma_i)$ encodes the singular values $\sigma_i$ in decreasing order. It is well-known that optimal solutions of $A$ and $B$ are given by $A = U_{d'}{\Sigma_{d'}}^{\frac{1}{2}}$ and $B = V_{d'}{\Sigma_{d'}}^{\frac{1}{2}}$ where $\Sigma_{d'} = \textup{diag}(\sigma_1,\cdots, \sigma_{d'})$ and $U_{d'}$ and $V_{d'}$ encode the corresponding singular vectors. When $A$ and $B$ are fixed, we update $w_i$ as
% $$
% w_i = \frac{\delta^{2-\alpha}}{\big(\delta^2 + \|AB^T-A_i B_i^T\|_{\set{F}}^2\big)^{2-\alpha}}.
% $$
\begin{align*}
 w_i=1/(\varepsilon^2+\| AB^T-A_iB_i^T\|_{\set{F}}^2)^{\frac{2-\alpha}{2}}.   
\end{align*}

\noindent \textbf{Evidence for the Existence of Subspaces}. Fig.~\ref{Figure:SVD:Spectrum} illustrates the singular values of $C$ among the self-attention matrices of the 17-th layer of VGGT when fine-tuned on a synthetic dataset with texture variations, where $d=16$ for each LoRA. We can see that there is indeed a significant drop between the 17-th singular value and the 16-th singular value of $C$, indicating the shared LoRA subspace. However, we also observed that this spectral gap varies between different layers and with respect to different variations. We will discuss such phenomena in Sec.~\ref{Sec:Analysis}.

\noindent \textbf{Leveraging Subspaces for Fine-tuning}. After obtaining the extracted subspaces for all attributes $\{A_iB_i^T\}_{i\in\Lambda}$, we can incorporate them after orthogonalization into our subspace LoRA fine-tuning: 
\begin{align*}
    \overline{A}M\overline{B}=(\Vert_{i\in\Lambda} A_i)\cdot (\Vert_{i\in\Lambda} \{B_i\})^T.
\end{align*}

% \begin{figure}
% \vspace{1.5in}
% \caption{Correlation to effective rank.}
% \label{Figure:Effective_Rank}
% \end{figure}

% As shown in Fig.~\ref{Figure:Effective_Rank}, we further observed a correlation between the singular value decay of $C$ and the effective rank of the original weight matrix $W$. We define this Effective Rank Metric (ERM) using its Frobenius and spectral norms as:
% \begin{align*}
%     \operatorname{ERM}(W)=\sqrt{\operatorname{Effective Rank}(W)}=
%     \frac{\|W\|_{\set{F}}}{\|W\|_2}.
% \end{align*}
% Therefore, the target subspace dimension $d'$ for different layers can be allocated dynamically based on this measure. Specifically, the layer-wise subspace size $d'_l$ is determined by:
% \begin{align*}
%     d'_l=d\times\left\lfloor
%     \frac{\operatorname{ERM}(W_l)}{\operatorname{Average ERM}}
%     \right\rfloor,
% \end{align*}
% where $d$ is the global budget for the subspace size.

%% file: sec/04_vggt_analysis.tex
\section{Subspaces of VGGT}
\label{Sec:Analysis}

This section details the procedure for extracting LoRA subspaces with respect to different types of variation. We begin with the generation of the controlled datasets. 
We then present analysis of subspaces with respect to each type of variation by applying the approach in Sec.~\ref{Sec:Shared:LoRA:Subspaces}.
Finally, we analyze the correlations between these different subspaces.

\subsection{Controlled Dataset Generation}
\label{Subsec:Controlled:Dataset:Gen}

Our datasets were constructed using MegaSynth~\cite{jiang2025megasynth}, which allows control over variations in texture, geometry, camera, and lighting. To extract LoRA subspaces that correspond to different types of variation, we generated multiple specialized datasets. As shown in  Fig.~\ref{Fig:Synthetic:Datasets}, we generate each dataset in each type by varying the corresponding attribute while fixing the remaining attribute types. For different datasets of the same type, the remaining fixed attributes are randomized. The motivation is that we can decompose the corresponding LoRA into two components, where the first component is shared (the attribute of interest) and the second component is random noise (remaining attributes). We can then apply the approach described in Sec.~\ref{Sec:Shared:LoRA:Subspaces} to extract the first component, which is desired. 

Motivated by the principles of domain randomization~\cite{DBLP:conf/iros/TobinFRSZA17}, we maximize the variations of each target type to the extreme. The core motivation here is to ensure that the variations are larger than the real-synthetic domain gap. By doing so, we aim to learn more robust subspaces that exhibit generalization capabilities to real-world data.
% However, we also find that it is crucial to maintain some small variations for other types: this prevents the resulting LoRAs from overfitting to unwanted variations. 
The detailed procedures for generating these synthetic datasets are deferred to the supp. material.

\subsection{Interpreting Individual LoRA Sub-Spaces}
\label{Subsec:Individual:LoRA}

In the following, we analyze the resulting shared LoRA subspaces, focusing on their behaviors across different layers of each type of variation and their properties across different types of variation. 

As illustrated in Fig.~\ref{Fig:Subspace:Analysis}, there are four types of spectral behavior of the LoRA subspace $C$ among different matrices in different layers, which are colored black, blue, red, and green. 
Similarly to Fig.~\ref{Figure:SVD:Spectrum}, the black curves correspond to the layers in which there is a drop at $r$, which is the rank of individual LoRA. We observe that these layers correspond to deep attention layers in VGGT or fully connected layers. This is expected as deep attention or fully connected layers of VGGT focus on global patterns in the input of one attribute and are insensitive to differences across the inputs introduced by relatively small variations with respect to other attributes. 

\begin{figure*}
\setlength\tabcolsep{2pt}
\begin{tabular}{cccc}
\includegraphics[width=0.24\textwidth]{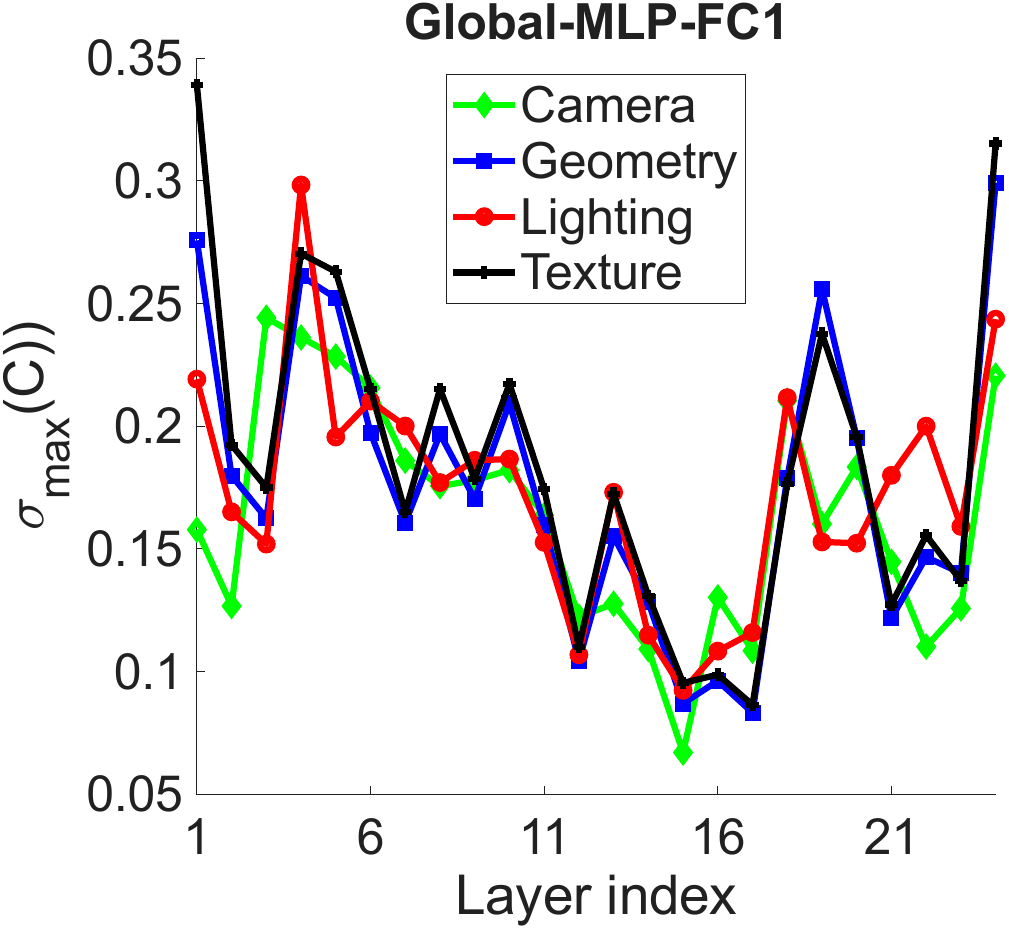}     & 
\includegraphics[width=0.24\textwidth]{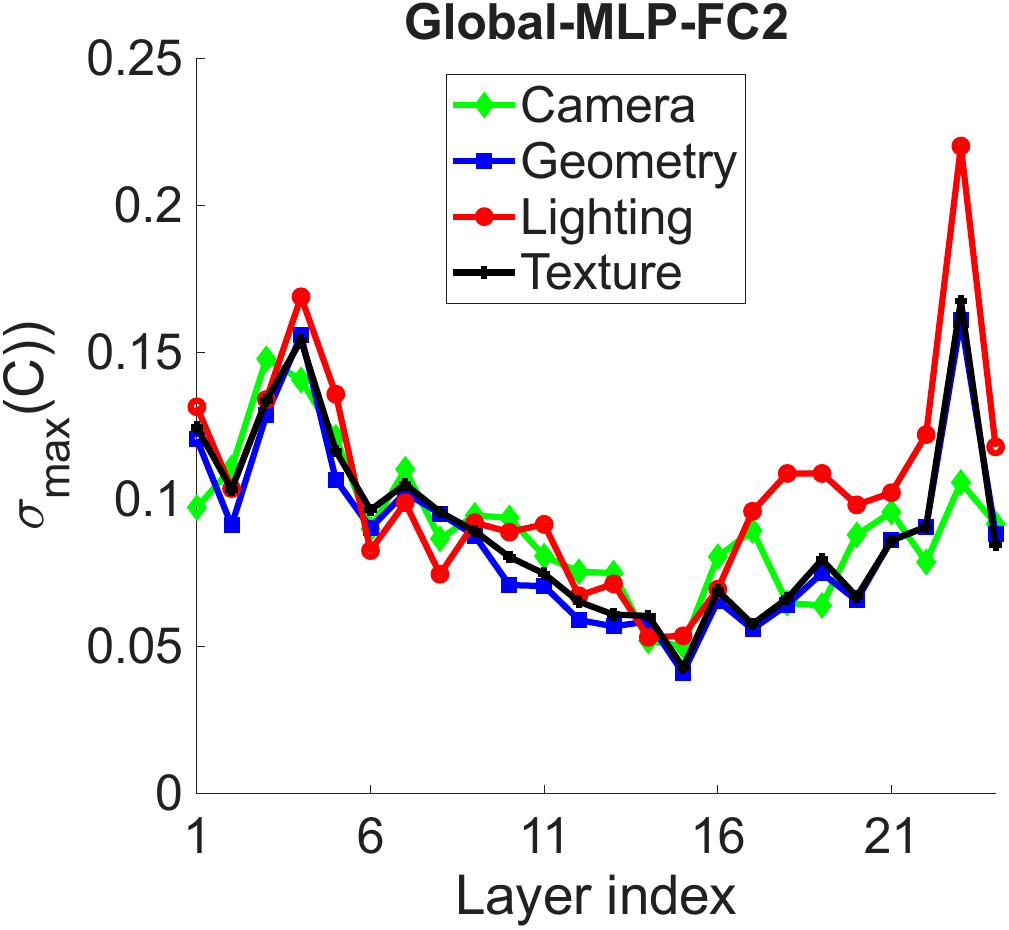}
&
\includegraphics[width=0.24\textwidth]{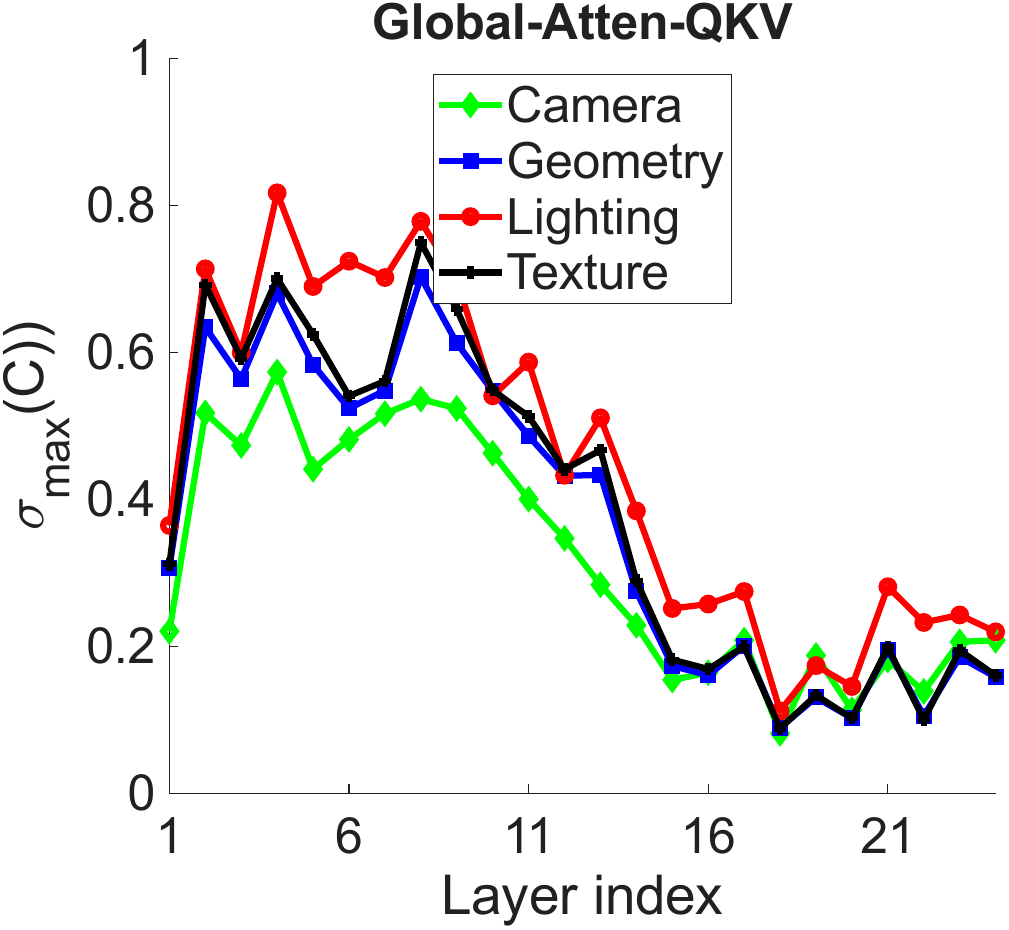}
&
\includegraphics[width=0.24\textwidth]{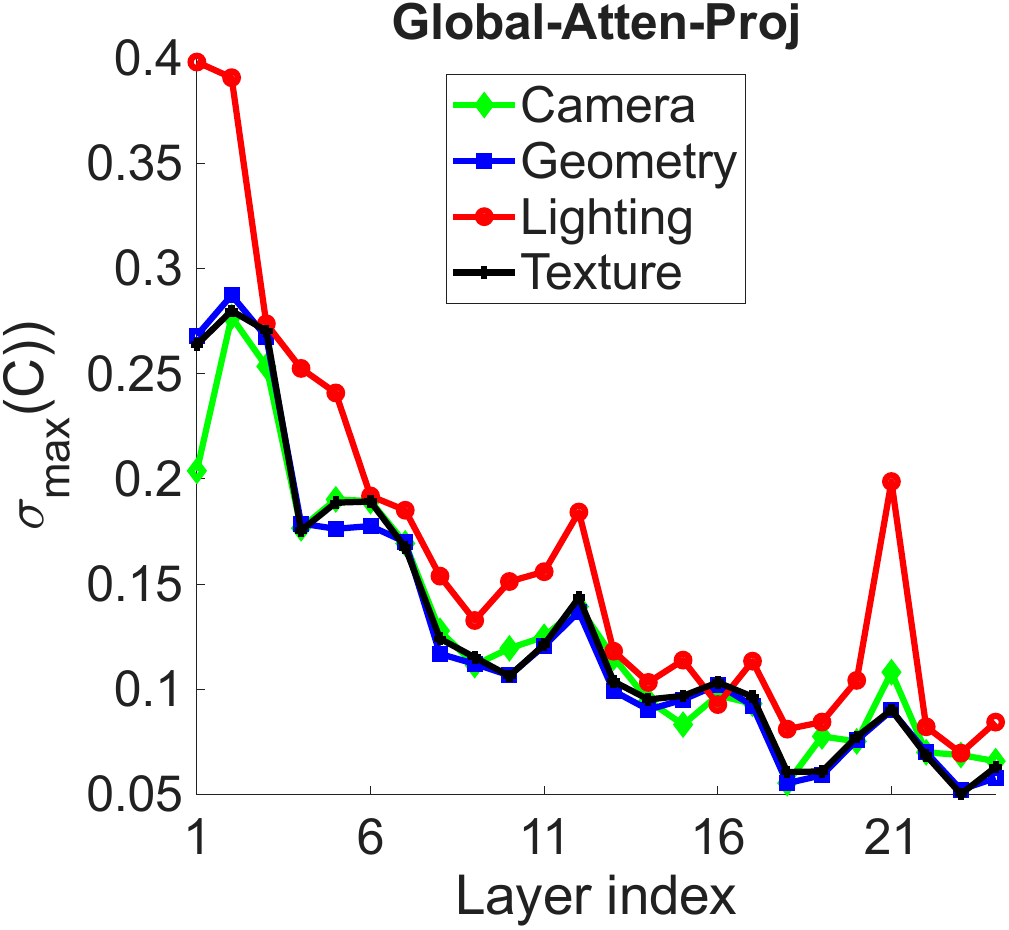}      \\
\includegraphics[width=0.24\textwidth]{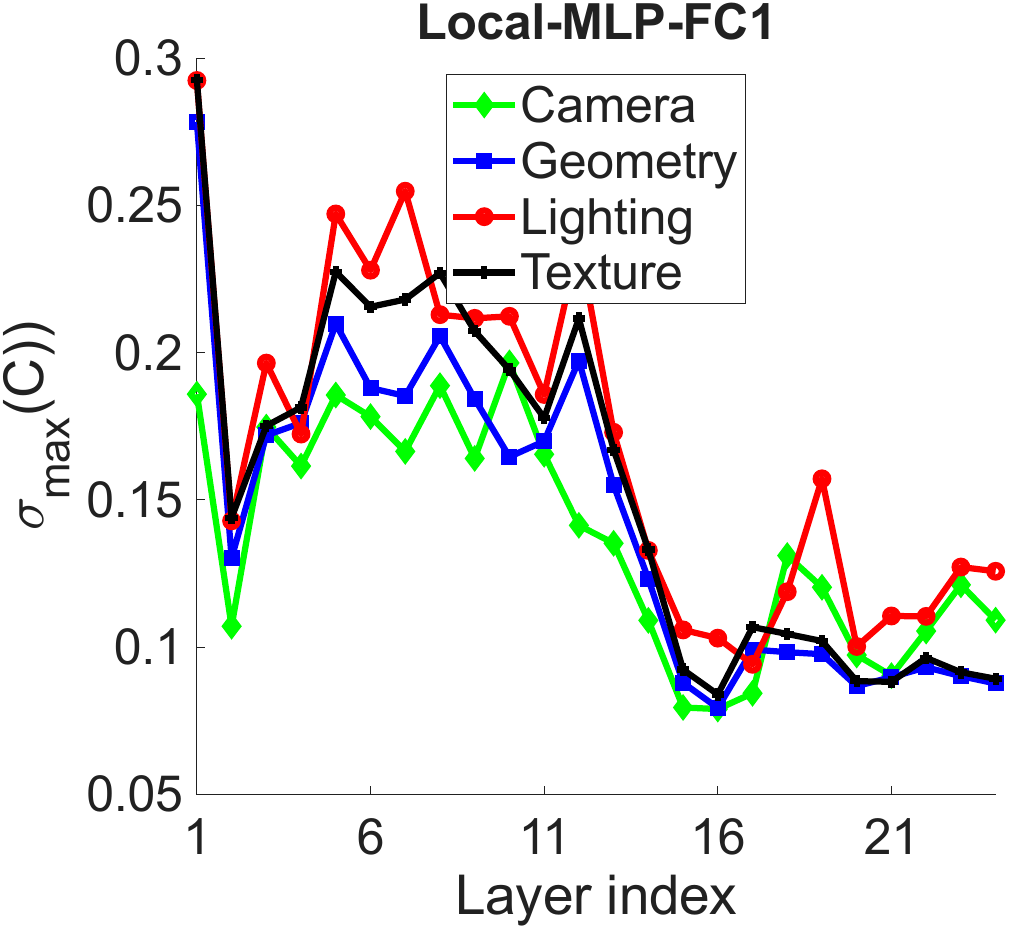}     & 
\includegraphics[width=0.24\textwidth]{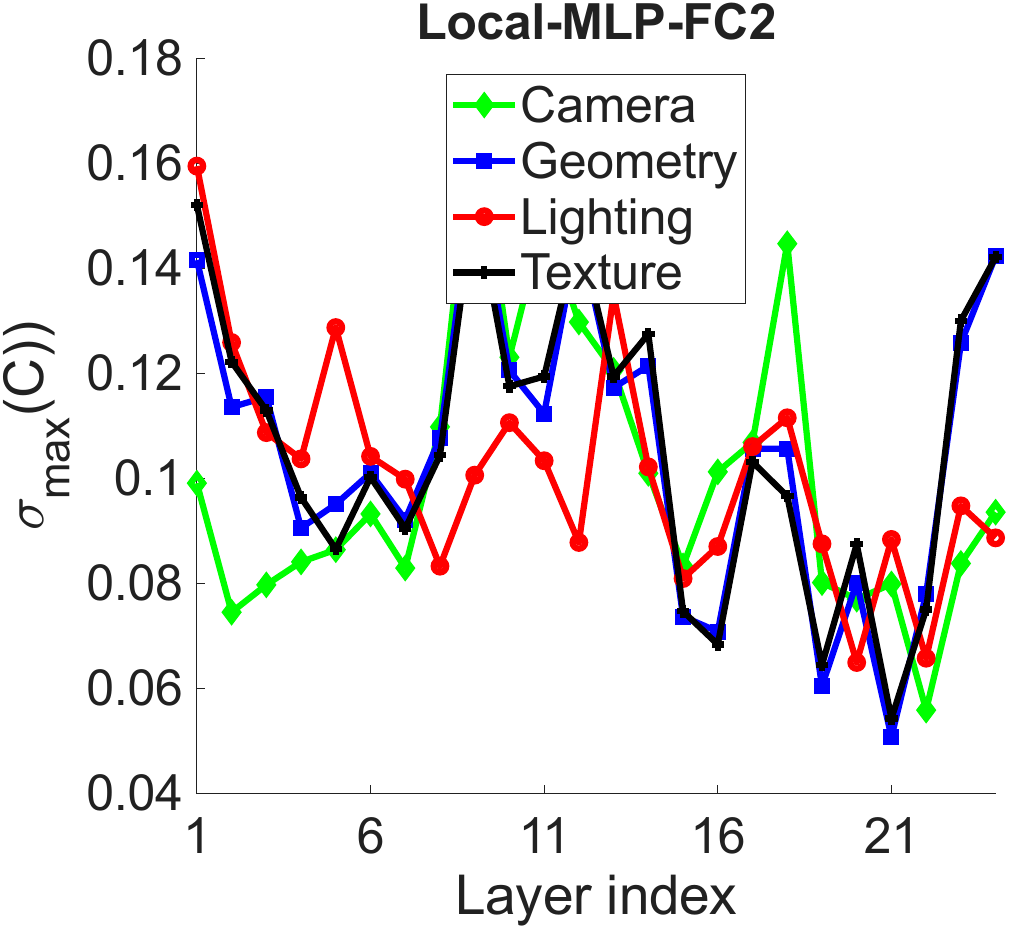}
&
\includegraphics[width=0.24\textwidth]{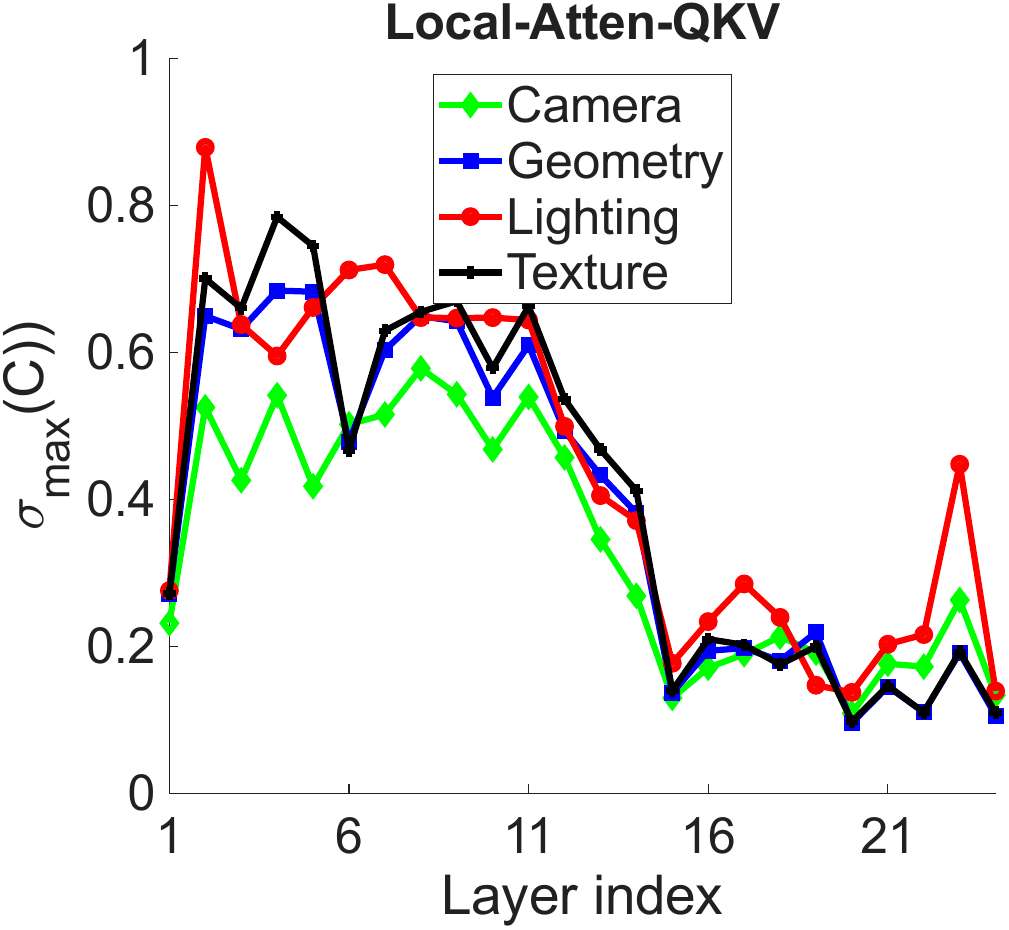}
&
\includegraphics[width=0.24\textwidth]{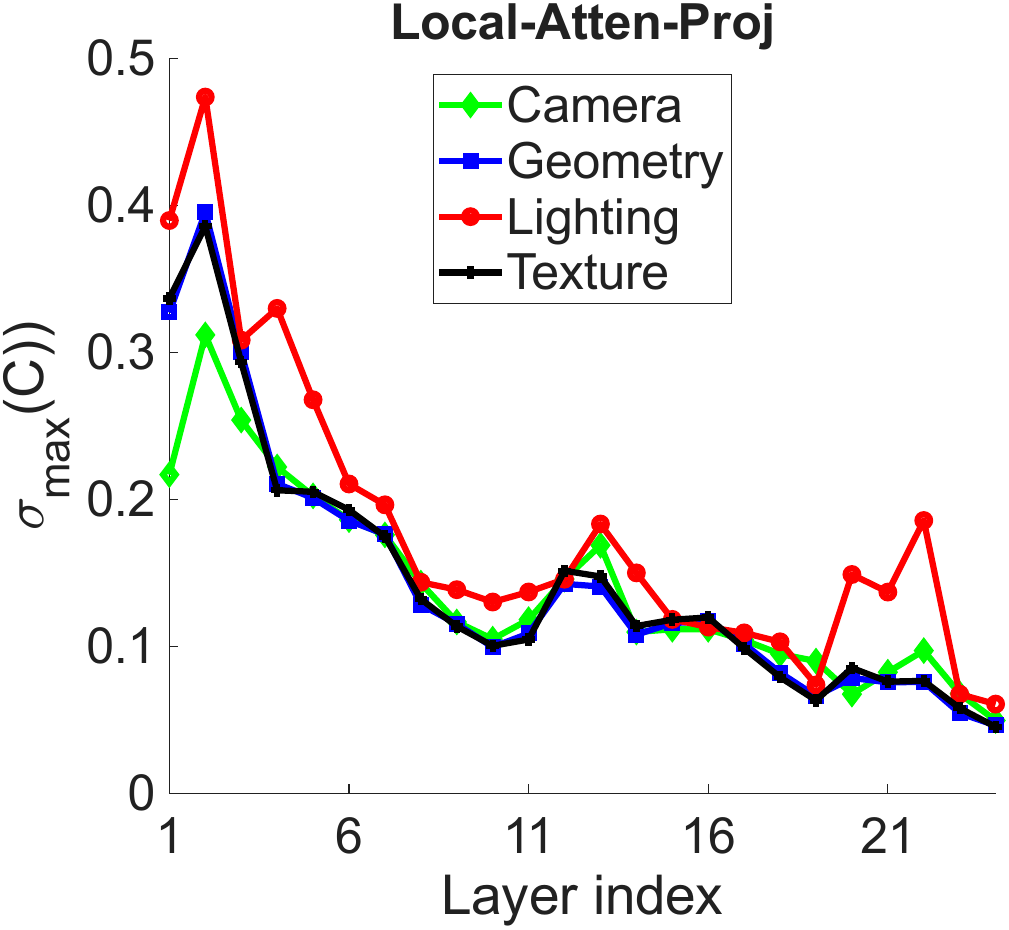}
\end{tabular}
\vspace{-0.1in}
\caption{Maximum singular values of each shared LoRA subspace. Linear layers, QKV, and projection matrices of global self-attention and local self-attention are plotted in the top and bottom rows, respectively. We show four curves in each plot, which correspond to texture, geometry, camera, lighting variations. }
\vspace{-0.2in}
\label{Fig:Subspace:Magnitudes}    
\end{figure*}
The blue curves correspond to the layers in which we still observe a drop in singular values, but the transition point is larger than $d$. Those layers are typically early layers (e.g., 4-7 in VGGT). This can be understood as the fact that these layers tend to capture more local patterns in the collection of synthetic datasets, which also include patterns that do not belong to the attribute of interest. The red curves show two transitions in singular values. They correspond to the layers between those of the blue curves and those of the black curves (e.g., 12-17 in VGGT). In contrast, the black curves show no transitions in singular values. We observe that they correspond to the first 1-3 layers in VGGT. This is expected as they record all local patterns in the synthetic datasets, whose size grows as the number of datasets increases. 

We then analyze the relative spectral properties between different types of variations. 
Although their spectral properties are different, the matrix $C$ is still considered low-rank as $\sigma_{2d}(C)/\sigma_{\max}(C) < 10^{-3}$. 
Fig.~\ref{Fig:Subspace:Magnitudes} plots the maximum singular values of the QKV, projection, and MLP matrices in different layers. We observe three behaviors. First, the magnitudes of the attention matrices drop for deep layers. In other words, global patterns in pre-trained models are generalizable to data variations including synthetic data. In contrast, early layers require significant adjustments in response to local patterns in synthetic datasets. Second, the magnitudes of the matrices in MLP are relatively stable. This again can be understood as the fact that they encode relational pre-trained patterns that are generalizable across real and synthetic datasets. 

Moreover, the relative magnitudes between different variations change drastically, particularly among the MLP layers. This means that it is important to extract an individual subspace for each type variation, as extracting a shared subspace from all types may discard useful subspaces. 

\subsection{Do Subspaces Disentangle? Yes!}

We proceed to study the relation between the extracted subspaces that correspond to different variations. The subspace of each variation type in each layer is encoded as $S=AB^T$. Therefore, we first introduce the distance between two subspaces $S$ and $S'$. This is non-trivial in our context for two reasons. First, $AB^T$ is invariant if we transform $A\rightarrow A X$ and $B\rightarrow  B{X^{-1}}^T$. In other words, we cannot compare $A$ and $A'$ or $B$ and $B'$ directly. We address this issue by enforcing that $A = U\Sigma^{\frac{1}{2}}$ and $B = V\Sigma^{\frac{1}{2}}$, where $U$, $V$, $\Sigma$ come from SVD of $AB^T = U\Sigma V^T$. Second, $S=AB^T$ encodes the same subspace under scaling $aS$. Note that although scaling each column of $S$ encodes the same linear space, it changes the absolute strength of each column, which is useful for characterizing the orthogonality between two subspaces. To address this issue, we define the angle between two matrices $S$ and $S'$ as

\vspace{-1em}
\begin{equation}
\resizebox{0.9\hsize}{!}{$
d(S,S') = \min\limits_{\bs{x},\bs{x}'}\frac{\|S\bs{x}-S'\bs{x}'\|^2}{\|S\bs{x}\|^2 + \|S'\bs{x}'\|^2}=  \min\limits_{\bs{x},\bs{x}'}\frac{\|(S,-S')(\bs{x};\bs{x}')\|^2}{\|\textup{diag}(S,S')(\bs{x};\bs{x}')\|^2}.
$}
\label{Eq:d:A:Aprime}
\end{equation}
\vspace{-1em}

It is clear that if $\bs{x}$ and $\bs{x}'$ is an optimal solution to $d(S,S')$, then $\bs{x}$ and $\frac{1}{a}\bs{S}'$ are an optimal solution to $d(S,aS')$. Therefore, $d(S,S')$ is invariant when scaling $S$ and $S'$. It is easy to see that the optimal solution to Eq.~(\ref{Eq:d:A:Aprime}) is given by the smallest generalized eigen-vector problem:

\vspace{-1em}
\begin{equation}
\resizebox{0.9\hsize}{!}{$
\left(
\begin{array}{cc}
S^TS & -S^T S' \\
-{S'}^T S & {S'}^T S'
\end{array}
\right)\left(
\begin{array}{c}
\bs{x}\\
\bs{x}'
\end{array}
\right) \nonumber \\
= \lambda \left(
\begin{array}{cc}
S^TS & 0 \\
0 & {S'}^T S'
\end{array}
\right)\left(
\begin{array}{c}
\bs{x}\\
\bs{x}'
\end{array}
\right).
\label{Eq:Gen:Eigenvalue}
$}
\end{equation}
\vspace{-1em}

\begin{figure*}
\setlength\tabcolsep{2pt}
\begin{tabular}{cccccc}
\includegraphics[width=0.16\textwidth]{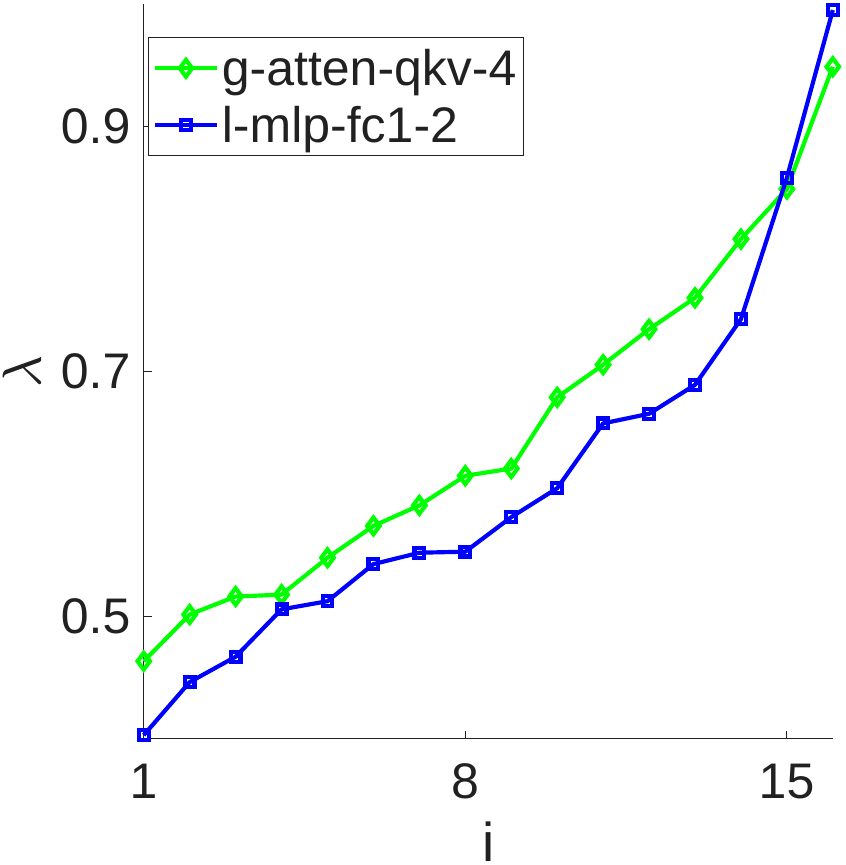}     & 
\includegraphics[width=0.16\textwidth]{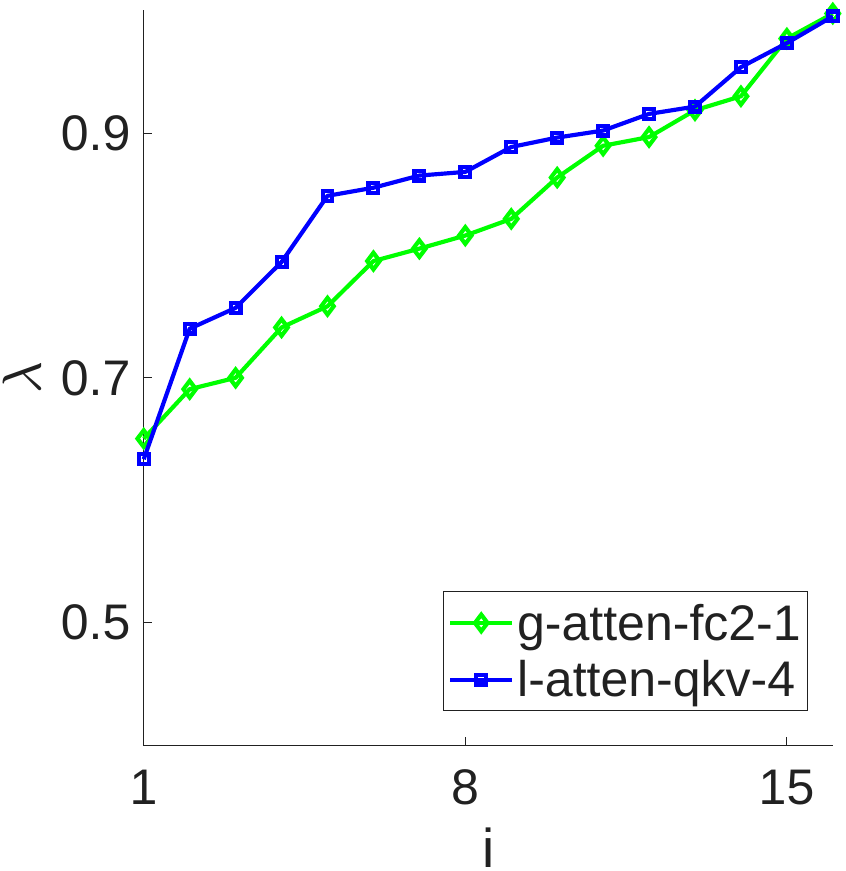}
&
\includegraphics[width=0.16\textwidth]{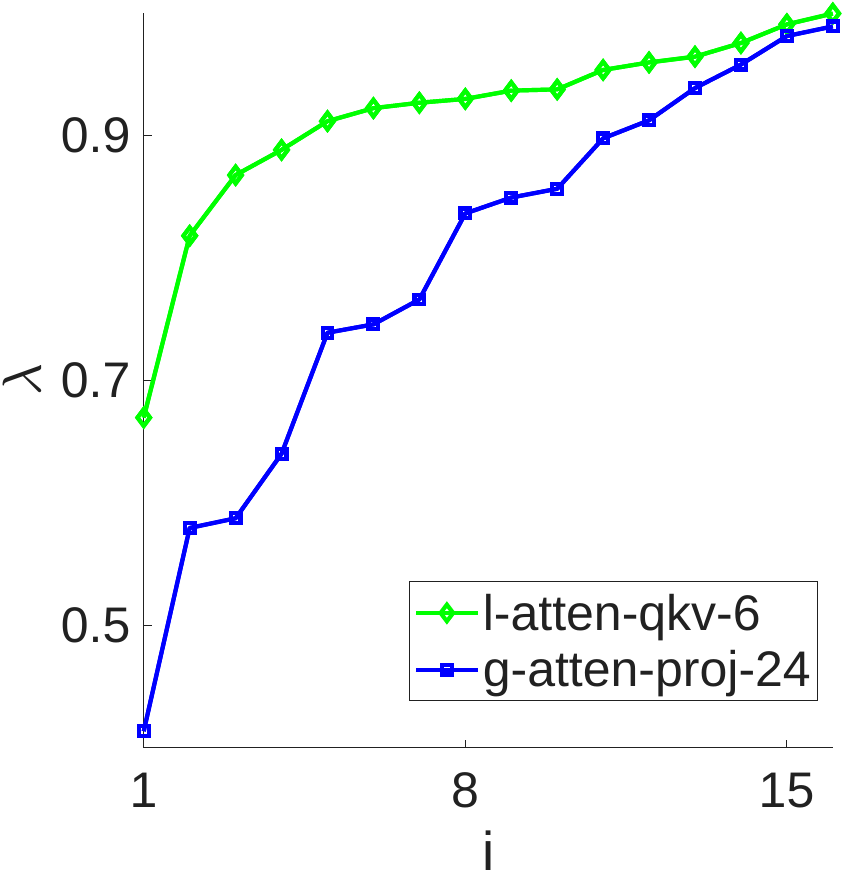}
&
\includegraphics[width=0.16\textwidth]{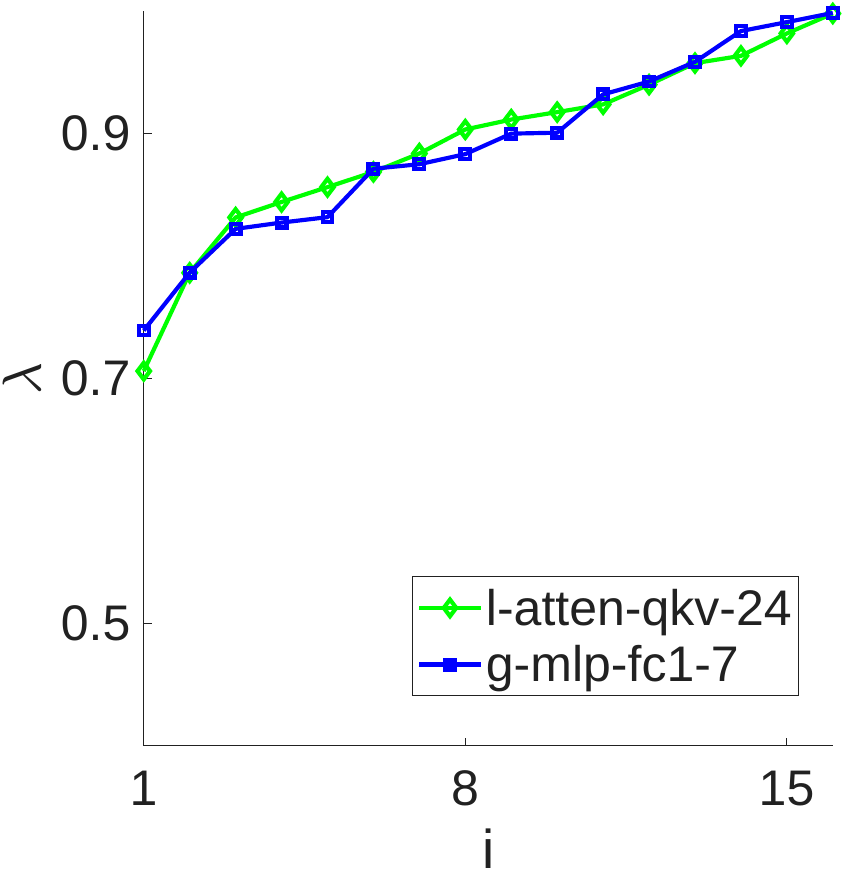}
&
\includegraphics[width=0.16\textwidth]{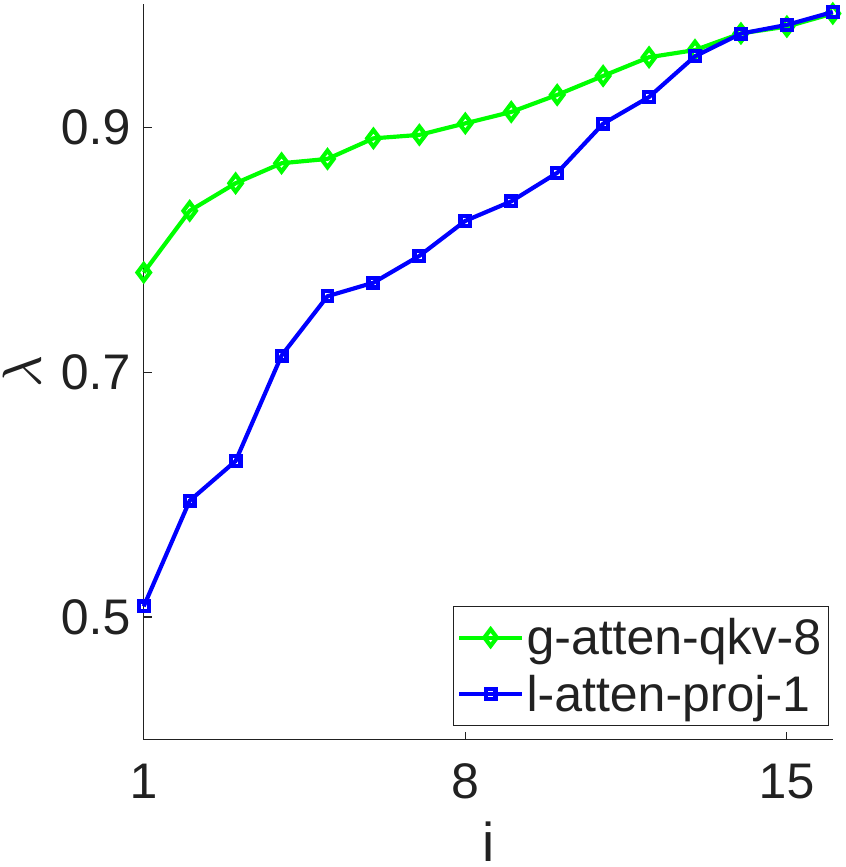}
& 
\includegraphics[width=0.16\textwidth]{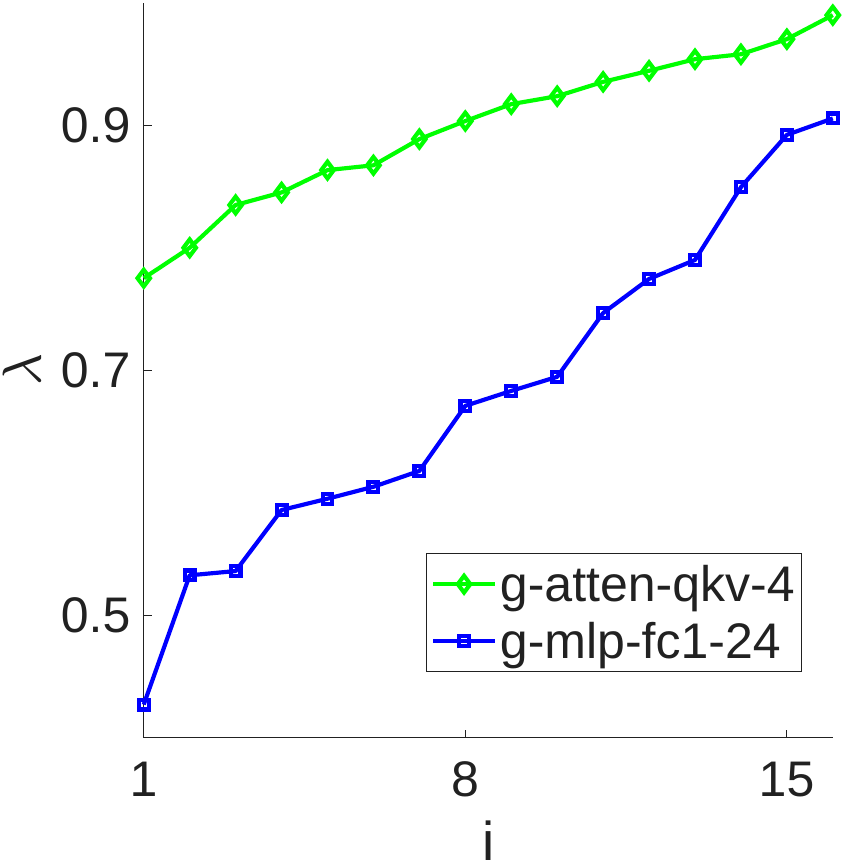} \\
Geo.+Tex.   & Geo. + Cam. & Geo.+Lig. & Tex. + Cam. & Tex. + Lig.& Cam. + Lig.  
\end{tabular}
\vspace{-1em}
\caption{We show overlap ratios between six pairs of four subspaces that correspond to variations in geometry, texture, camera, and lighting. We show three representative layers of the global QKV attention. This metric, akin to reprojection error, reveals that each pair of subspaces is approximately orthogonal. Notably, the orthogonality is most significant between texture and camera, as well as between geometry and camera.}
\vspace{-1em}
\label{Fig:Disentanglement}
\end{figure*}
Fig.~\ref{Fig:Disentanglement} shows the eigvenvalues in Eq.~\ref{Eq:Gen:Eigenvalue} for typical layers between six pairs of subspaces $(S, S')$. In general, most of the smallest eigenvalues are above $0.5$, indicating that the learned subspaces are disentangled ($1$ means that they are orthogonal). Moreover, both the geometry and texture subspaces show strong orthogonality with the camera subspace. 

Although images and corresponding learned features shall be a complex non-linear function of attribution variations, recent results in deep learning theory, neural tangent kernels~\cite{DBLP:conf/icml/DuLL0Z19} and diffusion generalization~\cite{DBLP:conf/iclr/KadkhodaieGSM24}, have shown that this non-linear function can be approximated by a linear function defined by the Jacobian of the network. Intuitively, this linear relationship promotes disentanglement while high-order residuals characterize correlations. We will leave a rigorous analysis of this matter for future work.

%% file: sec/06_evaluation.tex
\section{Experimental Evaluation}

We adopt VGGT~\cite{DBLP:conf/cvpr/WangCKV0N25} as our base model.
We first conducted 2D face anti-spoofing experiments to validate the effectiveness of the extracted subspaces. To evaluate generalization, we further conducted clothed human reconstruction experiments, demonstrating that subspaces extracted from synthetic data can generalize to real-world data. Finally, we evaluated our method on transparent object reconstruction, showing that the learned subspaces remain effective even under challenging settings with complex materials and limited training data.

\noindent\textbf{Baselines.}
We compare our subspace-based fine-tuning strategy with full fine-tuning, and several representative PEFT methods, including LoRA~\cite{DBLP:conf/iclr/HuSWALWWC22} and PiSSA~\cite{meng2024pissa}.
For prediction, we use the depth head instead of the point head. During fine-tuning, we freeze the DINO encoder and maintain the same number of training steps across all experiments, employing a two-stage learning rate scheduler that combines linear and cosine decay.

\subsection{2D Face Anti-Spoofing}

For the subspaces, we apply the texture and geometry subspaces extracted on our created synthetic data under micro-baseline settings. Each subspace is derived from five LoRA adapters with a rank of 16.

\noindent\textbf{Datasets.}
We fine-tune on data of indoor scenes created by MegaSynth~\cite{jiang2025megasynth} and rendered under micro-baseline cameras, following the protocol in Sec.~\ref{Subsec:Controlled:Dataset:Gen}.
The test set of human face consists of two parts: one part consists of face images collected from the internet, which are used to render micro-baseline videos.
The other part includes real-world face data captured with an iPhone equipped with LiDAR hardware to estimate the depth of printed images~\cite{chugunov2023shakes, chugunov2022implicit}.
All test image sequences consist of 42 images, with one image selected every six frames for the synthetic evaluation, and one image selected every two frames for the real-world evaluation.
%\qixing{What is the meaning of first/second tasks?}

\noindent\textbf{Metrics.}
For the synthetic face dataset, we evaluate the quality of the point cloud using the Chamfer Distance ($\times10^{-3}$) and the normal consistency. 
For Chamfer Distance, we report its two directional components: \textit{Accuracy} and \textit{Completeness}.
Additionally, we first use SAM 2~\cite{ravi2024sam} to extract the face mask from the video, and the metrics are calculated based on this mask.
The predicted point maps are first aligned with the ground truth using the Kabsch-Umeyama algorithm~\cite{Lawrence2019} for an initial Sim(3) alignment, followed by refinement using the Iterative Closest Point (ICP) algorithm~\cite{besl1992method}. 
For the real-world face dataset, we evaluate the quality of depth estimation using the Absolute Relative Error ($\times10^{-2}$) and the prediction accuracy at a threshold of $\delta < 1.25$. 
These metrics are evaluated under joint scale and 3D translation alignment.

\begin{table}[t]
    \centering
    
    % \vspace{-0.1in}
    \resizebox{1.0\columnwidth}!{
    \begin{tabular}{lcccccc}
        \toprule
        {\multirow{3}{*}{\textbf{Method}}} &
        {\multirow{3}{*}{\shortstack{\textbf{\# Trainable}\\\textbf{Param.}}}} &
        \multicolumn{3}{c}{\textbf{Synthetic Face Dataset}} &
        \multicolumn{2}{c}{\textbf{Real Face Dataset}} \\
        \cmidrule(r){3-5} \cmidrule(r){6-7}
        & & 
        {Acc $\downarrow$} & {Comp $\downarrow$} & {NC $\uparrow$} &
        {Abs Rel $\downarrow$} & {$\delta<1.25$ $\uparrow$} \\
               
        \midrule
        VGGT
        & -         & 9.006 & 4.965 & 80.74 & 2.651 & 98.59 \\
        % VGGT+
        Full
        & 853.6 M   & 5.585 & 3.531 & 85.77 & 2.203 & 98.85 \\
        \midrule
        % VGGT+
        LoRA (rank=16)
        & 16.3 M    & 5.767 & 3.385 & 84.78 & \textbf{2.115} & 98.92 \\
        % VGGT+
        LoRA (rank=32)
        & 32.7 M    & 6.251 & 3.841 & 84.64 & 2.159 & 98.93 \\
        % VGGT+
        LoRA (rank=64)
        & 65.3 M    & 6.393 & 3.971 & 84.59 & 2.157 & \underline{98.94} \\
        % VGGT+
        LoRA (rank=128)
        & 130.7 M   & 6.590 & 4.242 & 84.92 & 2.162 & \textbf{98.95} \\
        % VGGT+
        % LoRA (rank=256)
        % & 261.4 M   & 6.661 & 4.318 & 84.38 
        % & 2.306 & \textbf{98.97} \\
        \midrule
        % VGGT+
        PiSSA (rank=16)
        & 16.3 M    & 5.729 & 3.532 & 85.30 & 2.433 & 98.81 \\
        % VGGT+
        PiSSA (rank=32)
        & 32.7 M    & 6.488 & 4.198 & 84.64 & 3.185 & 98.41 \\
        % VGGT+
        PiSSA (rank=64)
        & 65.3 M    & 6.526 & 4.407 & 84.84 & 3.106 & 98.60 \\
        % VGGT+
        PiSSA (rank=128)
        & 130.7 M   & 6.890 & 4.706 & 84.77 & 4.020 & 98.32 \\
        % VGGT+
        % PiSSA (rank=256)
        % & 261.4 M   & 6.144 & 4.089 & 84.86 
        % & 4.876 & 96.40 \\
        \midrule
        Ours ($d$=8)
        & 3.8 M     & 5.921 & \textbf{1.966} & 76.77 & 2.774 & 98.26 \\
        Ours ($d$=16)
        & 4.0 M     & \textbf{3.831} & \underline{2.037} & \textbf{86.65} & 2.170 & 98.92 \\
        Ours ($d$=32)
        & 4.7 M     & \underline{4.287} & 2.395 & \underline{86.43} & \underline{2.151} & \underline{98.94} \\
        \bottomrule
    \end{tabular}
    }
    \caption{
        \textbf{Evaluation on Synthetic and Real-World Human Face Datasets under Micro-Baseline Settings.} 
        The synthetic dataset contains 50 face images, which are used to render videos from nine different viewpoints. 
        The real-world dataset consists of 50 printed images, each captured in one long-burst bundle.
        \textbf{Bold}: best; \underline{underline}: second best.
    }
    \vspace{-0.2in}
    \label{tab:2D_eval_table}
\end{table}

As presented in Table \ref{tab:2D_eval_table}, 
%when the fine-tuning dataset consists solely of synthetic data (i.e., indoor scenes without semantic meaning), a significant domain gap exists from the test set. However, due to the similarity in the distribution of specific 3D attributes (such as camera motion under micro-baseline settings), the fine-tuned model can generalize to the test set by improving its robustness to new variations, rather than merely memorizing the results.
our proposed subspace-based fine-tuning method significantly outperforms other fine-tuning methods on the synthetic face test set. Our method achieves comparable results on the real-world data set, while utilizing fewer trainable parameters. In contrast, LoRA lacks interpretability, and PiSSA exhibits overfitting.

\begin{figure}[t]
    \centering
    \includegraphics[width=1.0\linewidth]{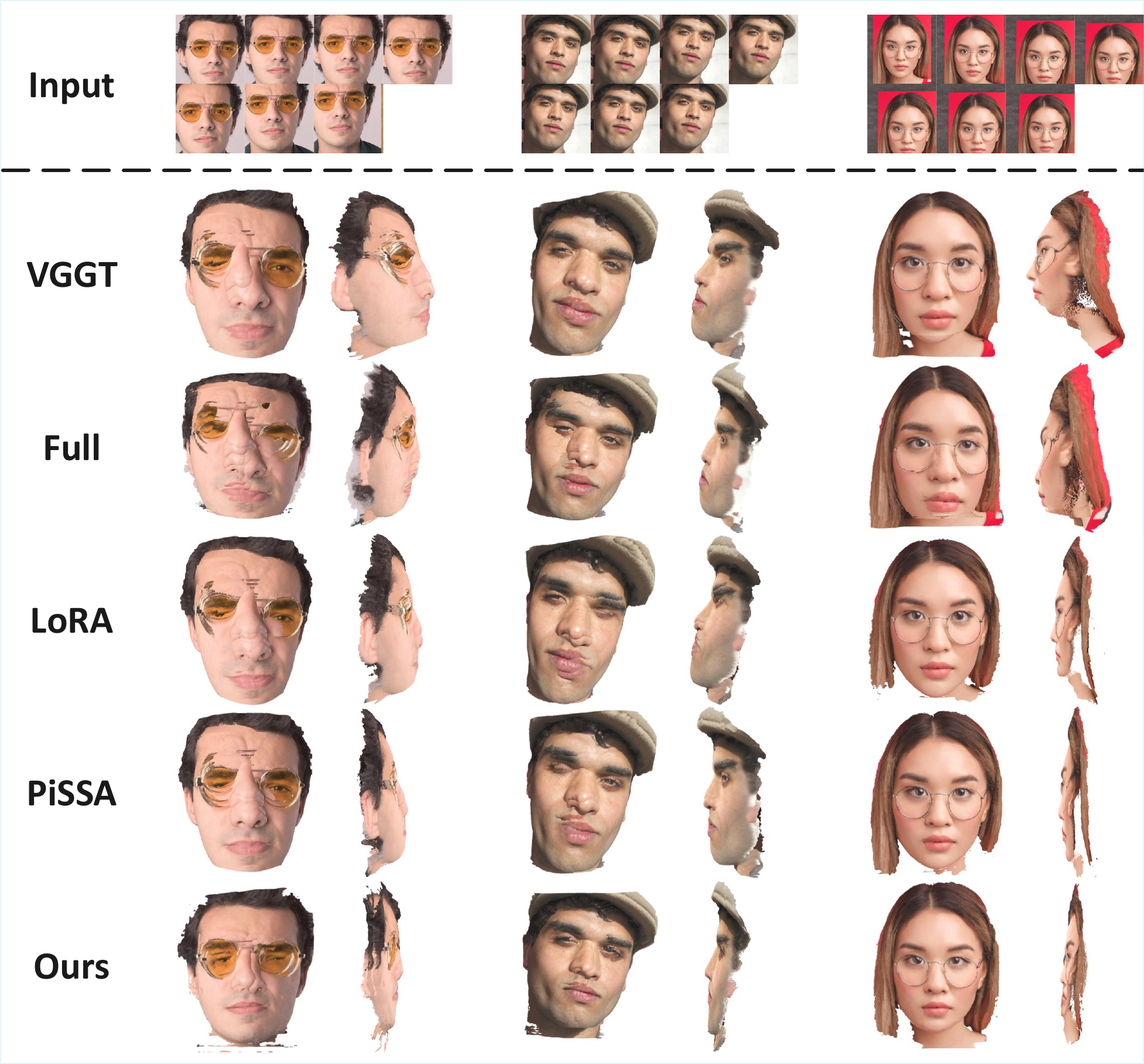}
    \vspace{-0.25in}
    \caption{\textbf{Qualitative Comparison of 2D Face Anti-Spoofing Tasks.}
    Compared to other fine-tune strategies, our method produces more accurate and robust reconstruction with fewer artifacts.}
    \label{fig:visual_2d}
    % \vspace{-1em}
\end{figure}

Qualitative visual results of point cloud reconstruction on the synthetic test dataset are shown in Fig.~\ref{fig:visual_2d}. The original VGGT model is heavily tricked by its visual semantic priors learned on normal real-world 3D face data, leading to scattered artifacts in the reconstructed faces. All fine-tuning strategies alleviate these issues to some degree by consuming micro-baseline data for fine-tuning, while our method produces the most accurate and robust reconstructions with noticeably fewer artifacts, showing the transferability of our method to out-of-distribution data.

\subsection{Clothed Human Reconstruction}

\begin{figure*}[t]
    \centering
    \includegraphics[width=1.0\linewidth]{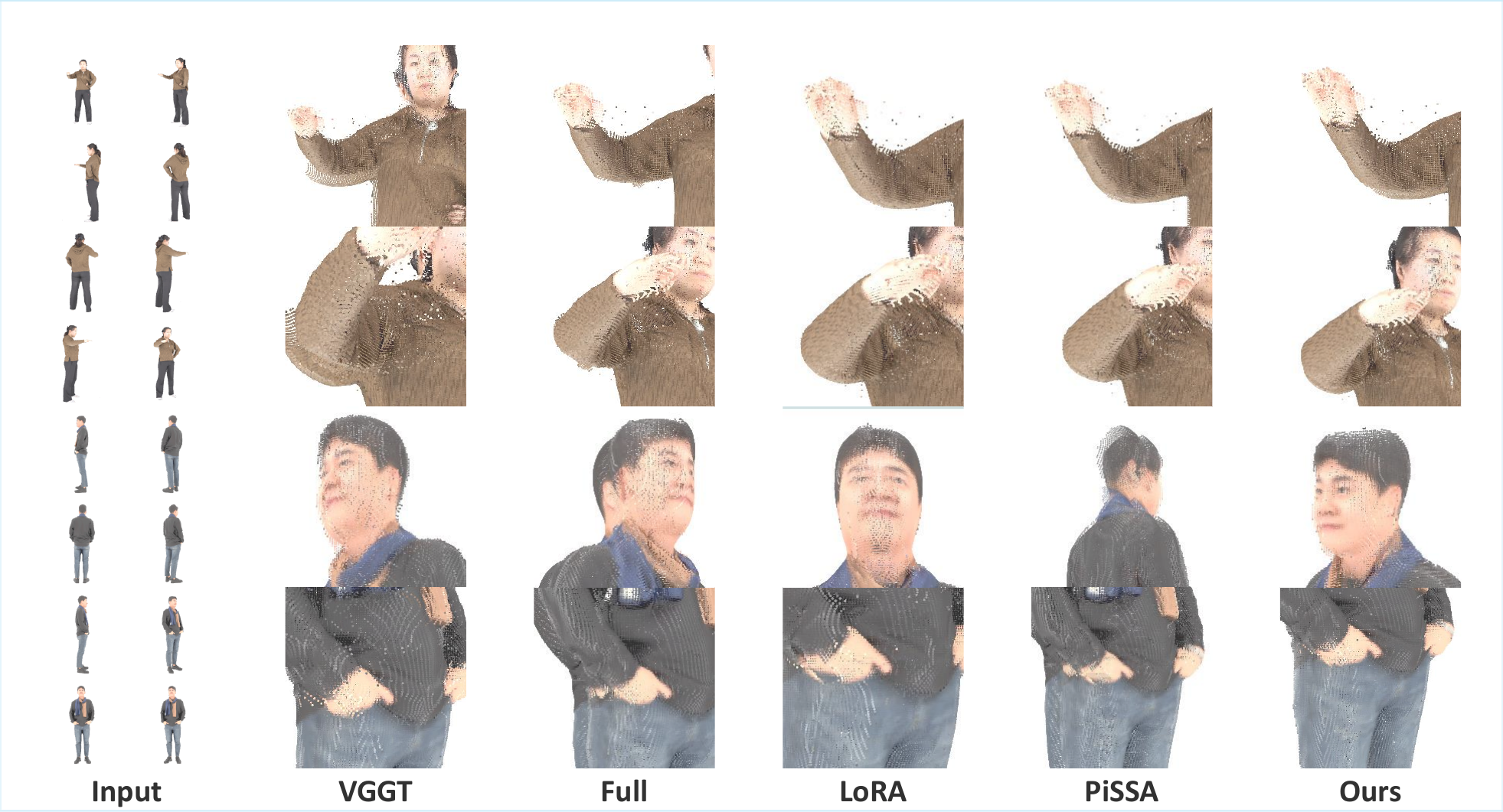}
    \caption{\textbf{Visual Results of Clothed Human Reconstruction Tasks.} Each input consists of eight different viewpoints. The first row is selected from the THuman 2.1 test split, while the second row is chosen from the 2K2K test set. Our model produces fewer artifacts on the object, but its performance on more detailed regions, such as the hands, is less ideal.}
    \vspace{-0.1in}
    \label{fig:visual_3d}
\end{figure*}

\noindent\textbf{Datasets.}
Following the approach in HART~\cite{chen2025hart}, we select 2,345 human scans from the THuman 2.1 dataset~\cite{yu2021function4d} as our fine-tuning dataset. The subjects are rendered from 96 distinct viewpoints along a 360-degree azimuthal trajectory. 
% To emphasize the foreground, we apply center cropping around the human masks, isolating the regions of interest.
We used two datasets for testing. One is the THuman 2.1 test split, which contains 100 subjects for in-domain evaluation. 
The second is the test set from the 2K2K dataset~\cite{han2023high}, which is used for cross-domain evaluation and offers greater age diversity.
All comparisons with baselines are conducted using a fixed setting of 8 input views.
We apply all four extracted subspaces derived in the object-centering settings. Each subspace is computed from a bundle of ten LoRA adapters, each having a rank of $r=16$.

\begin{table}[tbp]
    \centering
    
    % \vspace{-1em}
    \vspace{-0.1in}
    \resizebox{1.0\columnwidth}!{
    \begin{tabular}{lccccccc}
        \toprule
        {\multirow{3}{*}{\textbf{Method}}} &
        {\multirow{3}{*}{\shortstack{\textbf{\# Trainable}\\\textbf{Param.}}}} &
        \multicolumn{3}{c}{\textbf{THuman} (In-domain)} &
        \multicolumn{3}{c}{\textbf{2K2K} (Cross-domain)} \\
        \cmidrule(r){3-5} \cmidrule(r){6-8}
        & &
        {Acc $\downarrow$} & {Comp $\downarrow$} & {NC $\uparrow$} &
        {Acc $\downarrow$} & {Comp $\downarrow$} & {NC $\uparrow$} \\
               
        \midrule
        VGGT
        & -         & 2.816 & 1.911 & 91.51 & 3.103 & 2.122 & 92.81 \\
        % VGGT+
        Full
        & 853.6 M   & 3.053 & 1.932 & 91.17 & 3.655 & 2.213 & 92.25 \\
        \midrule
        % VGGT+
        LoRA (rank=16)
        & 16.3 M    & 3.195 & 2.089 & 91.63 & 2.717 & 1.829 & 92.73 \\
        % VGGT+
        LoRA (rank=32)
        & 32.7 M    & 2.849 & 1.922 & 91.85 & 2.633 & 1.773 & 93.00 \\
        % VGGT+
        LoRA (rank=64)
        & 65.3 M    & \underline{2.791} & \underline{1.902} & \underline{92.12} & 3.017 & 1.968 & 93.18 \\
        % VGGT+
        LoRA (rank=128)
        & 130.7 M   & 3.188 & 2.507 & \textbf{92.48} & 2.986 & 1.959 & \underline{93.86} \\
        % VGGT+
        LoRA (rank=256)
        & 261.4 M   & 3.521 & 4.978 & 91.81 & \underline{2.517} & \underline{1.769} & \textbf{93.99} \\
        \midrule
        % VGGT+
        PiSSA (rank=16)
        & 16.3 M    & 3.009 & 1.921 & 90.81 & 2.791 & 1.866 & 92.92 \\
        % VGGT+
        PiSSA (rank=32)
        & 32.7 M    & 3.228 & 2.028 & 90.59 & 2.991 & 1.972 & 92.59 \\
        % VGGT+
        PiSSA (rank=64)
        & 65.3 M    & 3.931 & 2.351 & 89.42 & 3.730 & 2.328 & 91.13 \\
        % VGGT+
        PiSSA (rank=128)
        & 130.7 M   & 4.052 & 2.416 & 90.10 & 3.745 & 2.250 & 91.52 \\
        % VGGT+
        PiSSA (rank=256)
        & 261.4 M   & 4.292 & 3.689 & 90.19 & 4.122 & 2.288 & 91.38 \\
        \midrule
        % Ours ($d$=4)
        % & 3.8 M     & 7.714 & 3.482 & 87.90 & 7.832 & 3.483 & 88.75 \\
        % Ours ($d$=8)
        % & 4.0 M     & 5.605 & 3.086 & 89.58 & 6.844 & 4.522 & 89.27 \\
        Ours ($d$=16)
        & 4.7 M     & 3.392 & 2.138 & 90.91 & 3.019 & 1.887 & 91.70 \\
        Ours ($d$=32)
        & 7.6 M     & 3.332 & 2.220 & 91.56 & 2.825 & 1.878 & 93.00 \\
        Ours ($d$=64)
        & 19.3 M    & \textbf{2.745} & \textbf{1.882} & 91.82 & \textbf{2.513} & \textbf{1.754} & 93.56 \\
        \bottomrule
    \end{tabular}
    }
    \caption{
        \textbf{Evaluation of Clothed Human Reconstruction: Point Map Estimation on THuman 2.1~\cite{yu2021function4d} and 2K2K~\cite{han2023high}.}
        Our method achieves the best performance across nearly all metrics.
    }
    \vspace{-2em}
    \label{tab:clothed_human_eval_table}
\end{table}

The point cloud reconstruction evaluation results are shown in Tab.~\ref{tab:clothed_human_eval_table}. 
We observe that all fine-tuning methods can, under certain configurations, degrade the performance of the original base model. 
For LoRA, a low rank ($r$) is insufficient to capture the new variations present in the target data. In contrast, when the rank is increased, LoRA tends to overfit the training data. 
PiSSA, which initializes LoRA using principal components derived from SVD, also shows a degradation as the rank increases.

Our proposed method, while achieving sub-optimal performance when the subspace dimension $d$ is small, shows a significant trend: as $d$ increases, the extracted subspace becomes increasingly robust. This robustness stabilizes the fine-tuning process and leads to superior generalization performance, ultimately achieving superior results across nearly all metrics with fewer parameters.

Fig.~\ref{fig:visual_3d} shows qualitative results. The VGGT base model produces noticeable artifacts in the presence of new variations, stemming from incorrect visual matching, particularly along object edges. %The model needs to account for the continuity of transitions between different viewpoints. 
After fine-tuning, the model's ability to handle these problems improves, but noise remains significant in detailed regions, such as the hands. On the 2K2K dataset, our approach shows the best generalization over all other methods. 
%Despite the dataset's variation in age and clothing, our method further reduces artifacts in the reconstruction.

\subsection{Transparent Object Reconstruction}

We also evaluate on ClearPose~\cite{chen2022clearpose}, a challenging real-world dataset designed for transparent object reconstruction. ClearPose comprises 51 real-world scenes, captured using Intel RealSense L515. They include 63  transparent objects (e.g., bottles and cups). We select 32 scenes for training and use the remaining scenes as the test set.
The results, with both \textit{Accuracy} and \textit{Completeness} reported in units of $10^{-2}$, are presented in Table~\ref{tab:clearpose}. We can see that with a similar number of trainable parameters, i.e., 16.3M, our approach outperforms all baselines. Note that this is achieved without introducing any sampling of transparent textures when learning subspaces. To achieve similar performance, LoRA and AdaLoRA~\cite{zhang2023adalora} require a much more number of parameters.

\begin{table}[h]
    \centering
    \vspace{-0.1in}
    \resizebox{1.0\columnwidth}!{
    \begin{tabular}{lcccc
    % ccc
    }
        \toprule
        {\multirow{3}{*}{\textbf{Method}}} &
        {\multirow{3}{*}{\shortstack{\textbf{\# Trainable}\\\textbf{Param.}}}} &
        \multicolumn{3}{c}{\textbf{In-domain}}
        % & \multicolumn{3}{c}{\textbf{Cross-domain}}
        \\
        \cmidrule(r){3-5} 
        % \cmidrule(r){6-8}
        & & 
        {Acc $\downarrow$} & {Comp $\downarrow$} & {NC $\uparrow$} 
        % & {Acc $\downarrow$} & {Comp $\downarrow$} & {NC $\uparrow$} 
        \\
        \midrule
        VGGT
        & -         & 3.123 & 3.271 & 67.77
                    % & 3.175 & 3.461 & 69.83 
                    \\
        % VGGT+
        Fully fine-tuned
        & 853.6 M   & 1.653 & 2.559 & 74.18
                    % & 1.773 & 2.688 & 73.97 
                    \\
        \midrule
        % VGGT+
        LoRA (rank=16)
        & 16.3 M    & 1.808 & 2.522 & 73.74 
                    % & 1.829 & 2.640 & 73.82 
                    \\
        % VGGT+
        LoRA (rank=32)
        & 32.7 M    & 1.811 & 2.562 & 73.76
                    % & 1.851 & 2.671 & 73.79 
                    \\
        % VGGT+
        LoRA (rank=64)
        & 65.3 M    & 1.787 & 2.580 & \underline{73.81} 
                    % & 1.826 & 2.688 & \textbf{73.97} 
                    \\
        % VGGT+
        LoRA (rank=128)
        & 130.7 M   & \textbf{1.753} & 2.571 & 73.80 
                    % & 1.798 & 2.673 & 73.96 
                    \\
        % VGGT+
%        LoRA (rank=256)
%        & 261.4 M   & \textbf{1.710} & 2.540 & %\textbf{73.86} 
                    % & \textbf{1.746} & \textbf{2.621} & 73.96 
 %                   \\
        \midrule
        AdaLoRA (rank=16)
        & 15.0 M    & 1.859 & 2.479 & 72.67 
                    % & 1.762 & 2.558 & 73.82 
                    \\
        AdaLoRA (rank=32)
        & 29.9 M    & 1.832 & 2.464 & 72.89
                    % & 1.748 & 2.550 & 73.81 
                    \\
        AdaLoRA (rank=64)
        & 59.7 M    & 1.826 & \textbf{2.453} & 73.07
                    % & \textbf{1.740} & \textbf{2.533} & 73.75 
                    \\
        AdaLoRA (rank=128)
        & 119.4 M   & 1.836 & 2.465 & 73.20
                    % & 1.751 & 2.536 & \textbf{73.68} 
 %                   \\
 %       AdaLoRA (rank=256)
 %       & 238.3 M   & 1.844 & 2.485 & \textbf{73.26} 
                    % & 1.760 & 2.547 & 73.74 
                    \\ \midrule
        % Ours ($d$=16)
        % & 4.5 M     & 2.091 & 2.341 & 72.83 
        %             & 1.816 & \textbf{2.396} & 72.77 \\
        % Ours ($d$=32)
        % & 6.9 M     & 1.861 & \textbf{2.227} & 74.42
        %             & 1.825 & 2.541 & 73.42 \\
        Ours % ($d$=64)
        & 16.3 M    & \underline{1.764} & \underline{2.462} & \textbf{73.86} 
                    % & 1.786 & 2.565 & 73.91 
                    \\
        \bottomrule
    \end{tabular}
    }
    \caption{
        \textbf{ Evaluation of Transparent Object Reconstruction: Point Map Estimation on ClearPose~\cite{chen2022clearpose}.} Our method achieves comparable performance across all metrics.  
    }
    \vspace{-0.2in}
    \label{tab:clearpose}
\end{table}

\subsection{Ablation Study}

%In this section, we aim to address two key questions. First, we conduct experiments to empirically validate the effectiveness of the extracted subspaces in improving the quality of fine-tuning. Second, we investigate the relationship between the quality of the subspace and the LoRA rank.

\noindent\textbf{Effectiveness of extracted subspaces.}
In the first experiment, We can replace $\overline{A}$ and $\overline{B}$ in the LoRA parametrization $dW=\overline{A}M\overline{B}$ with the principal singular vectors of the original model's weights. As shown in Tab.~\ref{tab:ablation_table1}, our approach shows a clear advantage over this alternative when using the same rank. 

\begin{table}[htbp]
    \centering
    \vspace{-1em}
    \resizebox{1.0\columnwidth}!{
    \begin{tabular}{lcccccc}
        \toprule
        {\multirow{3}{*}{\textbf{Method}}} &
        \multicolumn{3}{c}{\textbf{THuman} (In-domain)} &
        \multicolumn{3}{c}{\textbf{2K2K} (Cross-domain)} \\
        \cmidrule(r){2-4} \cmidrule(r){5-7}
        &
        {Acc $\downarrow$} & {Comp $\downarrow$} & {NC $\uparrow$} &
        {Acc $\downarrow$} & {Comp $\downarrow$} & {NC $\uparrow$} \\
               
        \midrule
        % VGGT+
%        PSV ($d$=16)
%        & 8.201 & 5.915 & 87.18 & 9.648 & 5.069 & 88.02 \\
        % VGGT+
        PSV ($d$=32)
        & 5.605 & 3.086 & 89.58 & 6.844 & 4.523 & 89.27 \\
        % VGGT+
        PSV ($d$=64)
        & 4.066 & 2.423 & 90.65 & 3.805 & 2.202 & 90.84 \\
        % VGGT+
        PSV ($d$=128)
        & 3.709 & 2.394 & 91.10 & 3.290 & 2.128 & 92.50 \\
        % VGGT+
        PSV ($d$=256)
        & 2.785 & 1.904 & 91.76 & 2.542 & 1.762 & 93.42 \\
        \midrule
 %       Ours ($d$=4)
 %       & 7.714 & 3.482 & 87.90 & 7.832 & 3.483 & 88.75 \\
        Ours ($d$=8)
        & 5.839 & 3.363 & 89.33 & 5.712 & 2.602 & 89.98 \\
        Ours ($d$=16)
        & 3.392 & 2.138 & 90.91 & 3.019 & 1.887 & 91.70 \\
        Ours ($d$=32)
        & 3.332 & 2.220 & 91.56 & 2.825 & 1.878 & 93.00 \\
        Ours ($d$=64)
        & 2.745 & 1.882 & 91.82 & 2.513 & 1.754 & 93.56 \\
        \bottomrule
    \end{tabular}
    }
    % \vspace{-0.1in}
    \caption{
Computing $\overline{A}$ and $\overline{B}$ using our approach is superior to using the leading singular vectors of the original weight matrix. 
    }
    % \vspace{-0.1in}
    \label{tab:ablation_table1}
\end{table}

\noindent\textbf{Rank of LoRA Pairs.} In this experiment, we fix the dimension $d$ of the shared sub-space while changing the rank $r$ of the input LoRA pairs. As shown in Tab.~\ref{tab:ablation_table2}, the performance trends of the two subspaces are generally similar when varying $d$ with fixed $r$ are similar. The difference may stem from the number of LoRA pairs and the quality of the dataset used to extract the subspaces. This also highlights the importance of increasing the variance and diversity of synthetic datasets.

%As mentioned in Sec.~\ref{Sec:Shared:LoRA:Subspaces}, our algorithm extracts the shared subspace among pairs of LoRA adapters. Although the training dynamics differs for LoRA adapters with different ranks, the singular values of LoRA are sparse. Therefore, we expect that the shared space between LoRA adapters with different ranks should be similar.

\begin{table}[!t]
    \centering
    
    % \vspace{-1em}
    \vspace{-0.1in}
    \resizebox{1.0\columnwidth}!{
    \begin{tabular}{lcccccc}
        \toprule
        {\multirow{3}{*}{\textbf{Method}}} &
        \multicolumn{3}{c}{\textbf{THuman} (In-domain)} &
        \multicolumn{3}{c}{\textbf{2K2K} (Cross-domain)} \\
        \cmidrule(r){2-4} \cmidrule(r){5-7}
        &
        {Acc $\downarrow$} & {Comp $\downarrow$} & {NC $\uparrow$} &
        {Acc $\downarrow$} & {Comp $\downarrow$} & {NC $\uparrow$} \\
               
        \midrule
        % VGGT+
%        r=64 ($d$=4)
%        & 8.499 & 3.726 & 87.10 & 8.423 & 3.646 & 88.20 \\
        % VGGT+
        r=64 ($d$=8)
        & 6.621 & 3.092 & 88.76 & 6.070 & 3.299 & 88.94 \\
        % VGGT+
        r=64 ($d$=16)
        & 4.030 & 2.343 & 89.94 & 4.813 & 3.031 & 89.93 \\
        % VGGT+
        r=64 ($d$=32)
        & 3.797 & 2.482 & 91.31 & 3.246 & 2.094 & 92.43 \\
        % VGGT+
        r=64 ($d$=64)
        & 2.783 & 1.871 & 91.41 & 2.563 & 1.756 & 93.19 \\
        \midrule
%        r=16 ($d$=4)
%        & 7.714 & 3.482 & 87.90 & 7.832 & 3.483 & 88.75 \\
        r=16 ($d$=8)
        & 5.839 & 3.363 & 89.33 & 5.712 & 2.602 & 89.98 \\
        r=16 ($d$=16)
        & 3.392 & 2.138 & 90.91 & 3.019 & 1.887 & 91.70 \\
        r=16 ($d$=32)
        & 3.332 & 2.220 & 91.56 & 2.825 & 1.878 & 93.00 \\
        r=16 ($d$=64)
        & 2.745 & 1.882 & 91.82 & 2.513 & 1.754 & 93.56 \\
        \bottomrule
    \end{tabular}
    }
    % \vspace{-0.1in}
    \caption{
        \textbf{Comparison of Subspaces Extracted from LoRAs with Different Ranks.}
    Ablation study to validate our shared space hypothesis.
    }
    \vspace{-0.2in}
    \label{tab:ablation_table2}
\end{table}

%% file: sec/07_con.tex
\vspace{-0.02in}
\section{Conclusions and Future Work}
\vspace{-0.02in}

In this paper, we introduce the problem of extracting subspaces of a transformer-based 3D foundation model for LoRA-based fine-tuning. We show that such subspaces that correspond to variations in geometry, texture, camera motion, and lighting do exist, and they are approximately disentangled. We present an algorithm that computes them from synthetic datasets generated in a controlled manner. A striking message is that these subspaces lead to efficient fine-tuning procedures for downstream tasks, achieving better predictive accuracy than state-of-the-art approaches. 

We hope that our work inspires exploration into the understanding of 3D foundation models. There are ample opportunities for future research. First of all, we study only static scenes. An obvious extension is to add motion variations to understand recent 4D foundation models. Another direction is to study common and differences across different 3D foundation models. Finally, this paper focuses on the use of synthetic data for fine-tuning, and it is interesting to study how to combine large-scale synthetic and small-scale real datasets to enhance fine-tuning performance. 

\noindent\textbf{Acknowledgments.}  This project was supported by NSF-2047677, 2413161, 2504906, 2515626, GIFTs from Adobe and Google, and computing support on the Vista GPU Cluster through the Center for Generative AI (CGAI) and TACC  at UT Austin. 

%% file: sec/supp_results.tex
\section{Implementation Details}

During subspace extraction and fine-tuning for downstream applications, we fine-tune the model from the pretrained VGGT-1B checkpoint and freeze the DINO encoder to save memory. 
For the aggregator, depth head, and camera head, we use a cosine learning rate schedule with warm-up. During the warm-up phase, the learning rate linearly increases from $1\times10^{-8}$ to $1\times10^{-5}$ over the first 5\% of the total training steps. Following the warm-up, the learning rate then decays following a cosine schedule, dropping to $1\times10^{-8}$ over the remaining training steps.
To stabilize training, we apply gradient norm clipping at $0.5$.

During the fine-tuning phase, we randomly sample 4-16 views per scene. The network is trained for a total of 24,000 steps, with each step processing 32 images as input. The entire training process requires approximately 20 hours using a single NVIDIA H200 GPU.

The dataset used for subspace extraction consists of 200 generated scenes, each rendered as a sequence of 100 images.

\section{Additional Results}

\subsection{Qualitative Results}

\begin{figure*}[htbp]
\centering
\includegraphics[width=0.95\linewidth]{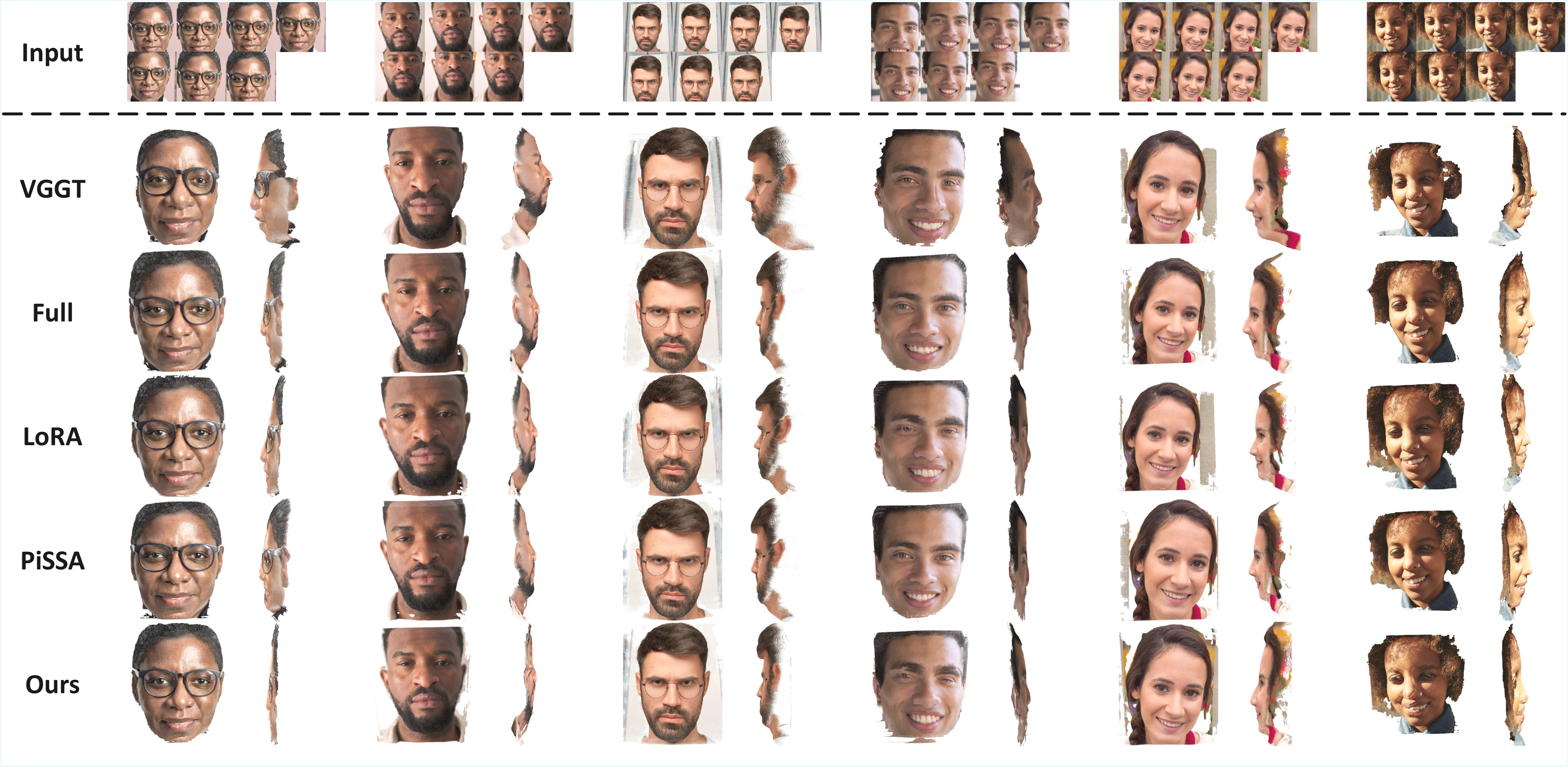}
\vspace{-1em}
\captionof{figure}{More visual comparison of 2D Face Anti-Spoofing Tasks.}
% \vspace{-2em}
\label{Fig:supp_2d}
\end{figure*}

We present additional qualitative results of point cloud reconstruction on the synthetic test dataset in Figure~\ref{Fig:supp_2d}, which further demonstrate the robustness of our method across different scenes. Our method produces the most accurate reconstructions with noticeably fewer artifacts, showing its transferability to out-of-distribution data.

\begin{figure*}[hp]
\centering
\includegraphics[width=0.95\linewidth]{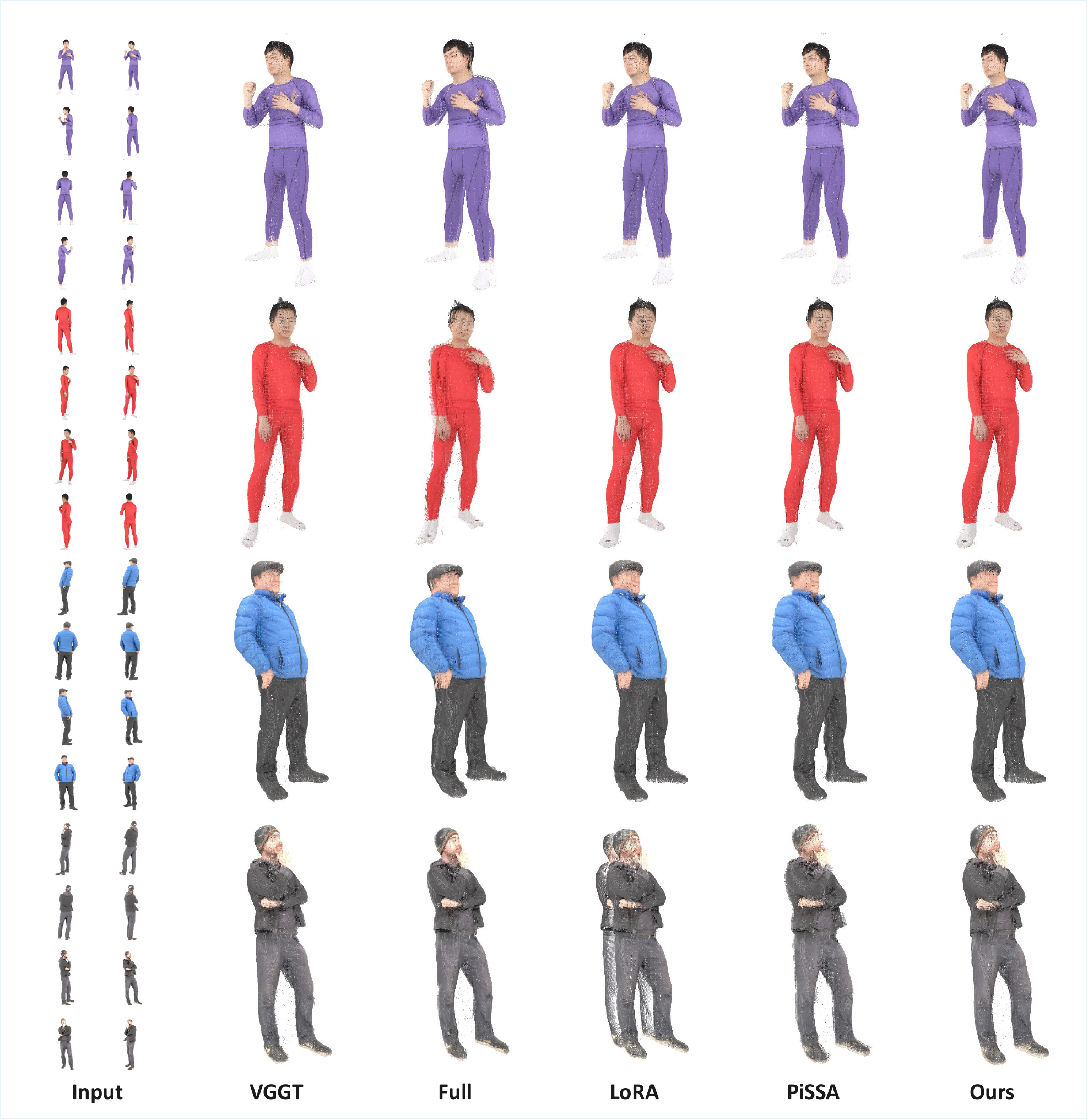}
\caption{More visual comparison of Clothed Human Reconstruction Tasks.}
\label{Fig:supp_3d}
\end{figure*}

We present more visualizations of clothed human reconstruction in Figure~\ref{Fig:supp_3d}. The first two rows are from the THuman dataset~\cite{yu2021function4d}, and the last two rows are from the 2K2K dataset~\cite{han2023high}. The comparison shows that our method exhibits strong robustness.

\subsection{Different Rank Allocation Strategies}

\begin{table}[htbp]
    \centering
    \caption{
        \textbf{Comparison Between Different Rank Allocation Strategies.}
        The overall trend is the same, with minimal performance differences.
    }
    % \vspace{-1em}
    \vspace{-0.1in}
    \resizebox{1.0\columnwidth}!{
    \begin{tabular}{lcccccc}
        \toprule
        {\multirow{3}{*}{\textbf{Method}}} &
        \multicolumn{3}{c}{\textbf{THuman} (In-domain)} &
        \multicolumn{3}{c}{\textbf{2K2K} (Cross-domain)} \\
        \cmidrule(r){2-4} \cmidrule(r){5-7}
        &
        {Acc $\downarrow$} & {Comp $\downarrow$} & {NC $\uparrow$} &
        {Acc $\downarrow$} & {Comp $\downarrow$} & {NC $\uparrow$} \\
        \midrule
        Uniform ($d$=16)
        & 3.392 & 2.138 & 90.91 & 3.019 & 1.887 & 91.70 \\
        Uniform ($d$=32)
        & 3.332 & 2.220 & 91.56 & 2.825 & 1.878 & 93.00 \\
        Uniform ($d$=64)
        & \textbf{2.745} & \textbf{1.882} & \textbf{91.82} & \textbf{2.513} & \textbf{1.754} & \textbf{93.56} \\
        \midrule
        Importance ($d$=16)
        & 4.088 & 3.783 & 88.81 & 3.822 & 5.900 & 90.29 \\
        Importance ($d$=32)
        & 3.778 & 2.444 & 91.23 & 3.165 & 2.041 & 92.41 \\
        Importance ($d$=64)
        & \textbf{2.766} & \textbf{1.912} & \textbf{92.03} & \textbf{2.481} & \textbf{1.762} & \textbf{93.54} \\
        
        \bottomrule
    \end{tabular}
    }
    \vspace{-1em}
    \label{tab:supp_different_rank_allocation}
\end{table}

In the Experiment section, we report the results of the method that applied the same $d$ to different layers. Another popular importance rank allocation strategy is based on the effective rank. The effective rank of a matrix $W$ is defined using its Frobenius and spectral norms as:
\begin{align*}
    \operatorname{Effective Rank}(W)=
    \left(\frac{\|W\|_\mathcal{F}}{\|W\|_2}\right)^2.
\end{align*}
Therefore, the target subspace dimension $d$ for each layer can be dynamically allocated based on this measure. Specifically, the layer-wise subspace size $d_l$ is determined by:
\begin{align*}
    d_l=d\times\left\lfloor
    \frac{\operatorname{Effective Rank}(W_l)}{\operatorname{Average Effective Rank}}
    \right\rfloor,
\end{align*}
where $d$ is the global budget for the subspace size.

We present the performance comparison between Uniform and Importance allocation strategies in Table~\ref{tab:supp_different_rank_allocation}. Uniform refers to the allocation strategy reported in the main text, while Importance is based on effective rank-based importance allocation. We observe that the overall trend is the same, and the performance differences are negligible.

\section{Details on Synthetic Dataset Generation}

In this section, we first briefly introduce the previous work, Megasynth~\cite{jiang2025megasynth}, and then describe the generation process of our synthetic datasets.

Megasynth is a pipeline designed for generating synthetic non-semantic datasets. Using scalability and controllability, we can synthesize datasets tailored to exhibit target 3D attribute variations. The generation process begins with creating the layout of indoor scenes by filling the space with boxes of varying sizes. Next, it generates the scene geometry and samples the textures. The geometry is constructed from primitives (such as ellipsoids, cubes, and cylinders) instantiated within each box. To maximize geometric variation, a height field is randomly assigned to each surface. Textures are sampled from the MatSynth texture dataset~\cite{vecchio2024matsynth}. During the rendering phase, the light sources are randomized, followed by a random sampling of both the camera distribution and the intrinsic camera parameters.

For the first experiment, 2D Face Anti-Spoofing, we utilized two distinct subspaces: texture and geometry. Each subspace was extracted from five different LoRA adapters. These ten datasets (for subspace extraction) and the datasets employed for fine-tuning were rendered under micro-baseline settings.

The micro-baseline setting emphasizes that camera movements were minimal. This was achieved by interpolating the camera's translation and rotation across control points. By ensuring that both the translational displacement and the angular differences between these control points remained within a predefined range, the overall movement of the camera trajectory was kept minimal.

In the second experiment, Clothed Human Reconstruction, we used four subspaces: texture, geometry, camera motion, and lighting. Each of these four subspaces was extracted from ten different LoRA adapters. These datasets were object-centered. To achieve this, we made slight modifications to the standard layout sampling strategy: we removed the walls, ceiling, and floor of the room, leaving only the synthetic boxes centrally arranged.

Next, we explain how we customized the dataset generation with respect to these four specific variations. For texture variation, we minimized geometry changes: the scene file was fixed entirely (in the first experiment), or the number of boxes and primitives was limited to introduce only slight geometry variance (in the second experiment), while textures were sampled broadly from the entire texture dataset. For geometry variation, we allowed texture sampling to repeatedly use a subset of the full texture dataset, with different subsets used across different datasets, while varying the box and primitive counts to maximize geometry diversity. To isolate camera movement, we uniformly sampled camera azimuths and elevations on a sphere and randomized distances to define control points. Then, spline interpolation was performed between these points to generate the camera’s movement trajectory; different datasets used varying ranges for these orientation targets and distances. Finally, for the lighting variation, we place some sunlight sources in the Blender environment and randomly assign their color and strength for each instance. All images were rendered at a resolution of $518\times518$ to align with DINO2's patchify process.

\section{Singular Values of Matrix $C$}

In this section, we will present the distribution of the singular values of the matrices $C$ during the first iteration of the subspace extraction process. Note that a logarithmic scale has been applied. The significant drop in these curves indicates the existence of shared subspaces.

\clearpage

\clearpage
\subsection{Texture Subspace}

\begin{figure}[htbp]
\centering
\includegraphics[width=0.92\linewidth]{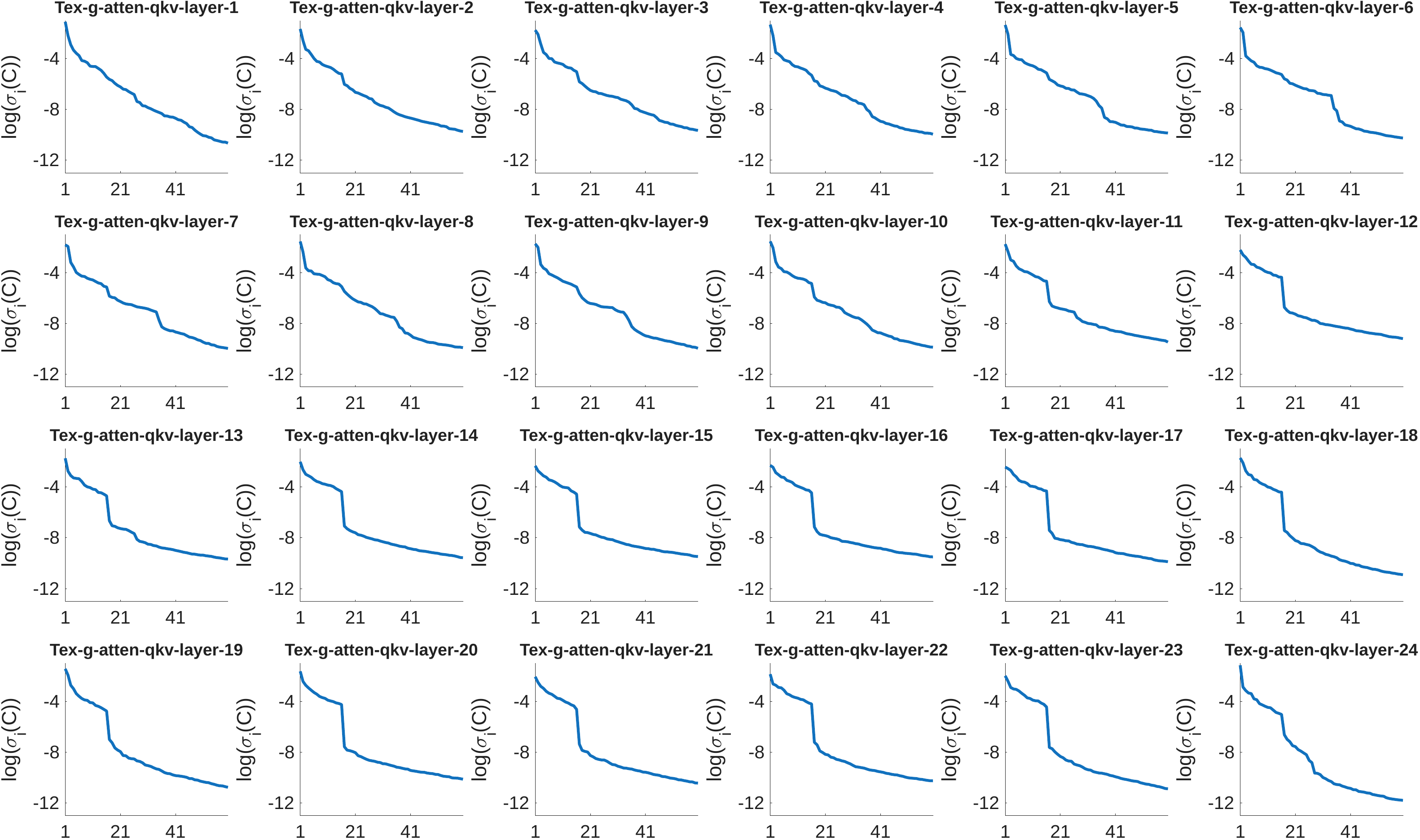}
\vspace{-1em} \caption{Singular values of the QKV matrix in the global attention layer with respect to texture variations.}
\label{Fig:Tex:g:atten:qkv}
\vspace{-2em} \end{figure}

\begin{figure}[htbp]
\centering
\includegraphics[width=0.92\linewidth]{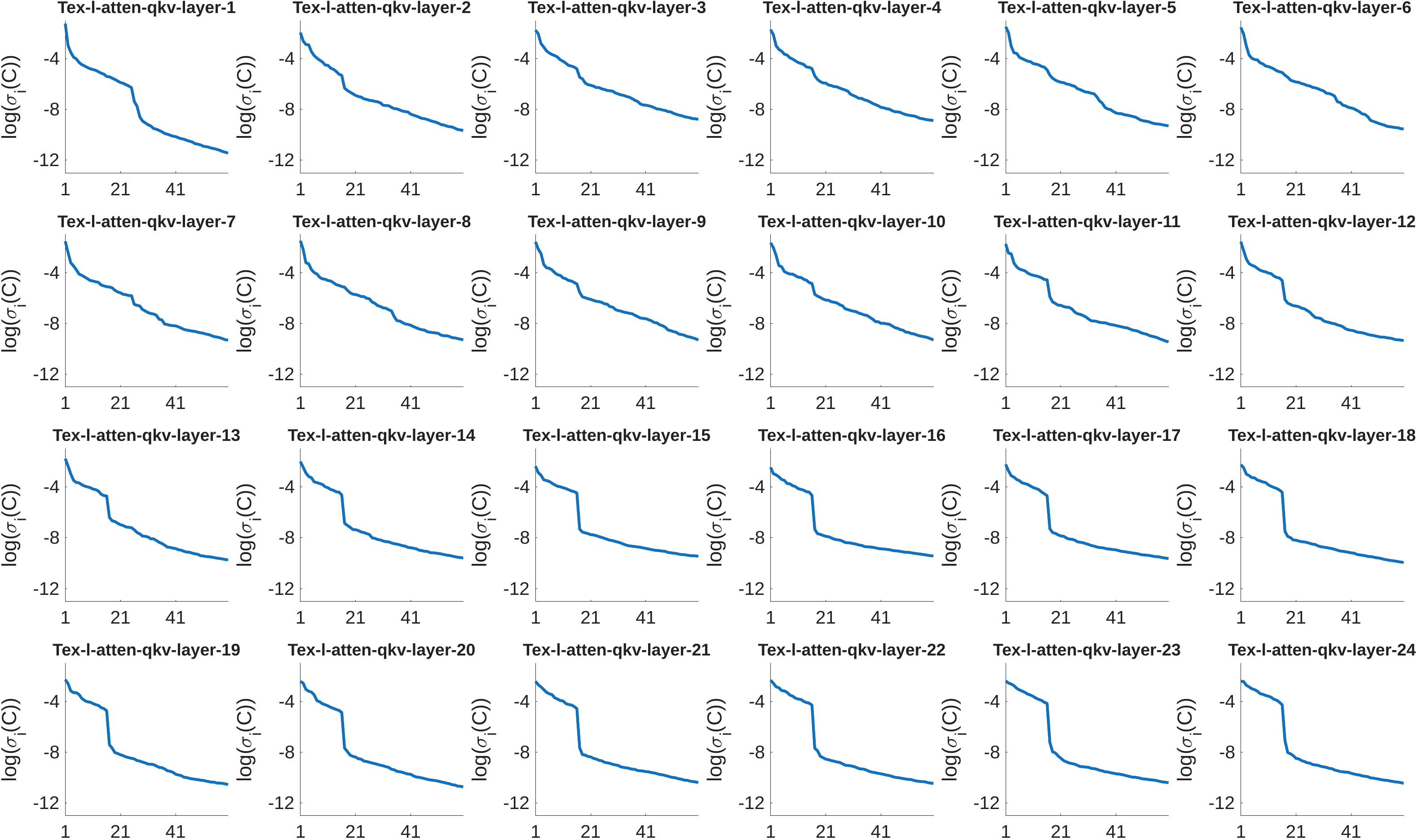}
\vspace{-1em} \caption{Singular values of the QKV matrix in the frame attention layer with respect to texture variations.}
\label{Fig:Tex:l:atten:qkv}
\vspace{-2em} \end{figure}

\begin{figure}[htbp]
\centering
\includegraphics[width=0.92\linewidth]{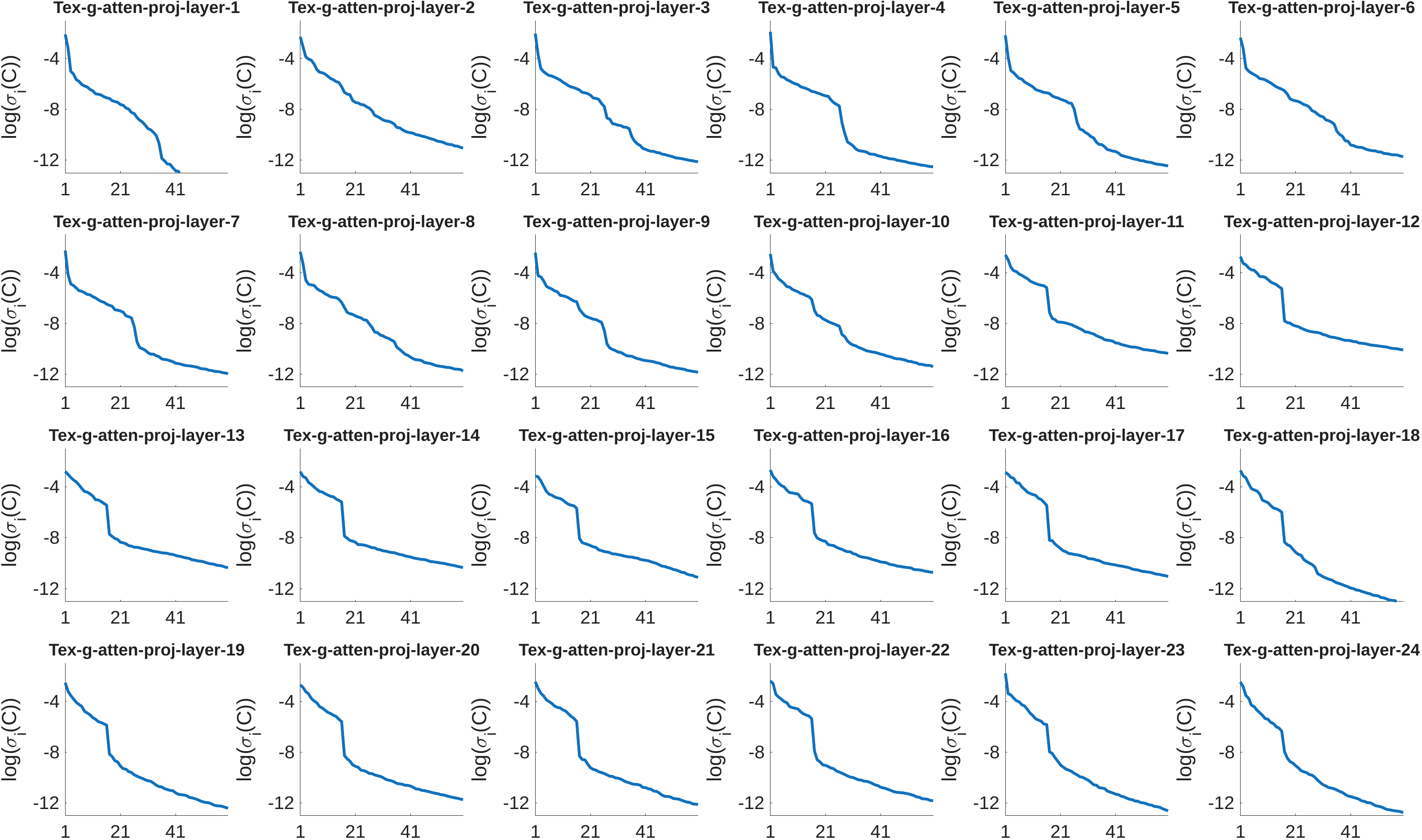}
\vspace{-1em} \caption{Singular values of the projection matrix in the global attention layer with respect to texture variations.}
\label{Fig:Tex:g:atten:proj}
\vspace{-2em} \end{figure}

\begin{figure}[htbp]
\centering
\includegraphics[width=0.92\linewidth]{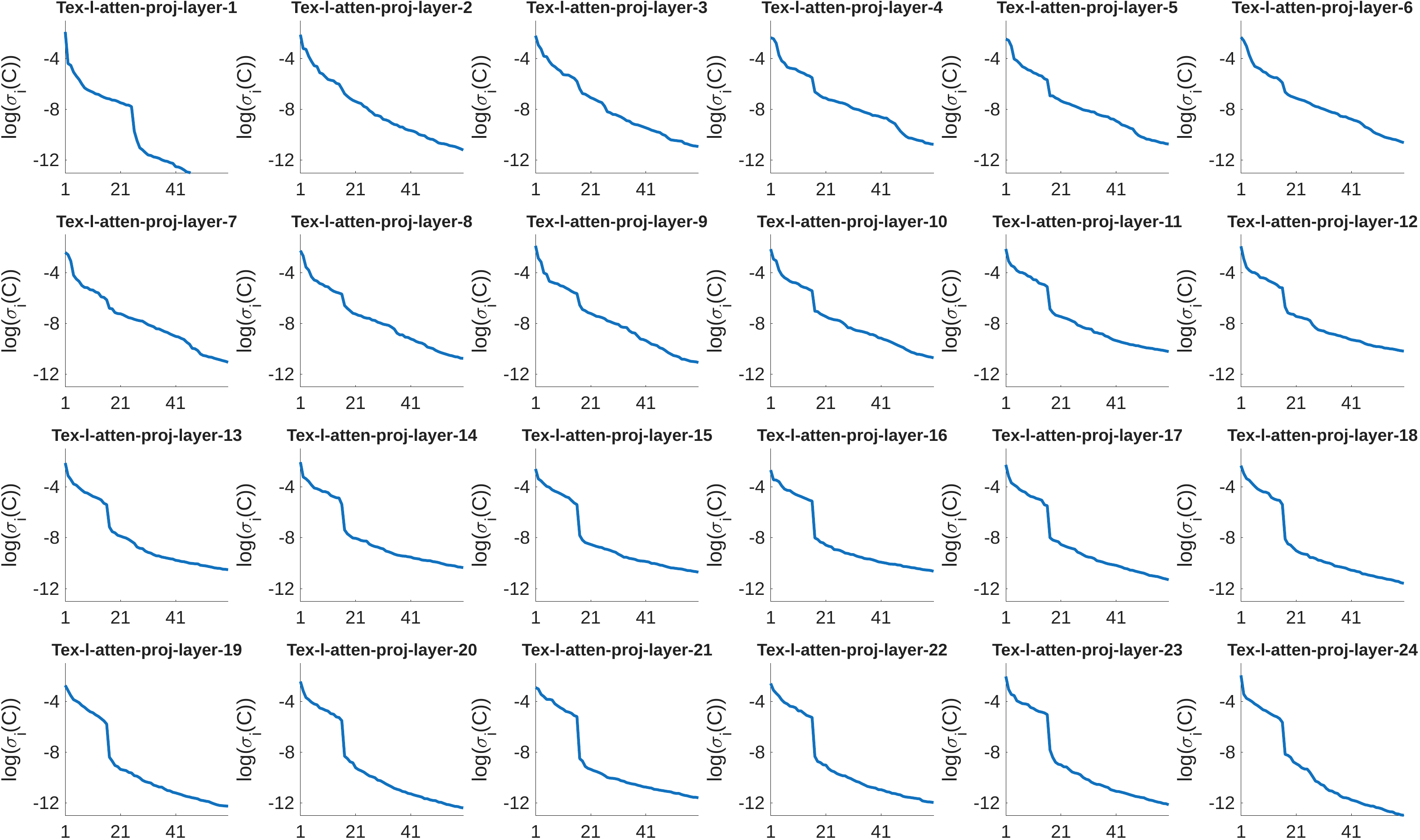}
\vspace{-1em} \caption{Singular values of the projection matrix in the frame attention layer with respect to texture variations.}
\label{Fig:Tex:l:atten:proj}
\vspace{-2em} \end{figure}

\begin{figure}[htbp]
\centering
\vspace{2em}
\includegraphics[width=0.92\linewidth]{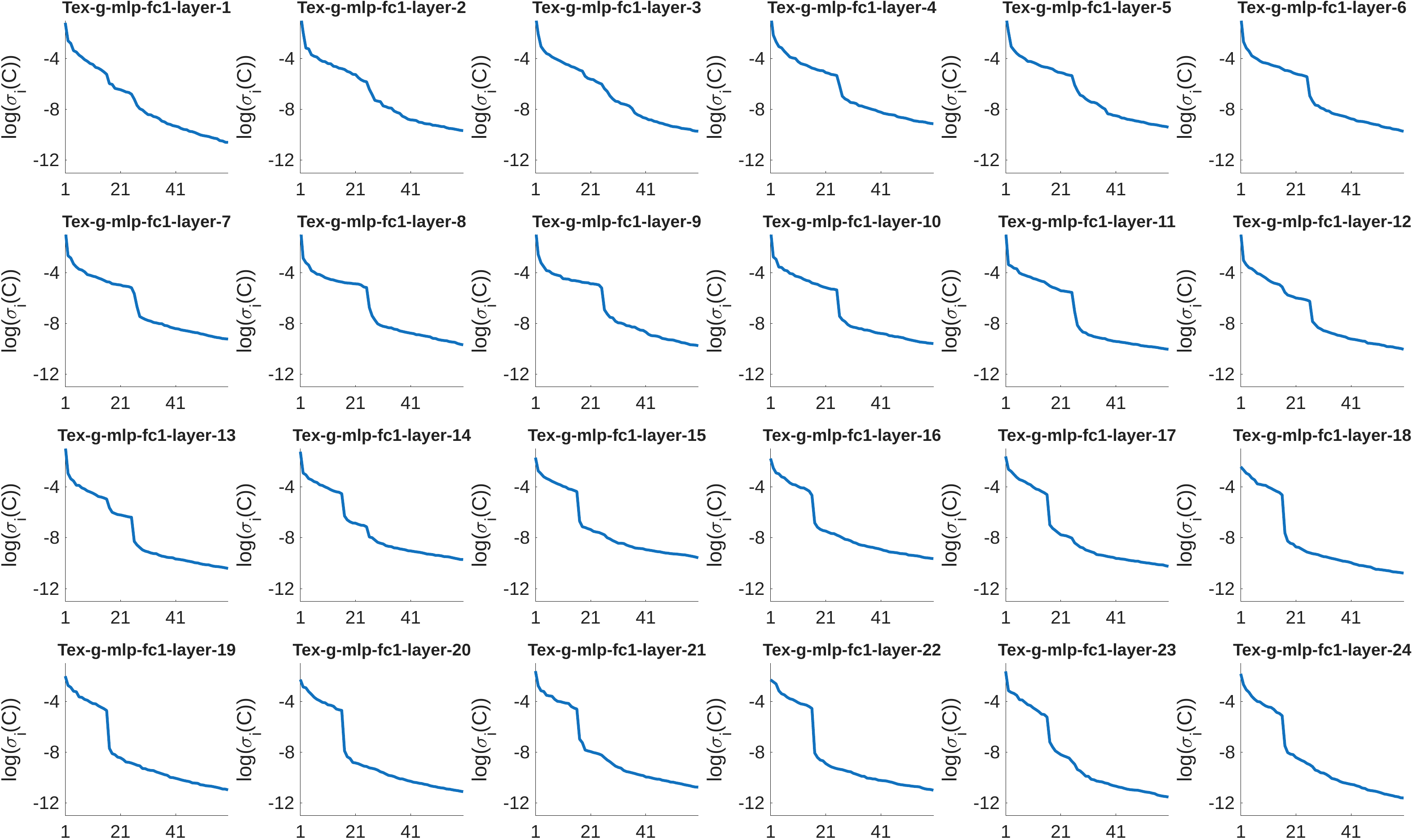}
\vspace{-1em} \caption{Singular values of the first fully connected matrix in the global attention layer with respect to texture variations.}
\label{Fig:Tex:g:mlp:fc1}
\vspace{-2em} \end{figure}

\begin{figure}[htbp]
\centering
\includegraphics[width=0.92\linewidth]{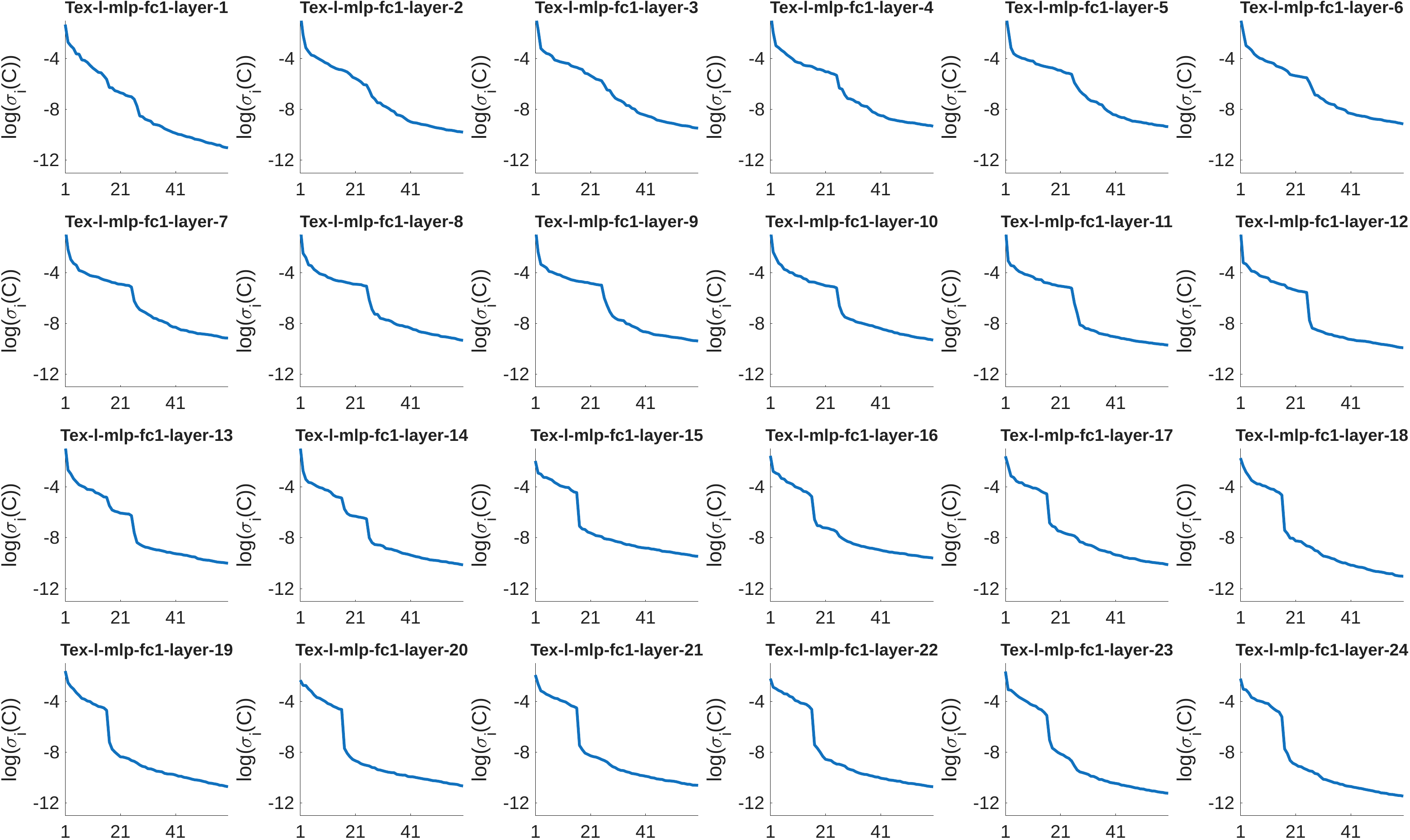}
\vspace{-1em} \caption{Singular values of the first fully connected matrix in the frame attention layer with respect to texture variations.}
\label{Fig:Tex:l:mlp:fc1}
\vspace{-2em} \end{figure}

\begin{figure}[htbp]
\centering
\includegraphics[width=0.92\linewidth]{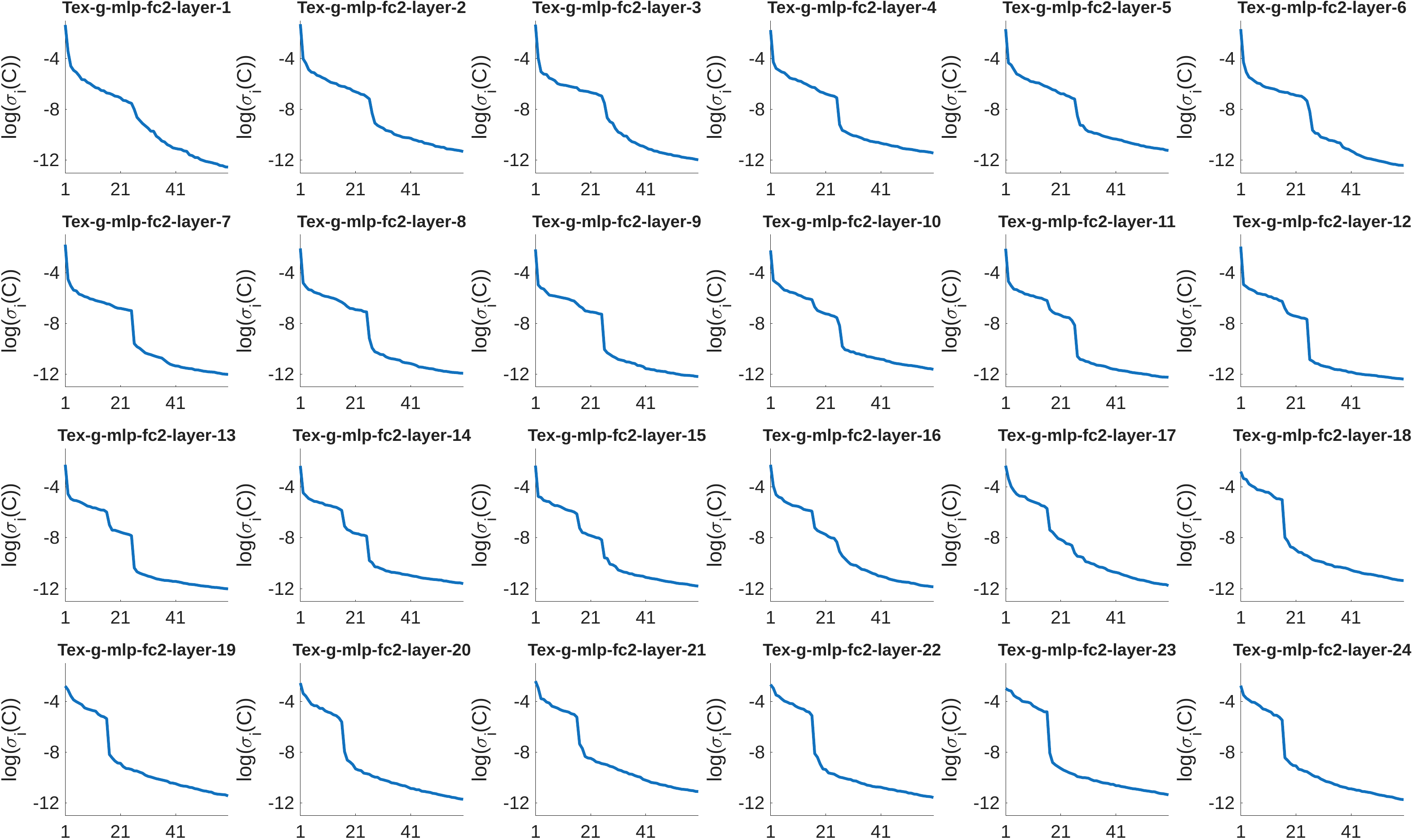}
\vspace{-1em} \caption{Singular values of the second fully connected matrix in the global attention layer with respect to texture variations.}
\label{Fig:Tex:g:mlp:fc2}
\vspace{-2em} \end{figure}

\begin{figure}[htbp]
\centering
\includegraphics[width=0.92\linewidth]{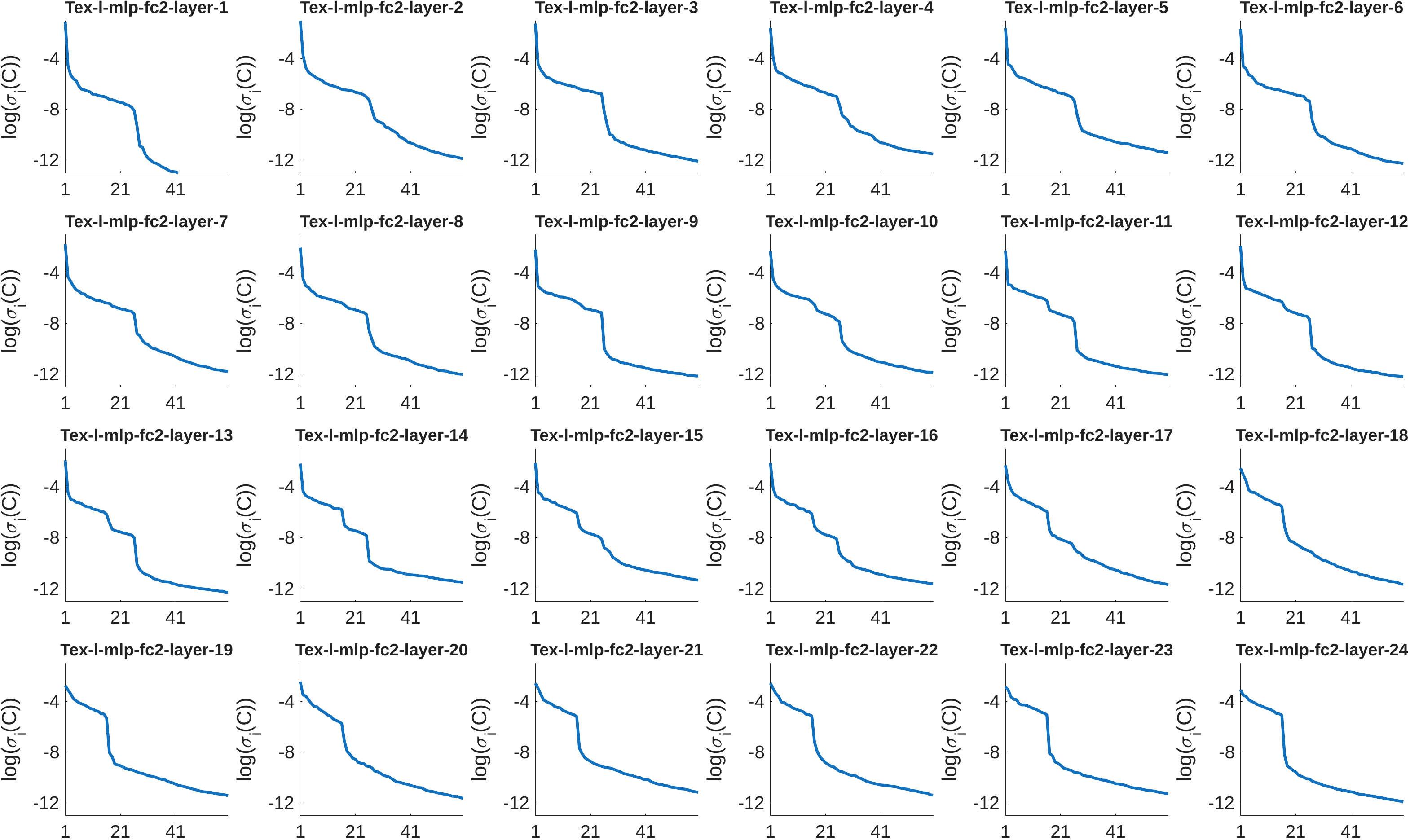}
\vspace{-1em} \caption{Singular values of the second fully connected matrix in the frame attention layer with respect to texture variations.}
\label{Fig:Tex:l:mlp:fc2}
\vspace{-2em} \end{figure}

\clearpage
\subsection{Geometry Subspace}

\begin{figure}[htbp]
\centering
\includegraphics[width=0.92\linewidth]{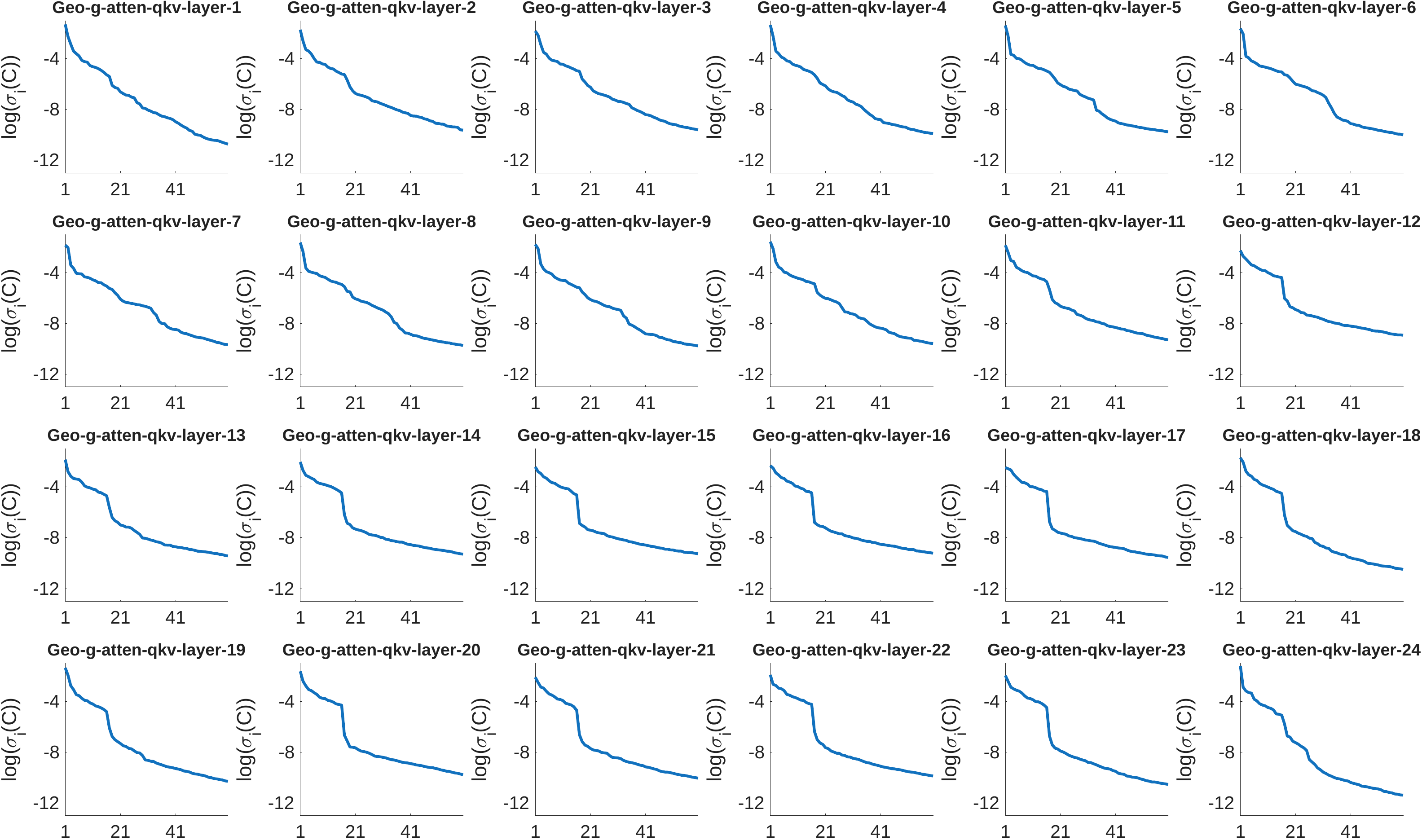}
\vspace{-1em} \caption{Singular values of the QKV matrix in the global attention layer with respect to geometry variations.}
\label{Fig:Geo:g:atten:qkv}
\vspace{-2em} \end{figure}

\begin{figure}[htbp]
\centering
\includegraphics[width=0.92\linewidth]{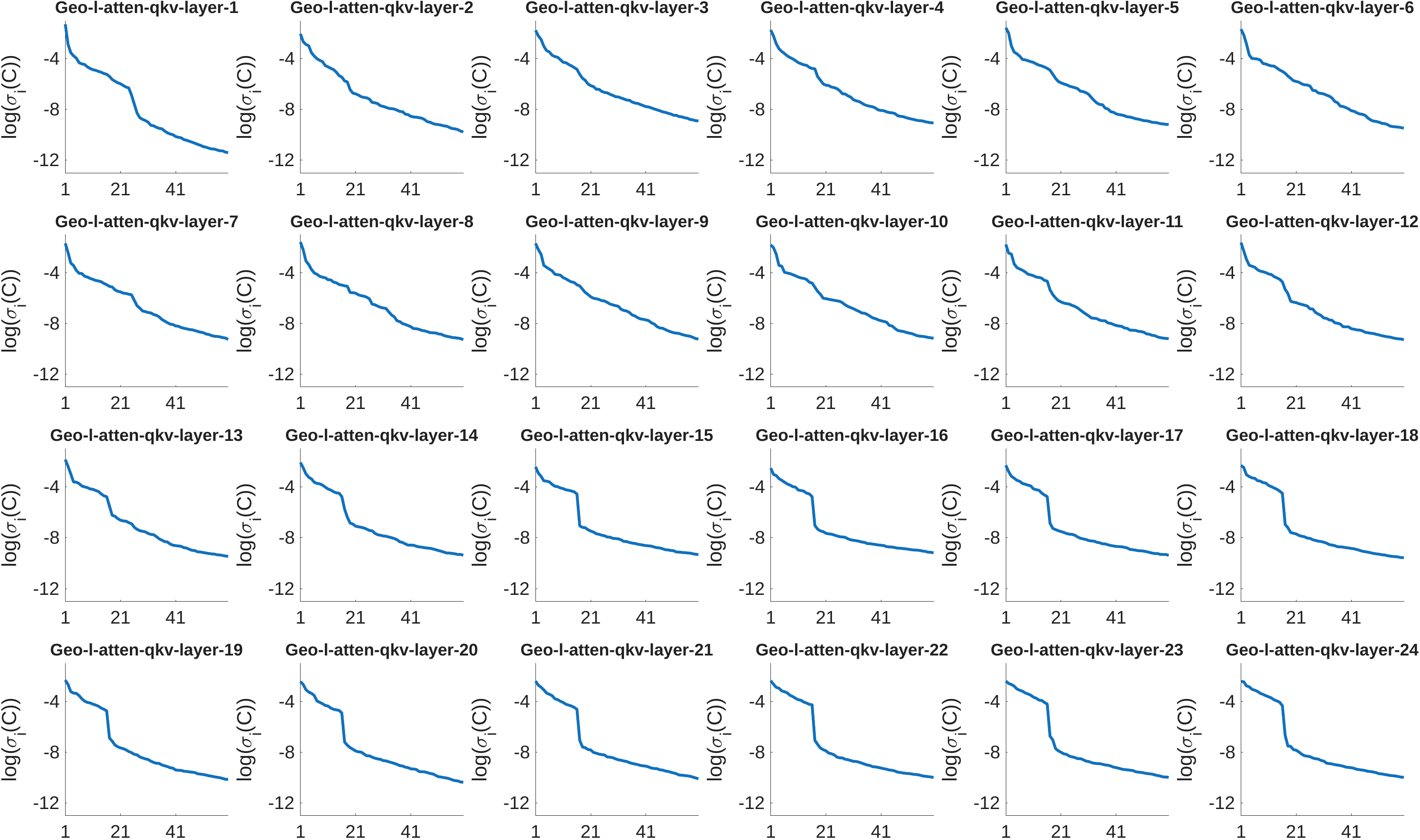}
\vspace{-1em} \caption{Singular values of the QKV matrix in the frame attention layer with respect to geometry variations.}
\label{Fig:Geo:l:atten:qkv}
\vspace{-2em} \end{figure}

\begin{figure}[htbp]
\centering
\includegraphics[width=0.92\linewidth]{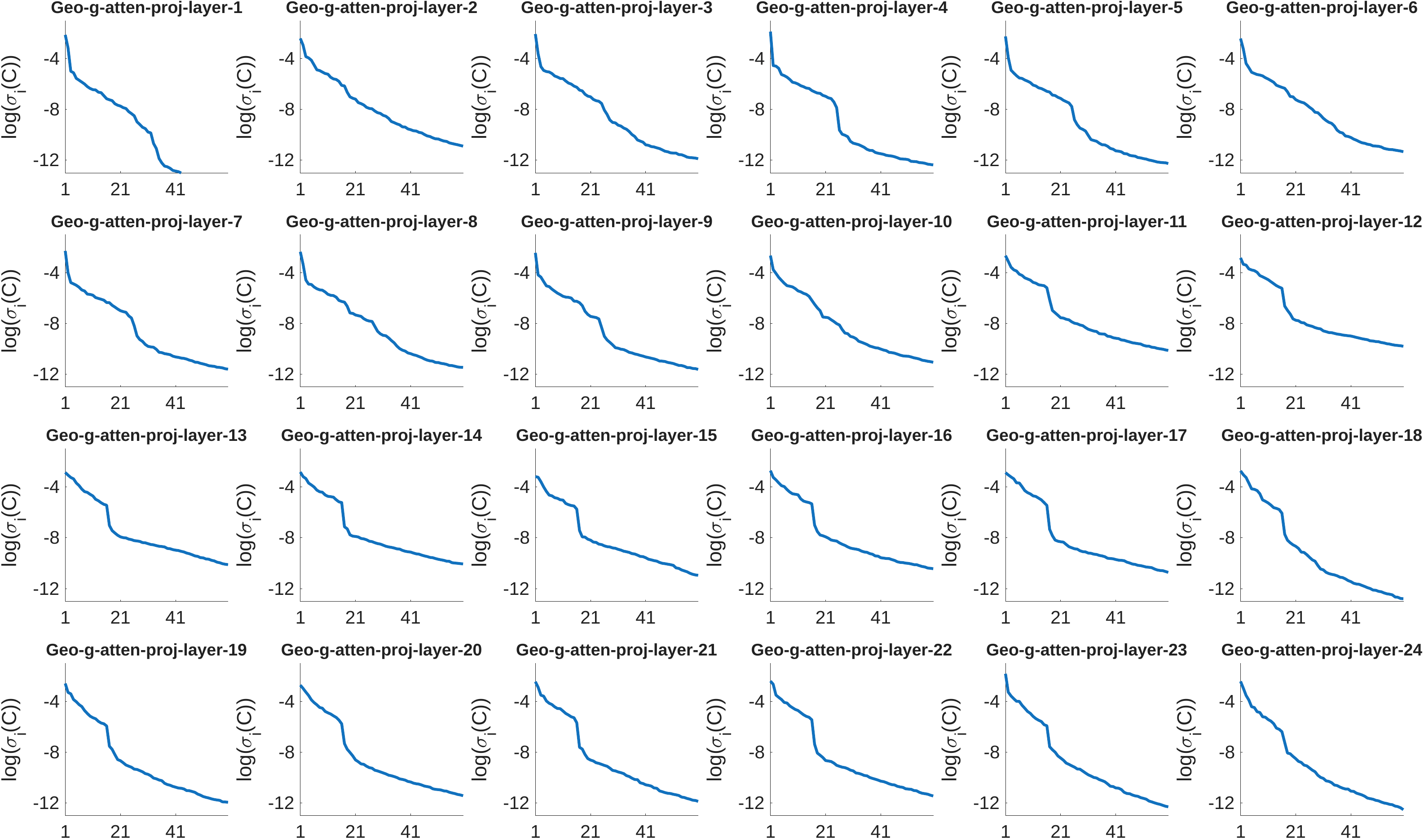}
\vspace{-1em} \caption{Singular values of the projection matrix in the global attention layer with respect to geometry variations. }
\label{Fig:Geo:g:atten:proj}
\vspace{-2em} \end{figure}

\begin{figure}[htbp]
\centering
\includegraphics[width=0.92\linewidth]{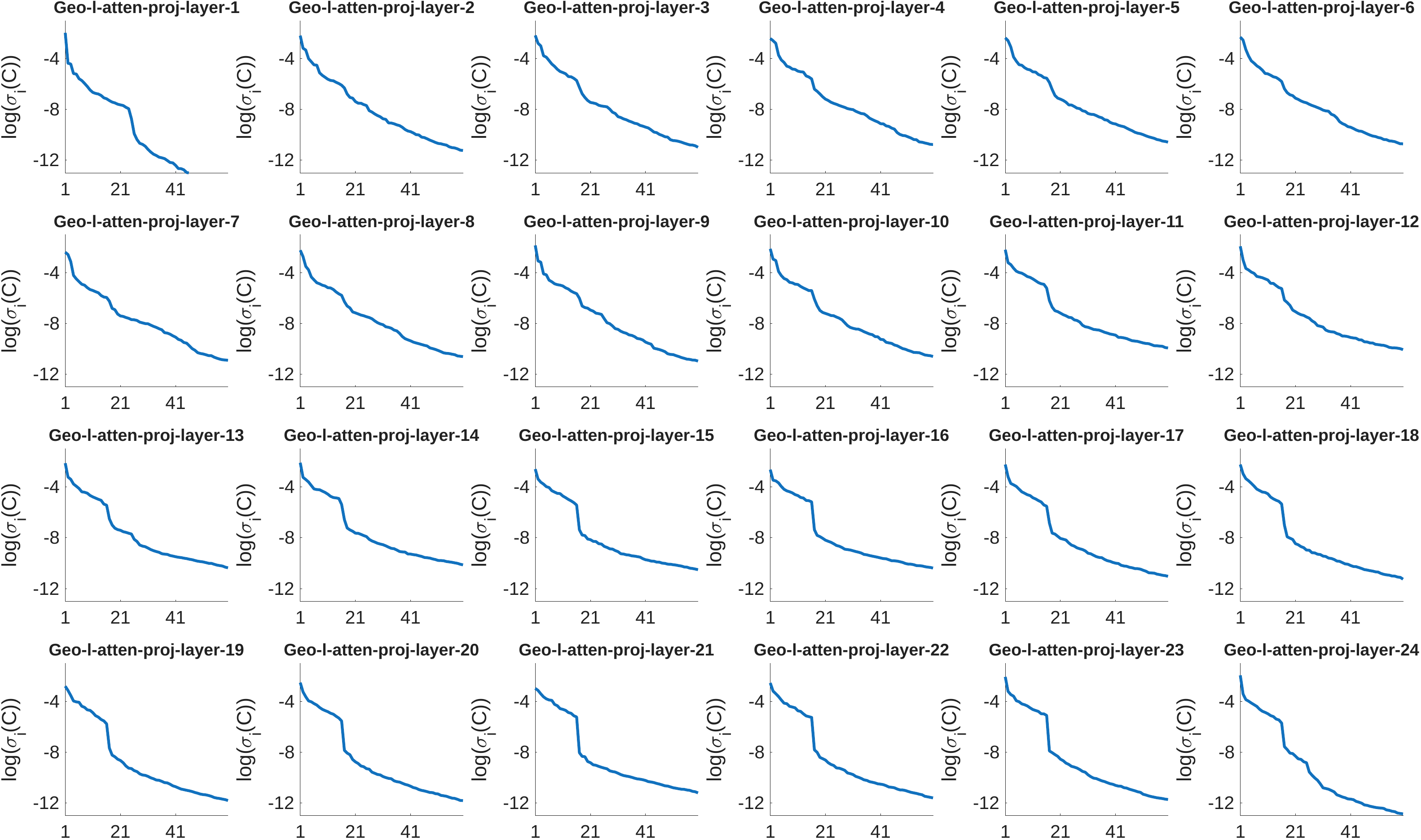}
\vspace{-1em} \caption{Singular values of the projection matrix in the frame attention layer with respect to geometry variations.}
\label{Fig:Geo:l:atten:proj}
\vspace{-2em} \end{figure}

\begin{figure}[htbp]
\centering
\vspace{2em}
\includegraphics[width=0.92\linewidth]{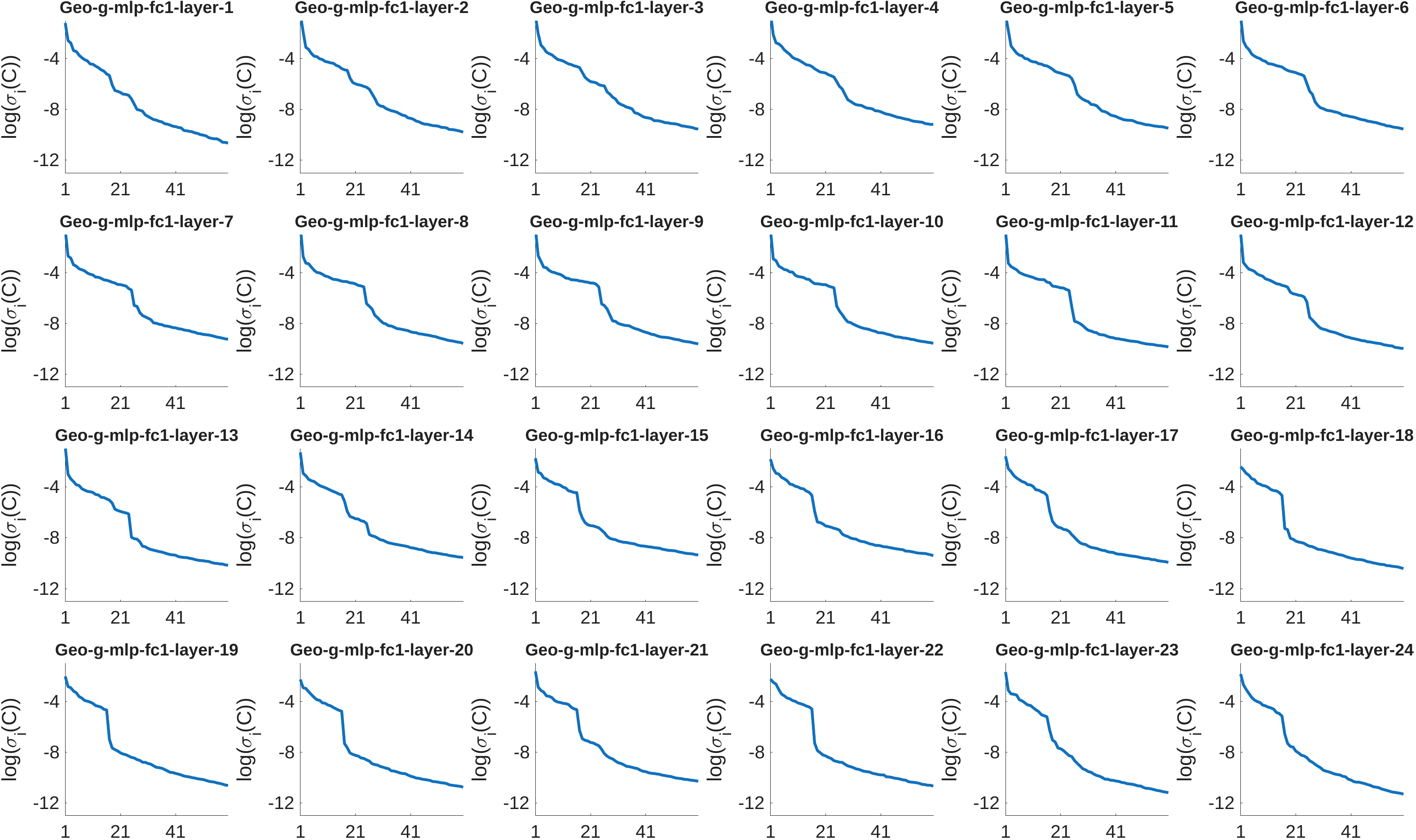}
\vspace{-1em} \caption{Singular values of the first fully connected matrix in the global attention layer with respect to geometry variations.}
\label{Fig:Geo:g:mlp:fc1}
\vspace{-2em} \end{figure}

\begin{figure}[htbp]
\centering
\includegraphics[width=0.92\linewidth]{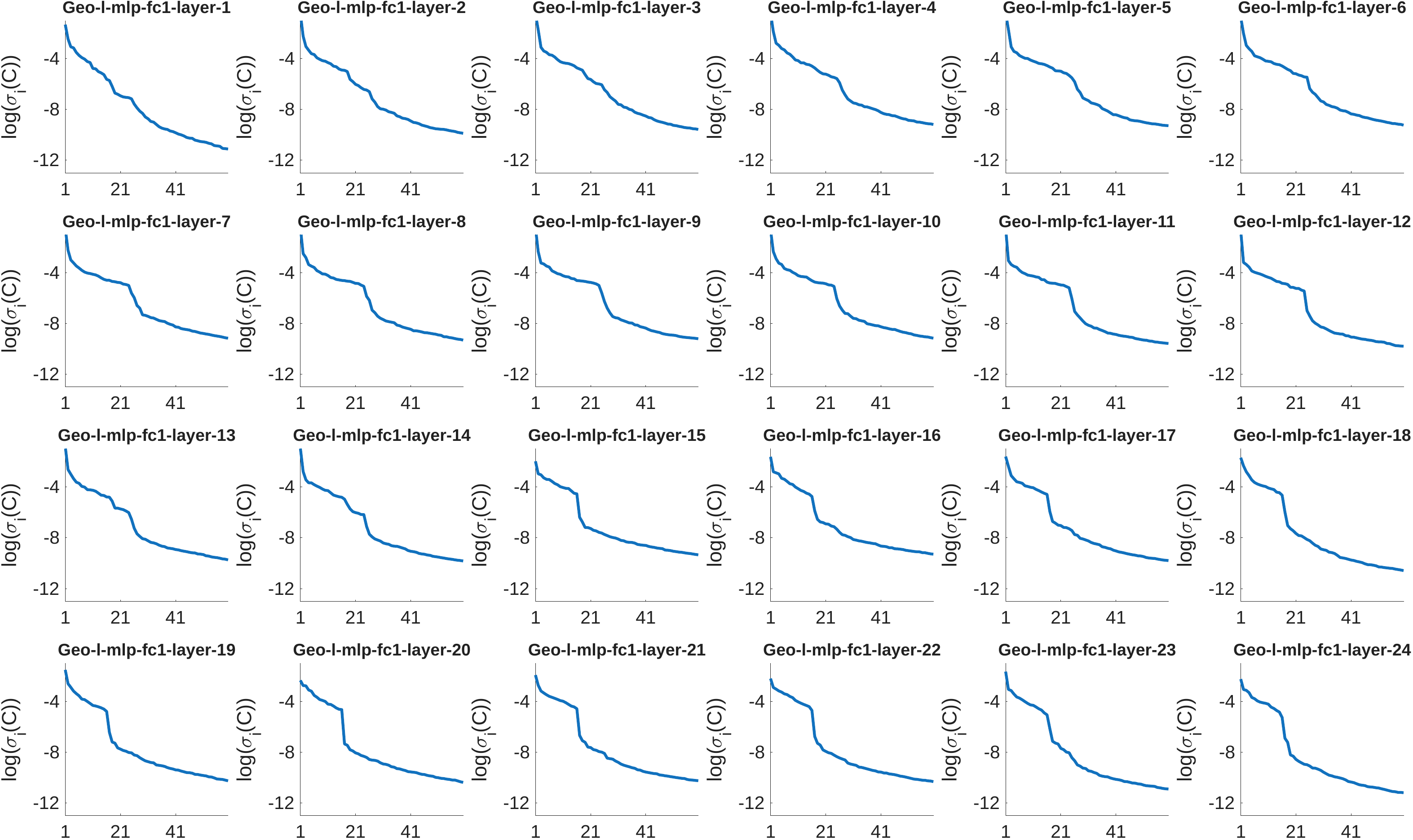}
\vspace{-1em} \caption{Singular values of the first fully connected matrix in the frame attention layer with respect to geometry variations.}
\label{Fig:Geo:l:mlp:fc1}
\vspace{-2em} \end{figure}

\begin{figure}[htbp]
\centering
\includegraphics[width=0.92\linewidth]{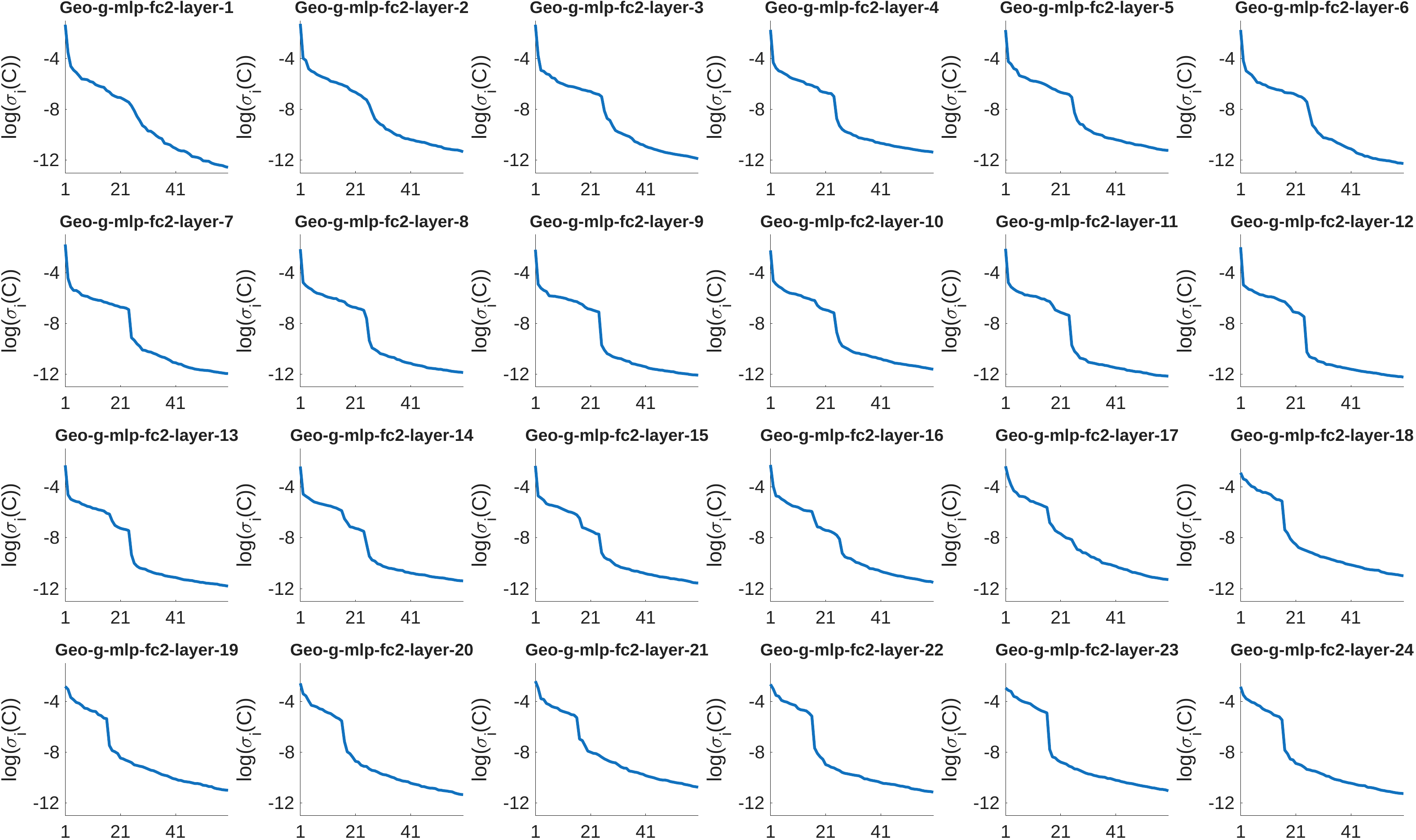}
\vspace{-1em} \caption{Singular values of the second fully connected matrix in the global attention layer with respect to geometry variations.}
\label{Fig:Geo:g:mlp:fc2}
\vspace{-2em} \end{figure}

\begin{figure}[htbp]
\centering
\includegraphics[width=0.92\linewidth]{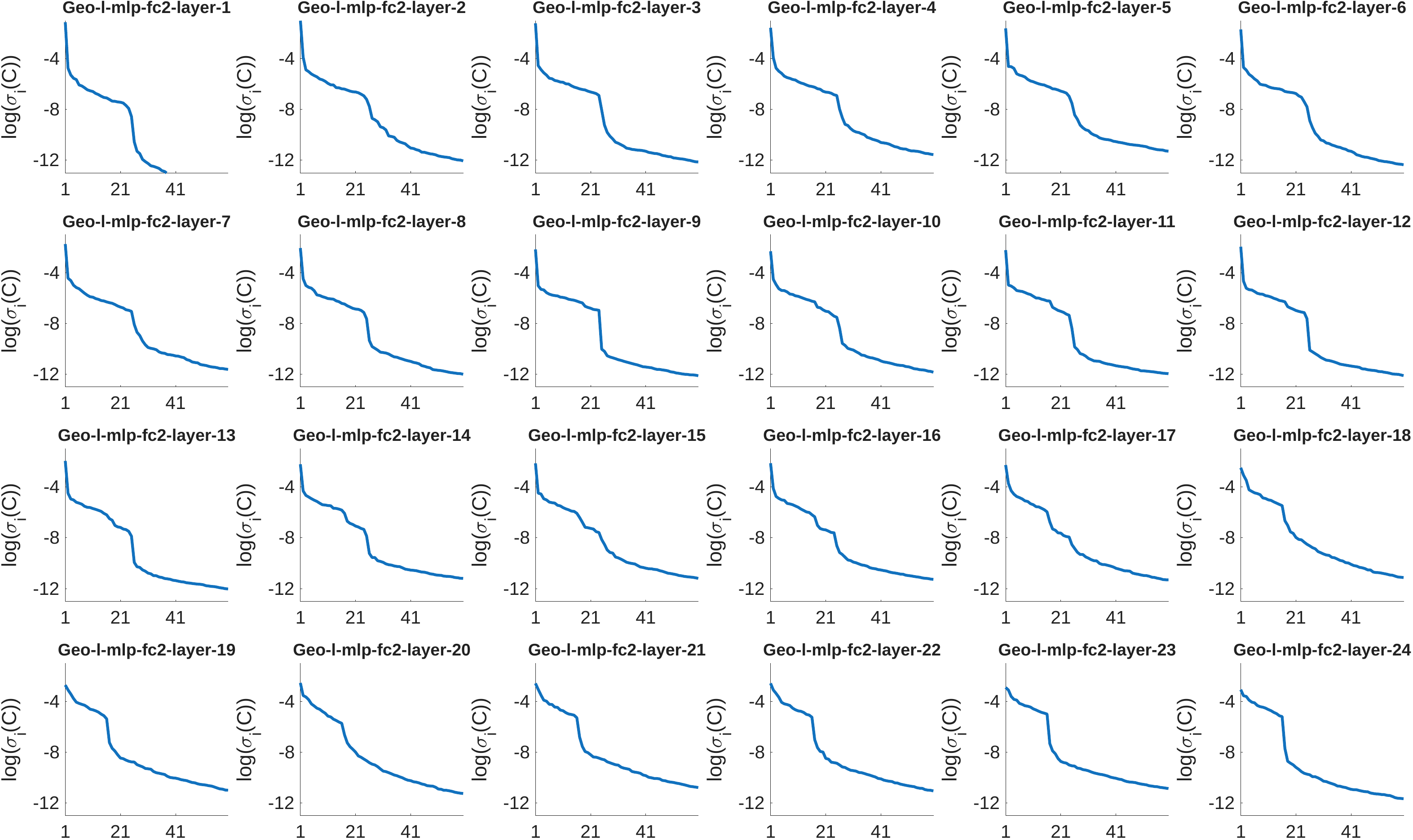}
\vspace{-1em} \caption{Singular values of the second fully connected matrix in the frame attention layer with respect to geometry variations.}
\label{Fig:Geo:l:mlp:fc2}
\vspace{-2em} \end{figure}

\clearpage
\subsection{Camera Motion Subspace}

\begin{figure}[htbp]
\centering
\includegraphics[width=0.92\linewidth]{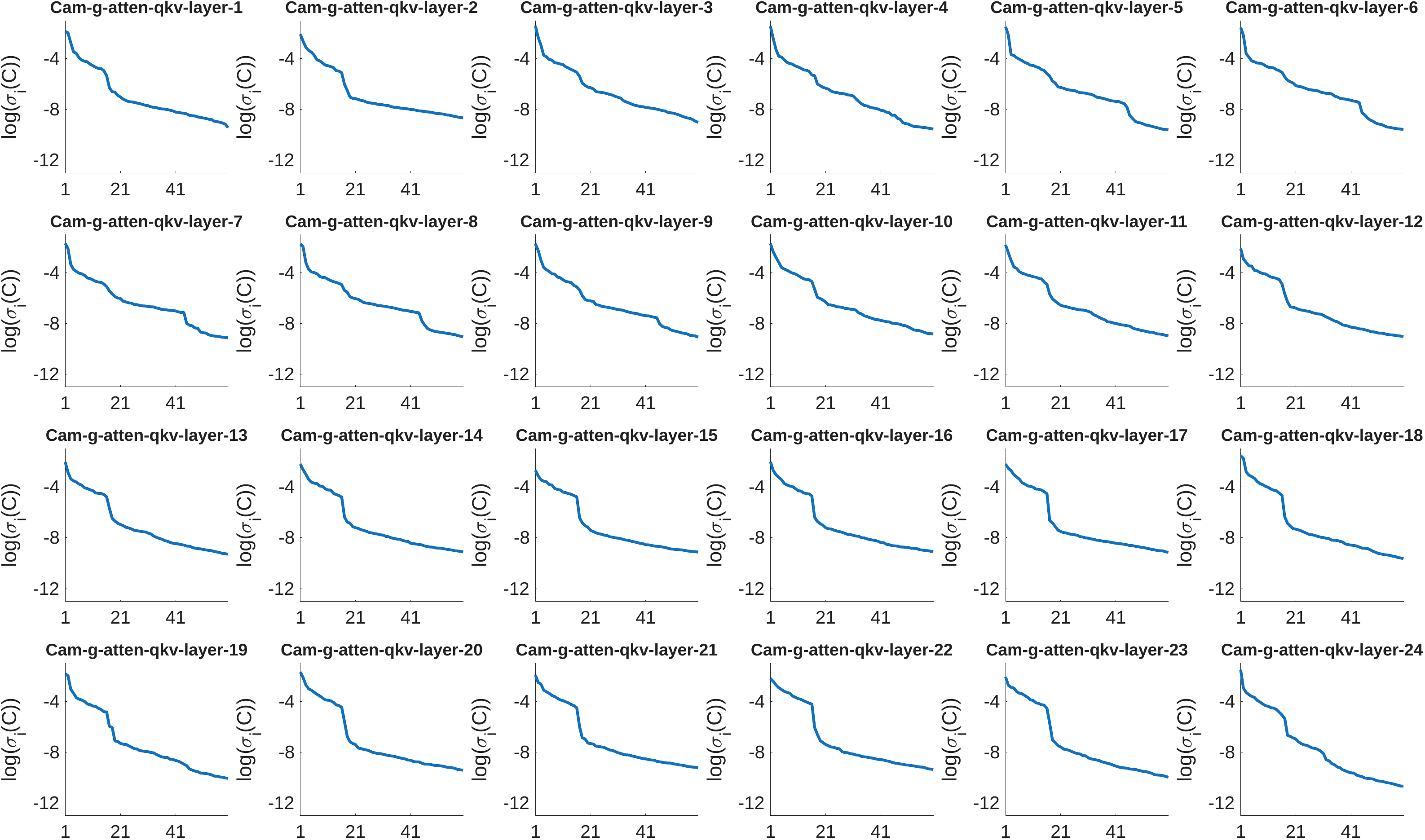}
\vspace{-1em} \caption{Singular values of the QKV matrix in the global attention layer with respect to camera variations.}
\label{Fig:Cam:g:atten:qkv}
\vspace{-2em} \end{figure}

\begin{figure}[htbp]
\centering
\includegraphics[width=0.92\linewidth]{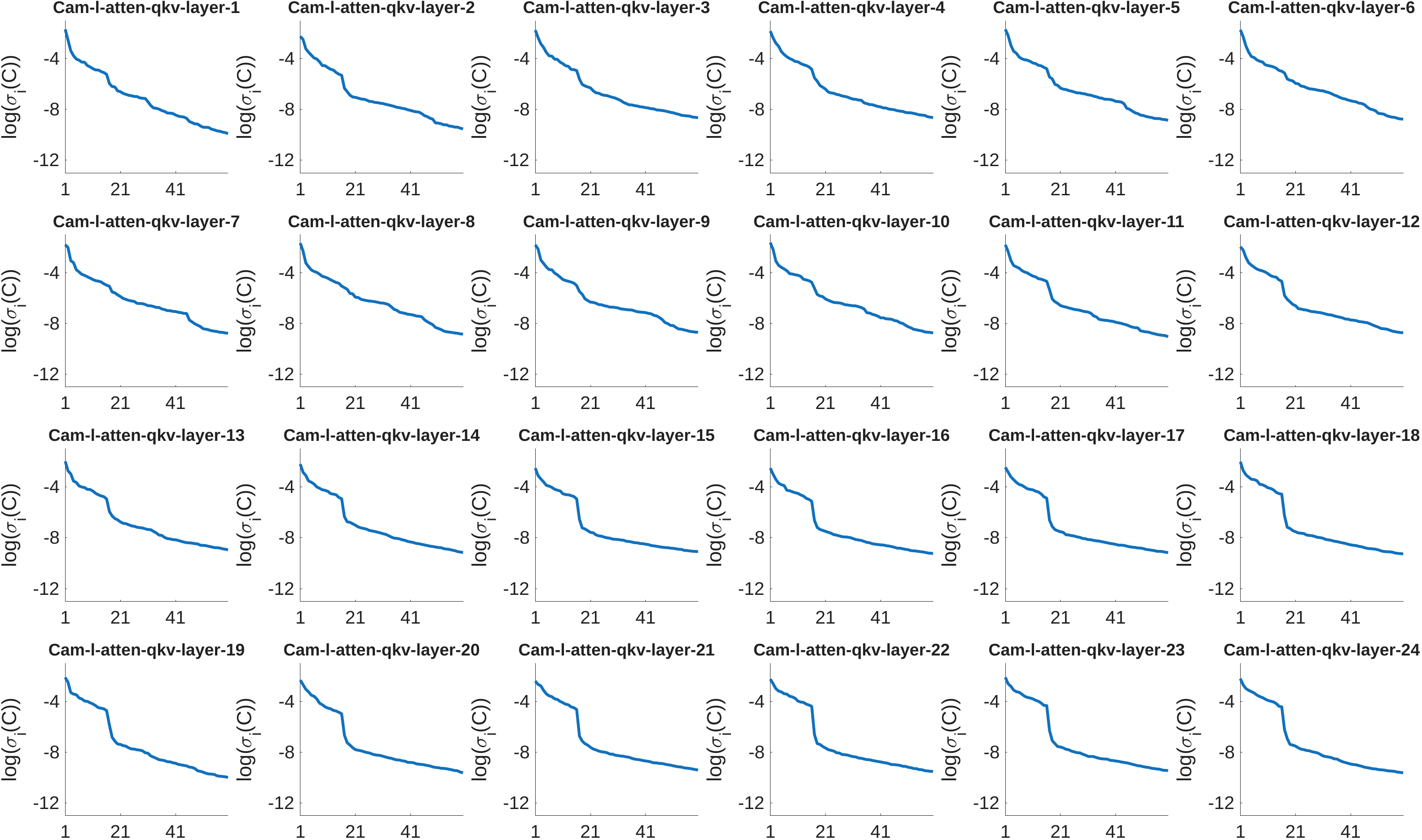}
\vspace{-1em} \caption{Singular values of the QKV matrix in the frame attention layer with respect to camera variations.}
\label{Fig:Cam:l:atten:qkv}
\vspace{-2em} \end{figure}

\begin{figure}[htbp]
\centering
\includegraphics[width=0.92\linewidth]{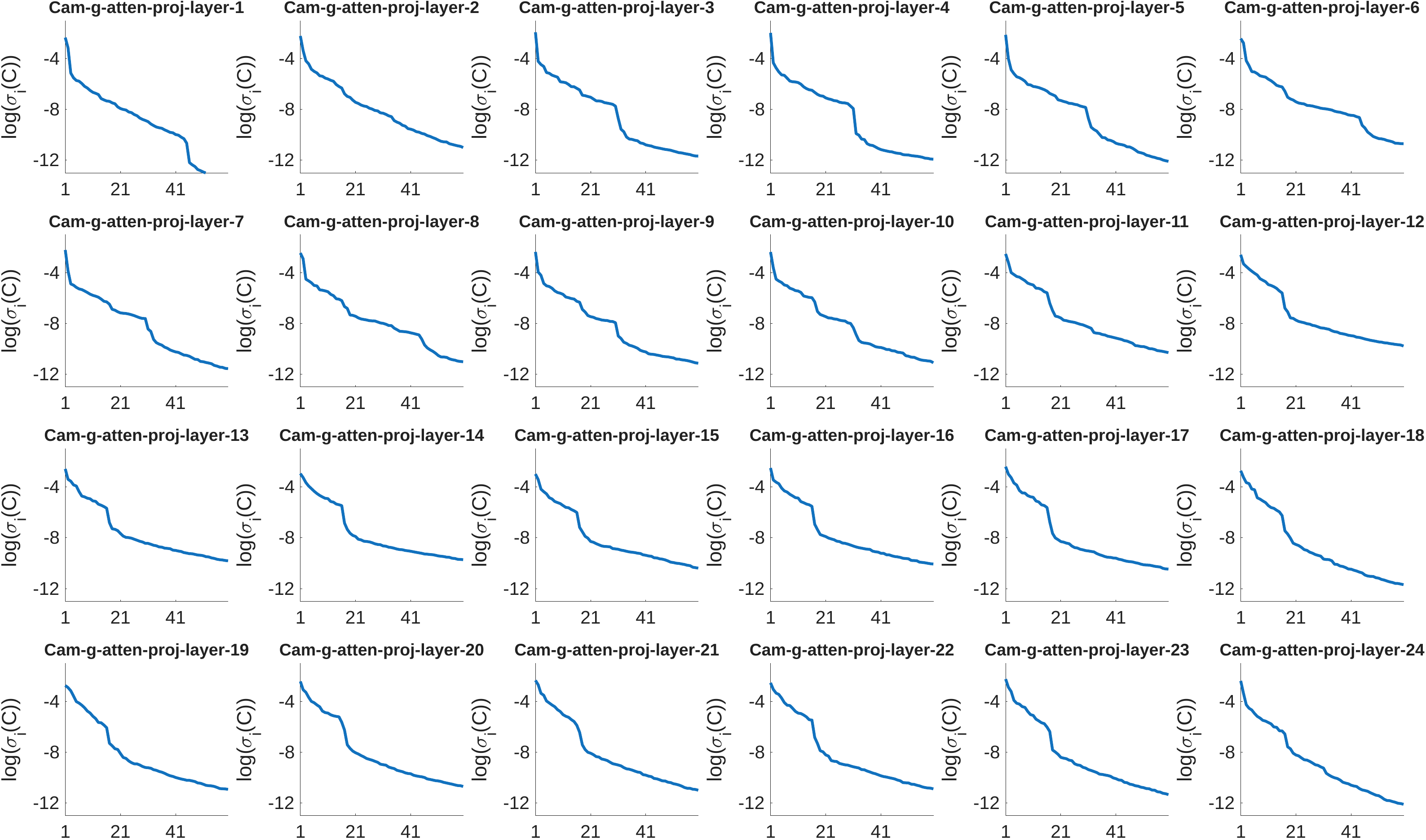}
\vspace{-1em} \caption{Singular values of the projection matrix in the global attention layer with respect to camera variations. }
\label{Fig:Cam:g:atten:proj}
\vspace{-2em} \end{figure}

\begin{figure}[htbp]
\centering
\includegraphics[width=0.92\linewidth]{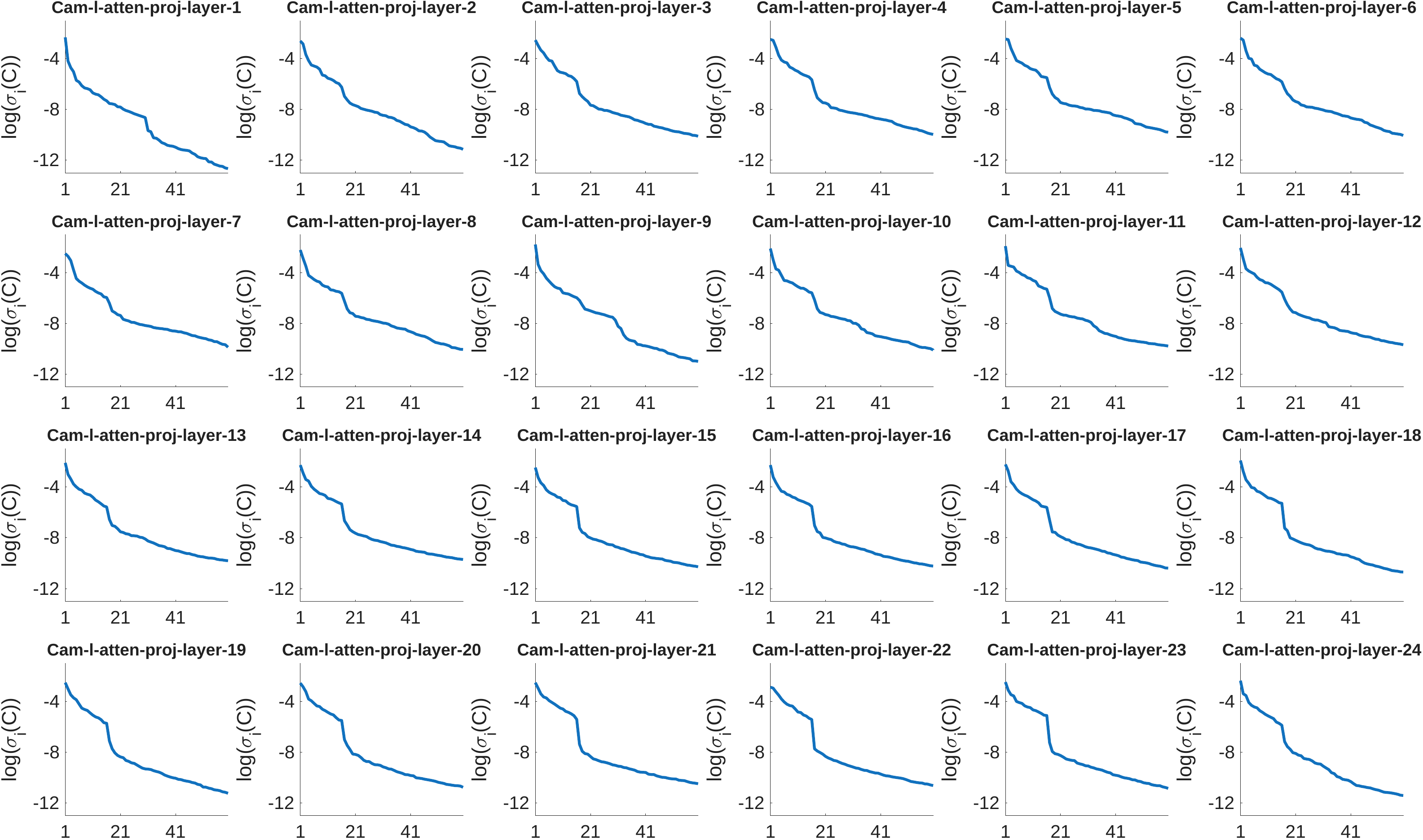}
\vspace{-1em} \caption{Singular values of the projection matrix in the frame attention layer with respect to camera variations.}
\label{Fig:Cam:l:atten:proj}
\vspace{-2em} \end{figure}

\begin{figure}[htbp]
\centering
\vspace{2em}
\includegraphics[width=0.92\linewidth]
{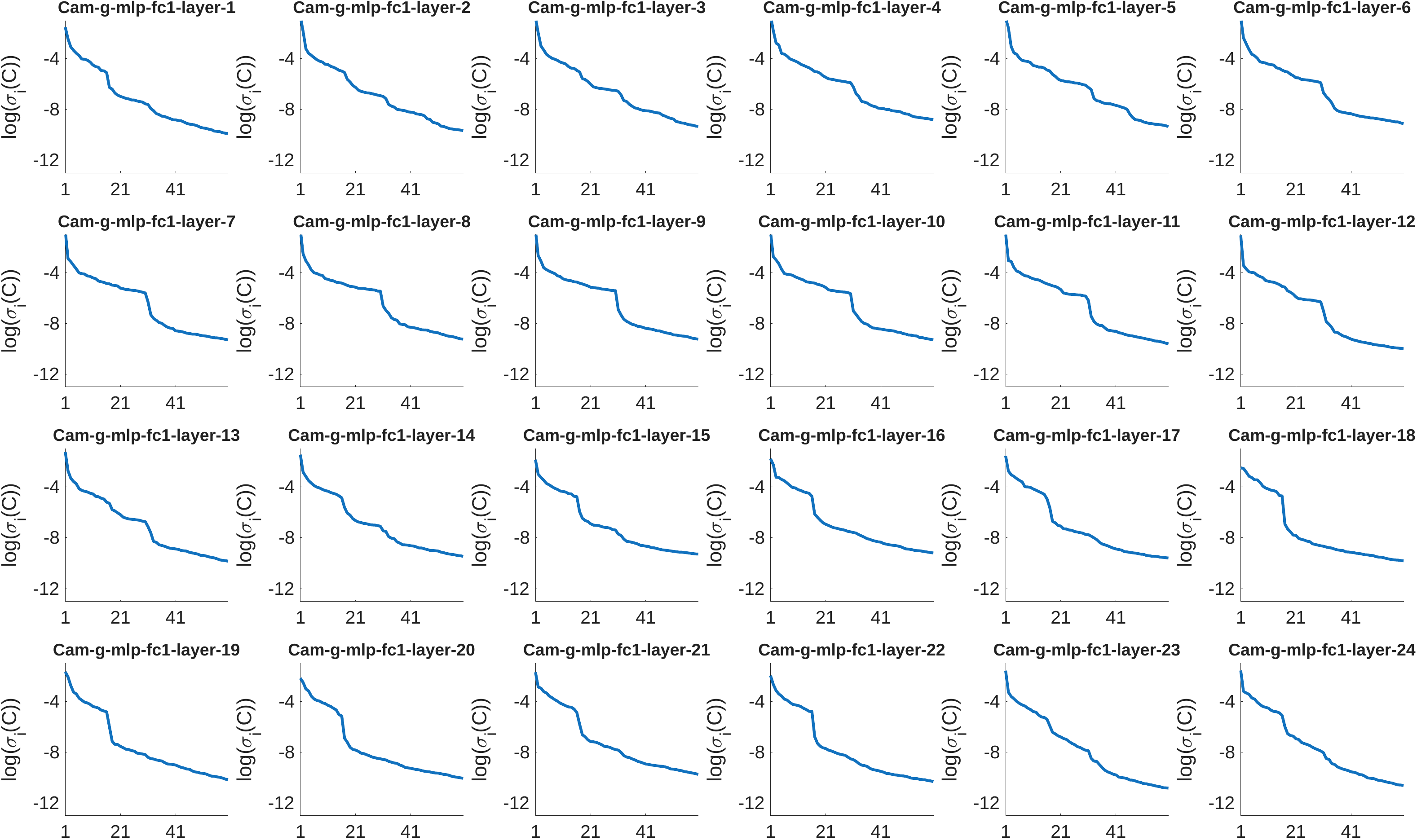}
\vspace{-1em} \caption{Singular values of the first fully connected matrix in the global attention layer with respect to camera variations.}
\label{Fig:Cam:g:mlp:fc1}
\vspace{-2em} \end{figure}

\begin{figure}[htbp]
\centering
\includegraphics[width=0.92\linewidth]{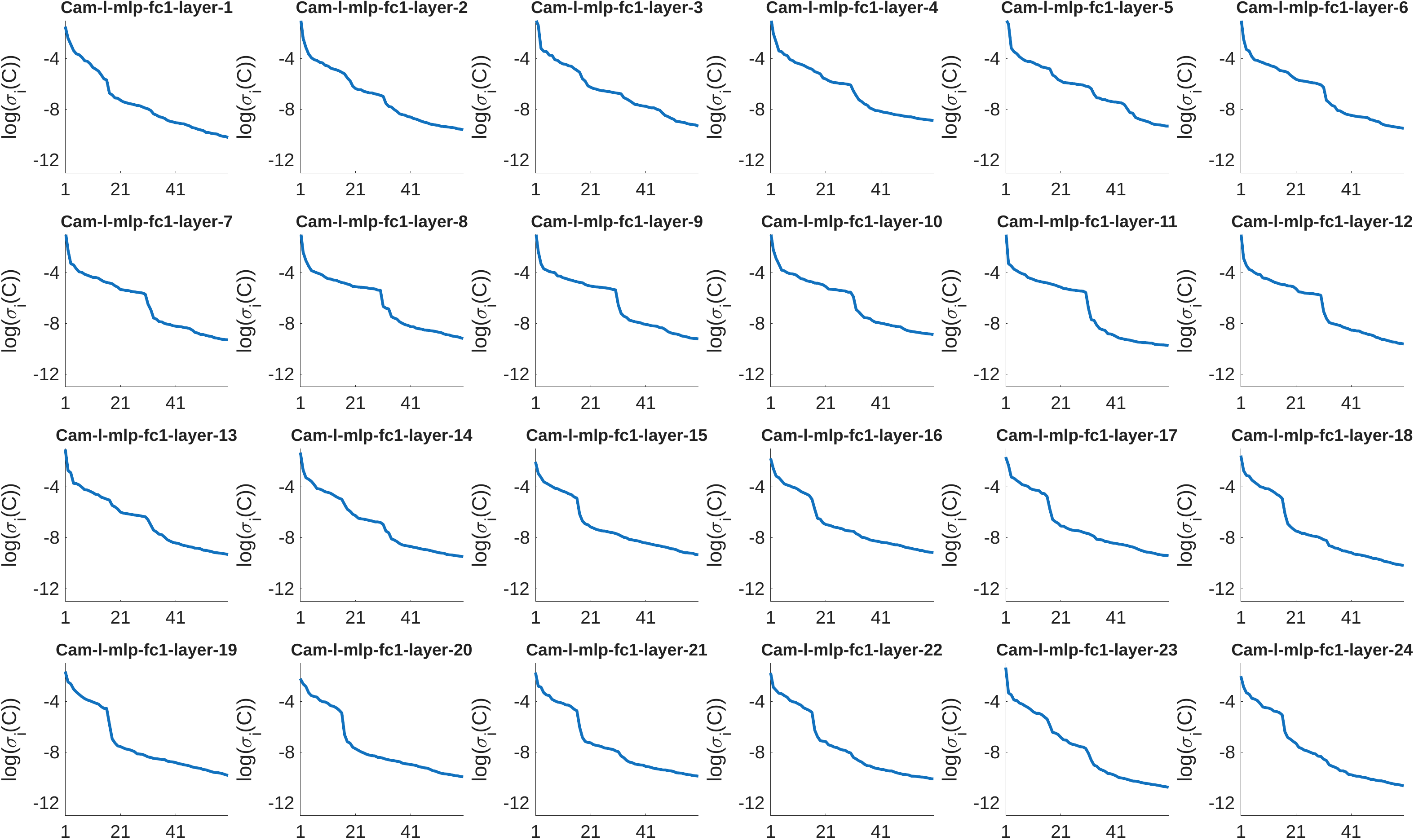}
\vspace{-1em} \caption{Singular values of the first fully connected matrix in the frame attention layer with respect to camera variations.}
\label{Fig:Cam:l:mlp:fc1}
\vspace{-2em} \end{figure}

\begin{figure}[htbp]
\centering
\includegraphics[width=0.92\linewidth]{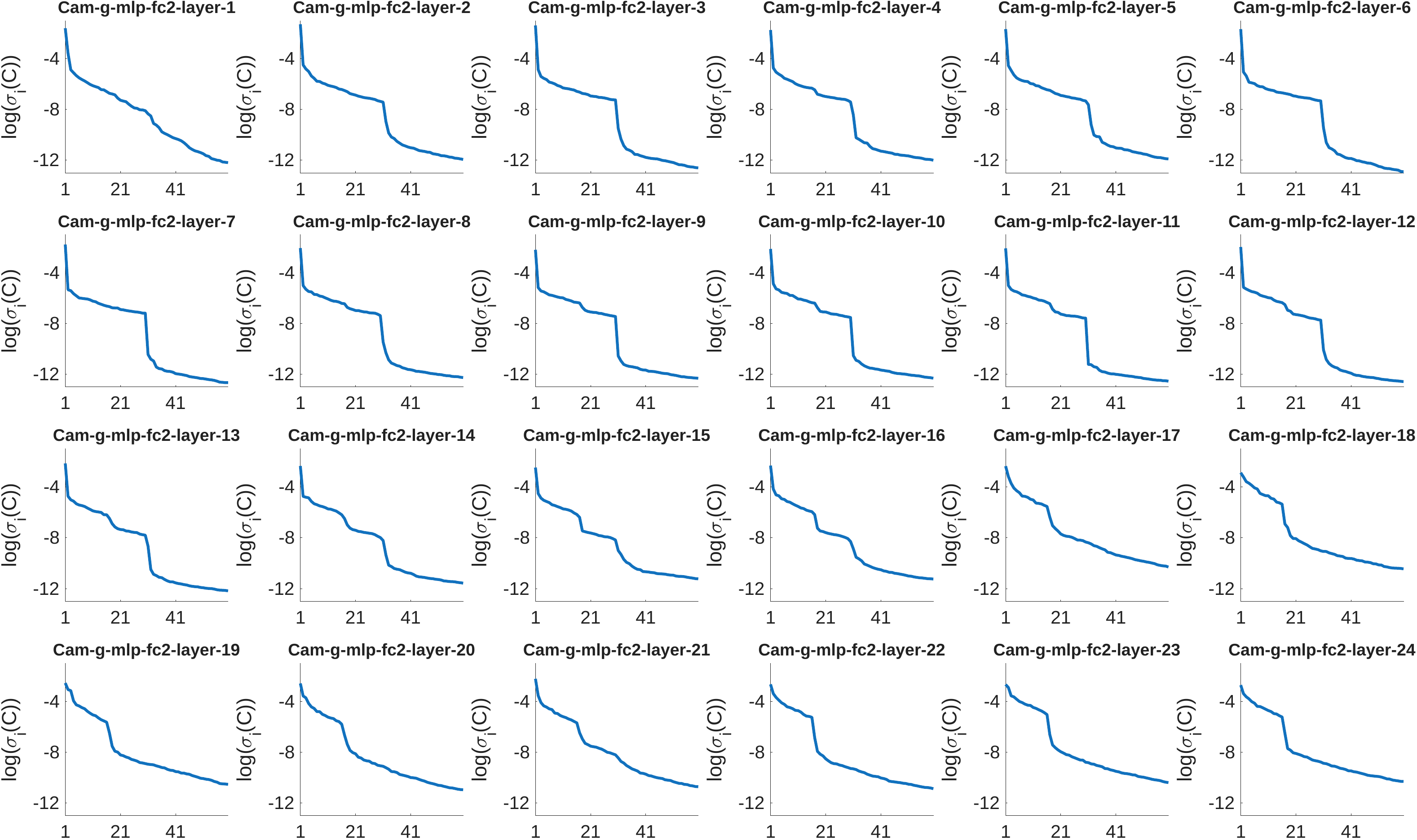}
\vspace{-1em} \caption{Singular values of the second fully connected matrix in the global attention layer with respect to camera variations.}
\label{Fig:Cam:g:mlp:fc2}
\vspace{-2em} \end{figure}

\begin{figure}[htbp]
\centering
\includegraphics[width=0.92\linewidth]{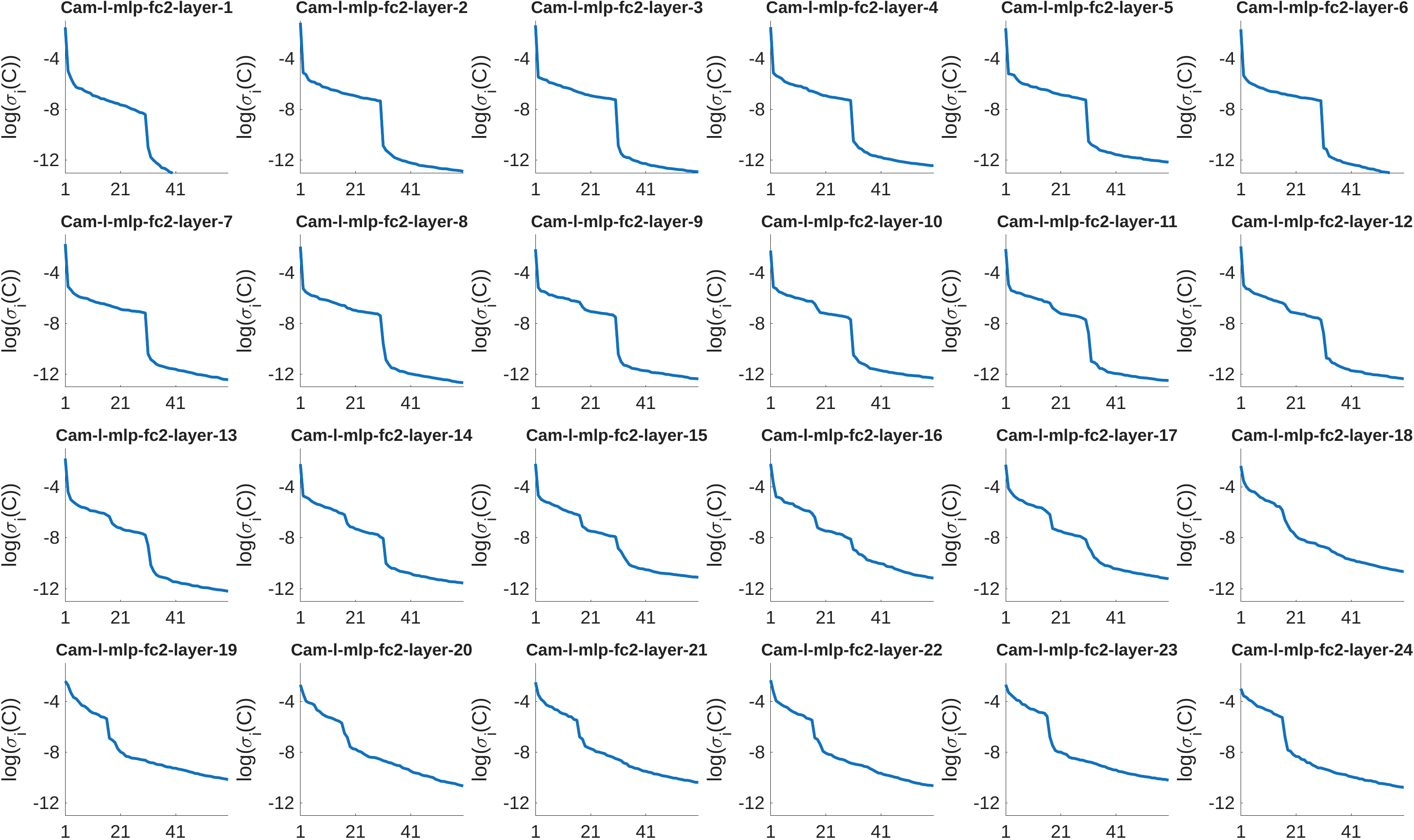}
\vspace{-1em} \caption{Singular values of the second fully connected matrix in the frame attention layer with respect to camera variations.}
\label{Fig:Cam:l:mlp:fc2}
\vspace{-2em} \end{figure}

\clearpage
\subsection{Lighting Subspace}

\begin{figure}[htbp]
\centering
\includegraphics[width=0.92\linewidth]{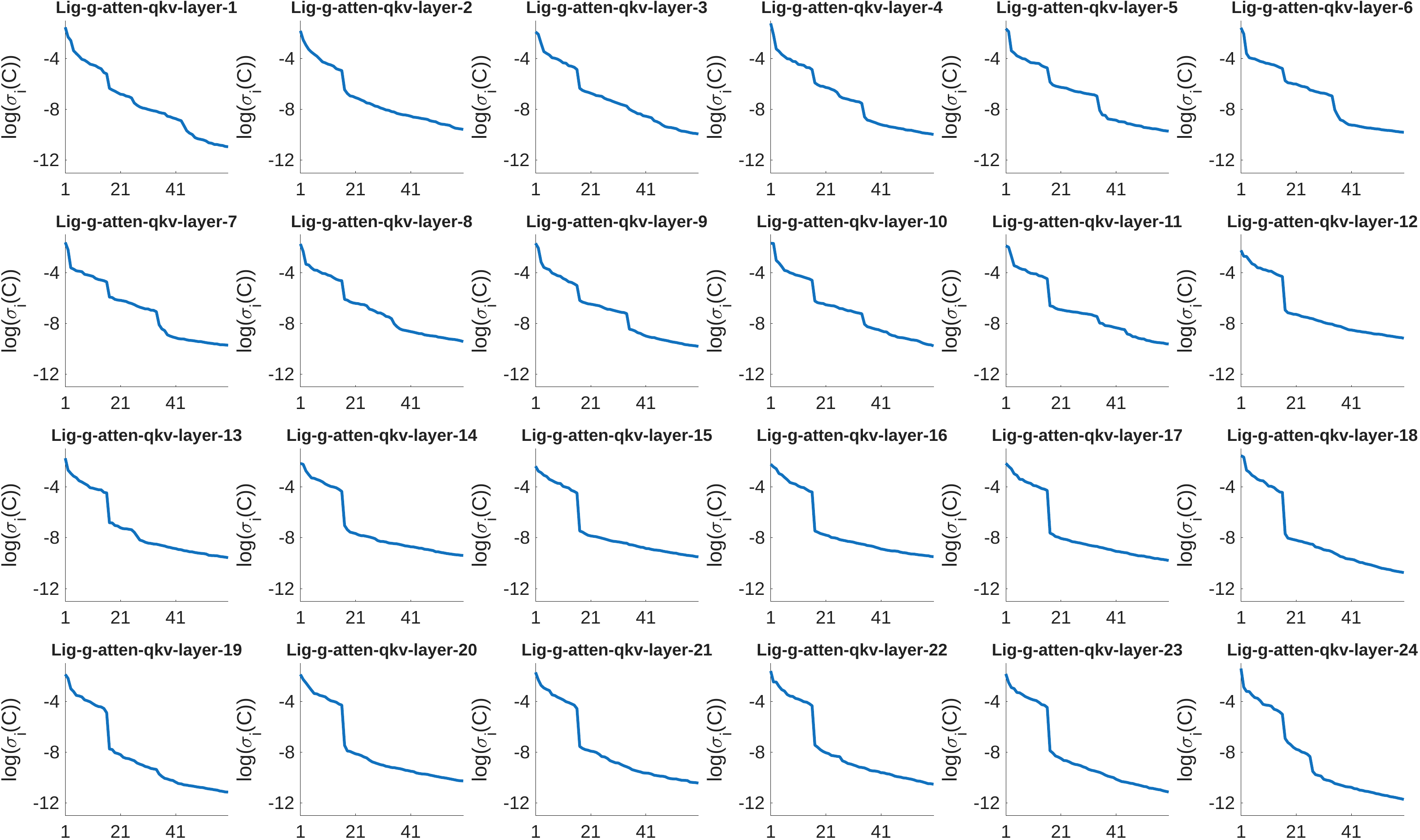}
\vspace{-1em} \caption{Singular values of the QKV matrix in the global attention layer with respect to lighting variations.}
\label{Fig:Lig:g:atten:qkv}
\vspace{-2em} \end{figure}

\begin{figure}[htbp]
\centering
\includegraphics[width=0.92\linewidth]{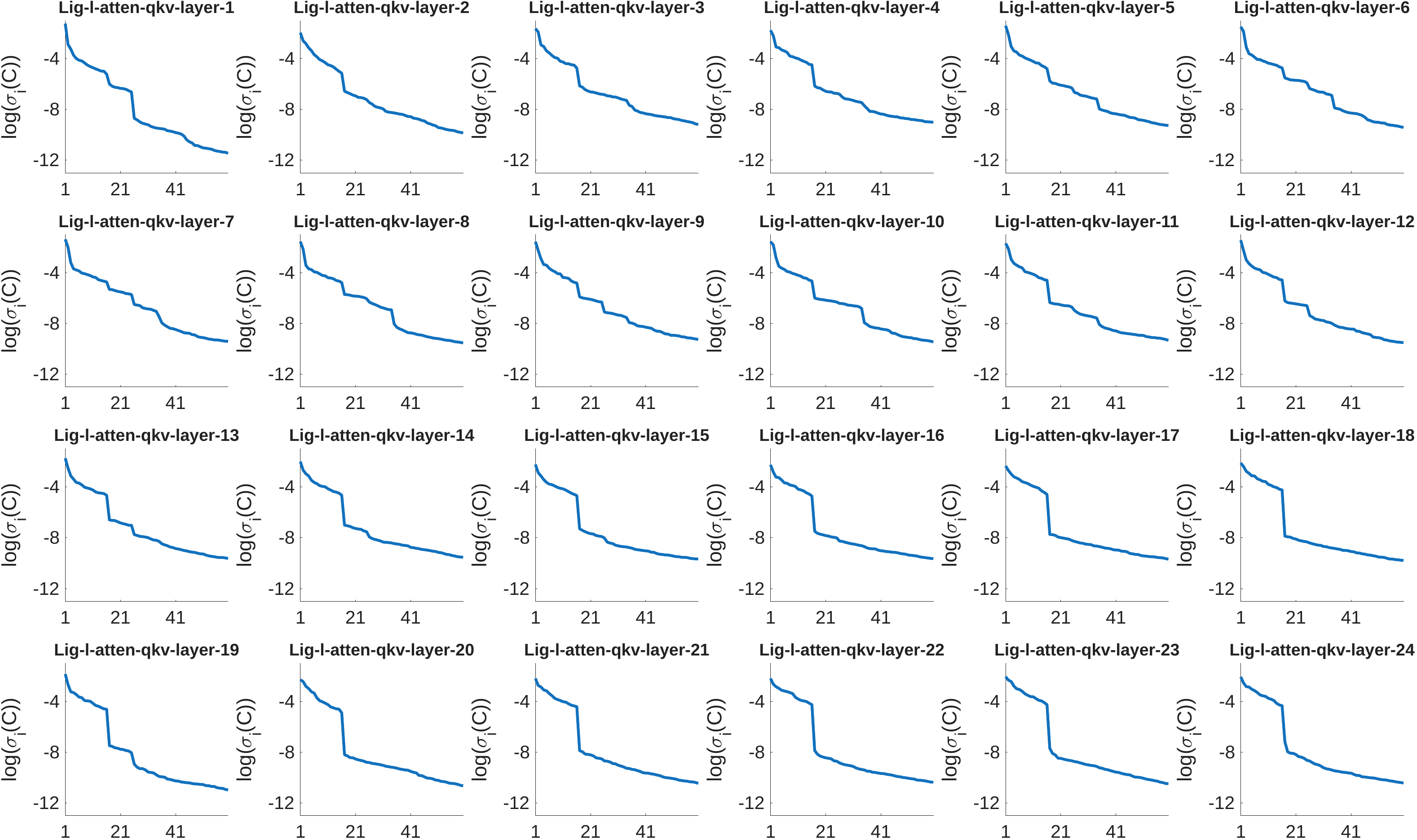}
\vspace{-1em} \caption{Singular values of the QKV matrix in the frame attention layer with respect to lighting variations.}
\label{Fig:Lig:l:atten:qkv}
\vspace{-2em} \end{figure}

\begin{figure}[htbp]
\centering
\includegraphics[width=0.92\linewidth]{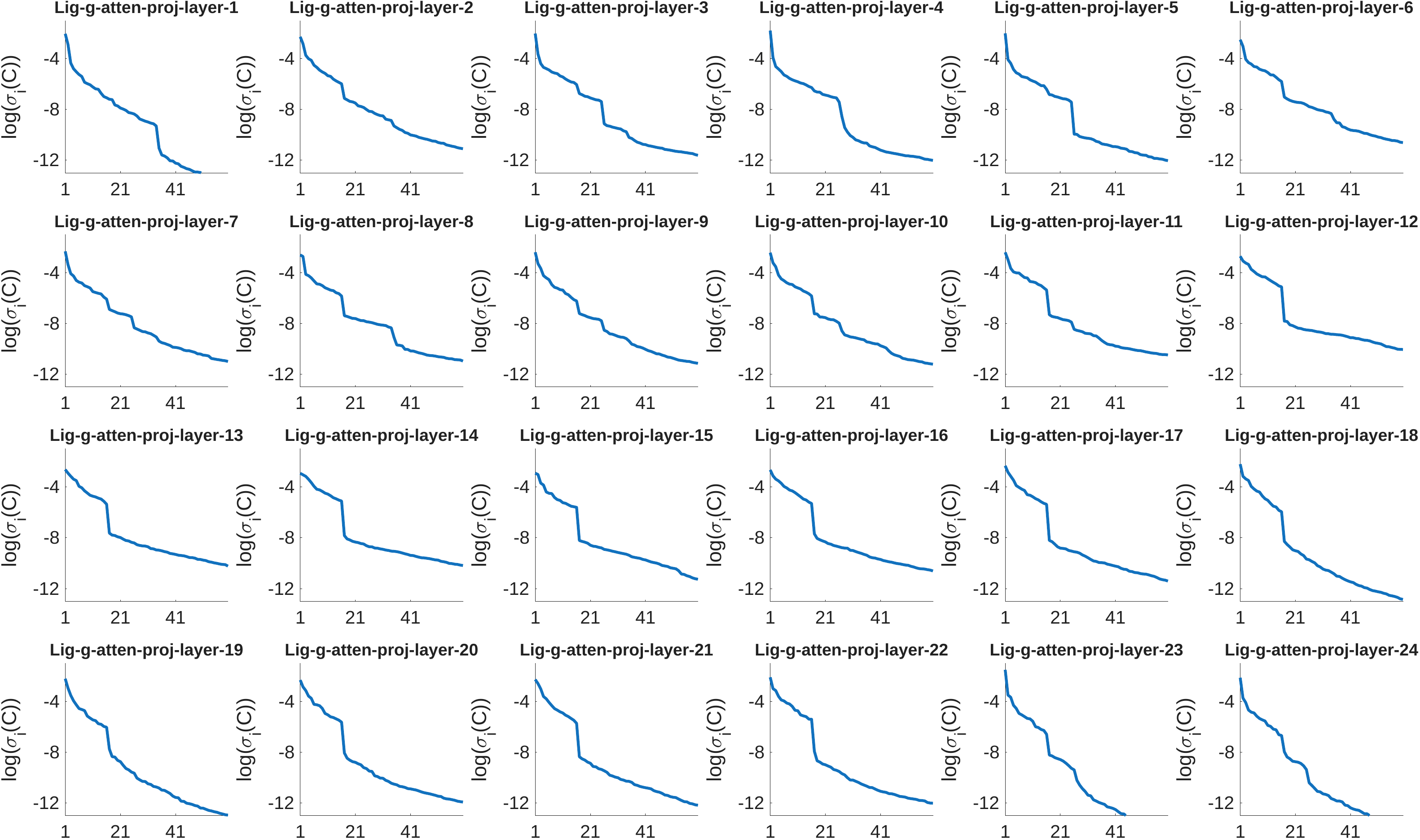}
\vspace{-1em} \caption{Singular values of the projection matrix in the global attention layer with respect to lighting variations.}
\label{Fig:Lig:g:atten:proj}
\vspace{-2em} \end{figure}

\begin{figure}[htbp]
\centering
\includegraphics[width=0.92\linewidth]{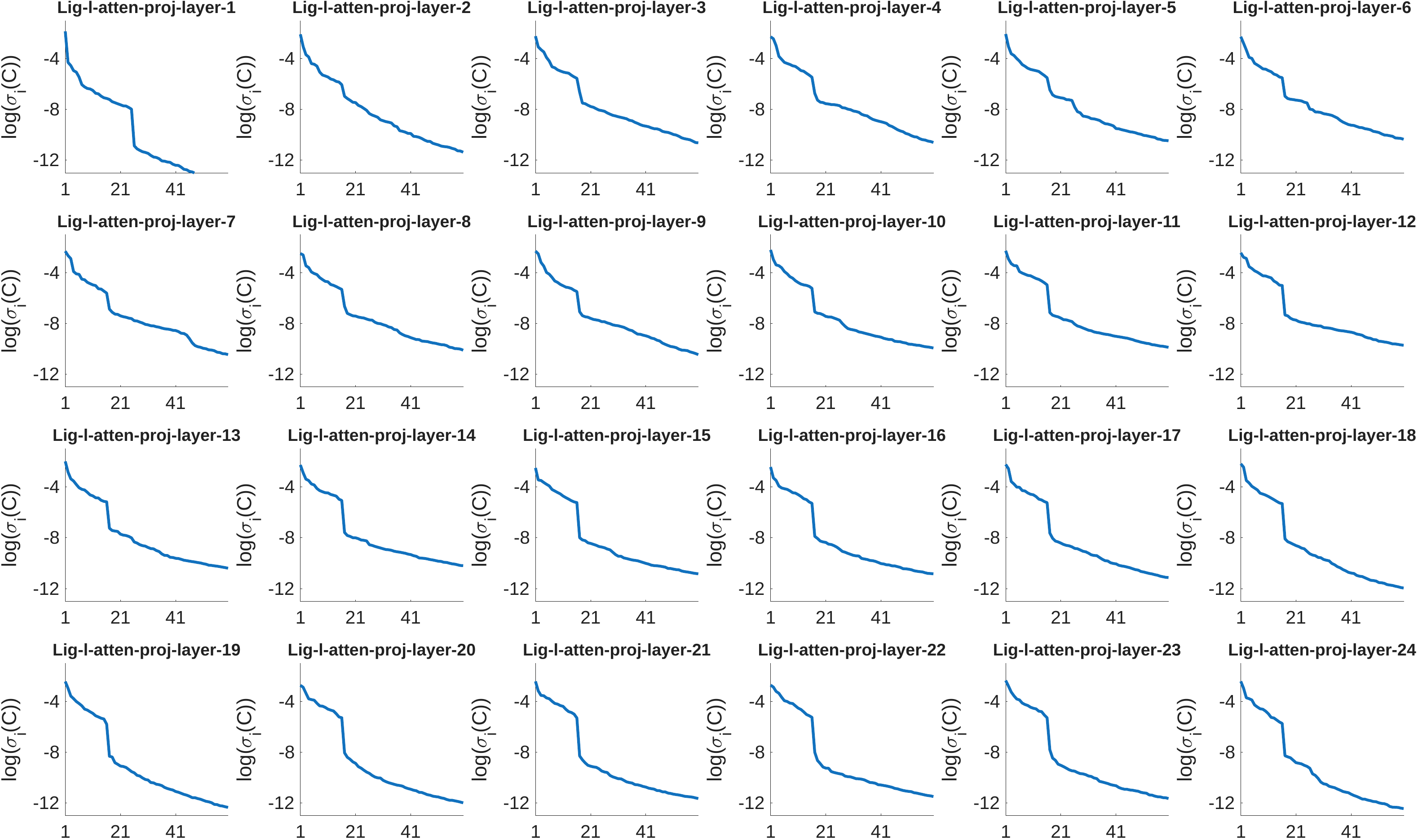}
\vspace{-1em} \caption{Singular values of the projection matrix in the frame attention layer with respect to lighting variations.}
\label{Fig:Lig:l:atten:proj}
\vspace{-2em} \end{figure}

\begin{figure}[htbp]
\centering
\vspace{2em}
\includegraphics[width=0.92\linewidth]{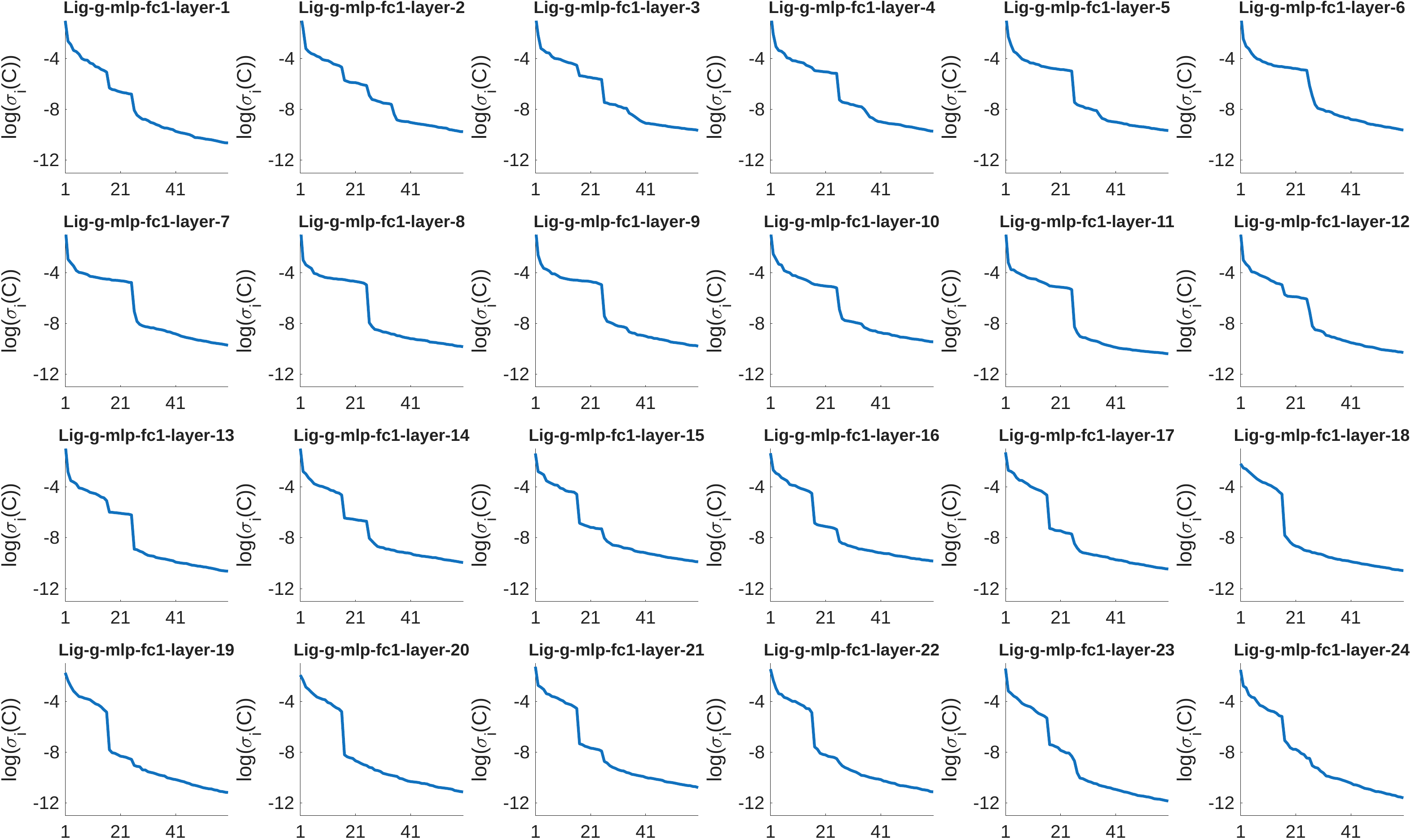}
\vspace{-1em} \caption{Singular values of the first fully connected matrix in the global attention layer with respect to lighting variations.}
\label{Fig:Lig:g:mlp:fc1}
\vspace{-2em} \end{figure}

\begin{figure}[htbp]
\centering
\includegraphics[width=0.92\linewidth]{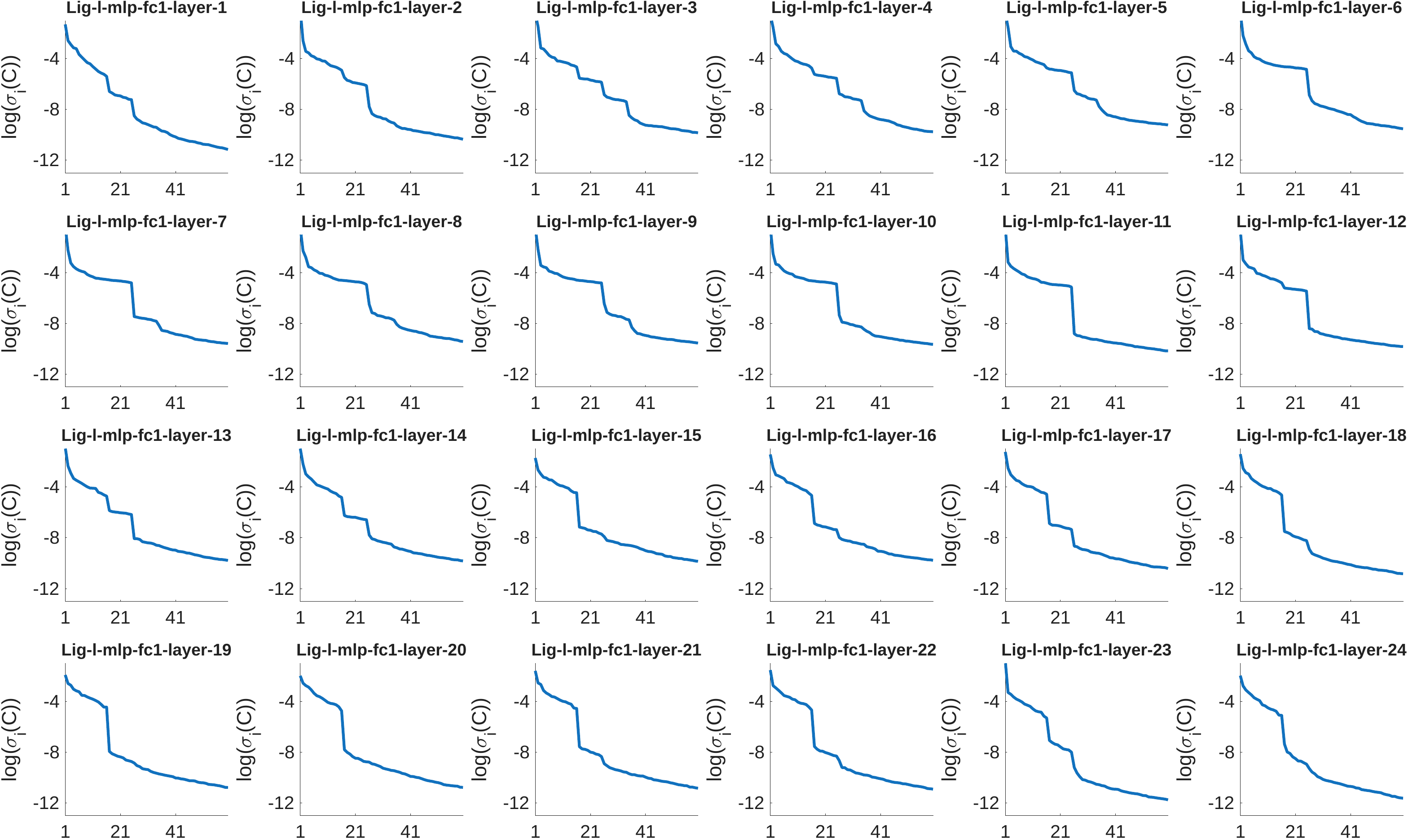}
\vspace{-1em} \caption{Singular values of the first fully connected matrix in the frame attention layer with respect to lighting variations.}
\label{Fig:Lig:l:mlp:fc1}
\vspace{-2em} \end{figure}

\begin{figure}[htbp]
\centering
\includegraphics[width=0.92\linewidth]{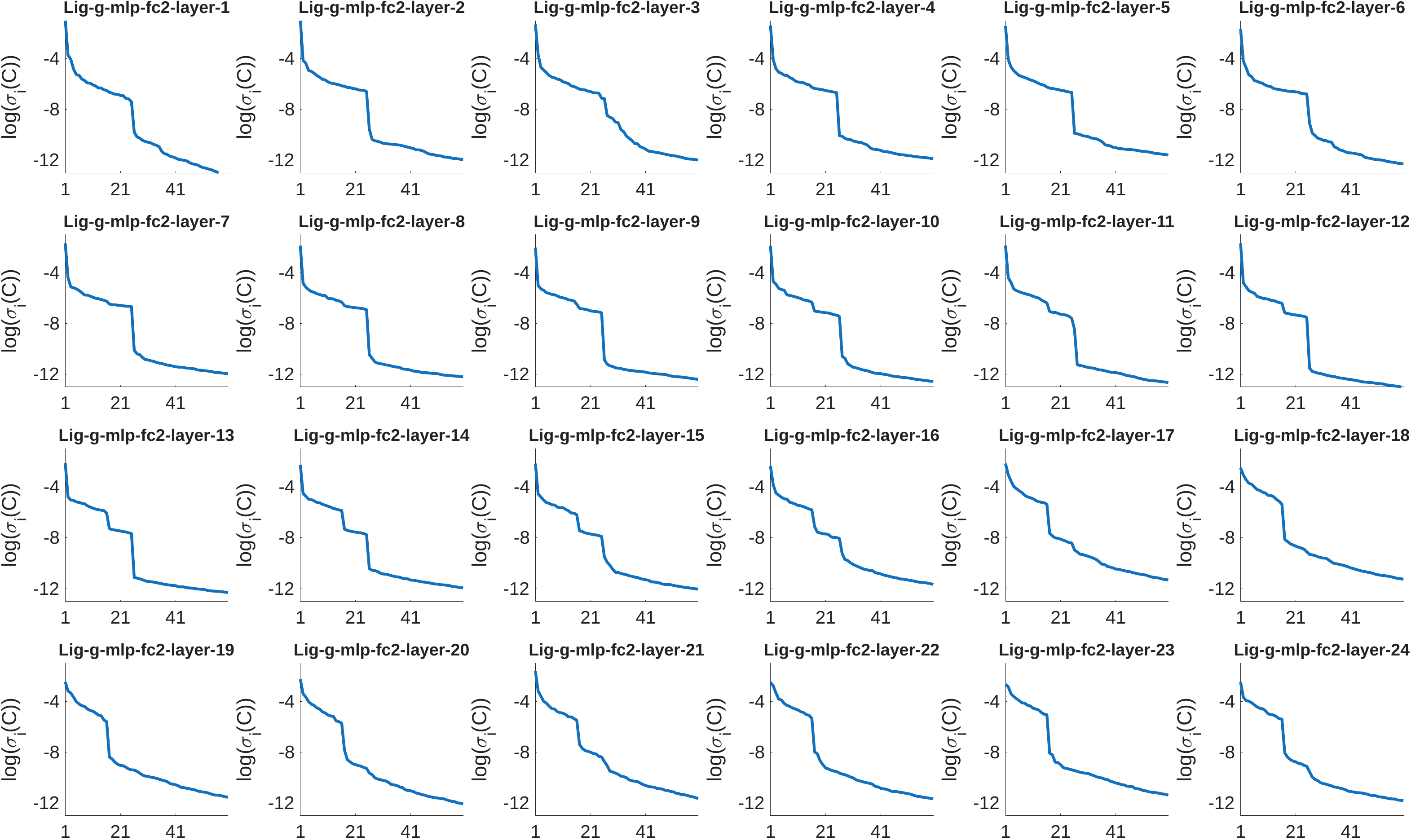}
\vspace{-1em} \caption{Singular values of the second fully connected matrix in the global attention layer with respect to lighting variations.}
\label{Fig:Lig:g:mlp:fc2}
\vspace{-2em} \end{figure}

\begin{figure}[htbp]
\centering
\includegraphics[width=0.92\linewidth]{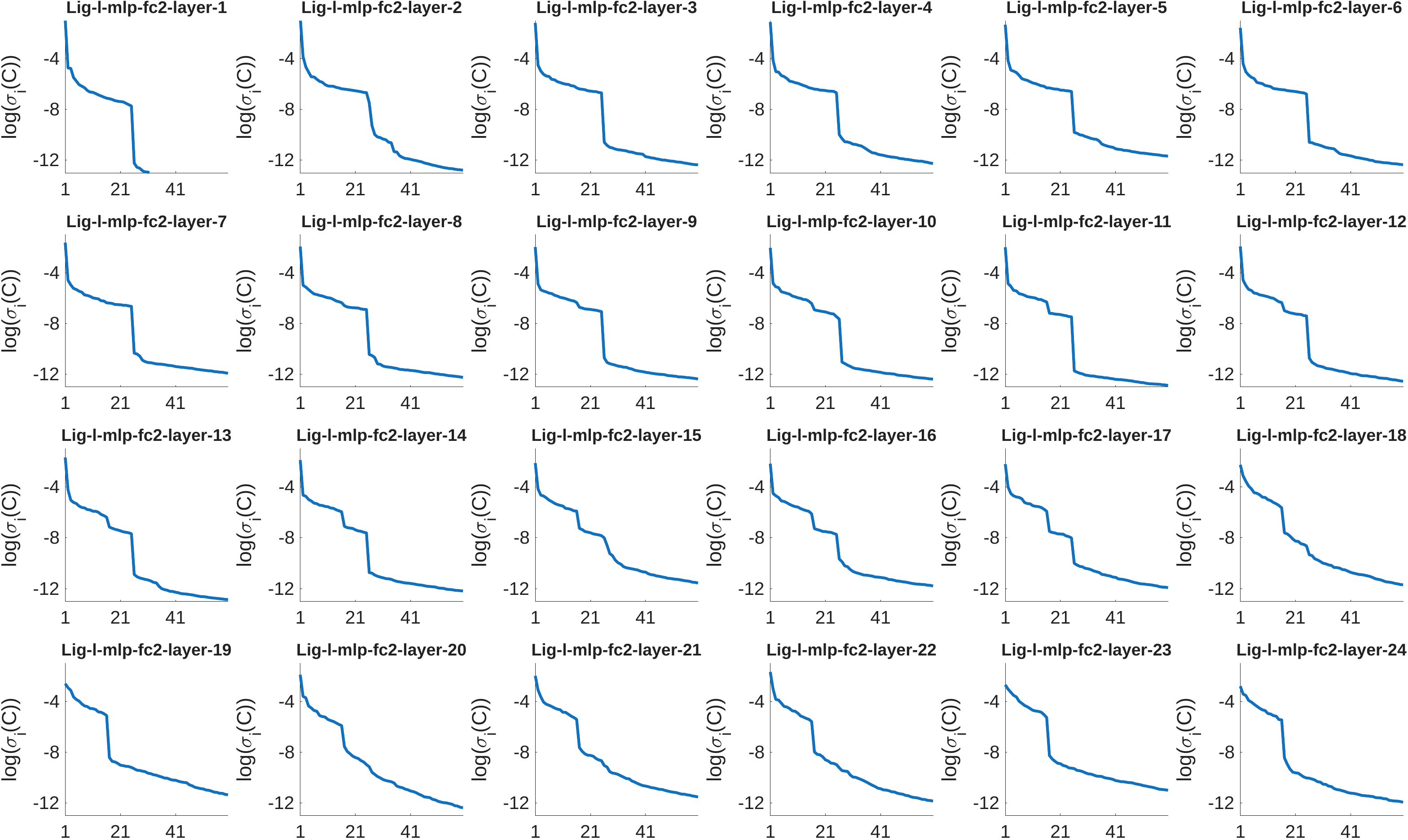}
\vspace{-1em} \caption{Singular values of the second fully connected matrix in the frame attention layer with respect to lighting variations.}
\label{Fig:Lig:l:mlp:fc2}
\vspace{-2em} \end{figure}

\clearpage

\section{Subspace Orthogonality Analysis}

In this section, we present the distribution of the generalized eigenvalues $\lambda$ used in our subspace orthogonality analysis. Results are shown for all six pairs of subspaces. The eigenvalue acts as a reprojection error, where values closer to 1 indicate greater orthogonality between two subspaces. The curves show that these six pairs of subspaces are approximately orthogonal to each other.

\clearpage

\subsection{Geometry vs Texture}

\begin{table*}[bp]
\centering
\setlength\tabcolsep{6pt}
\begin{tabular}{cccc}
\includegraphics[width=0.21\textwidth]{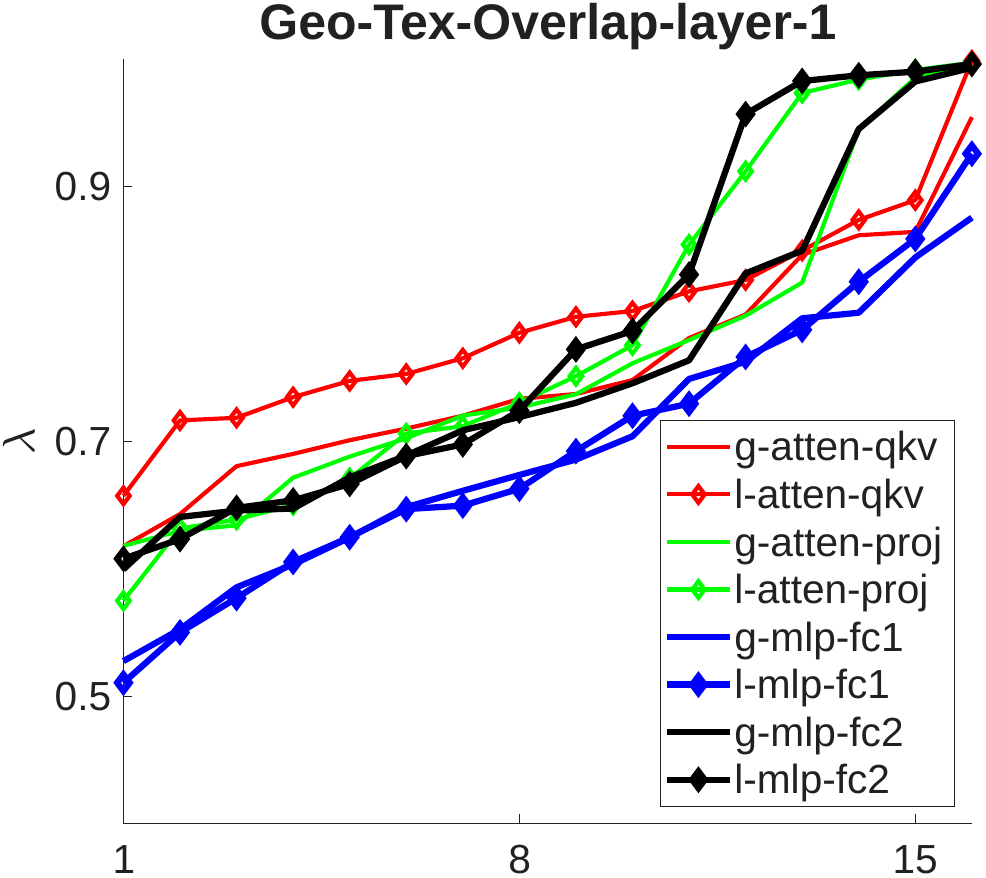}
& 
\includegraphics[width=0.21\textwidth]{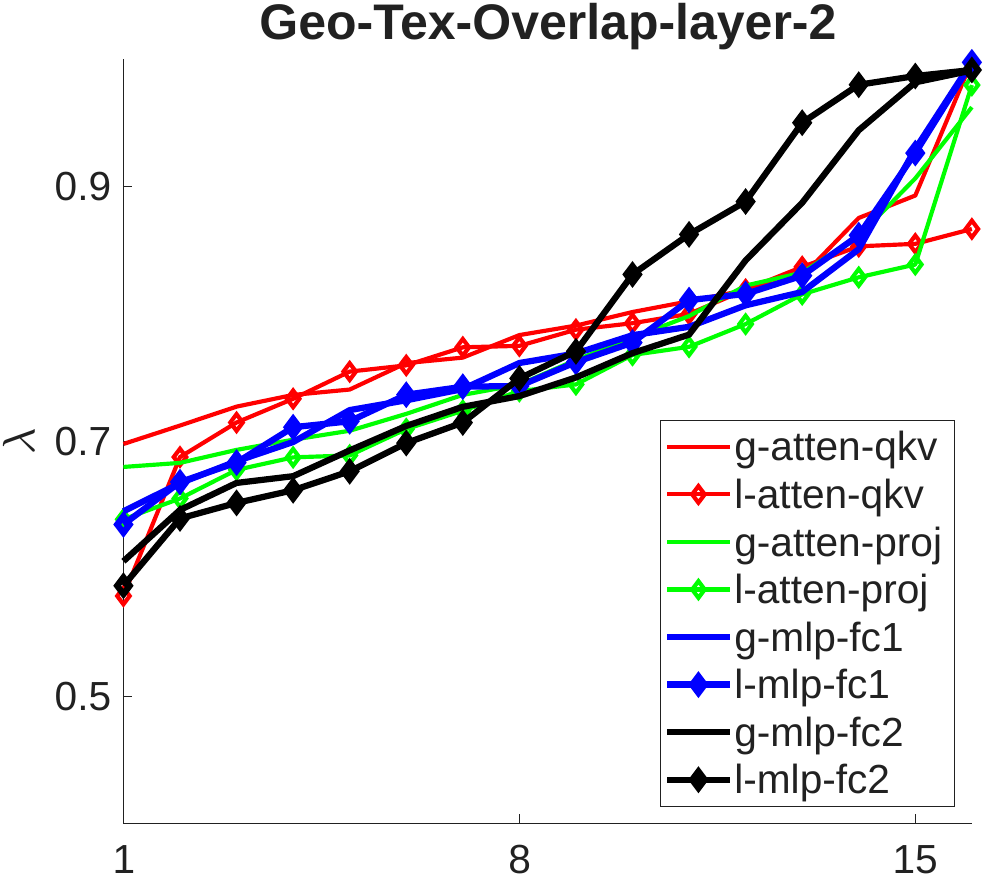}
&
\includegraphics[width=0.21\textwidth]{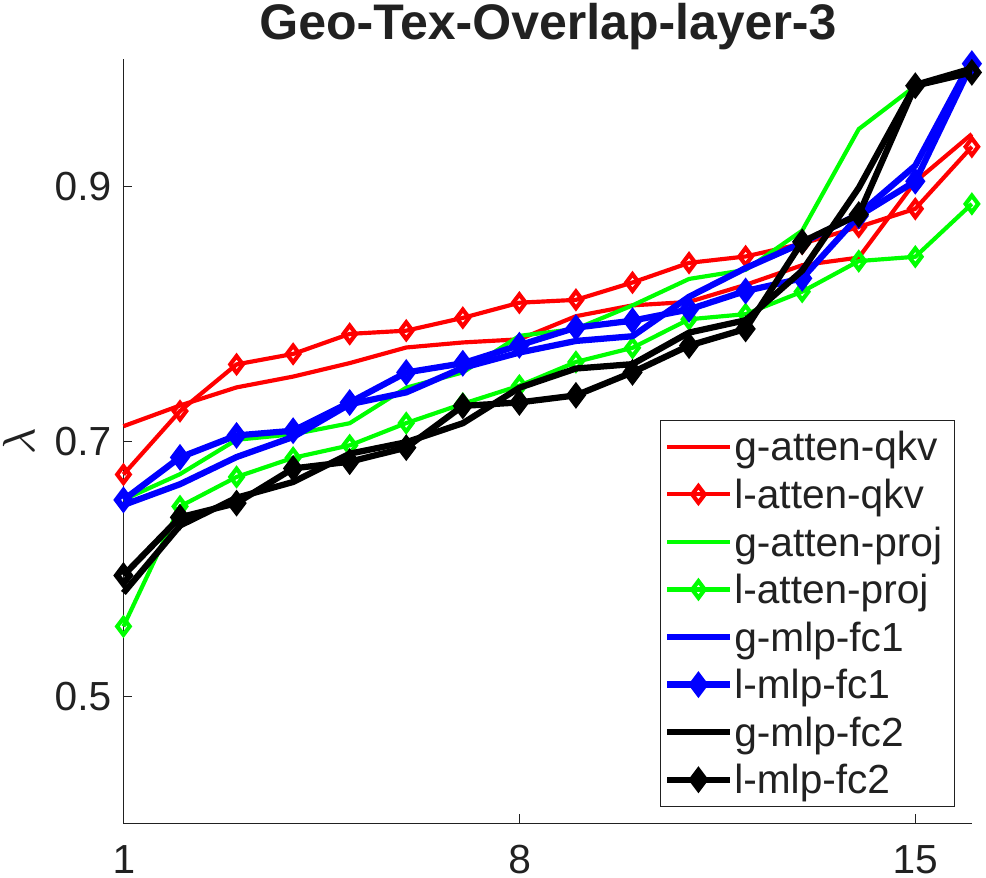}
&
\includegraphics[width=0.21\textwidth]{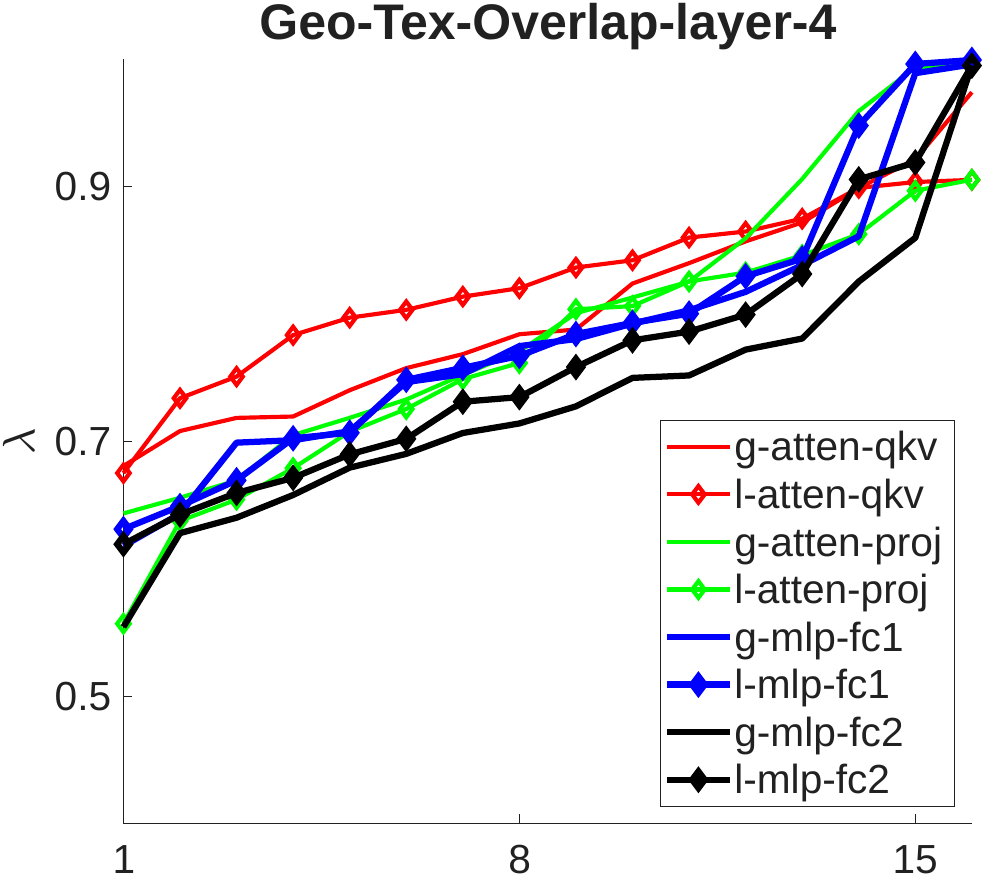}      
\\
\includegraphics[width=0.21\textwidth]{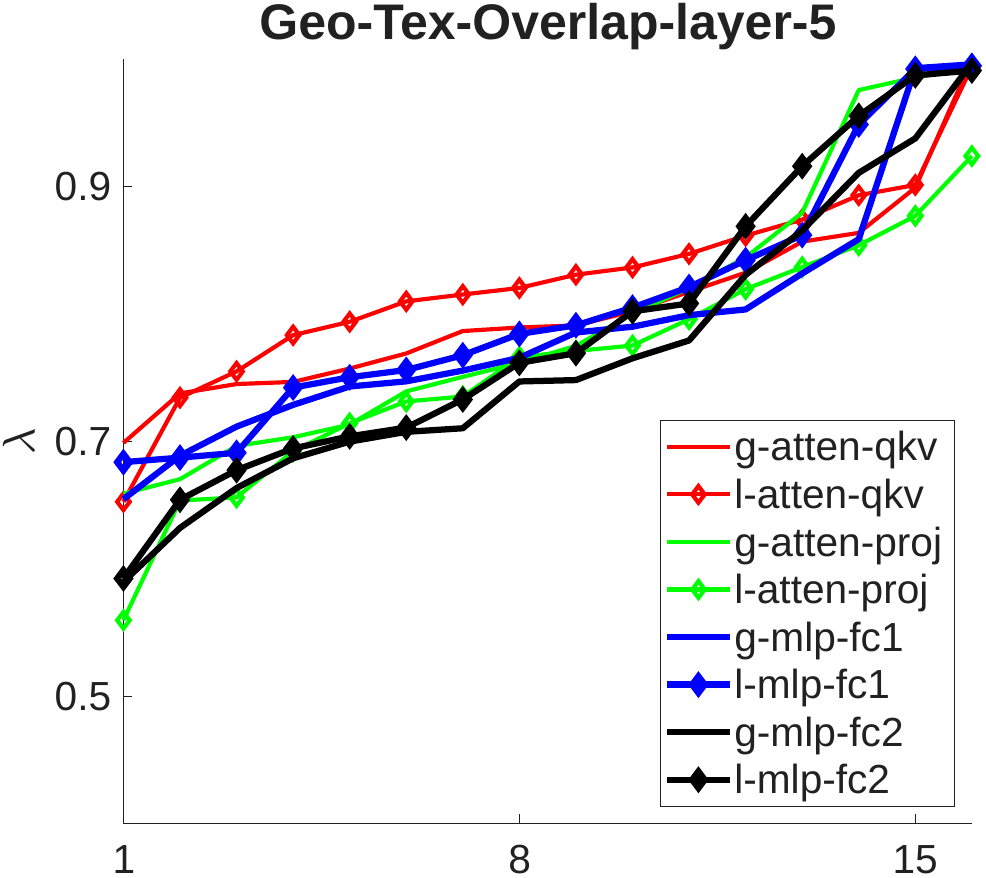}
& 
\includegraphics[width=0.21\textwidth]{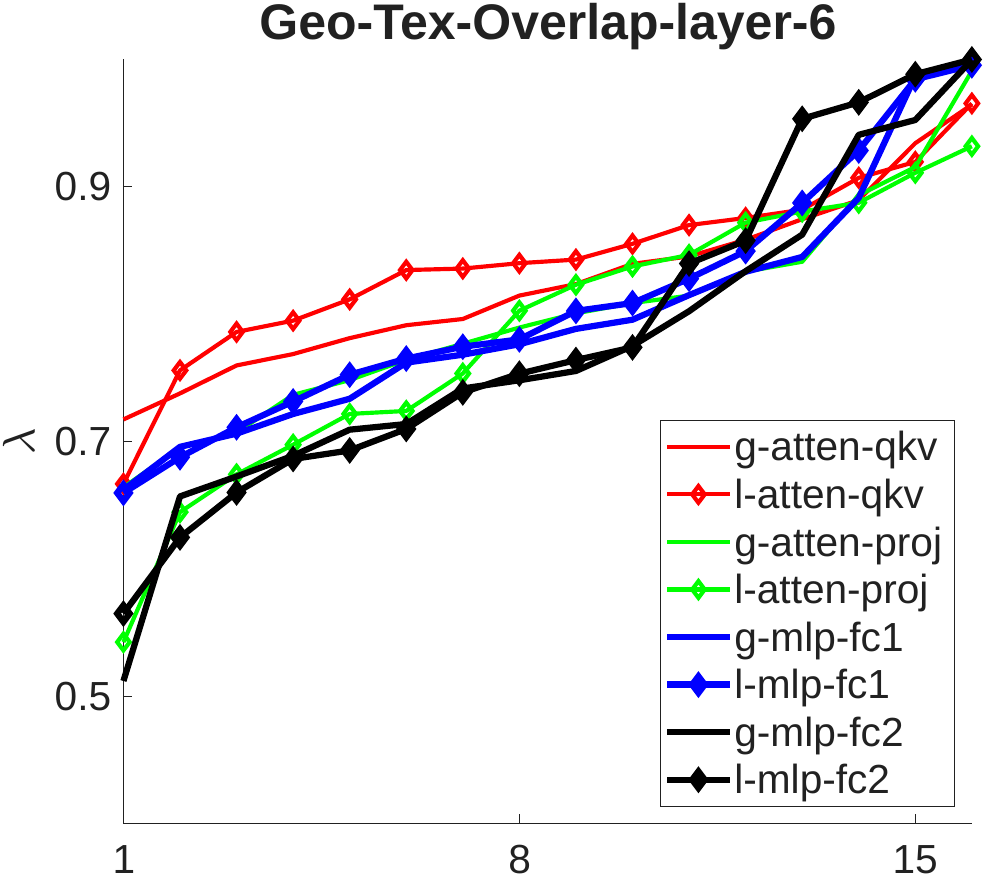}
&
\includegraphics[width=0.21\textwidth]{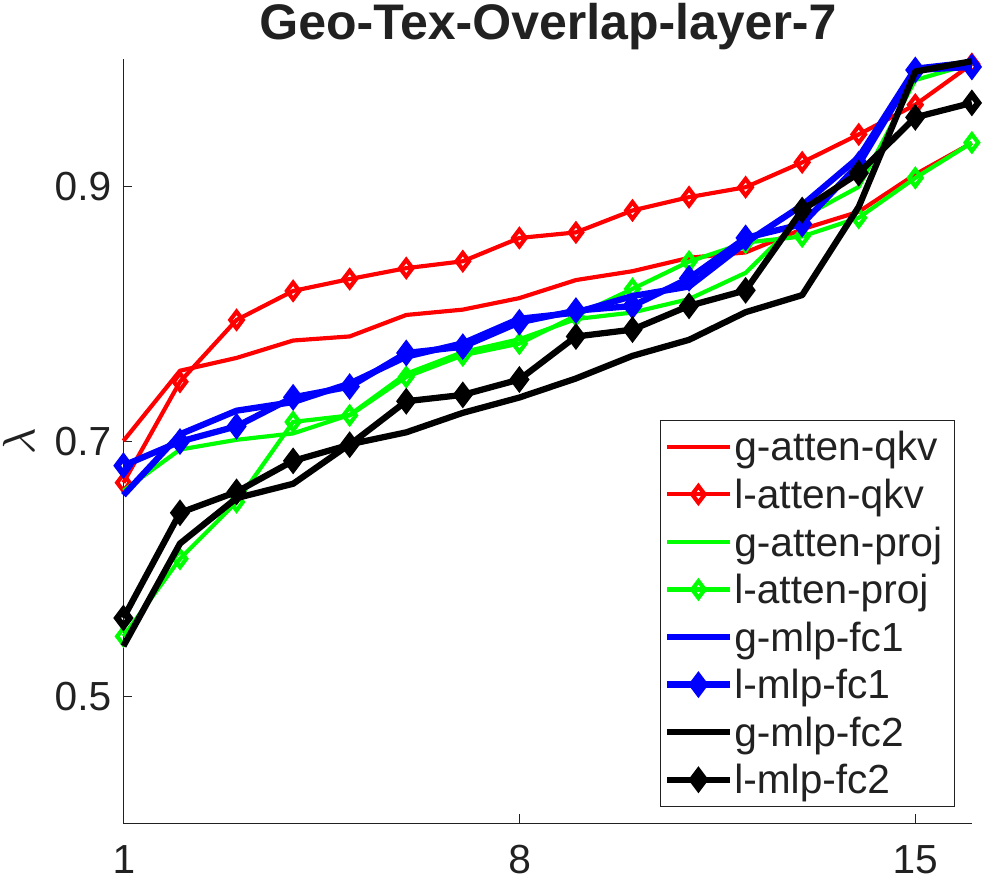}
&
\includegraphics[width=0.21\textwidth]{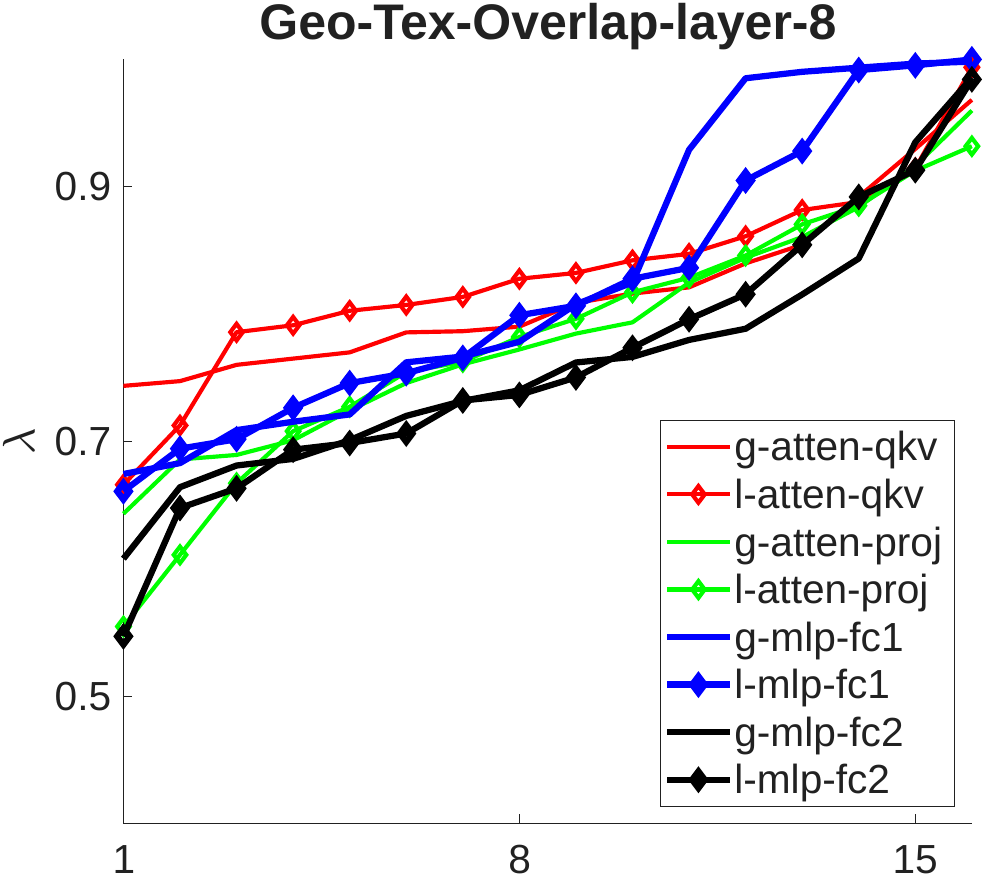}      
\\
\includegraphics[width=0.21\textwidth]{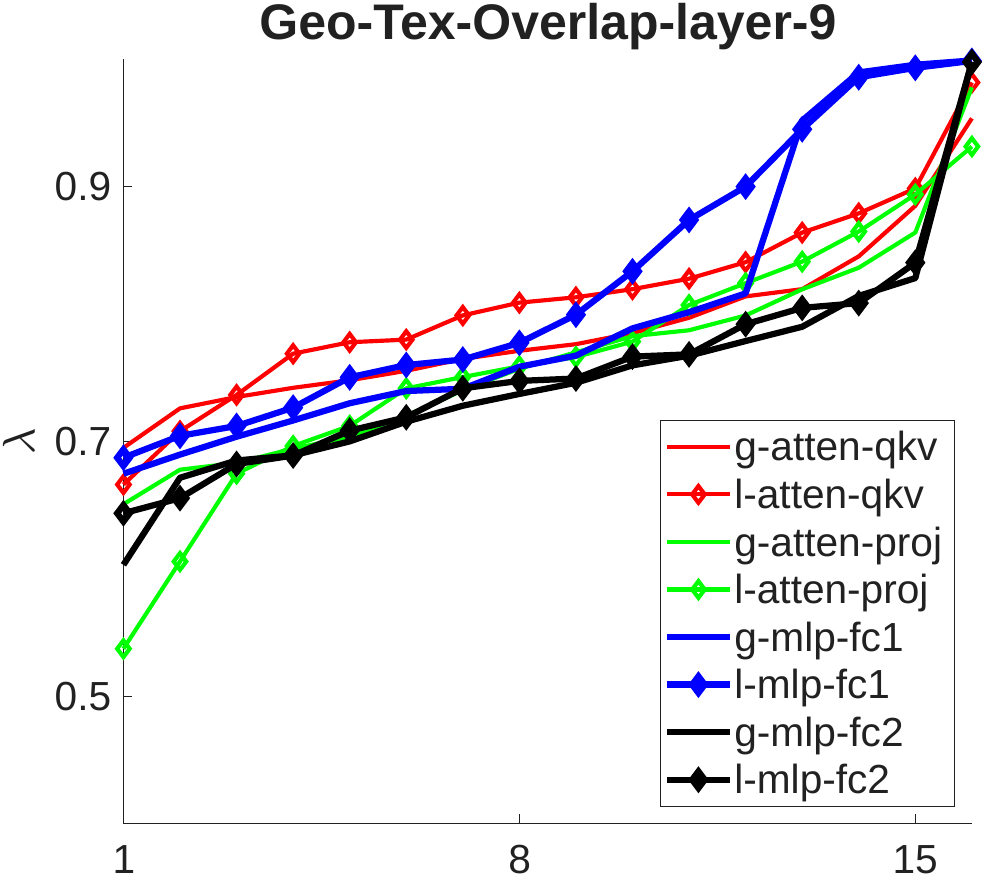}
& 
\includegraphics[width=0.21\textwidth]{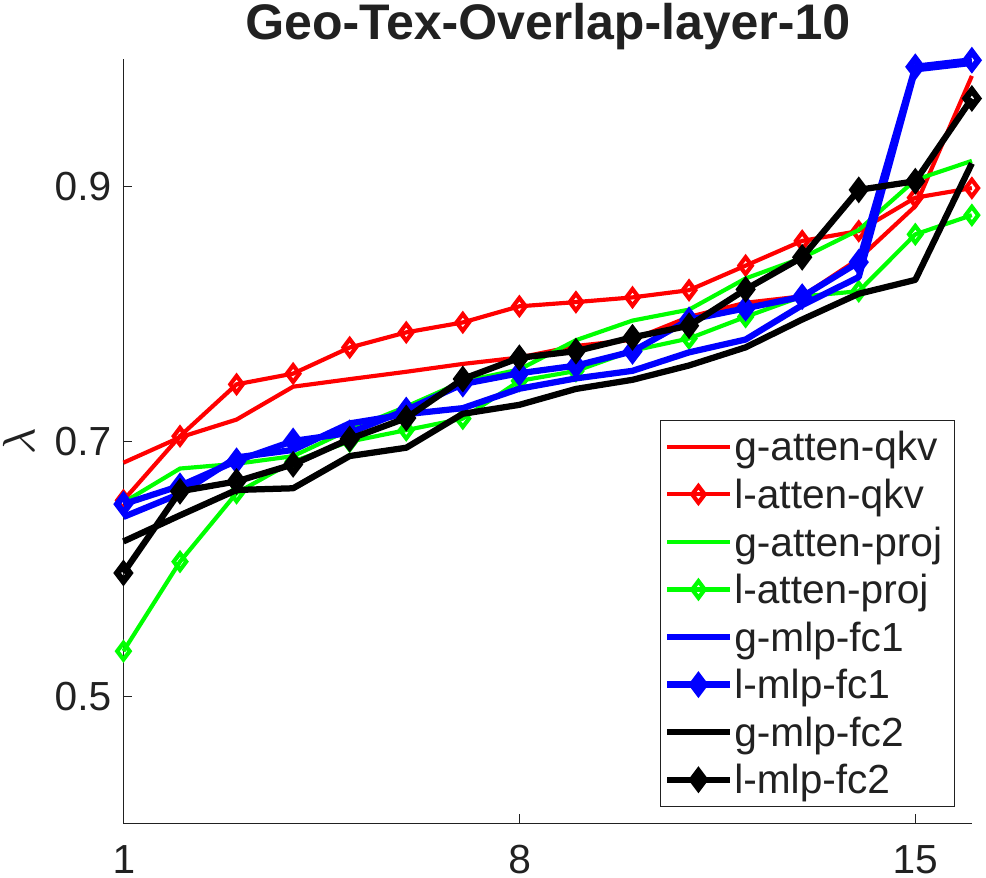}
&
\includegraphics[width=0.21\textwidth]{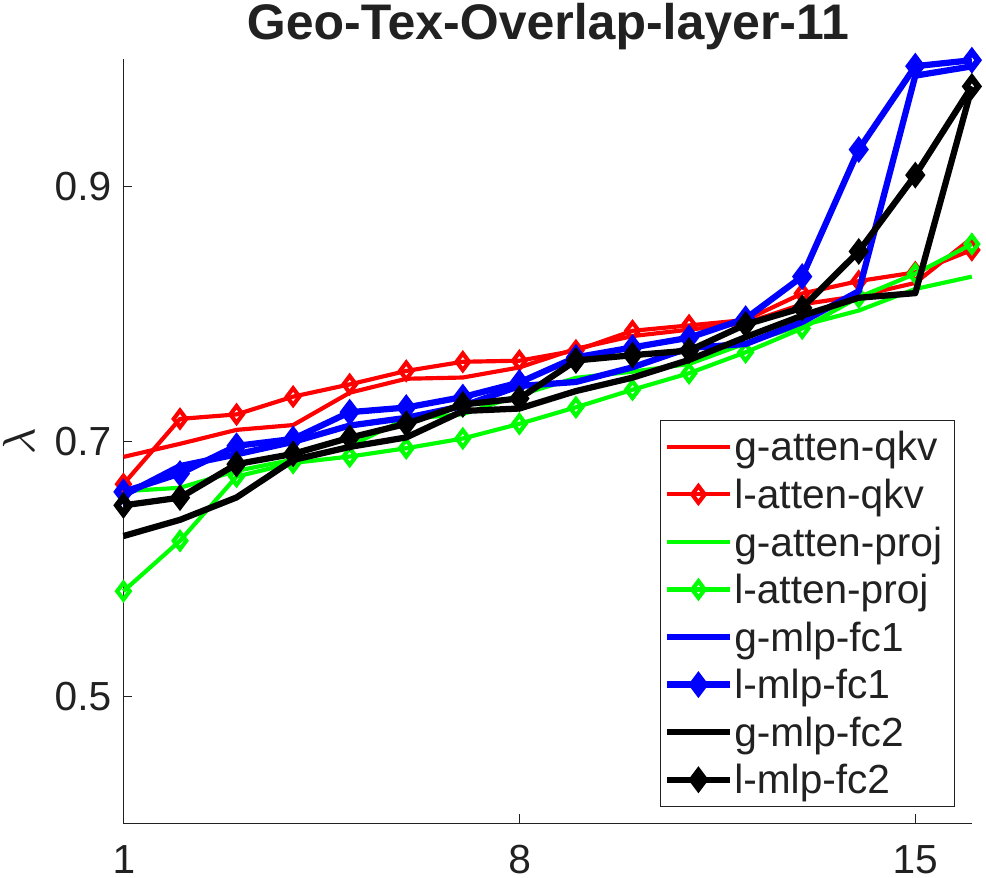}
&
\includegraphics[width=0.21\textwidth]{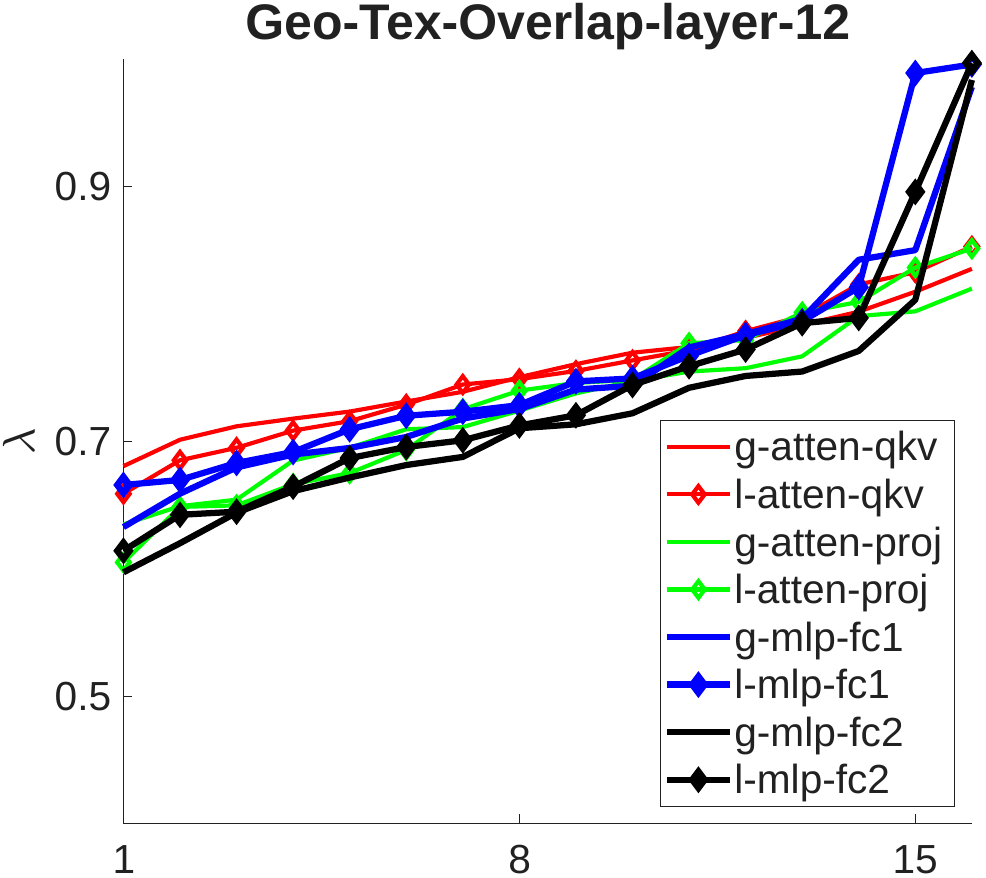}      
\\
\includegraphics[width=0.21\textwidth]{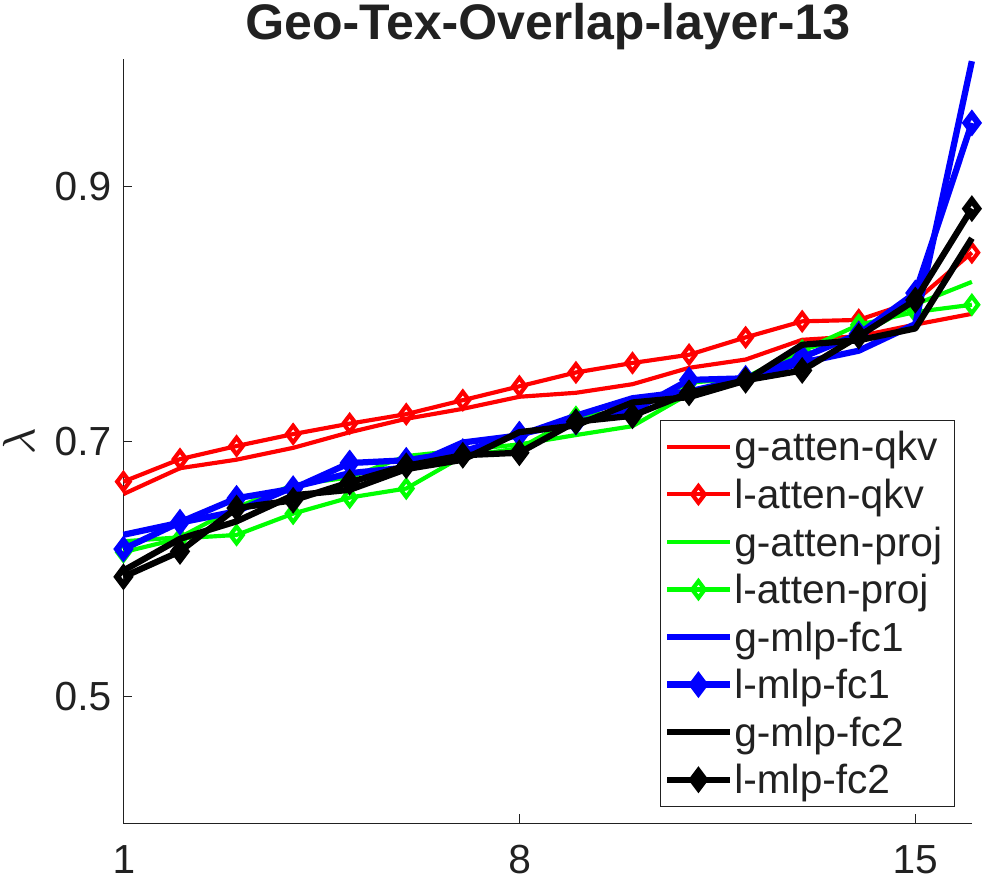}
& 
\includegraphics[width=0.21\textwidth]{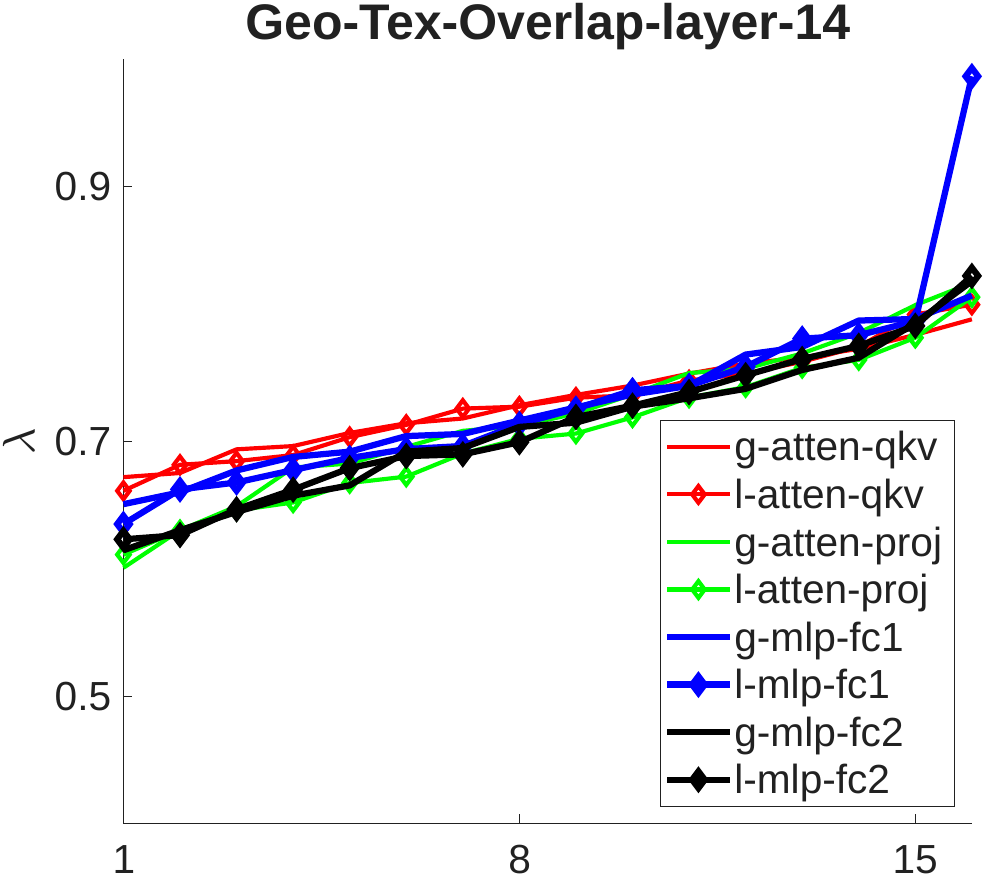}
&
\includegraphics[width=0.21\textwidth]{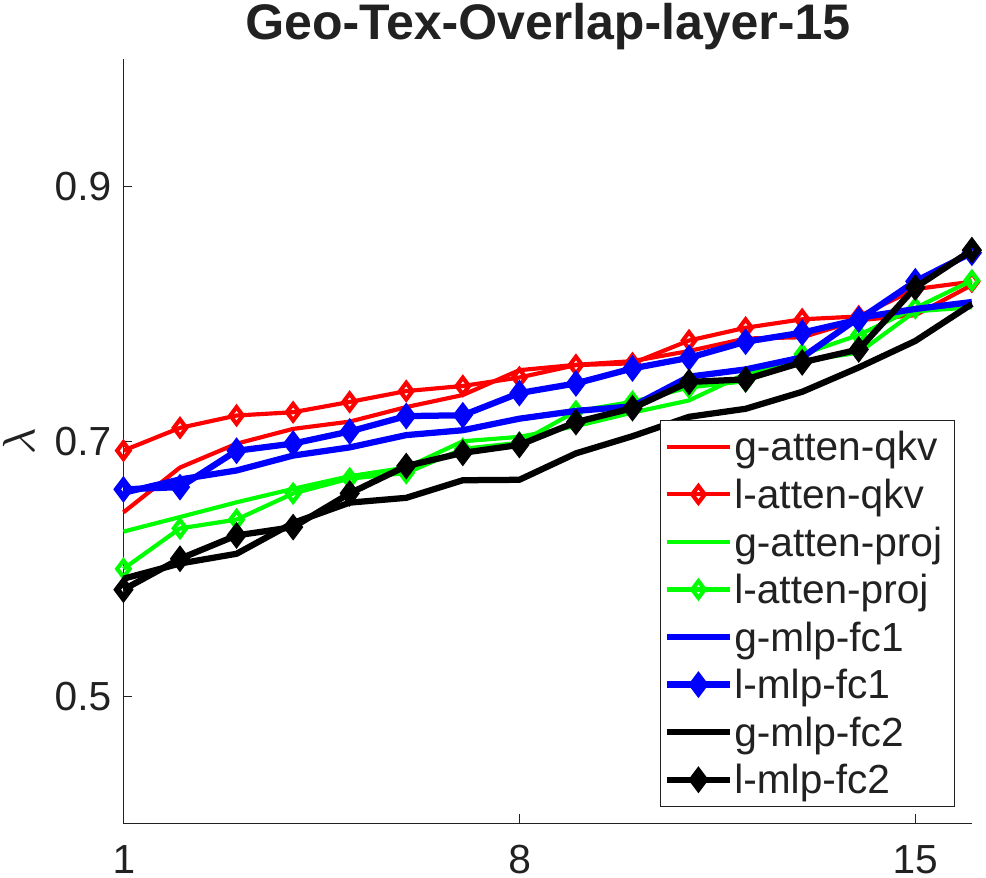}
&
\includegraphics[width=0.21\textwidth]{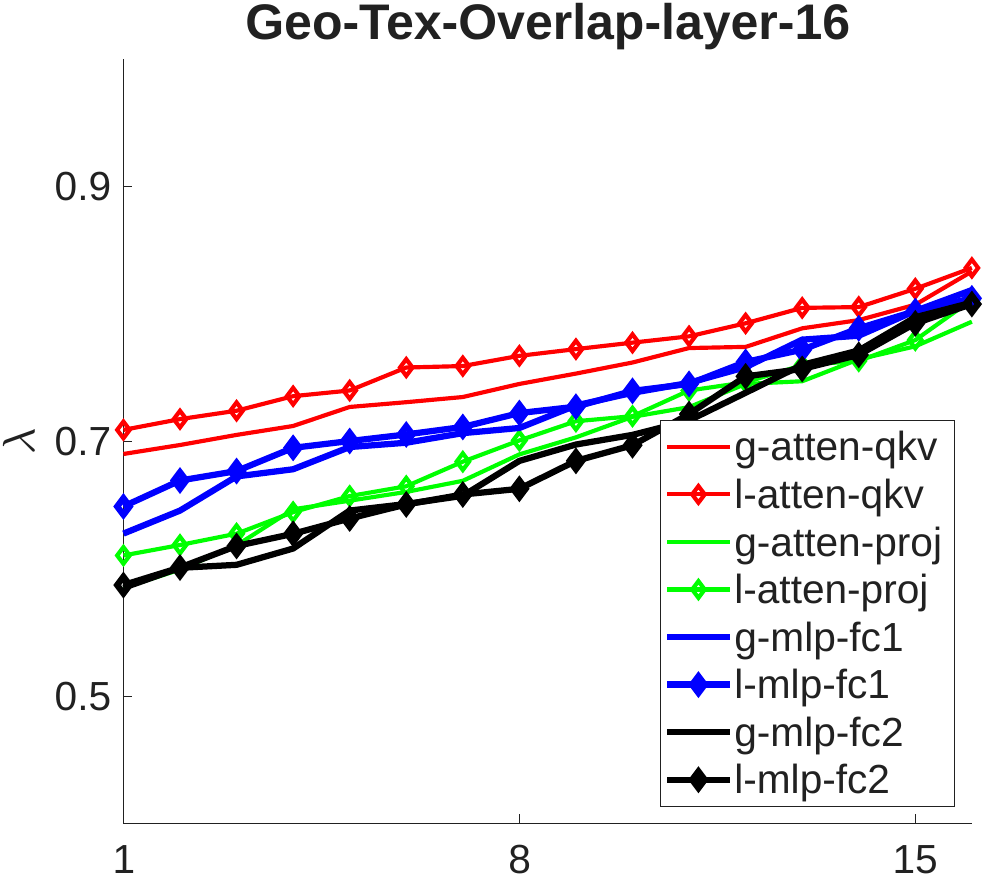}      
\\
\includegraphics[width=0.21\textwidth]{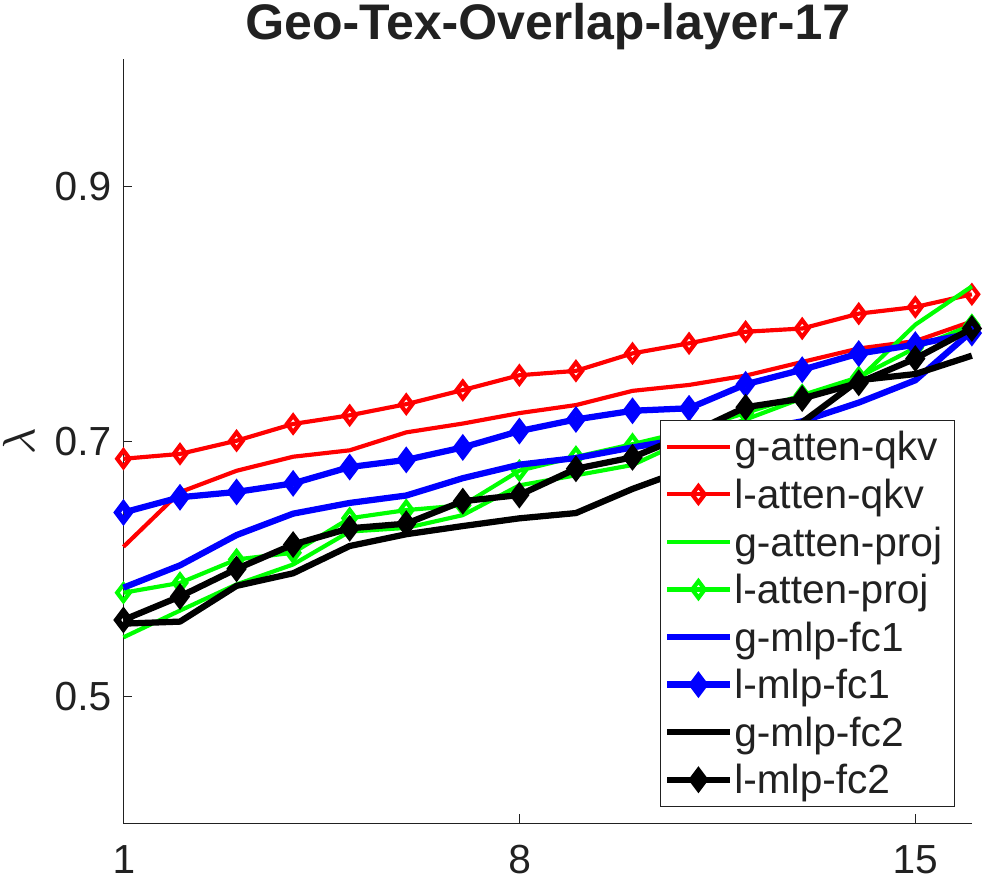}
& 
\includegraphics[width=0.21\textwidth]{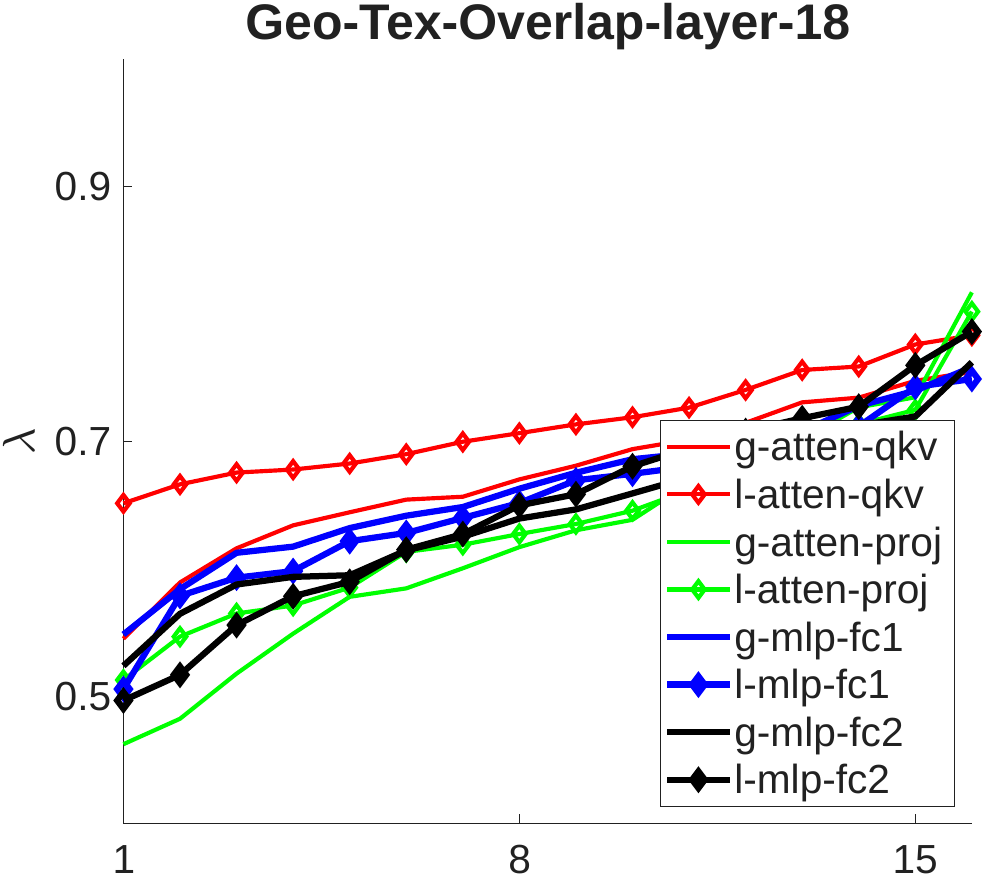}
&
\includegraphics[width=0.21\textwidth]{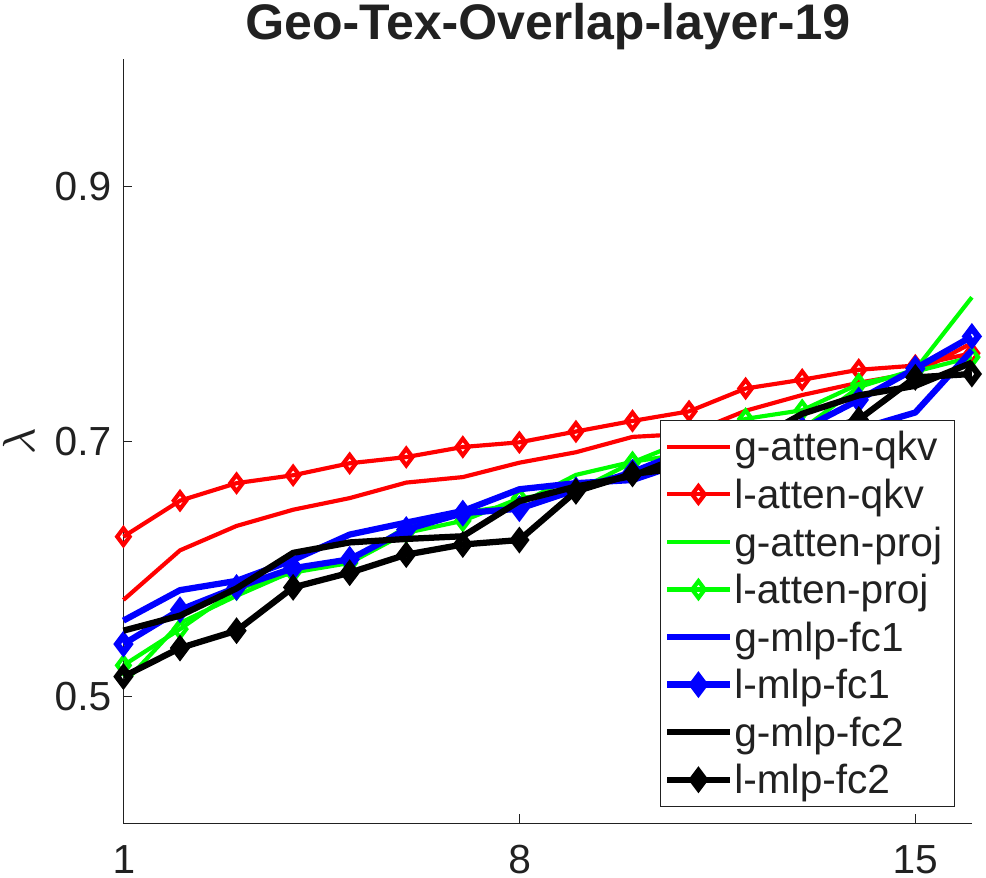}
&
\includegraphics[width=0.21\textwidth]{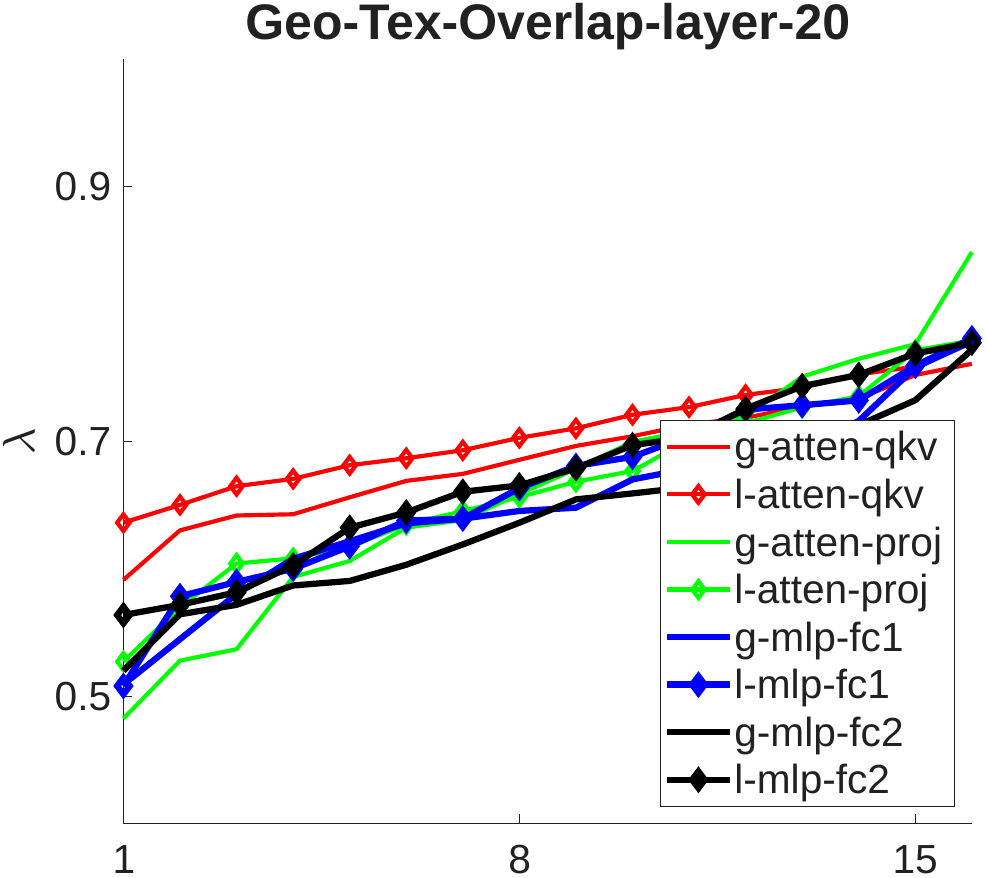}      
\\
\includegraphics[width=0.21\textwidth]{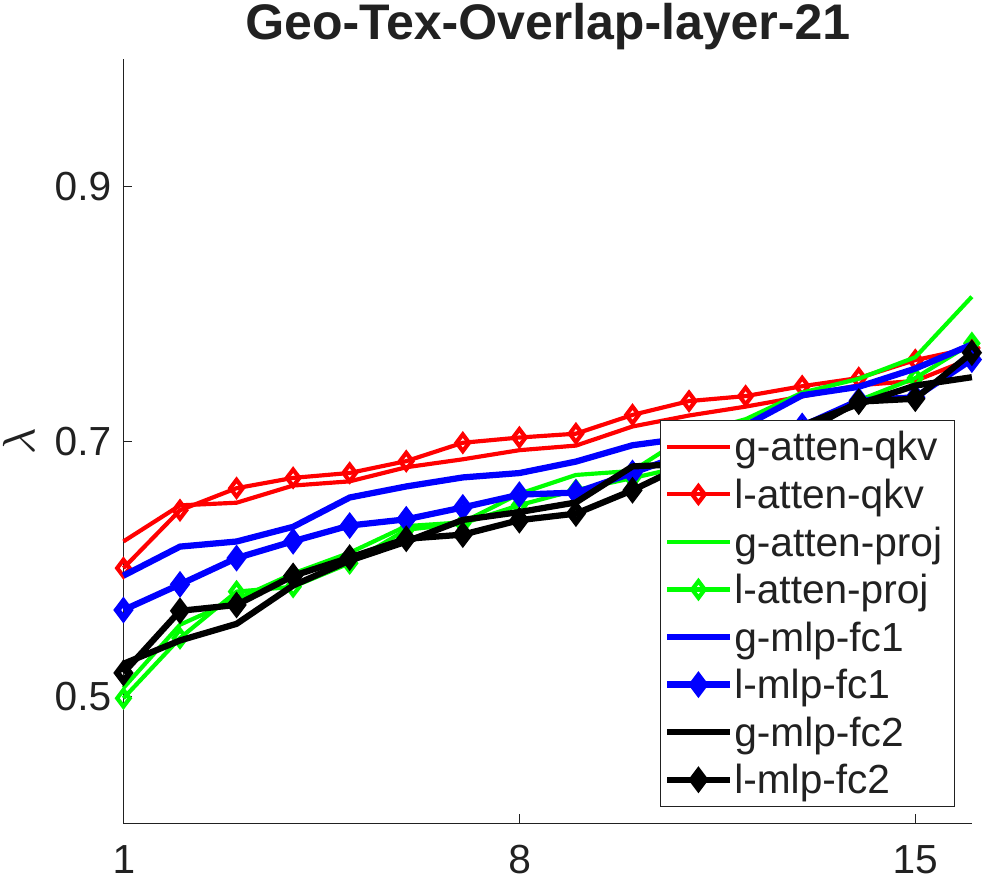}
& 
\includegraphics[width=0.21\textwidth]{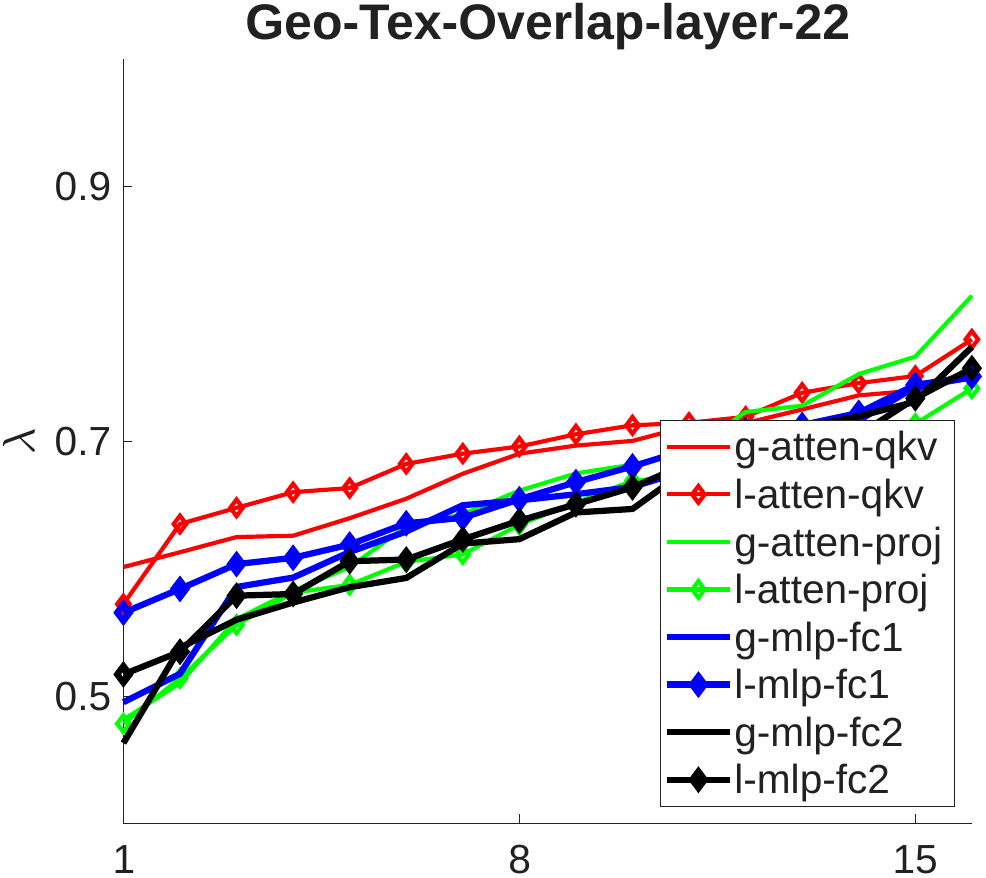}
&
\includegraphics[width=0.21\textwidth]{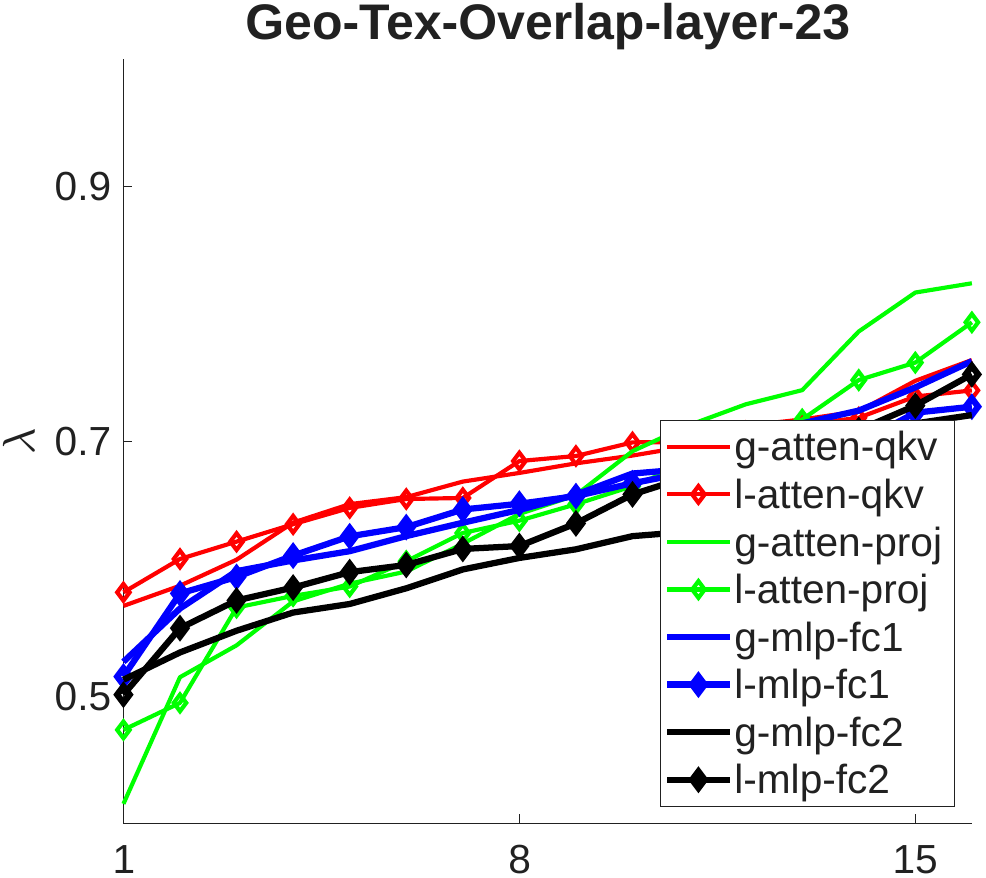}
&
\includegraphics[width=0.21\textwidth]{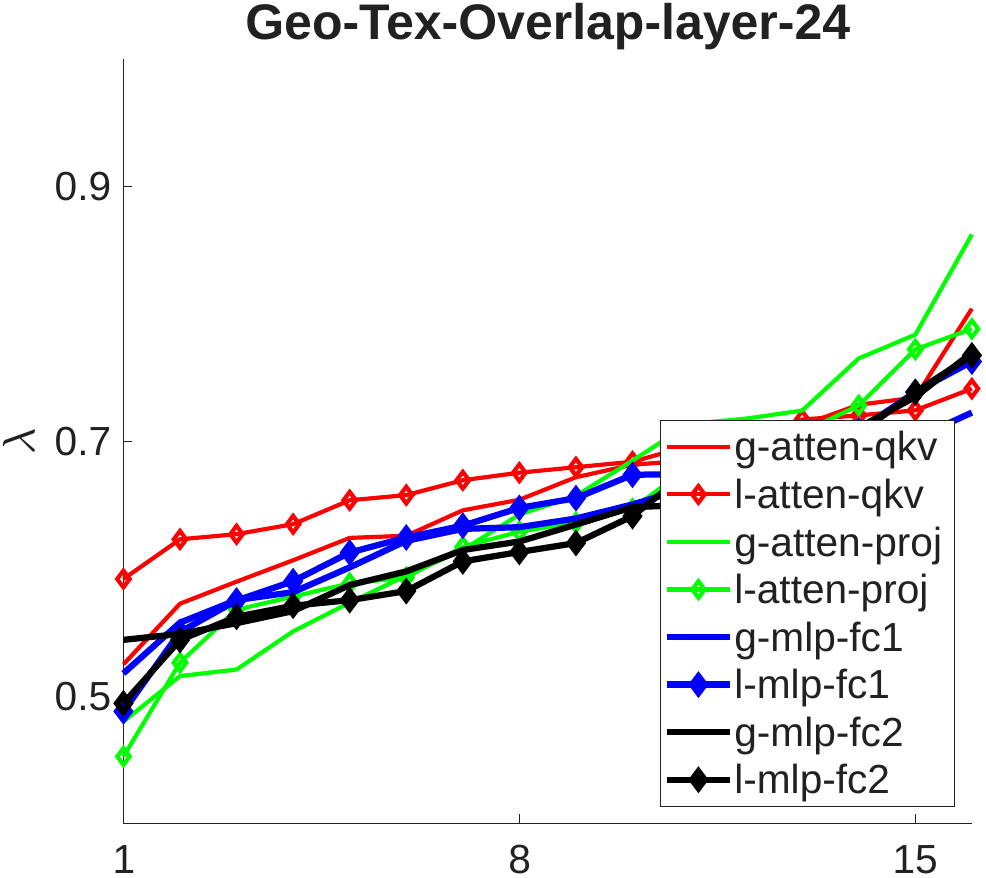}      

\end{tabular}

\captionof{figure}{The overlap ratio between subspaces that correspond to variations in geometry and texture. }

% \label{Fig:Subspace:Magnitudes}    
\vspace{-3em}
\end{table*}

\clearpage

\subsection{Geometry vs Camera}

\begin{table*}[bp]
\centering
\setlength\tabcolsep{6pt}
\begin{tabular}{cccc}
\includegraphics[width=0.21\textwidth]{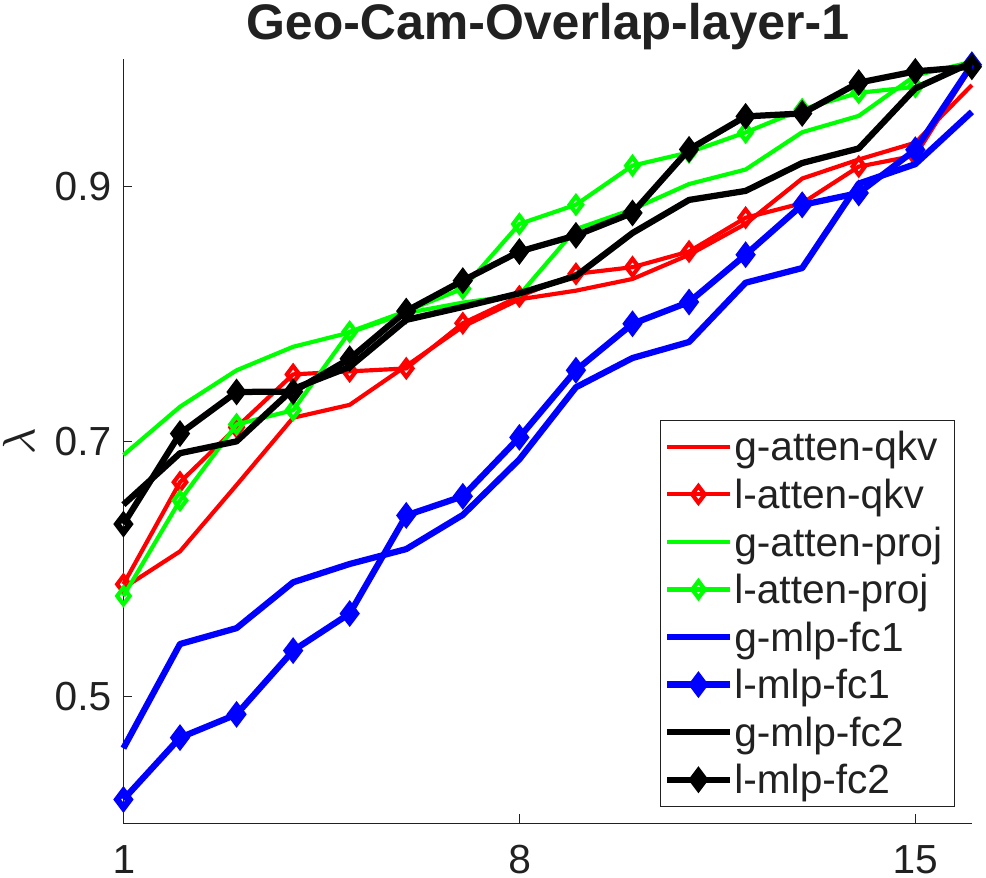}
& 
\includegraphics[width=0.21\textwidth]{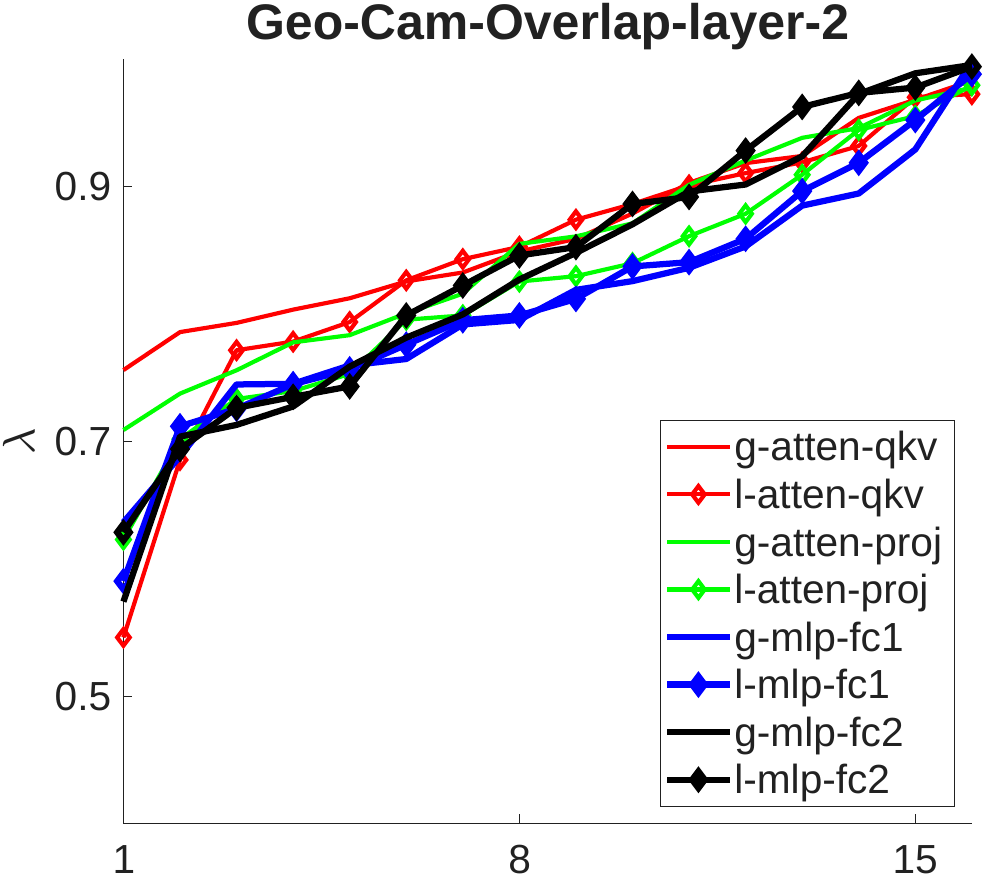}
&
\includegraphics[width=0.21\textwidth]{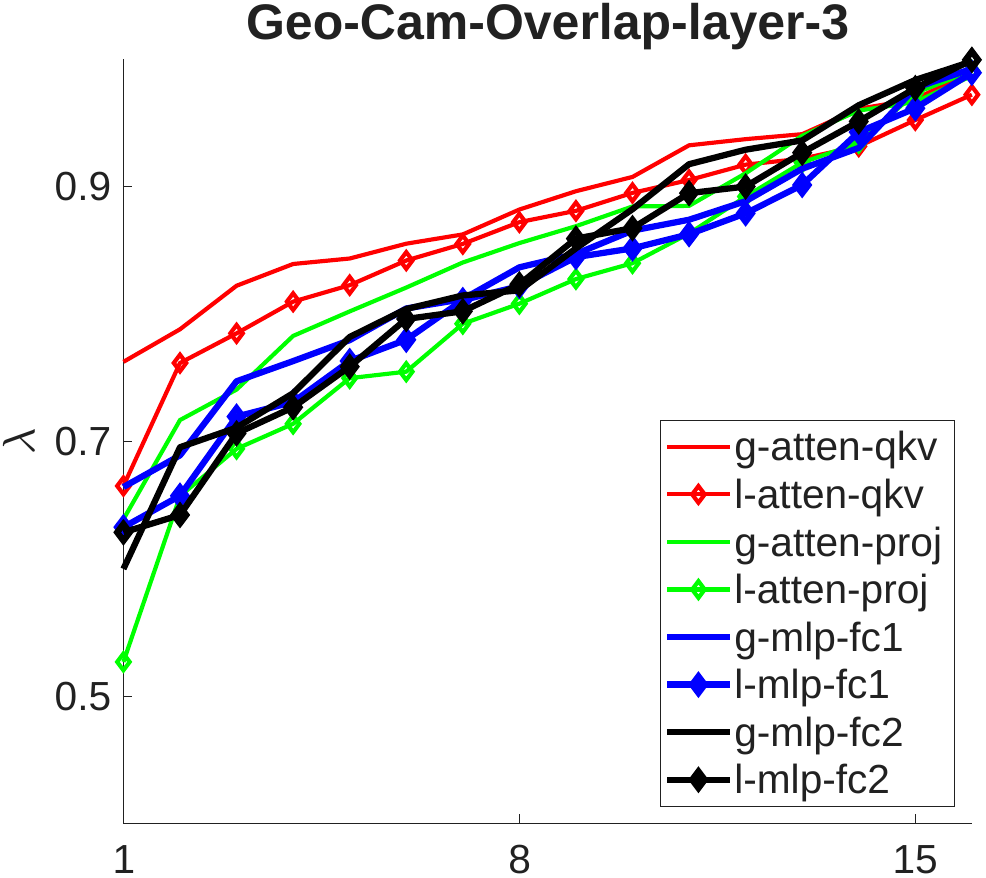}
&
\includegraphics[width=0.21\textwidth]{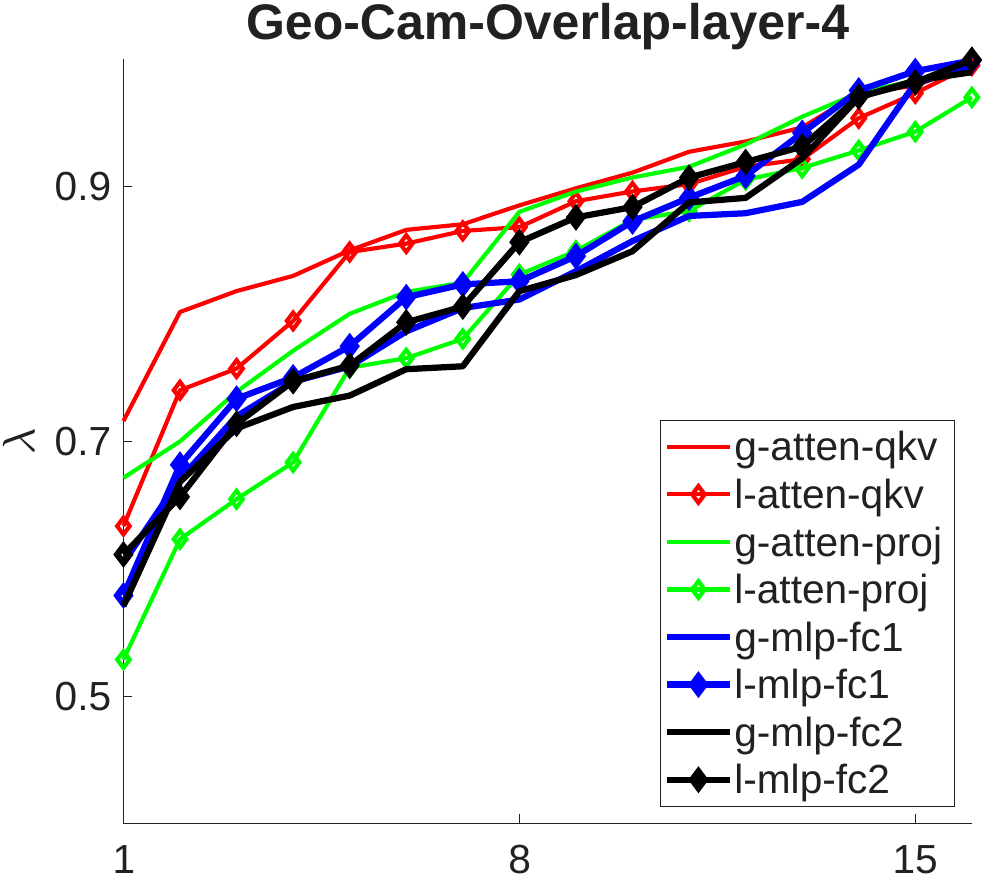}      
\\
\includegraphics[width=0.21\textwidth]{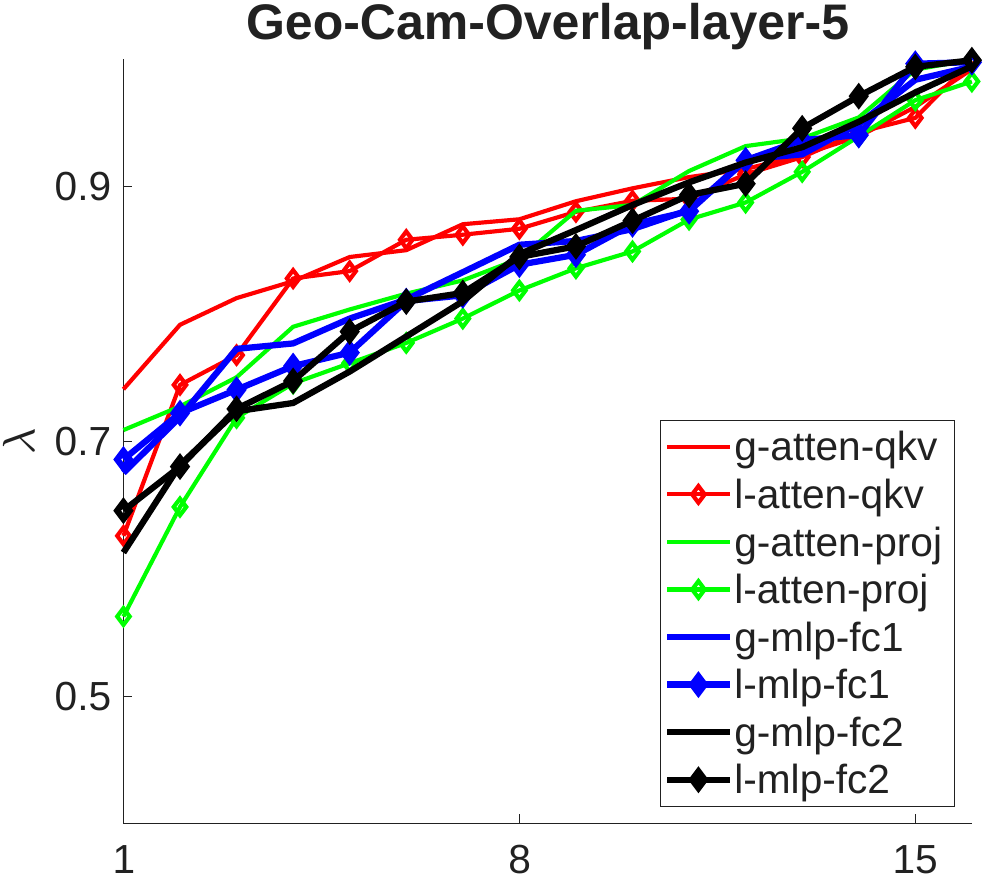}
& 
\includegraphics[width=0.21\textwidth]{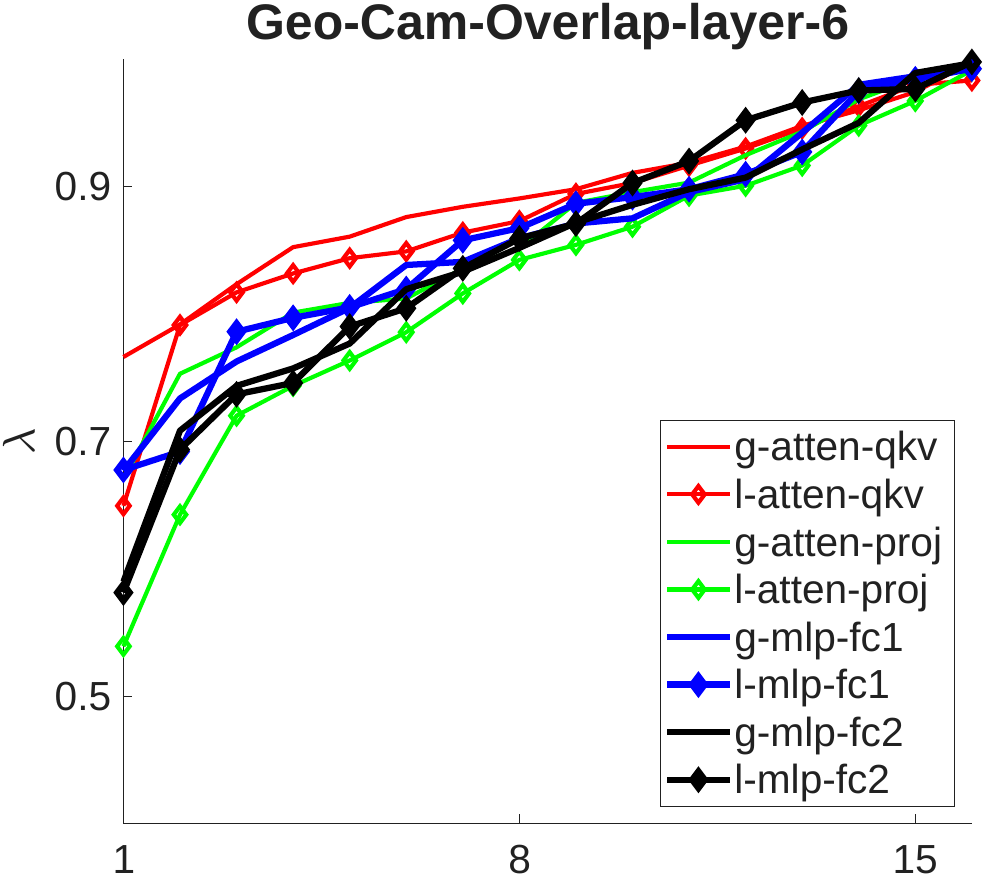}
&
\includegraphics[width=0.21\textwidth]{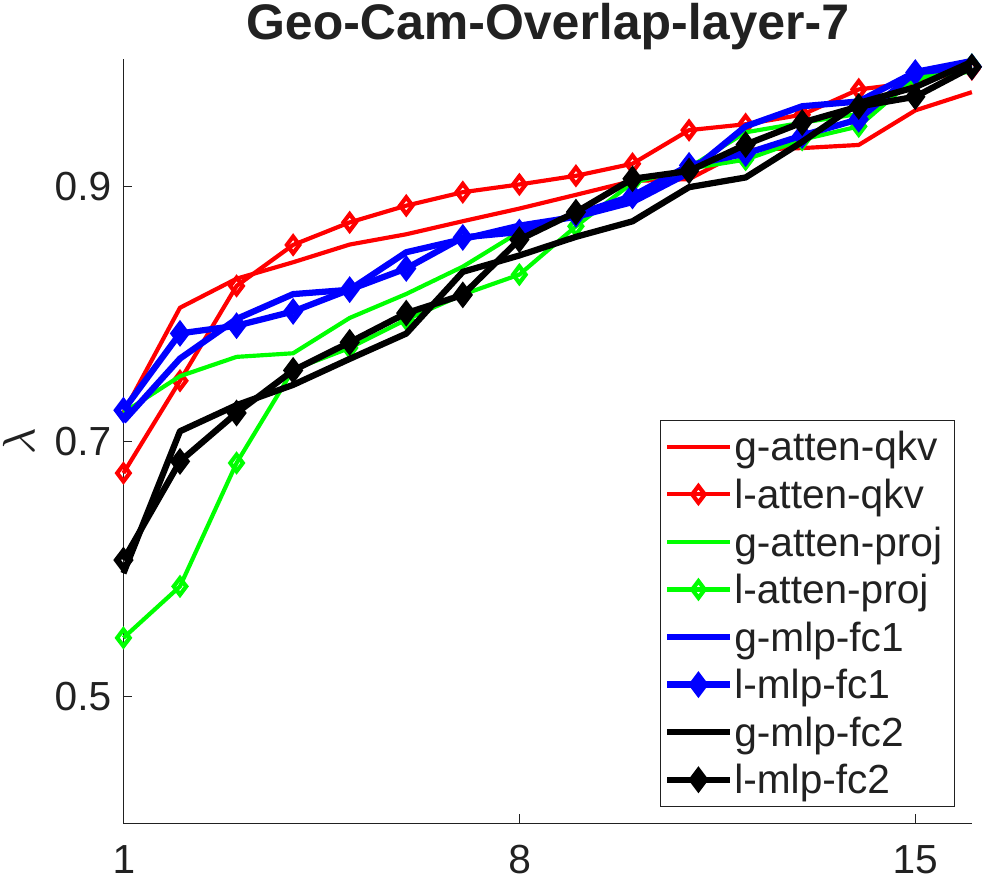}
&
\includegraphics[width=0.21\textwidth]{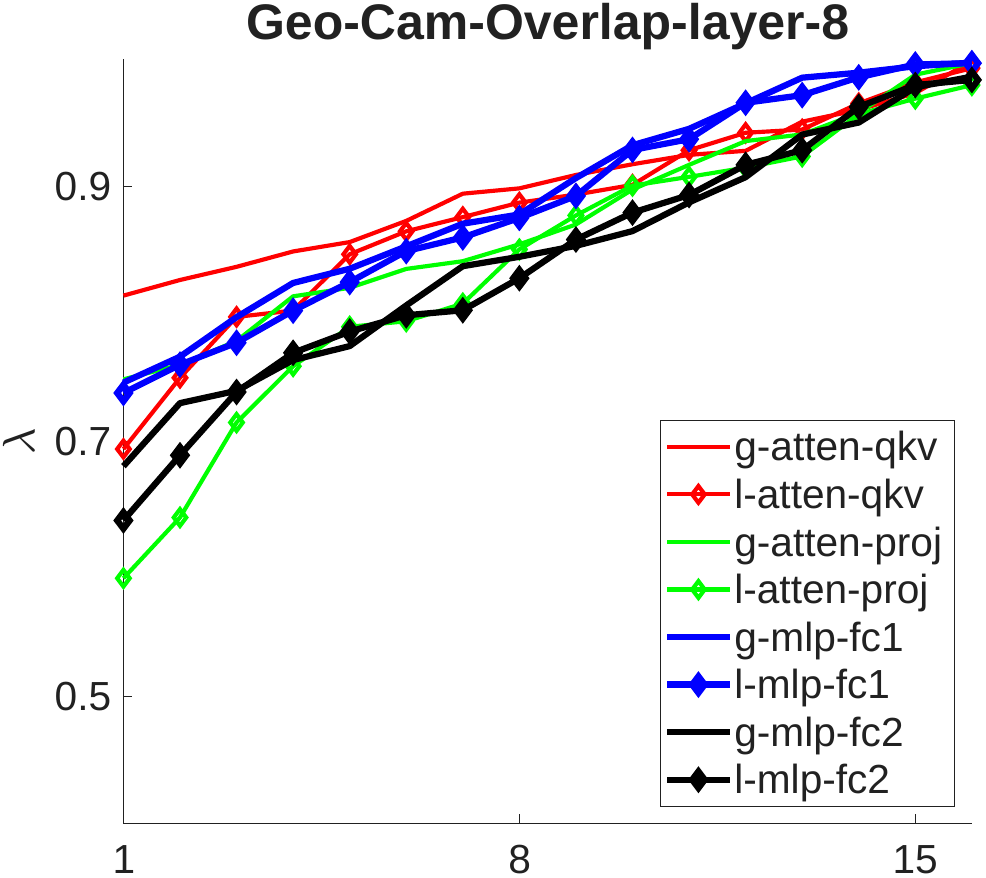}      
\\
\includegraphics[width=0.21\textwidth]{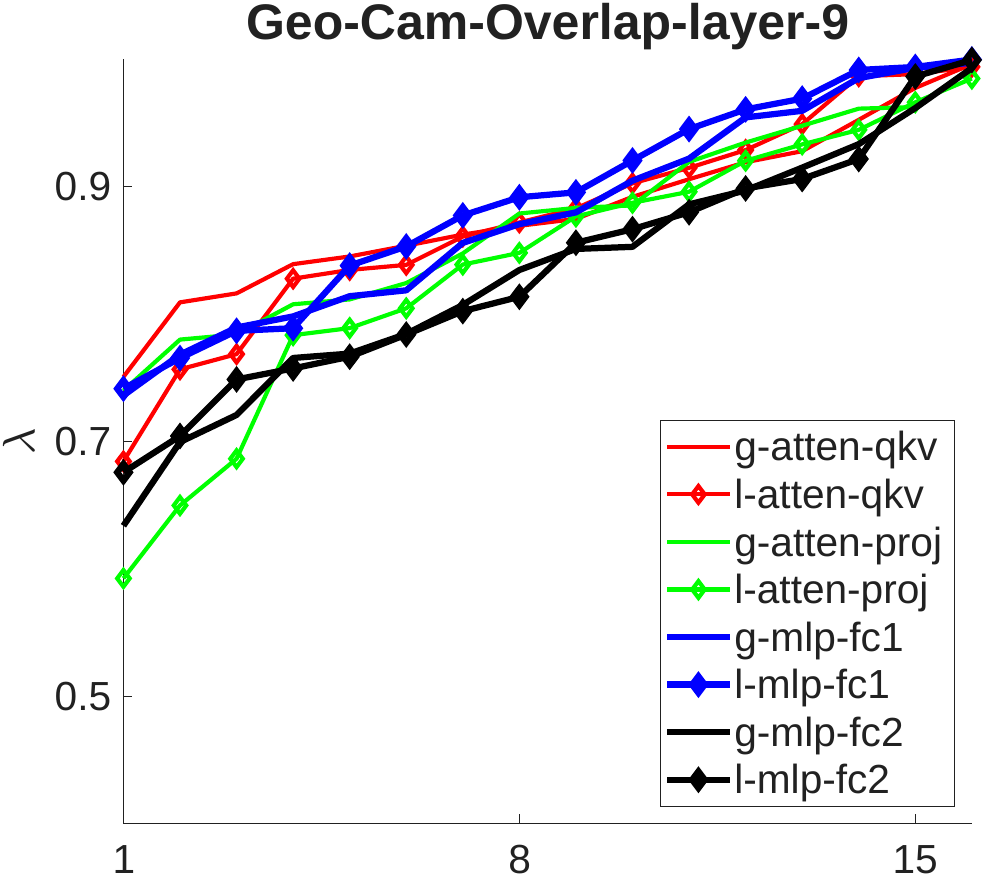}
& 
\includegraphics[width=0.21\textwidth]{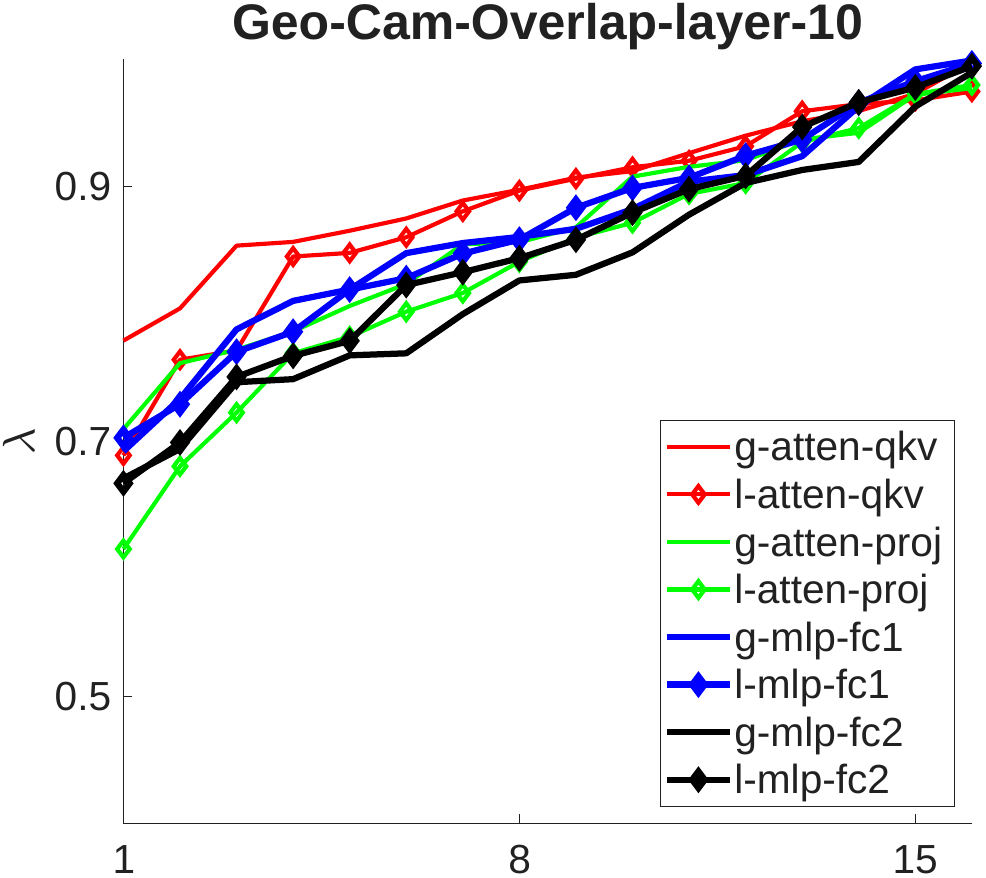}
&
\includegraphics[width=0.21\textwidth]{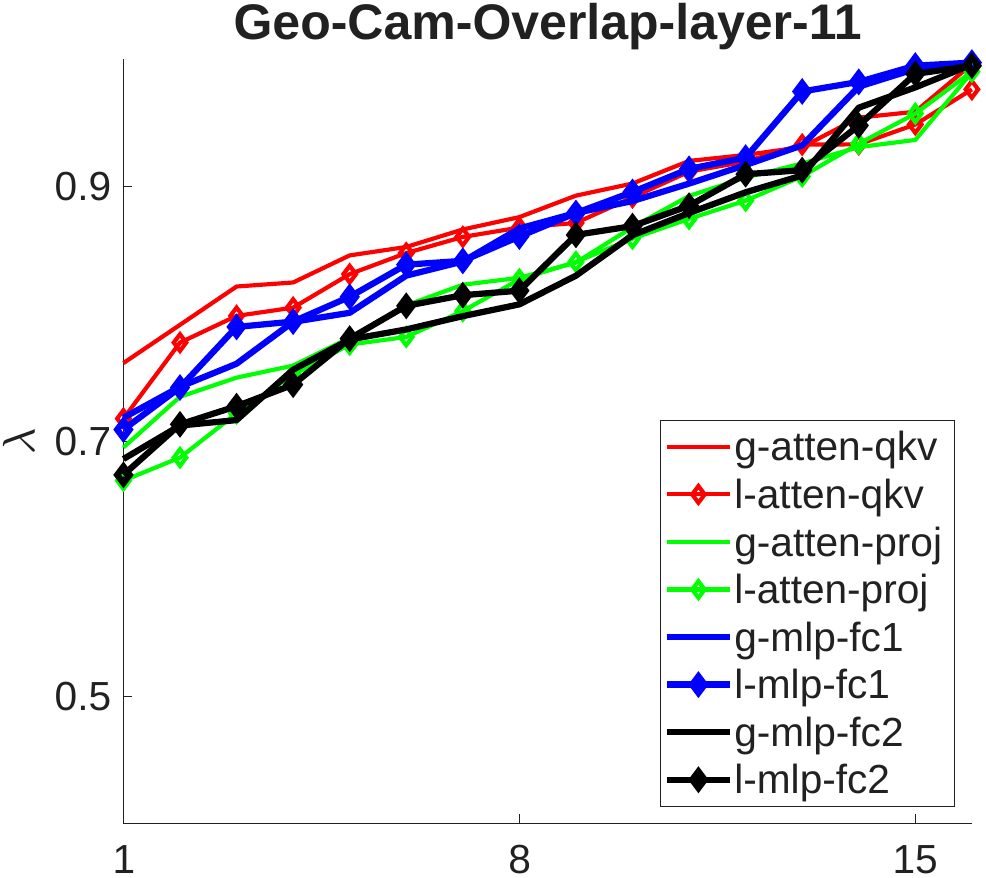}
&
\includegraphics[width=0.21\textwidth]{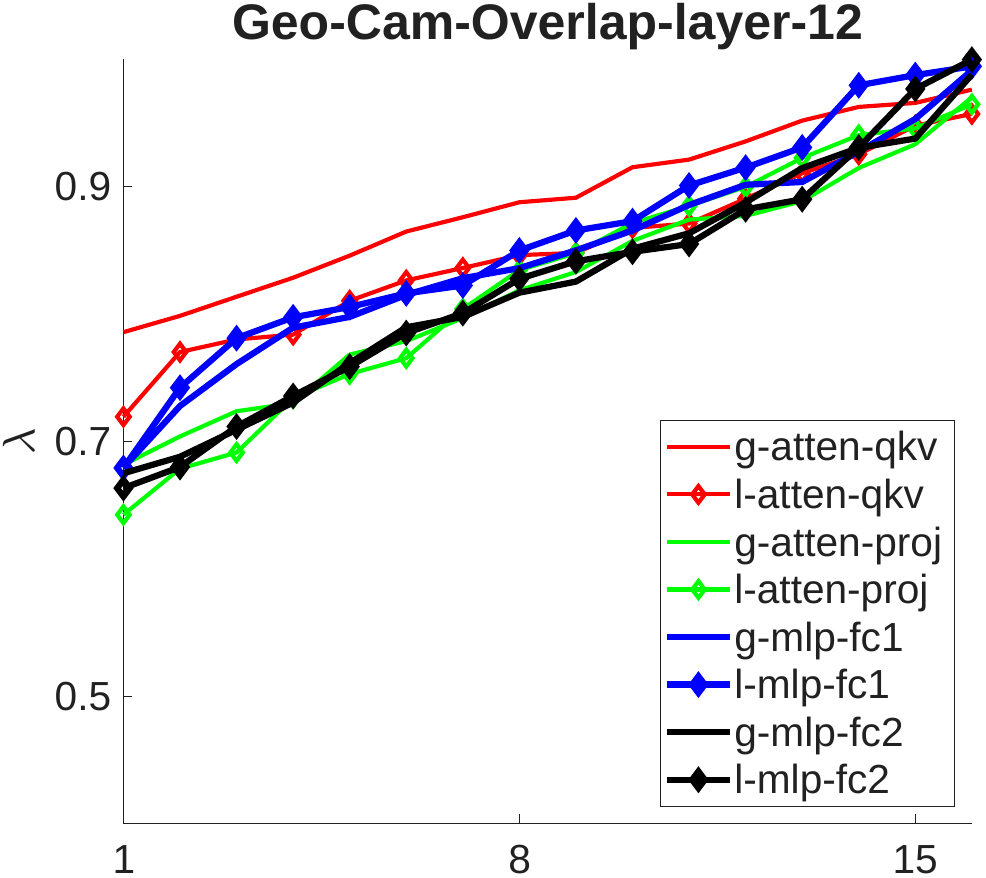}      
\\
\includegraphics[width=0.21\textwidth]{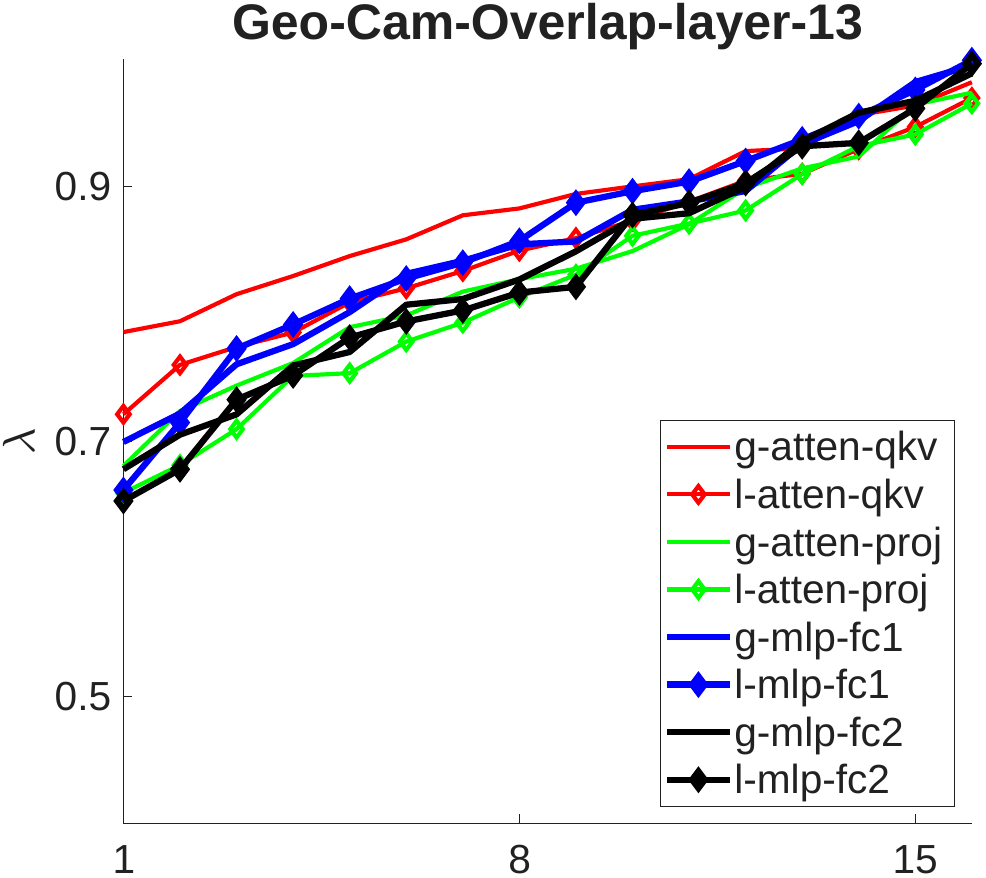}
& 
\includegraphics[width=0.21\textwidth]{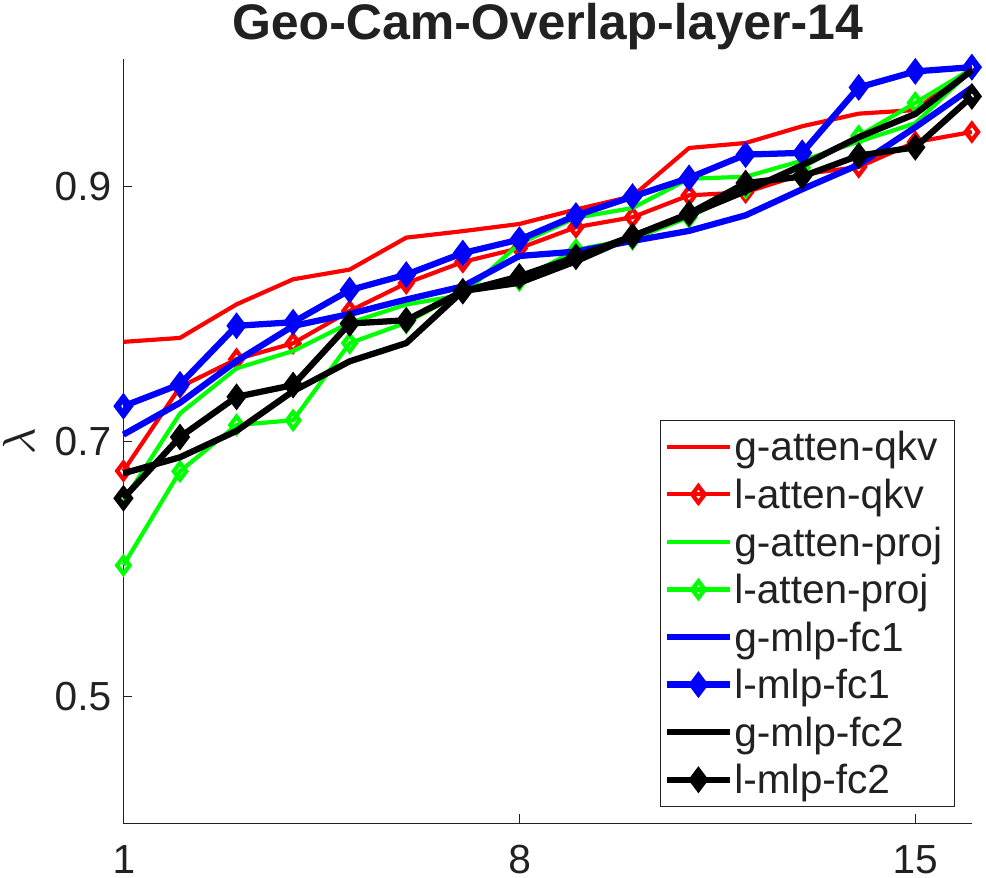}
&
\includegraphics[width=0.21\textwidth]{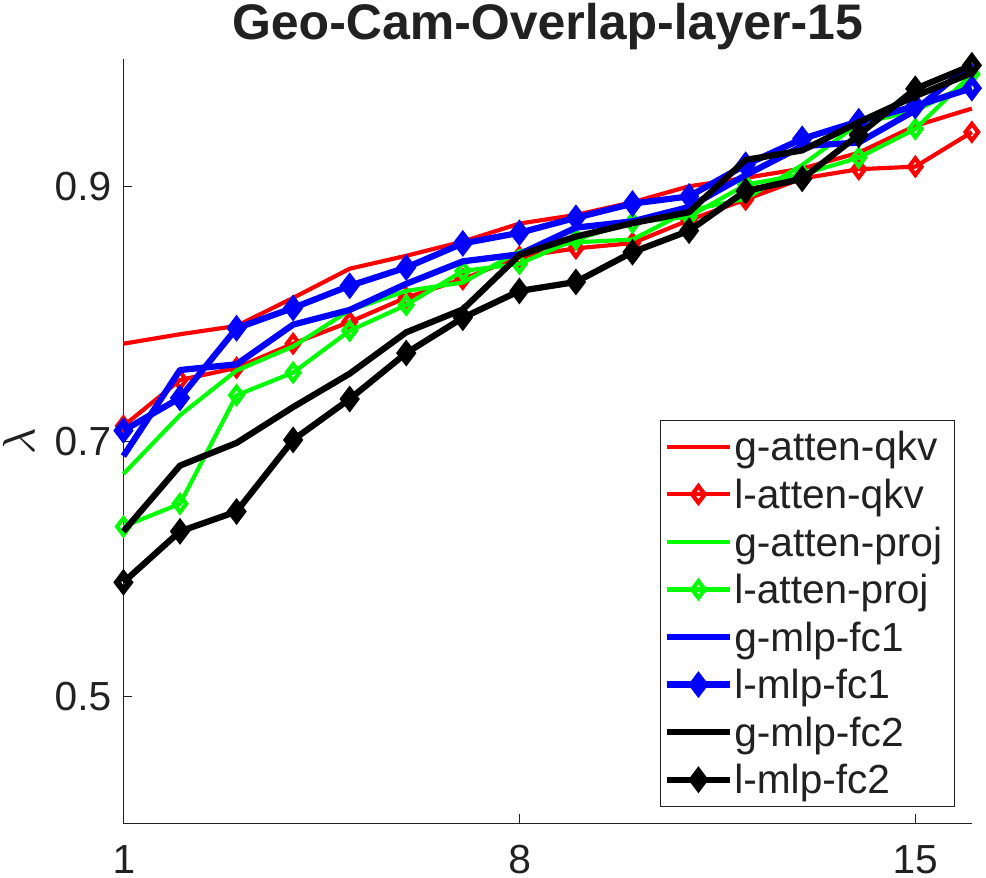}
&
\includegraphics[width=0.21\textwidth]{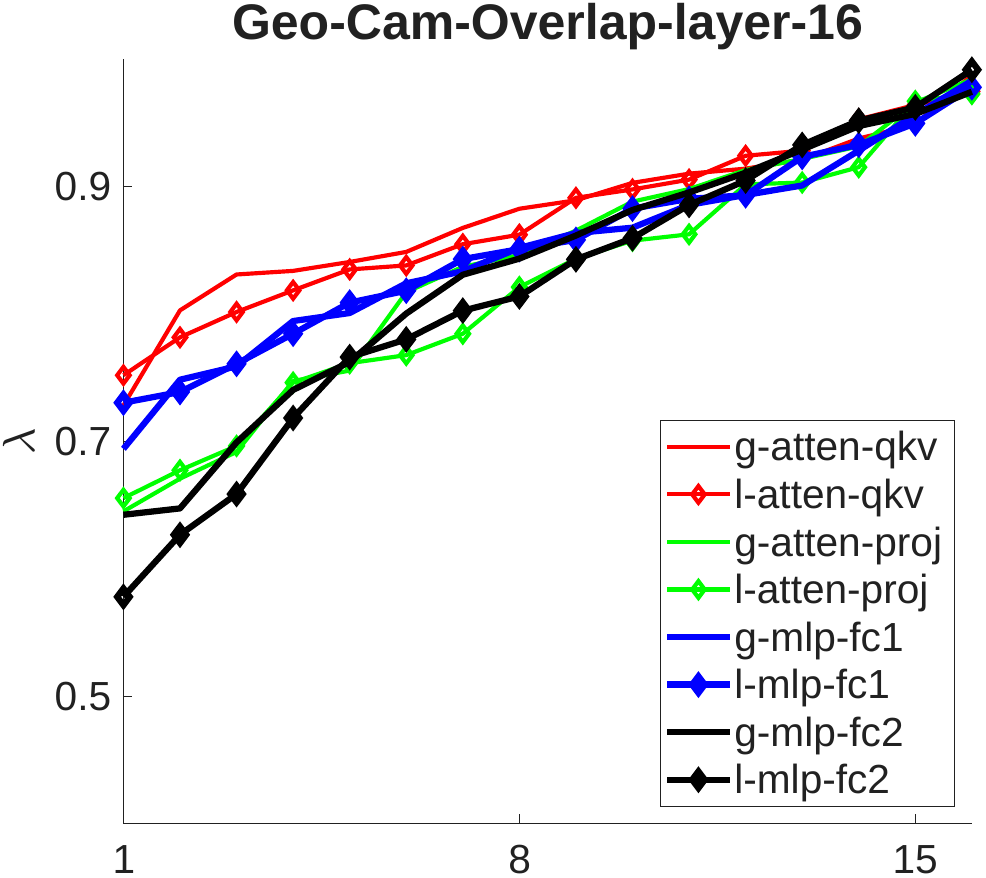}      
\\
\includegraphics[width=0.21\textwidth]{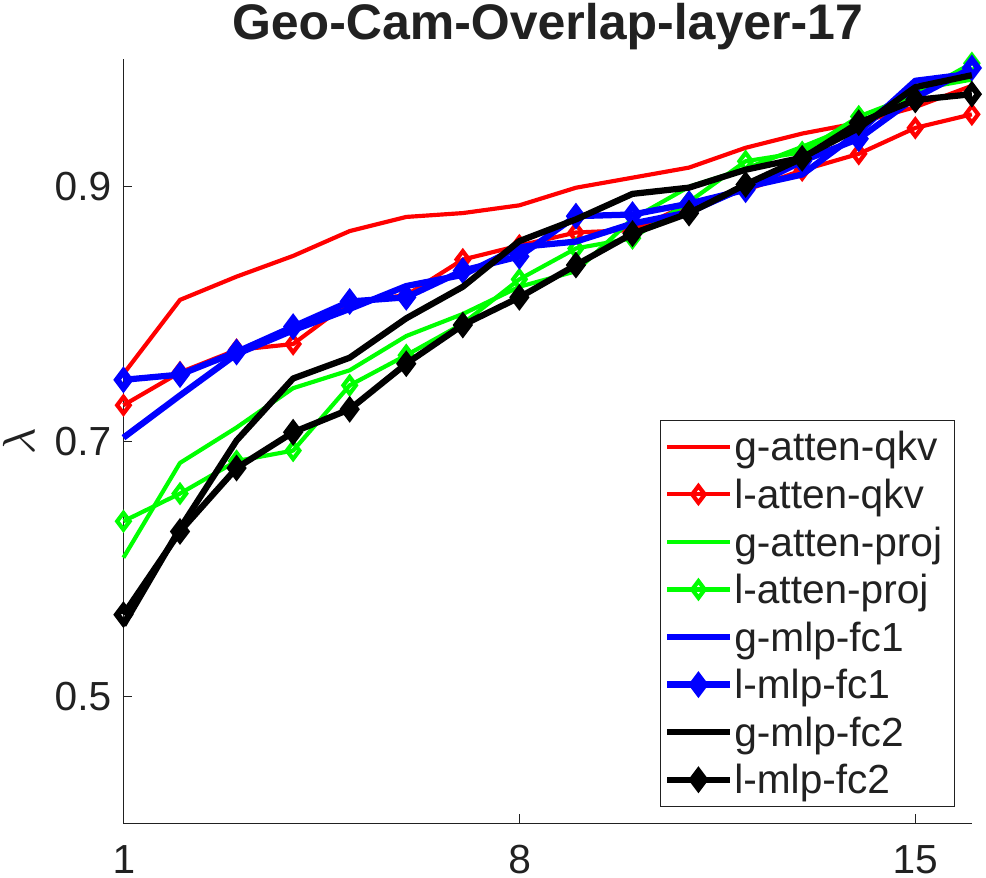}
& 
\includegraphics[width=0.21\textwidth]{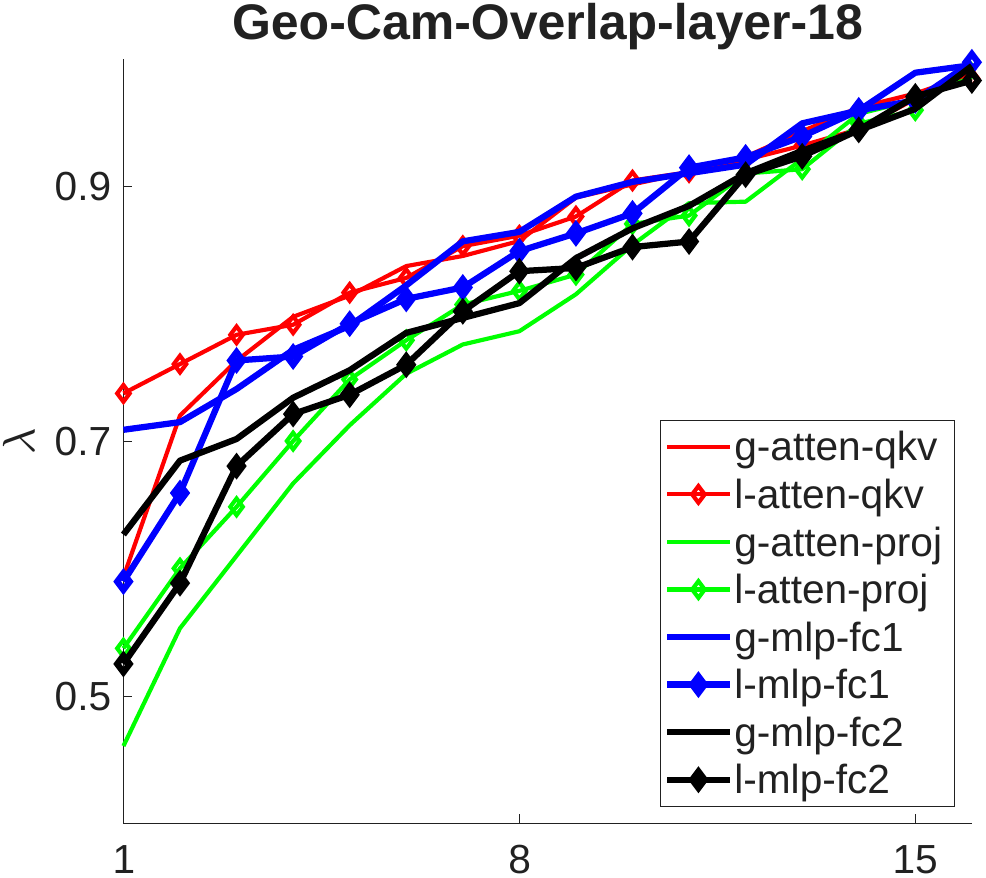}
&
\includegraphics[width=0.21\textwidth]{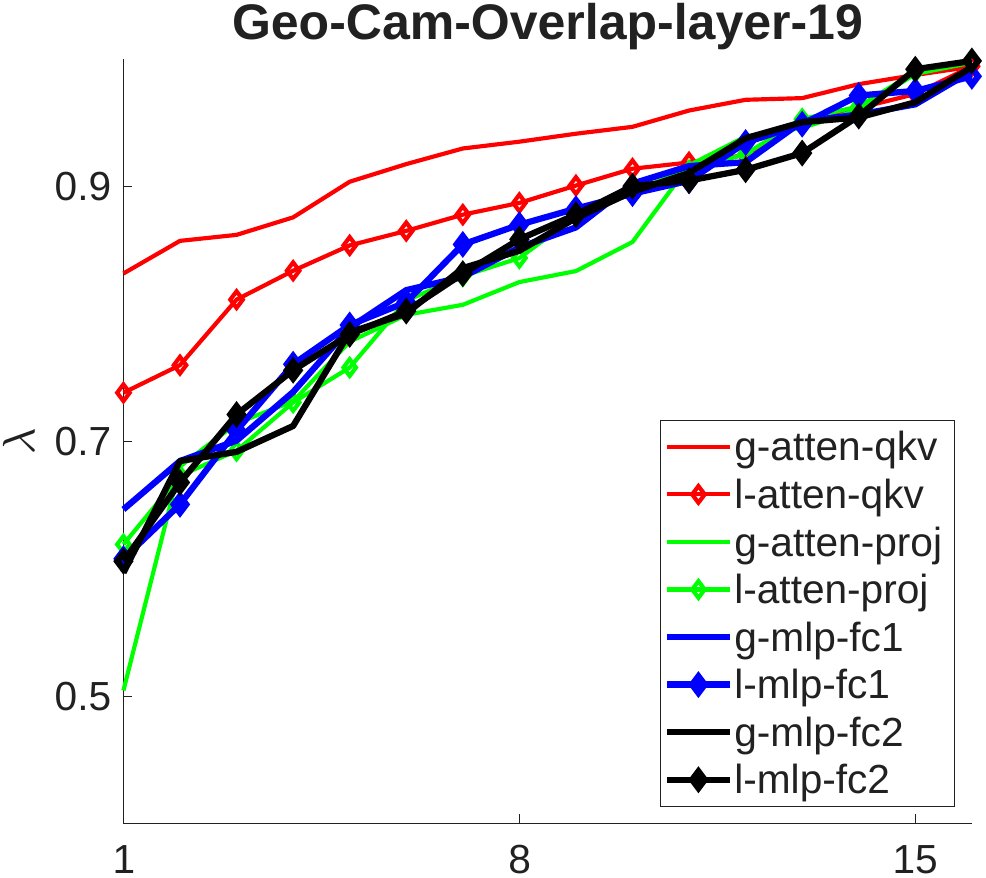}
&
\includegraphics[width=0.21\textwidth]{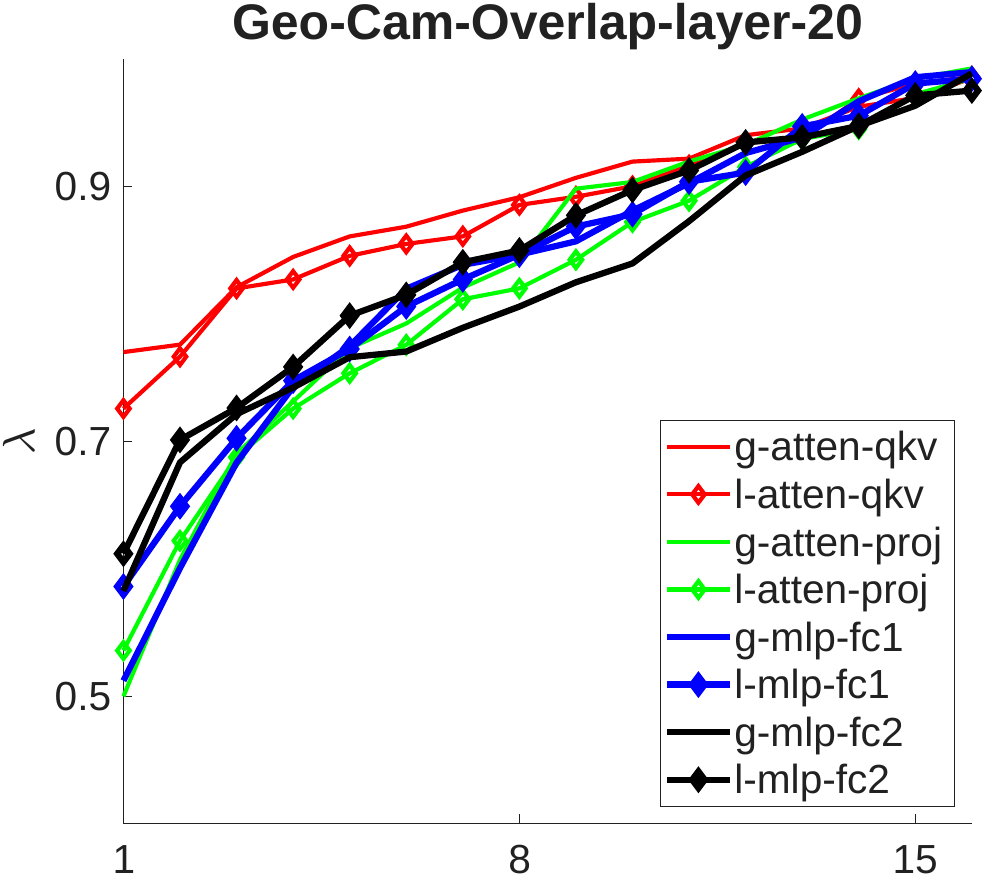}      
\\
\includegraphics[width=0.21\textwidth]{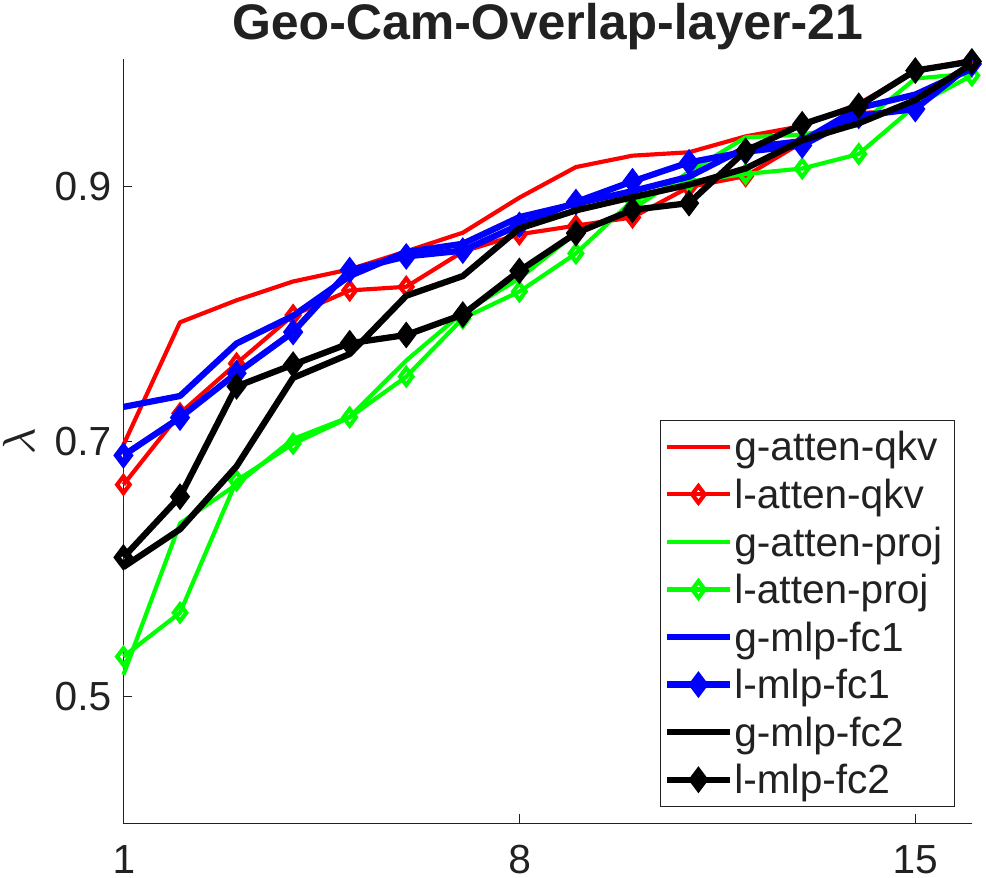}
& 
\includegraphics[width=0.21\textwidth]{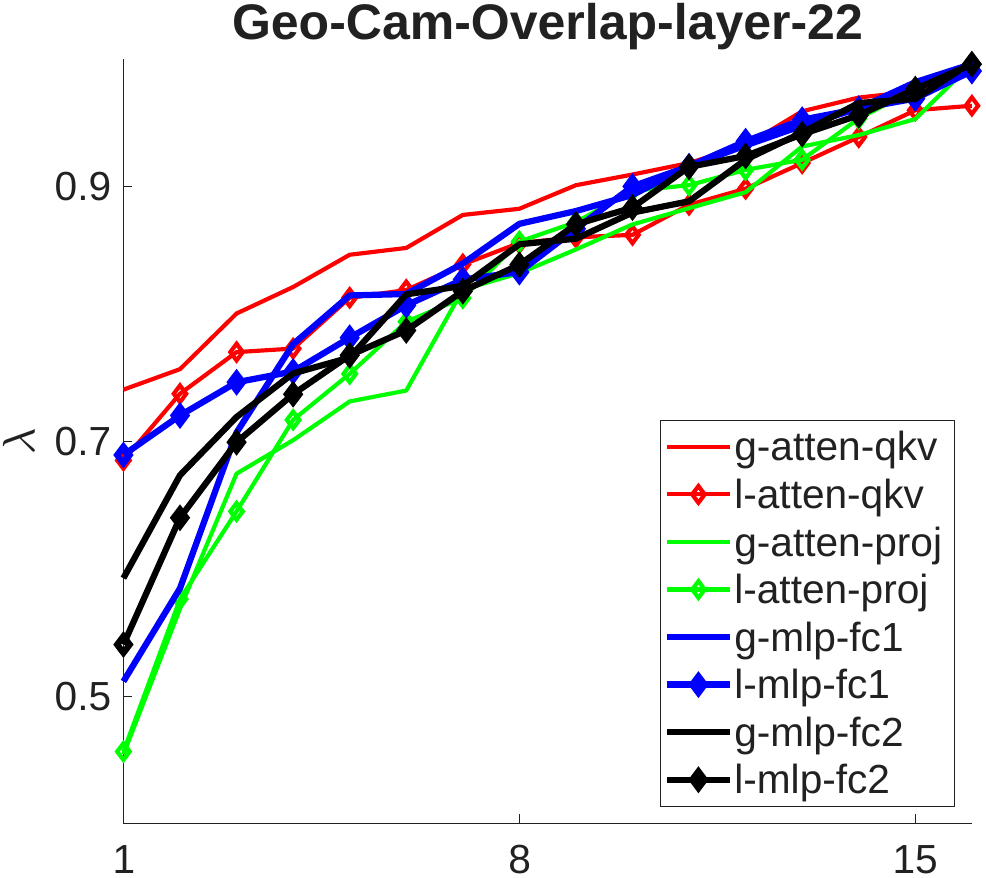}
&
\includegraphics[width=0.21\textwidth]{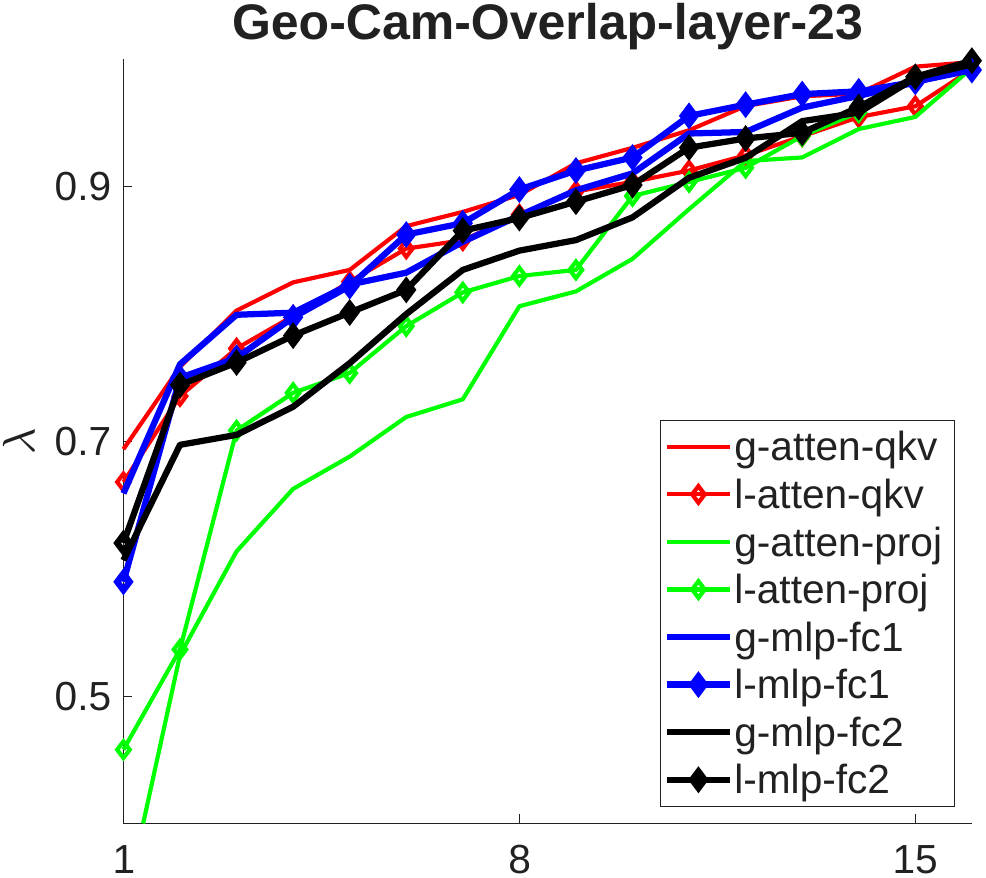}
&
\includegraphics[width=0.21\textwidth]{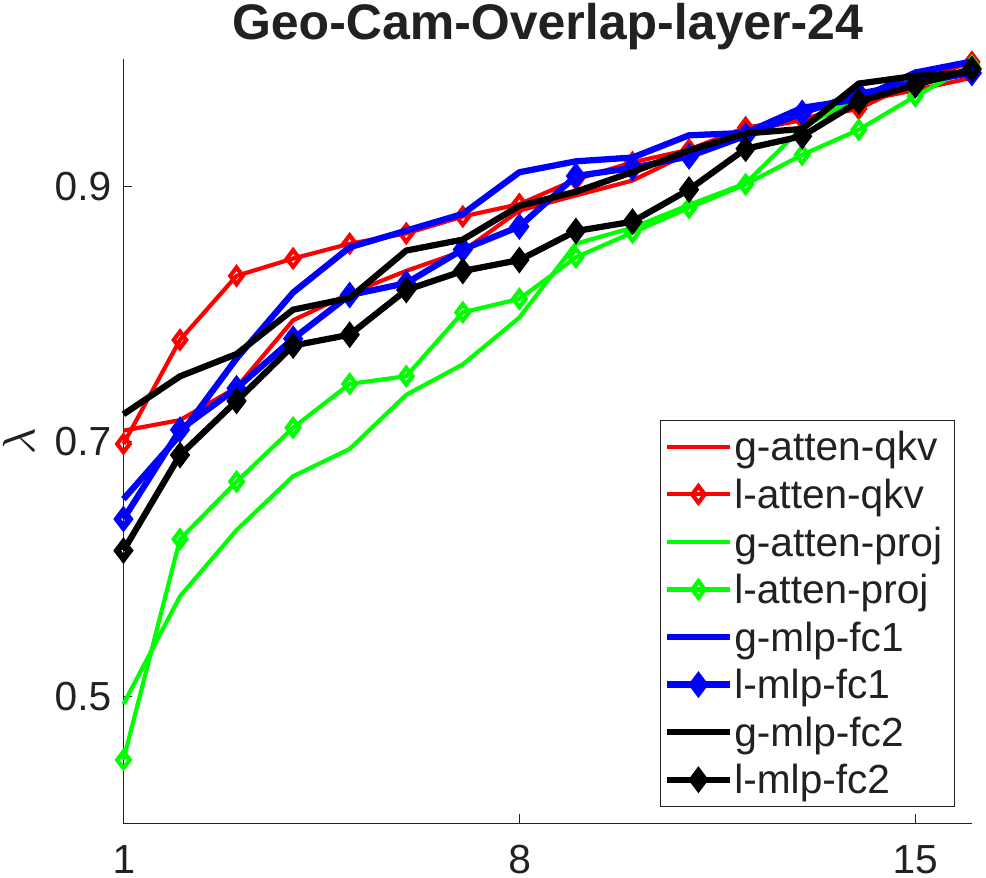}      

\end{tabular}

\captionof{figure}{The overlap ratio between subspaces that correspond to variations in geometry and camera motion. }

% \label{Fig:Subspace:Magnitudes}    
\vspace{-3em}
\end{table*}

\clearpage

\subsection{Geometry vs Lighting}

\begin{table*}[bp]
\centering
\setlength\tabcolsep{6pt}
\begin{tabular}{cccc}
\includegraphics[width=0.21\textwidth]{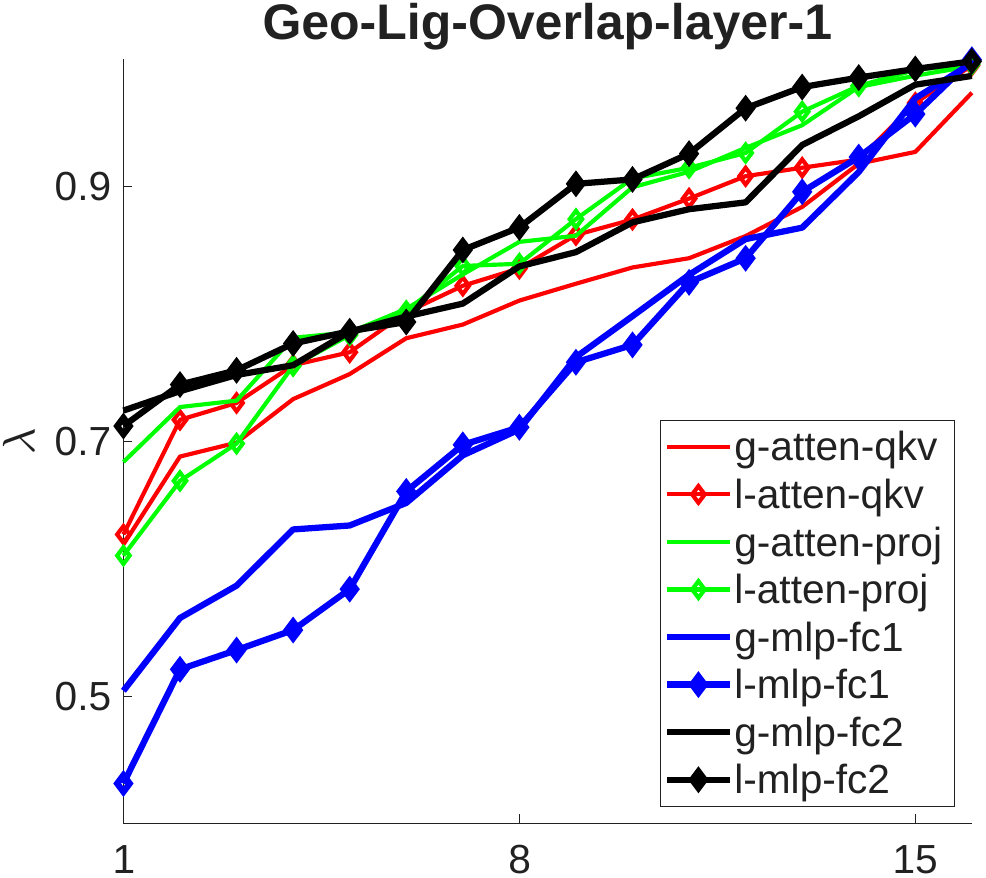}
& 
\includegraphics[width=0.21\textwidth]{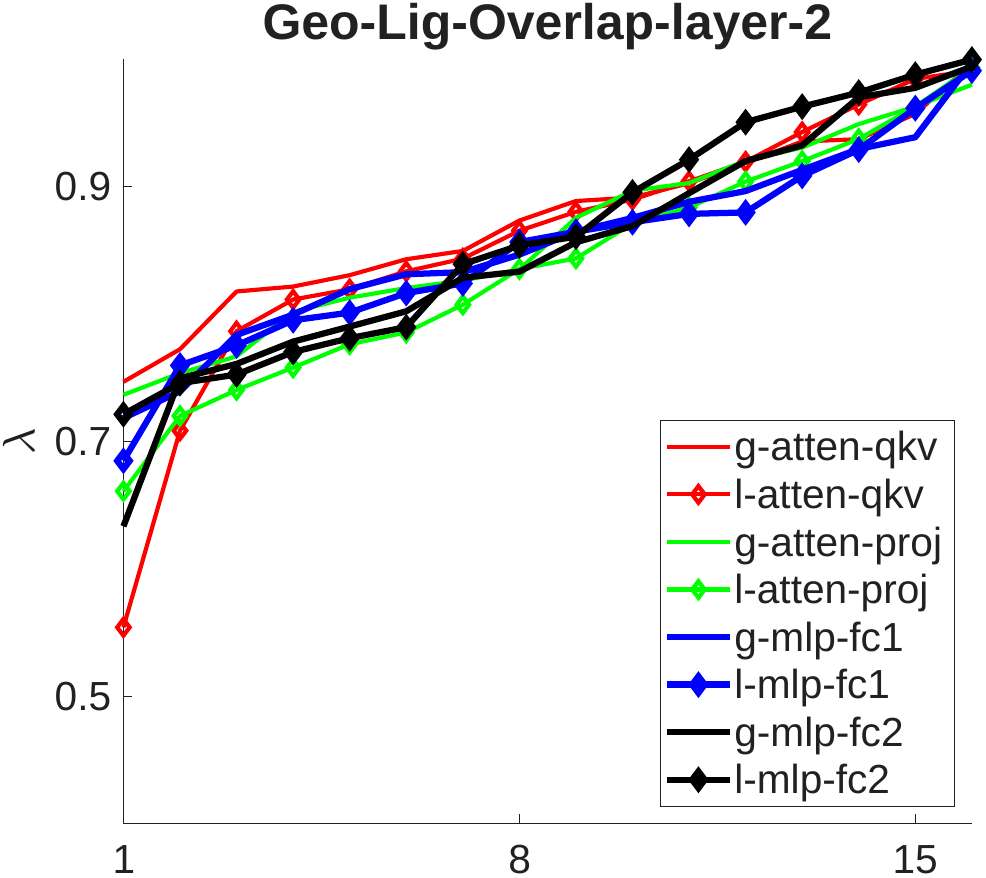}
&
\includegraphics[width=0.21\textwidth]{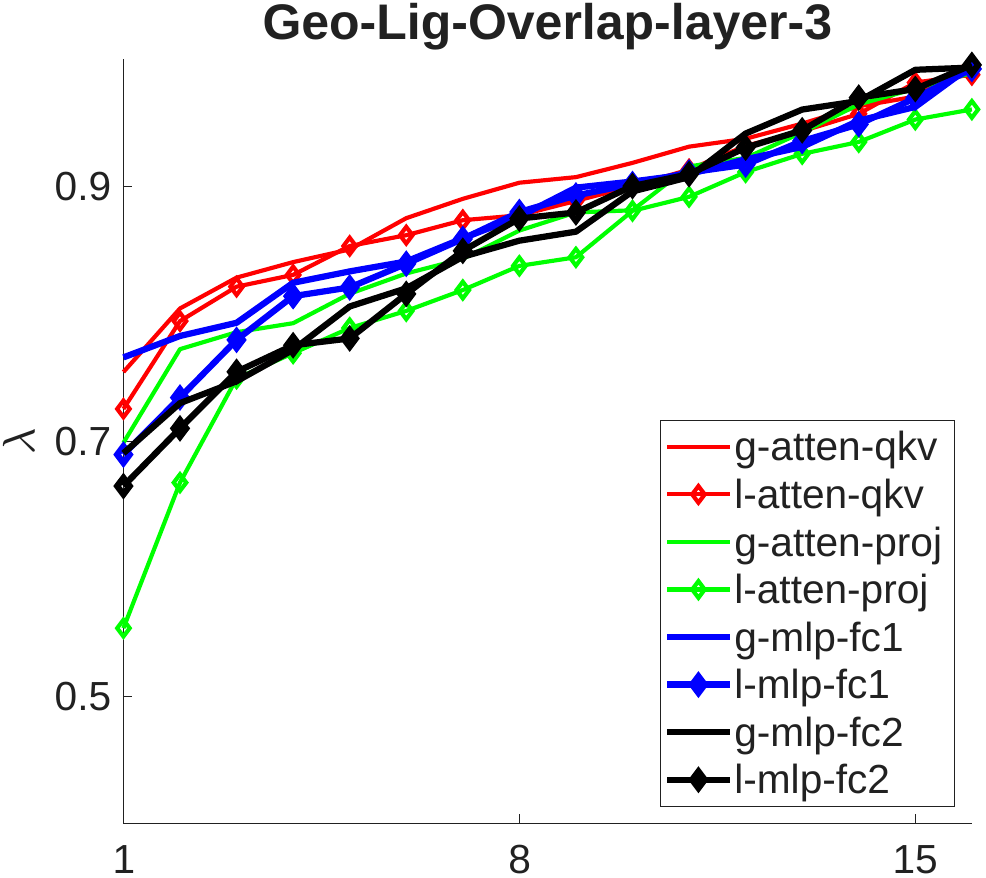}
&
\includegraphics[width=0.21\textwidth]{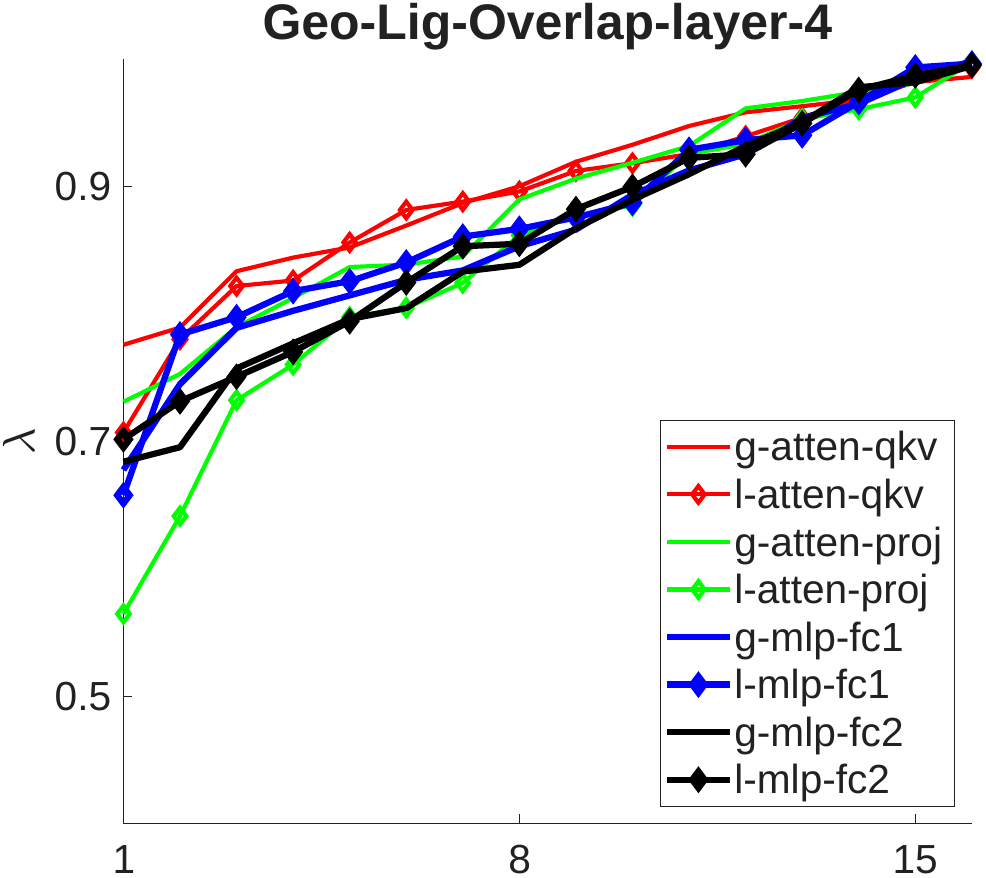}      
\\
\includegraphics[width=0.21\textwidth]{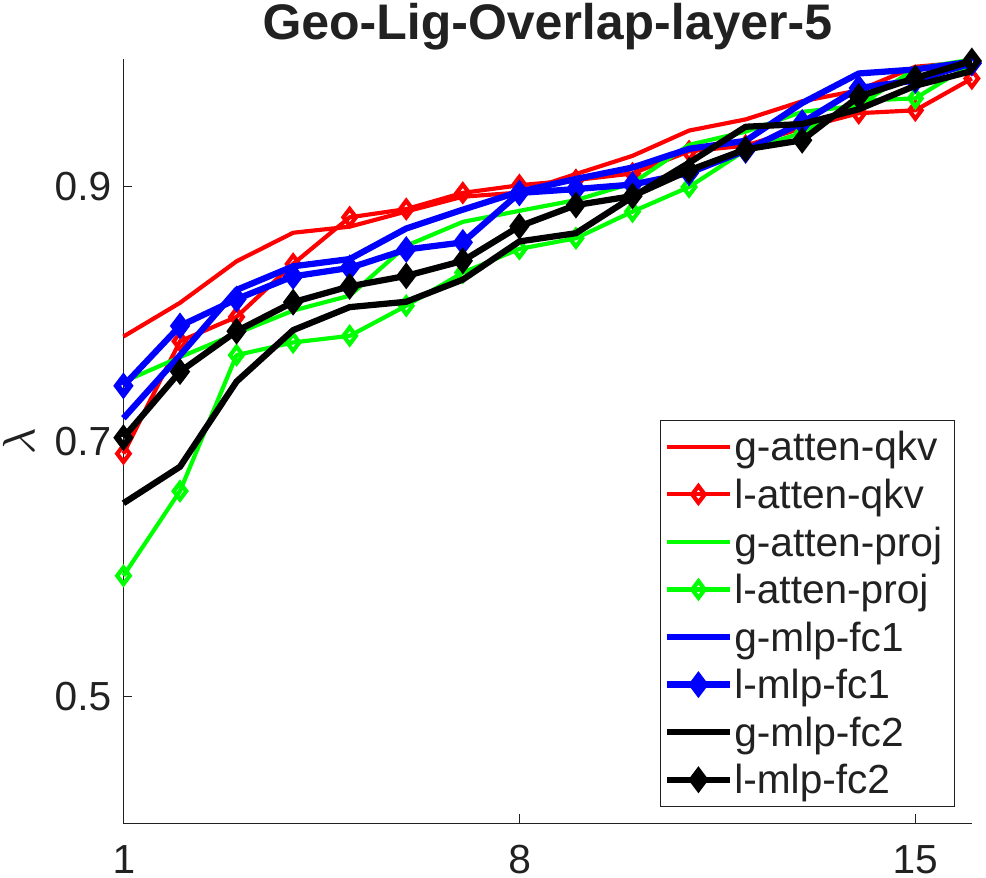}
& 
\includegraphics[width=0.21\textwidth]{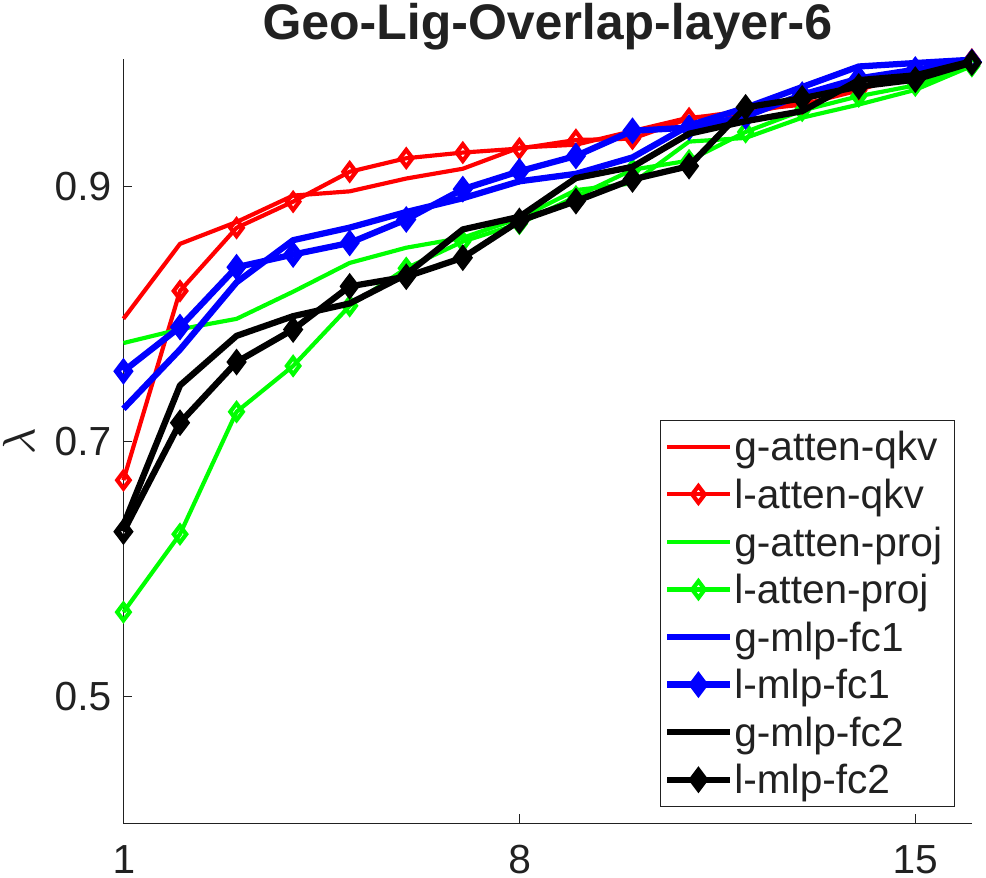}
&
\includegraphics[width=0.21\textwidth]{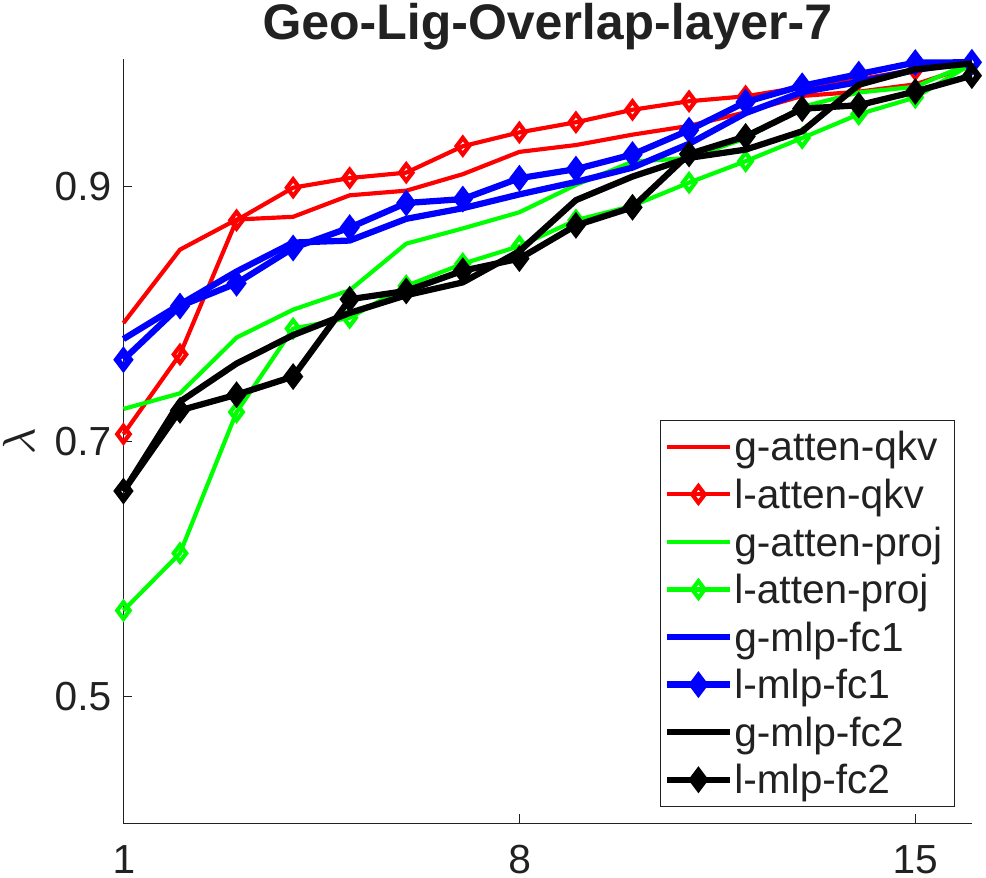}
&
\includegraphics[width=0.21\textwidth]{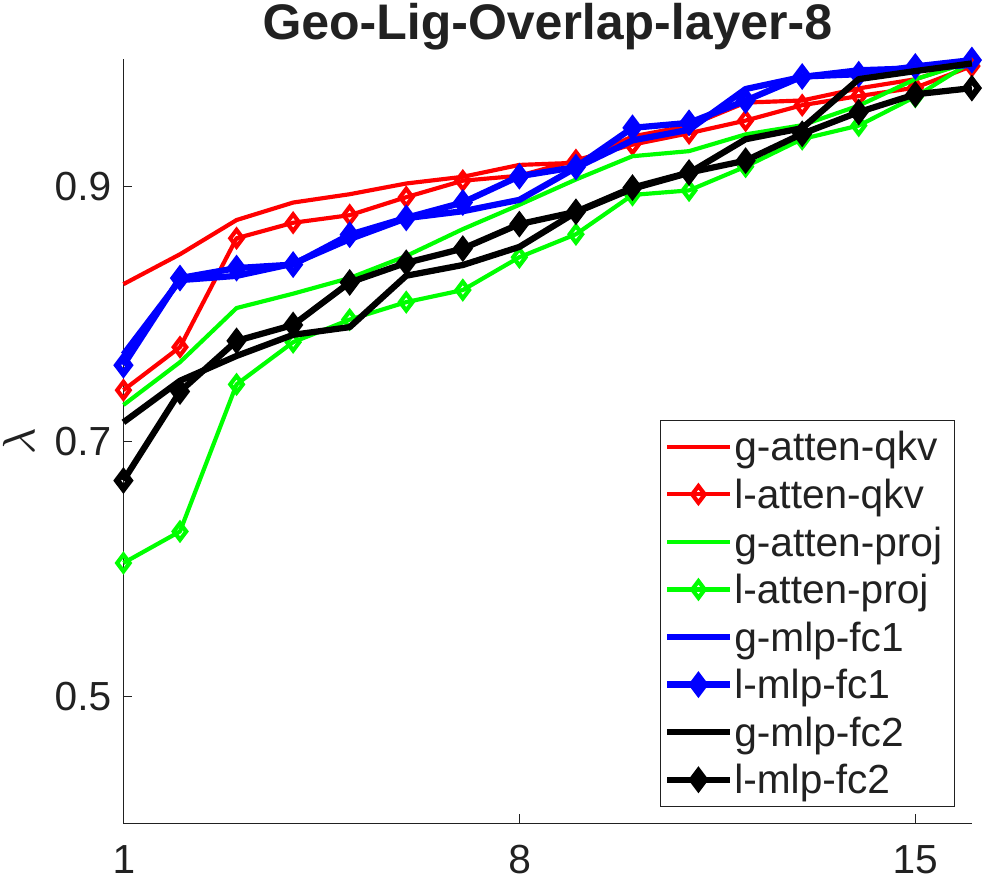}      
\\
\includegraphics[width=0.21\textwidth]{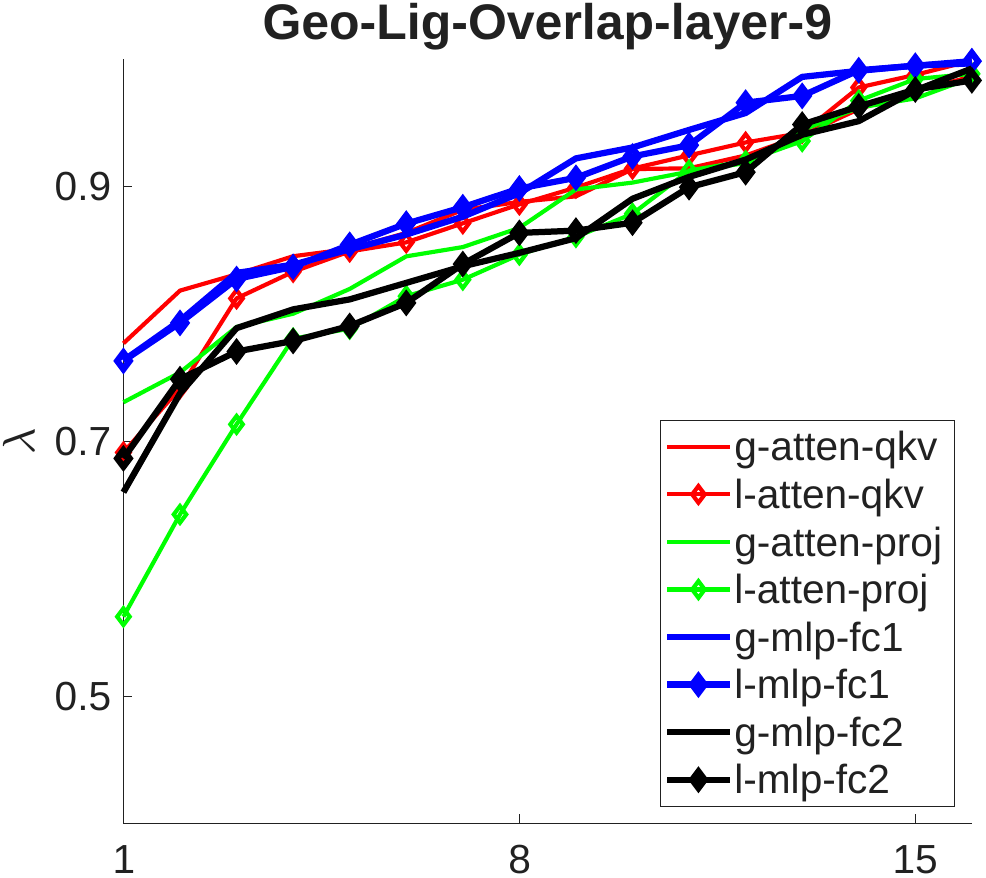}
& 
\includegraphics[width=0.21\textwidth]{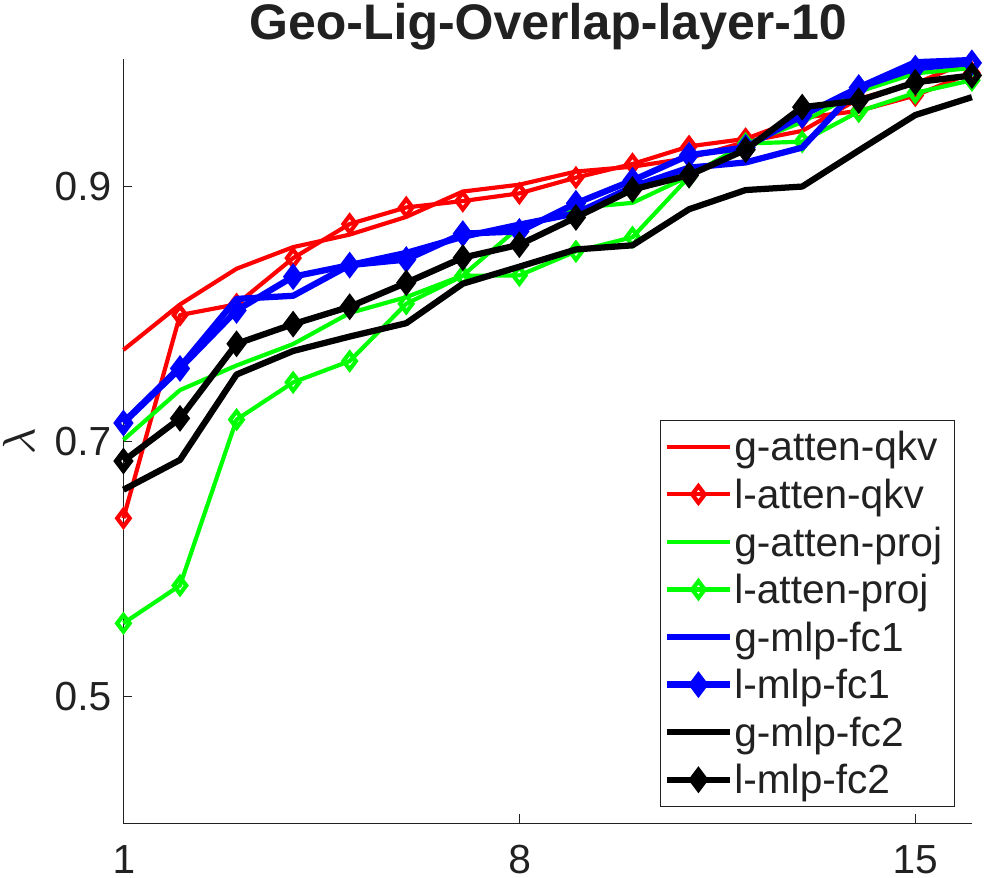}
&
\includegraphics[width=0.21\textwidth]{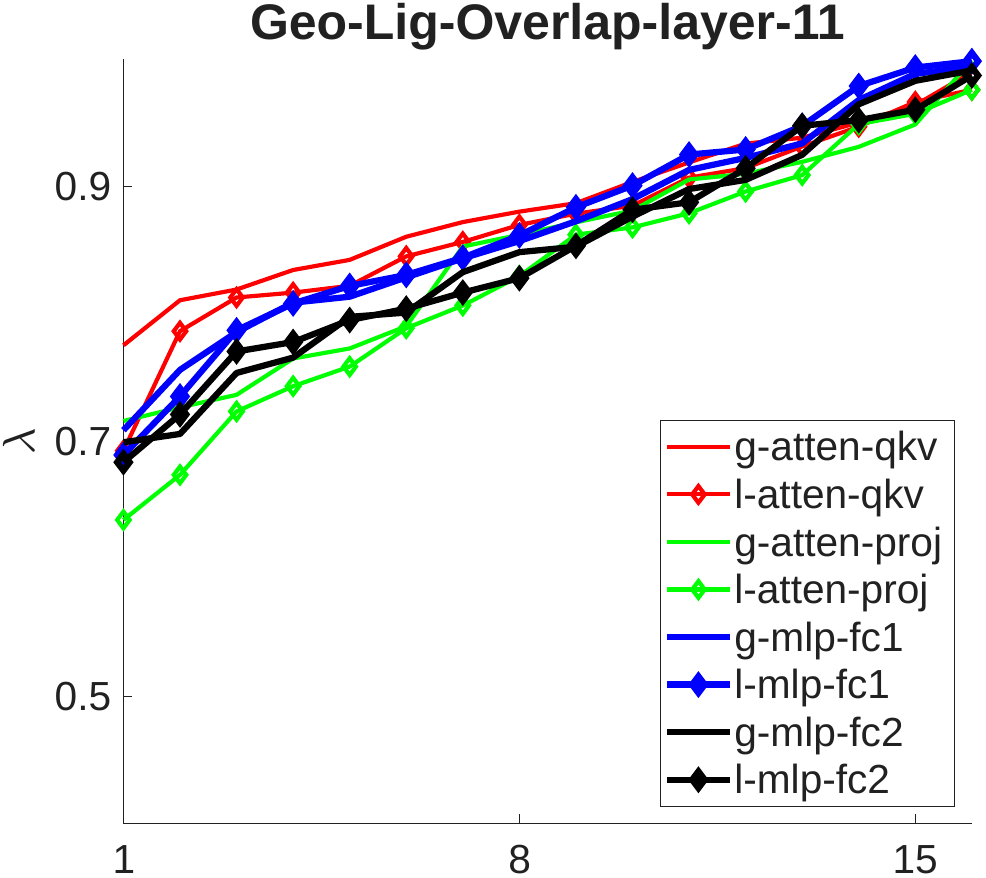}
&
\includegraphics[width=0.21\textwidth]{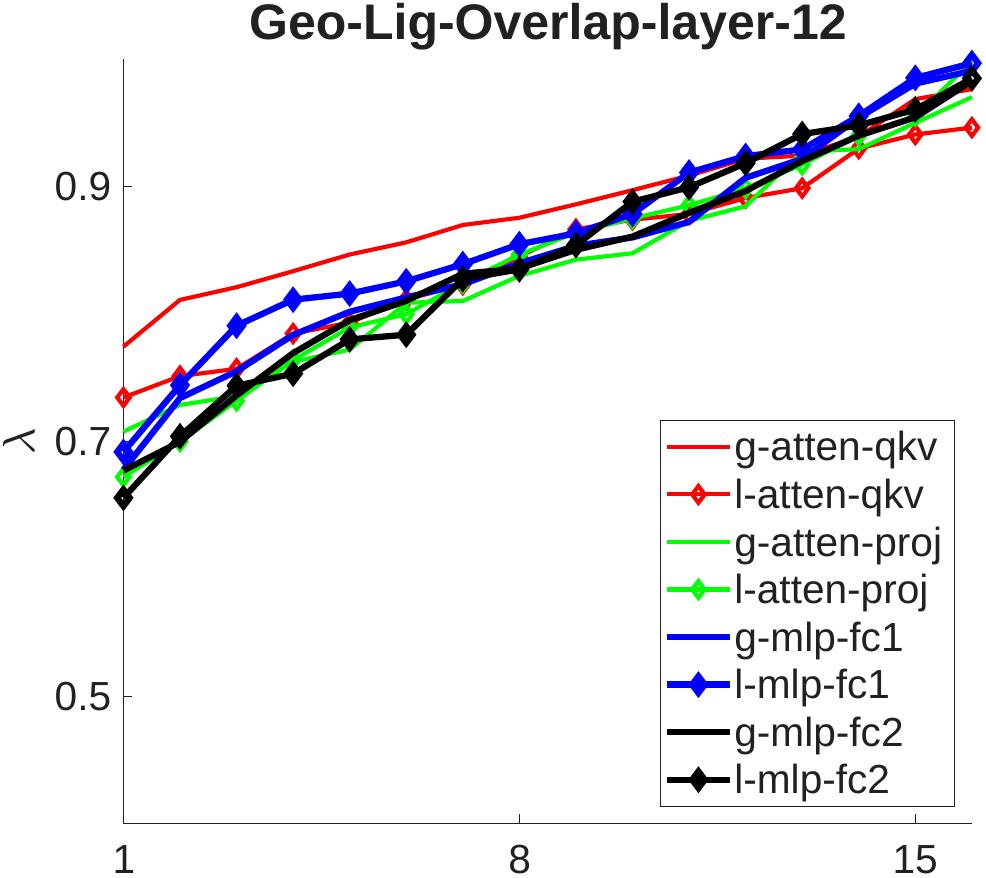}      
\\
\includegraphics[width=0.21\textwidth]{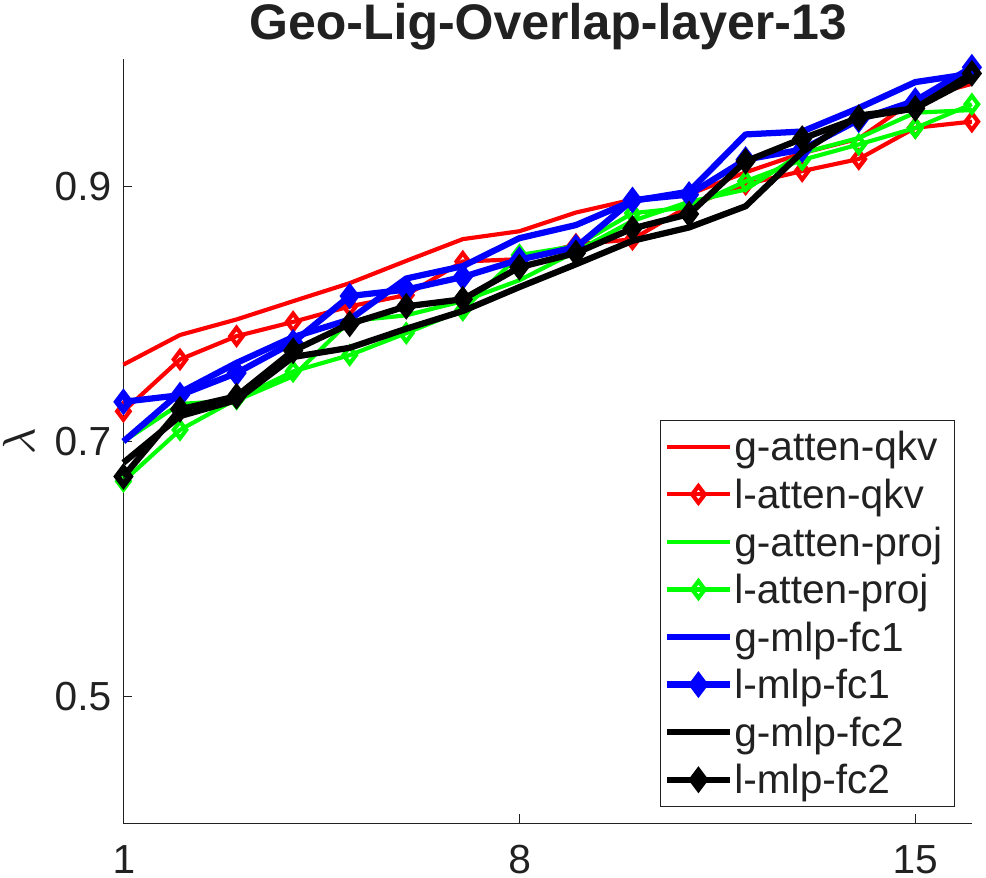}
& 
\includegraphics[width=0.21\textwidth]{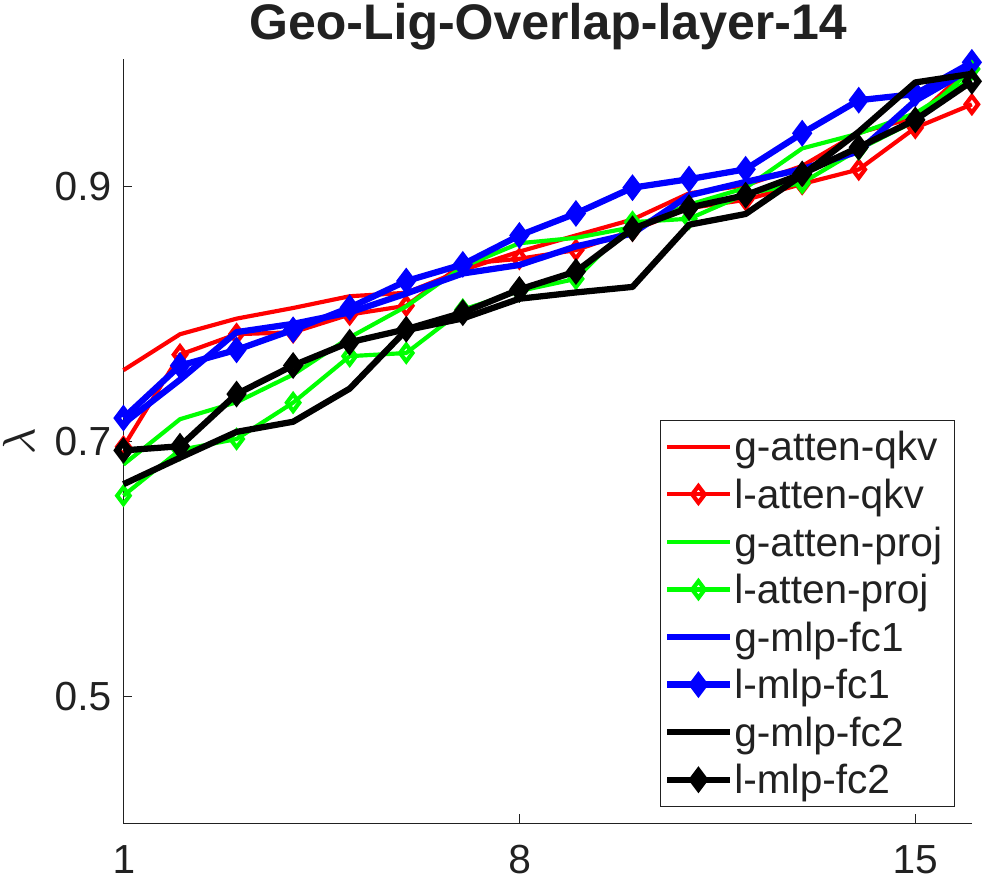}
&
\includegraphics[width=0.21\textwidth]{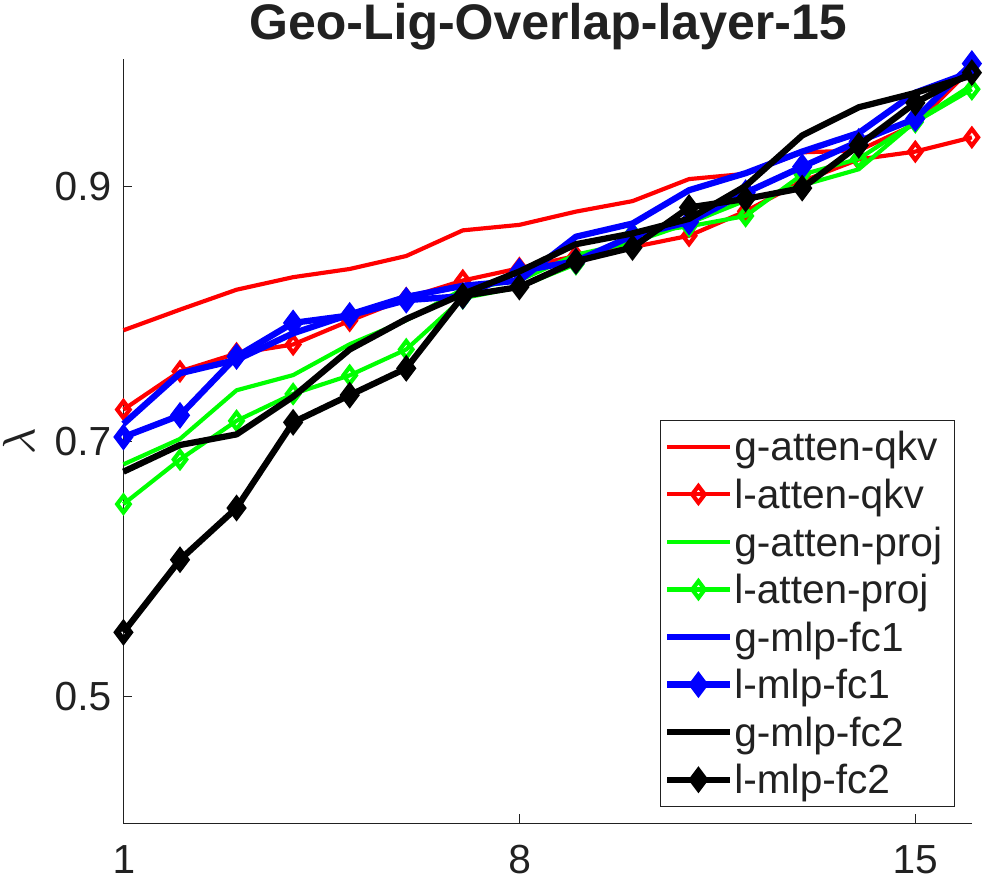}
&
\includegraphics[width=0.21\textwidth]{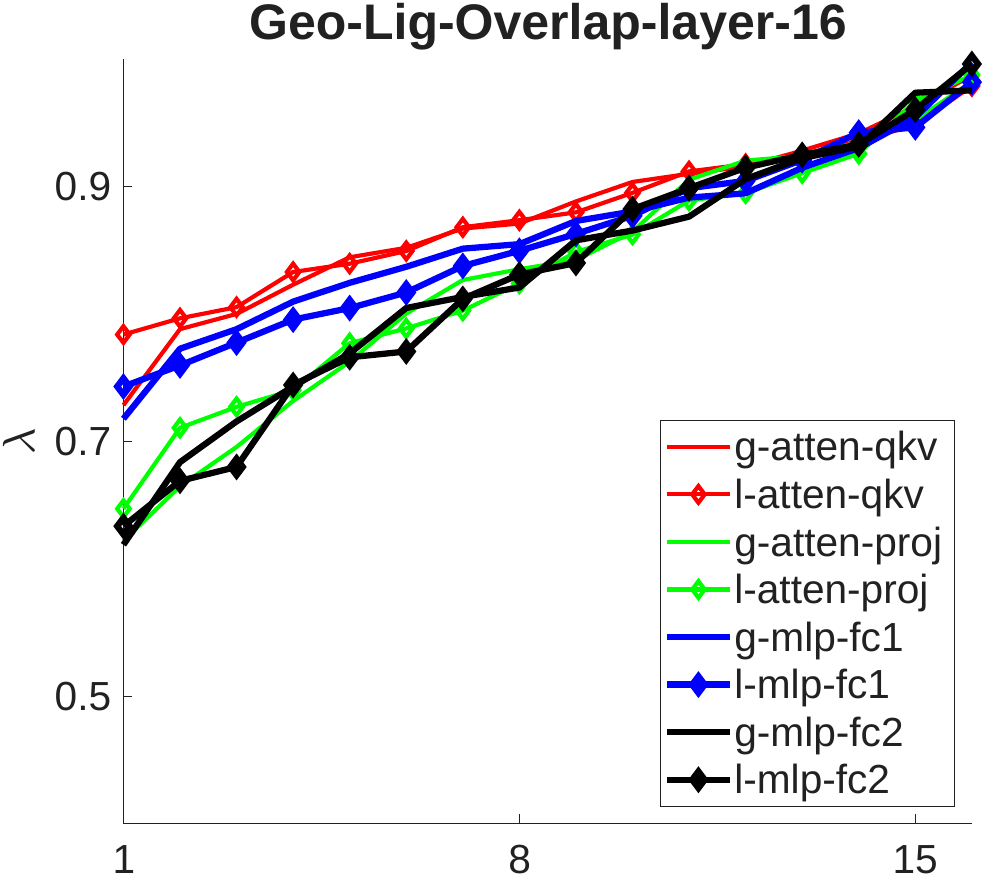}      
\\
\includegraphics[width=0.21\textwidth]{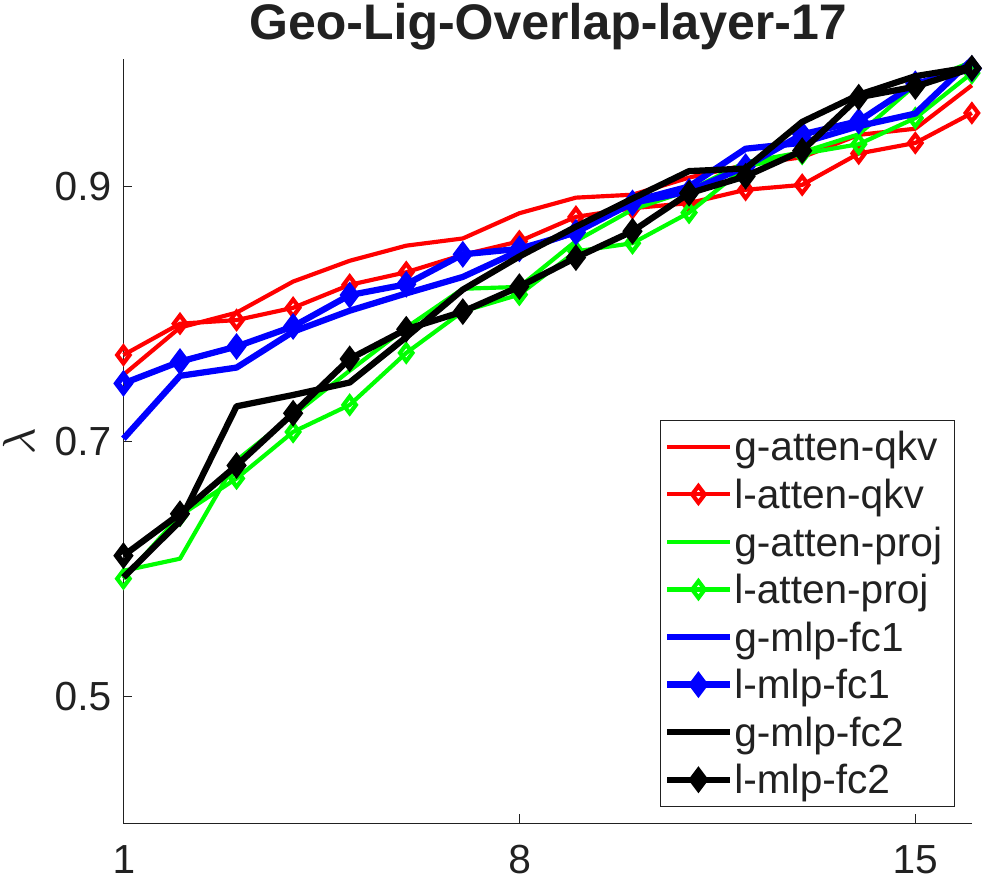}
& 
\includegraphics[width=0.21\textwidth]{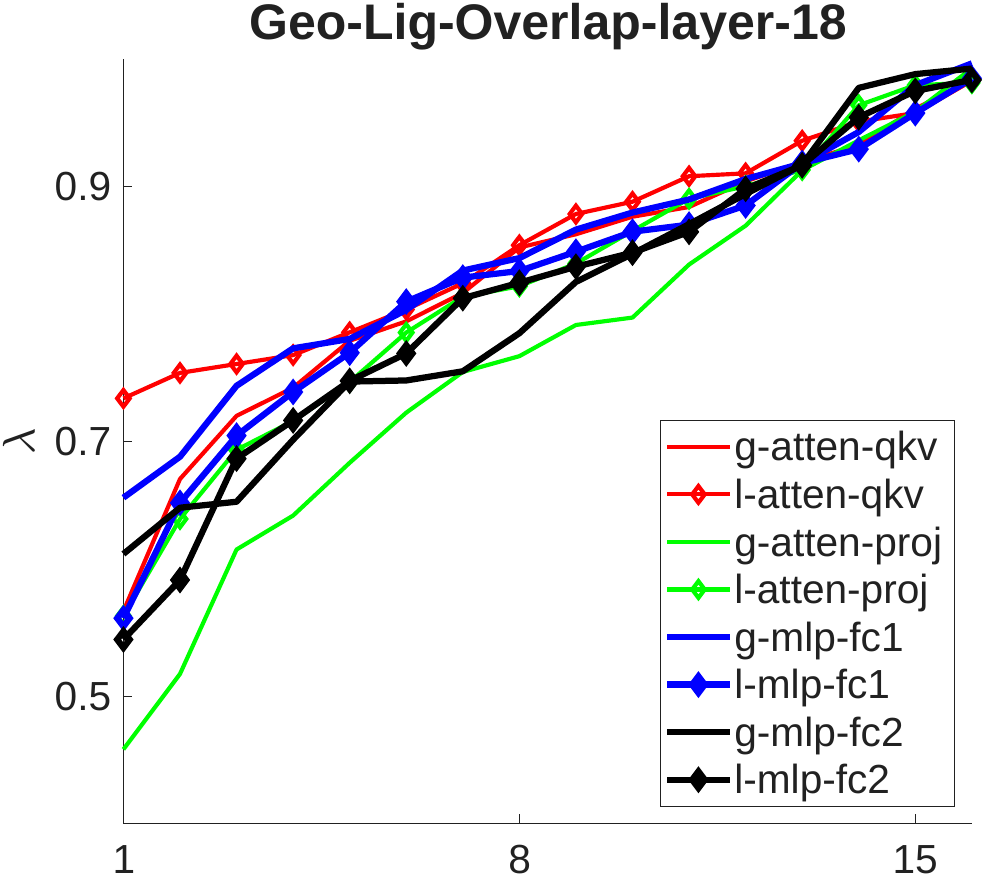}
&
\includegraphics[width=0.21\textwidth]{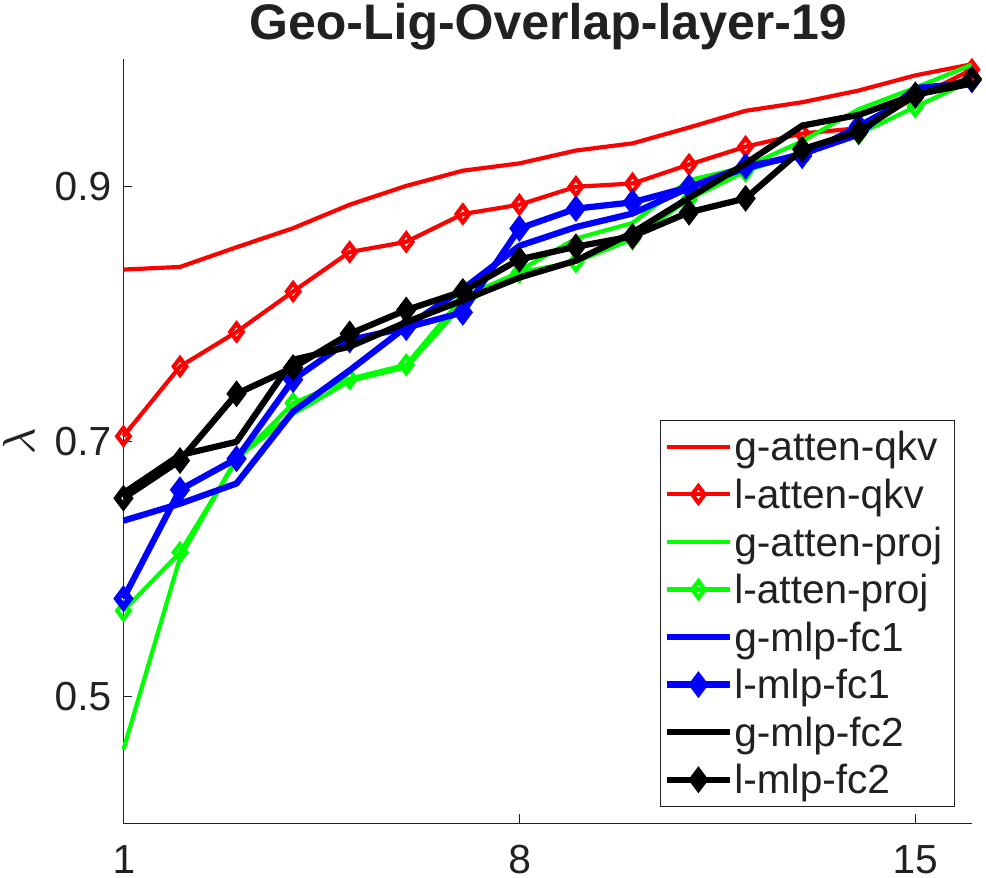}
&
\includegraphics[width=0.21\textwidth]{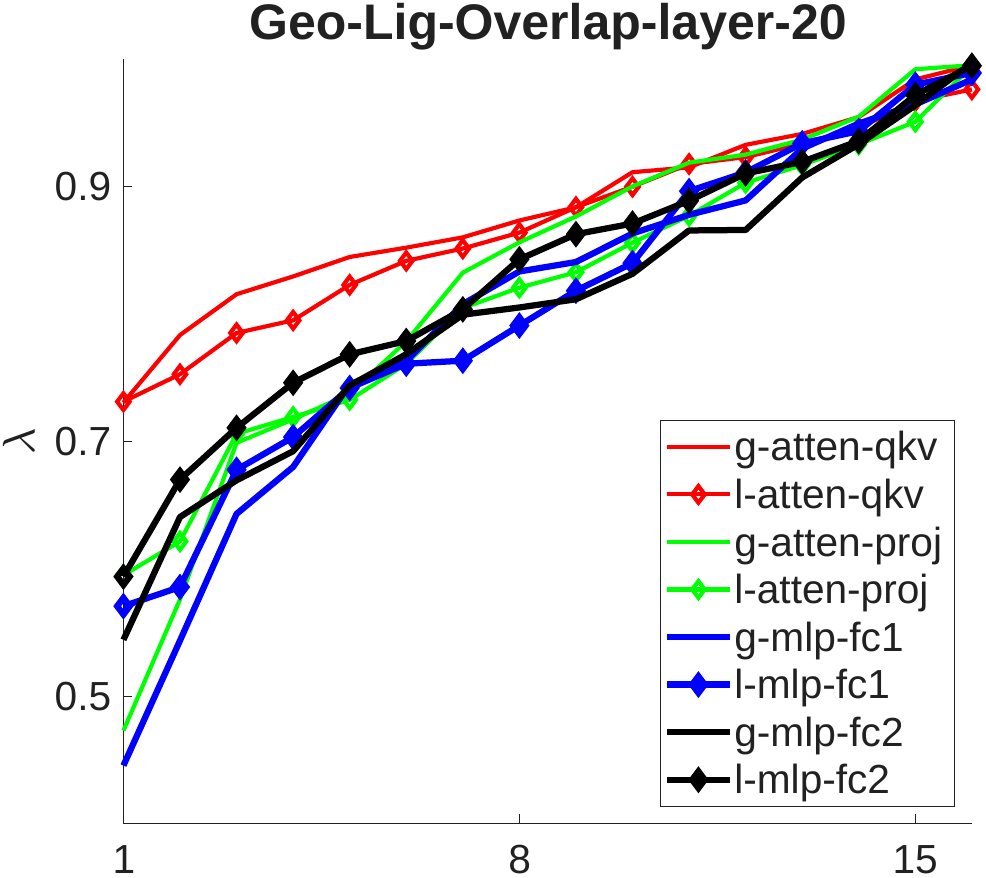}      
\\
\includegraphics[width=0.21\textwidth]{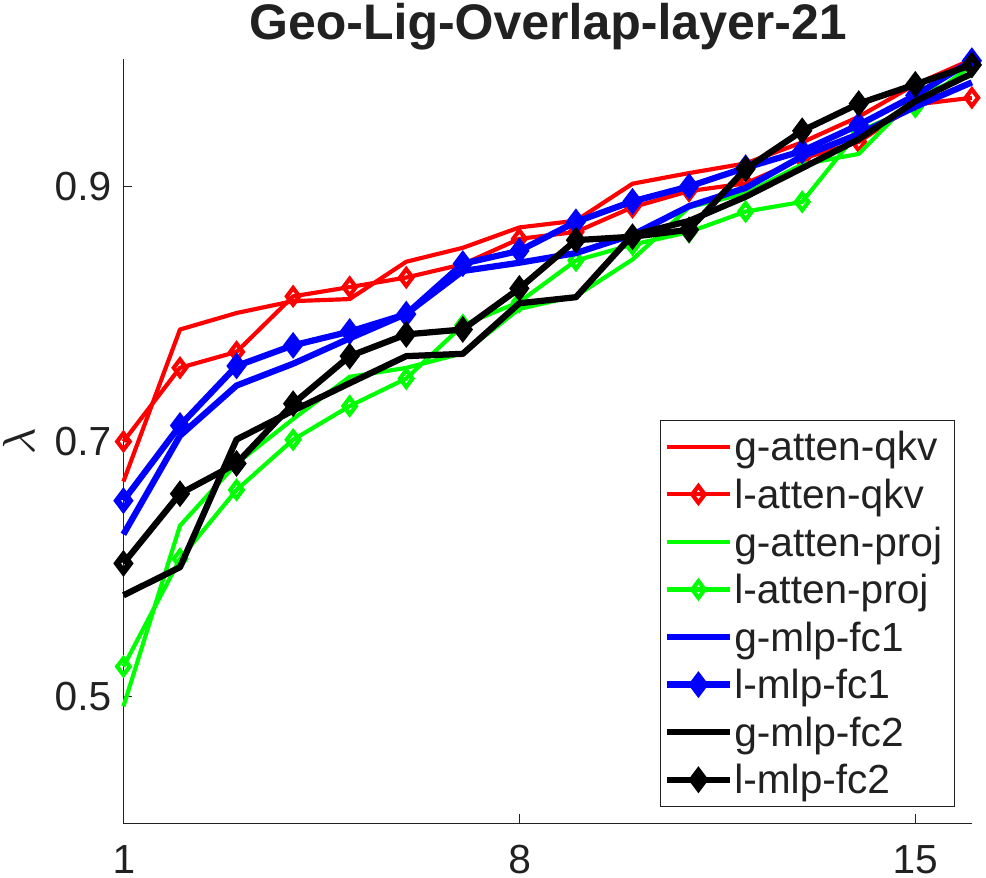}
& 
\includegraphics[width=0.21\textwidth]{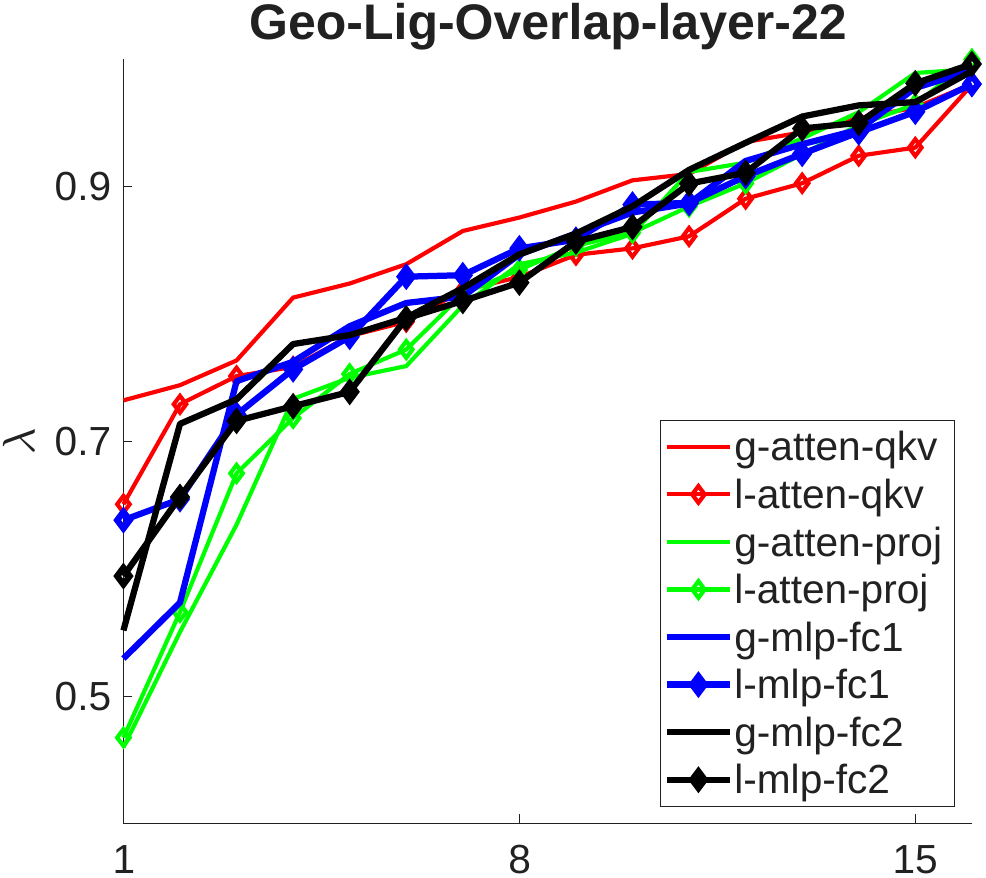}
&
\includegraphics[width=0.21\textwidth]{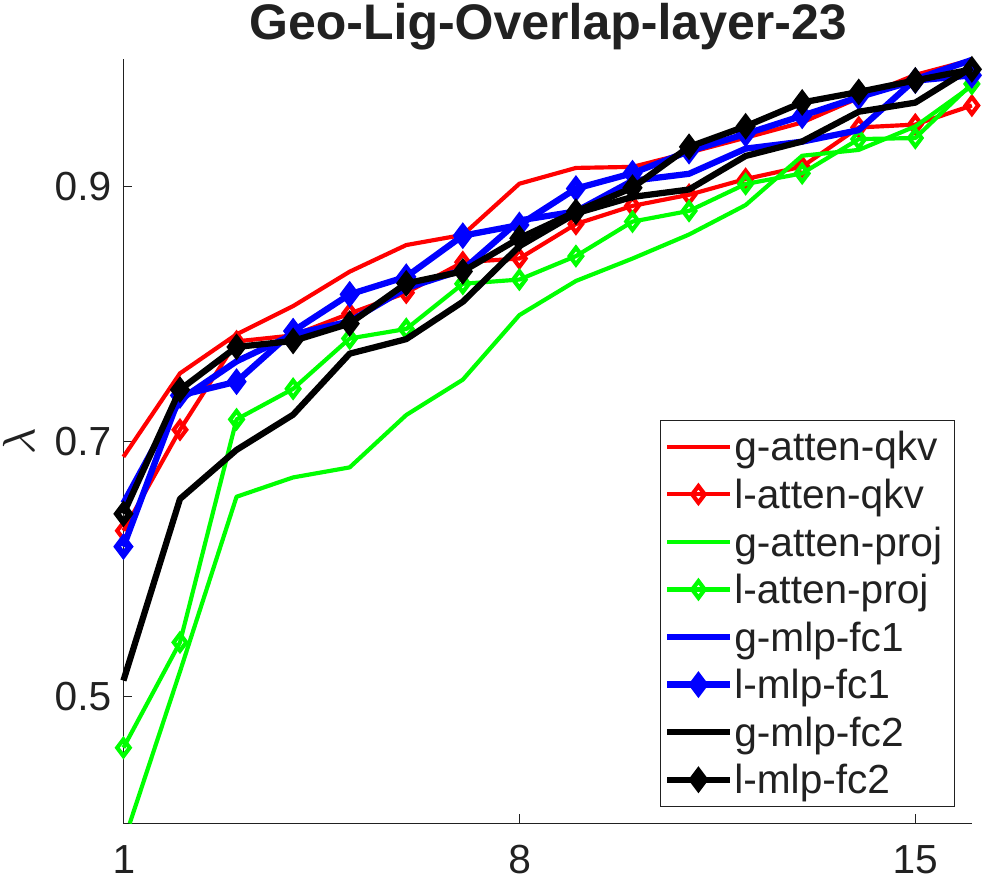}
&
\includegraphics[width=0.21\textwidth]{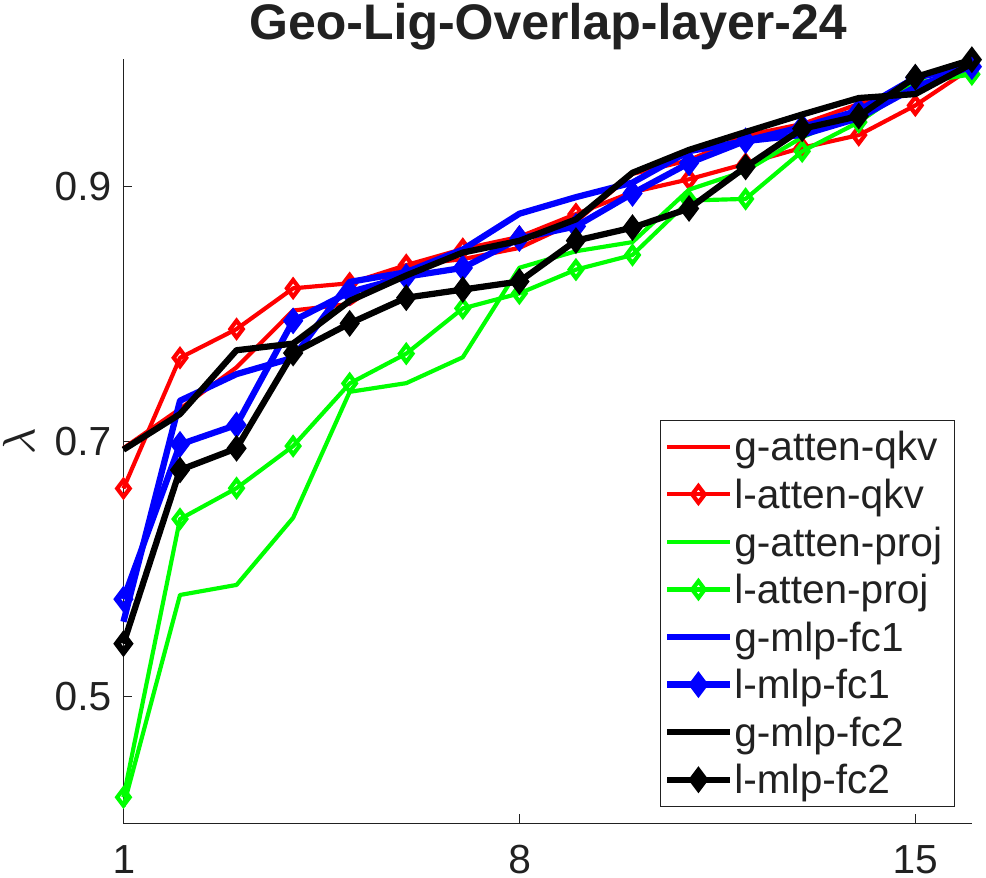}      

\end{tabular}

\captionof{figure}{The overlap ratio between subspaces that correspond to variations in geometry and lighting. }

% \label{Fig:Subspace:Magnitudes}    
\vspace{-3em}
\end{table*}

\clearpage

\subsection{Texture vs Camera}

\begin{table*}[bp]
\centering
\setlength\tabcolsep{6pt}
\begin{tabular}{cccc}
\includegraphics[width=0.21\textwidth]{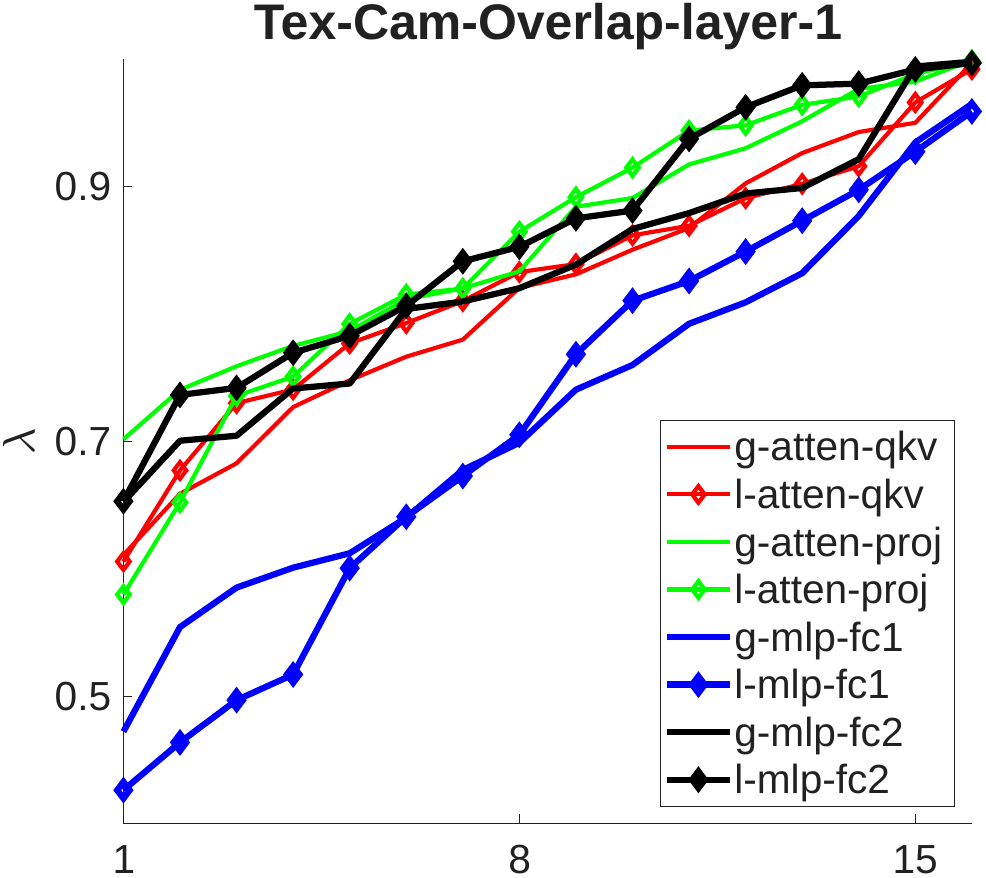}
& 
\includegraphics[width=0.21\textwidth]{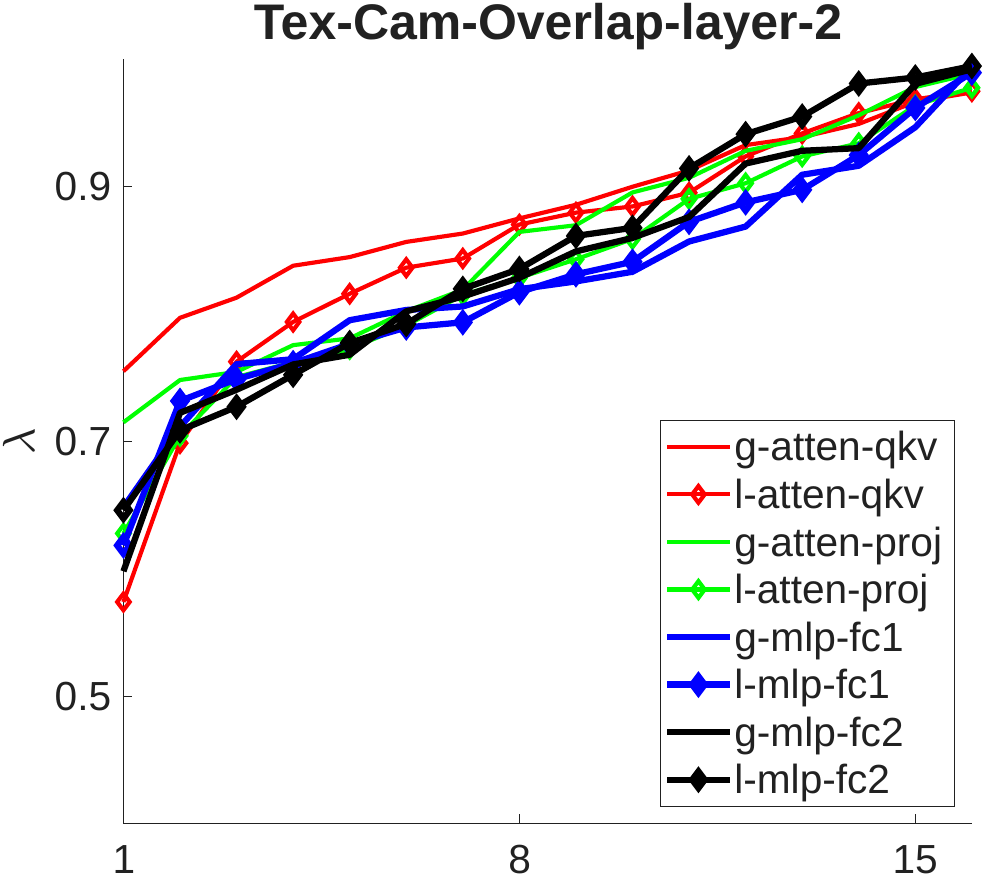}
&
\includegraphics[width=0.21\textwidth]{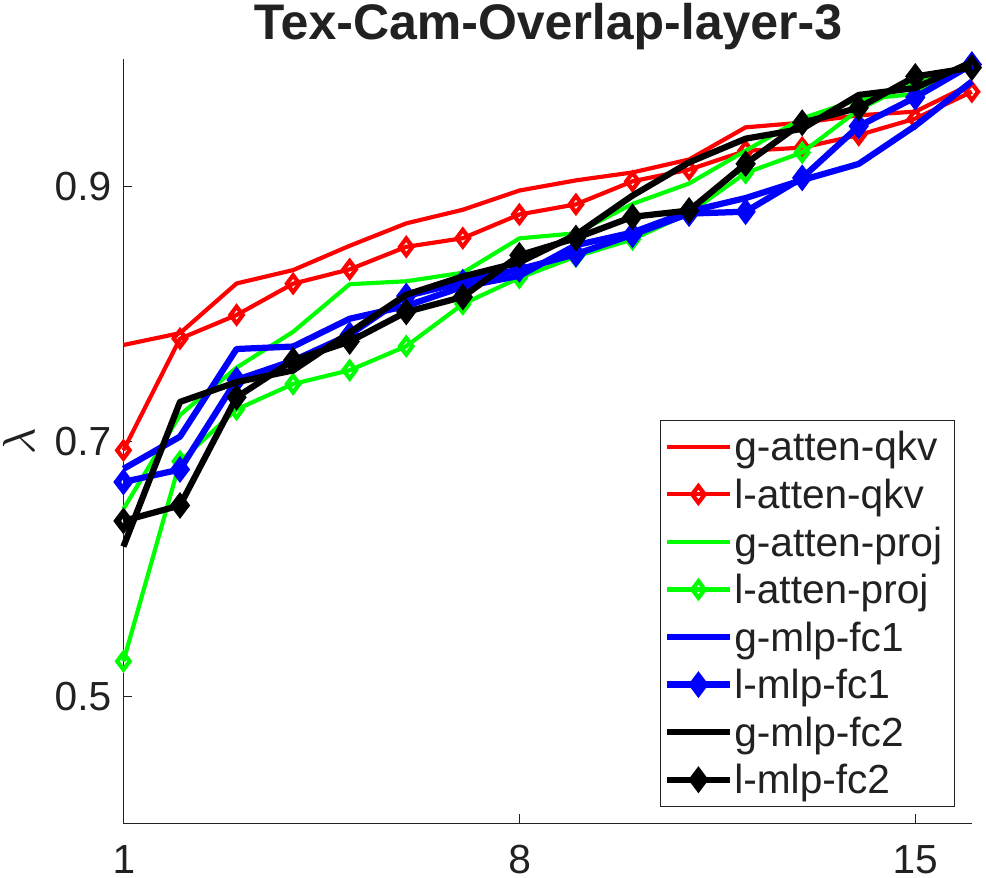}
&
\includegraphics[width=0.21\textwidth]{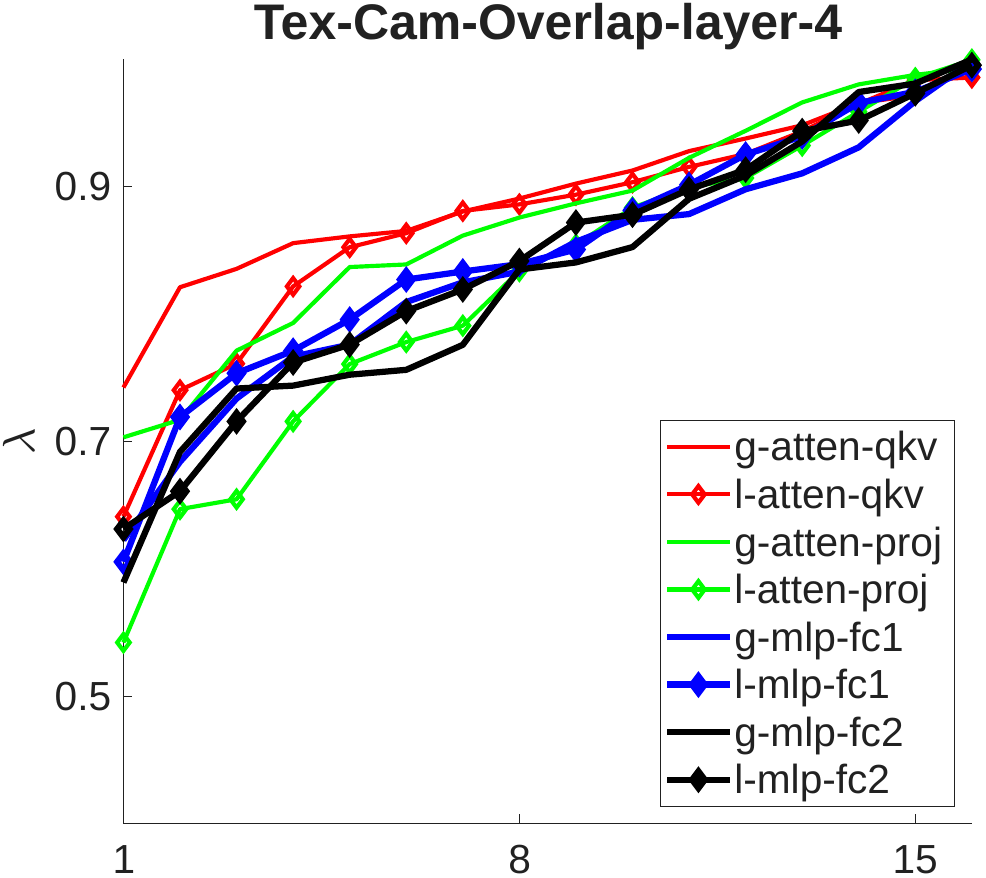}      
\\
\includegraphics[width=0.21\textwidth]{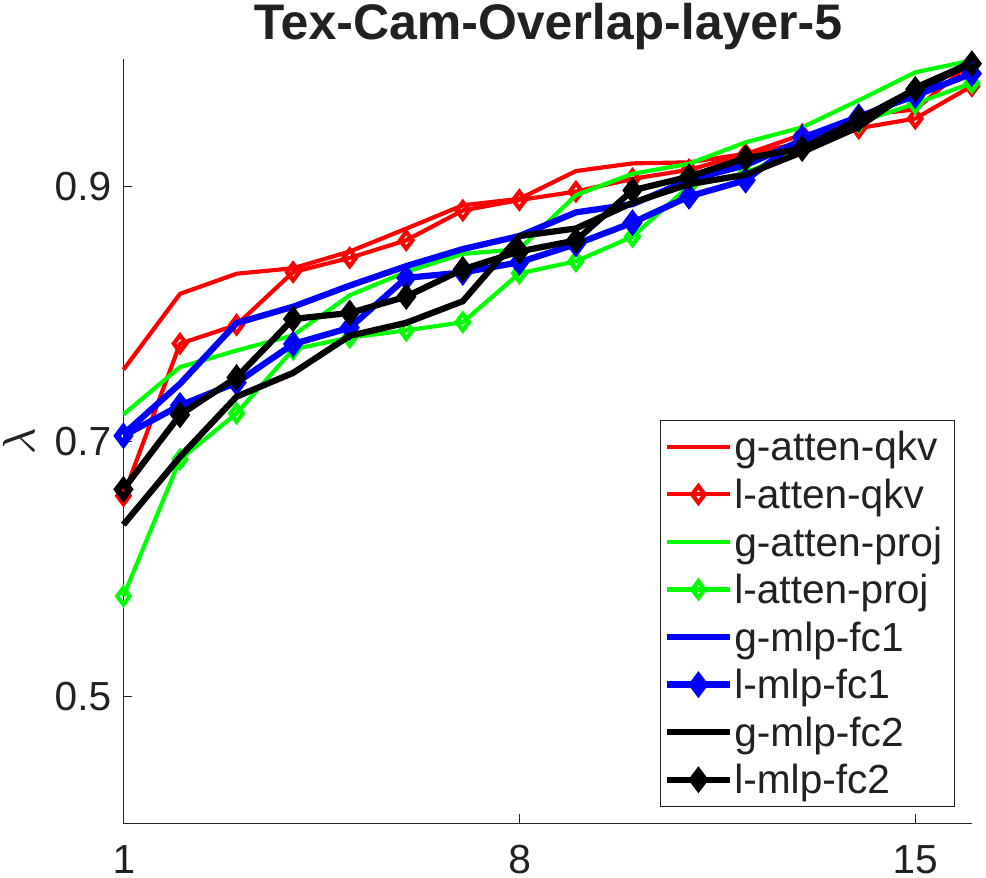}
& 
\includegraphics[width=0.21\textwidth]{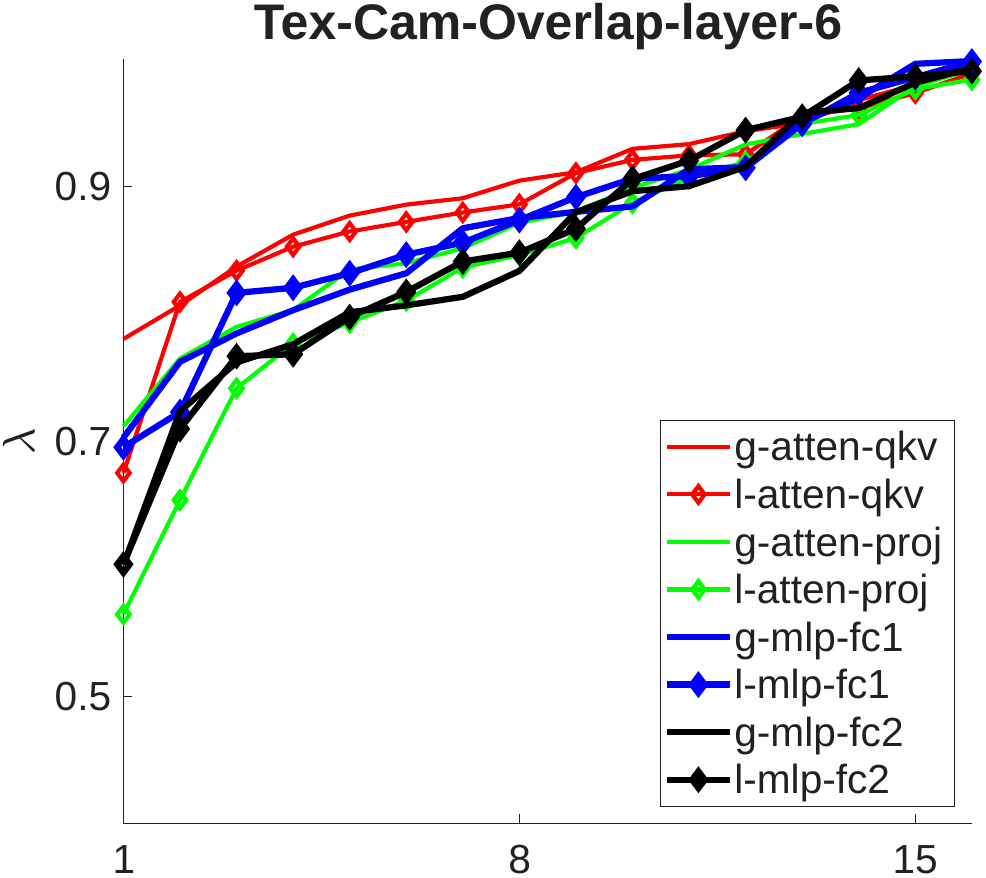}
&
\includegraphics[width=0.21\textwidth]{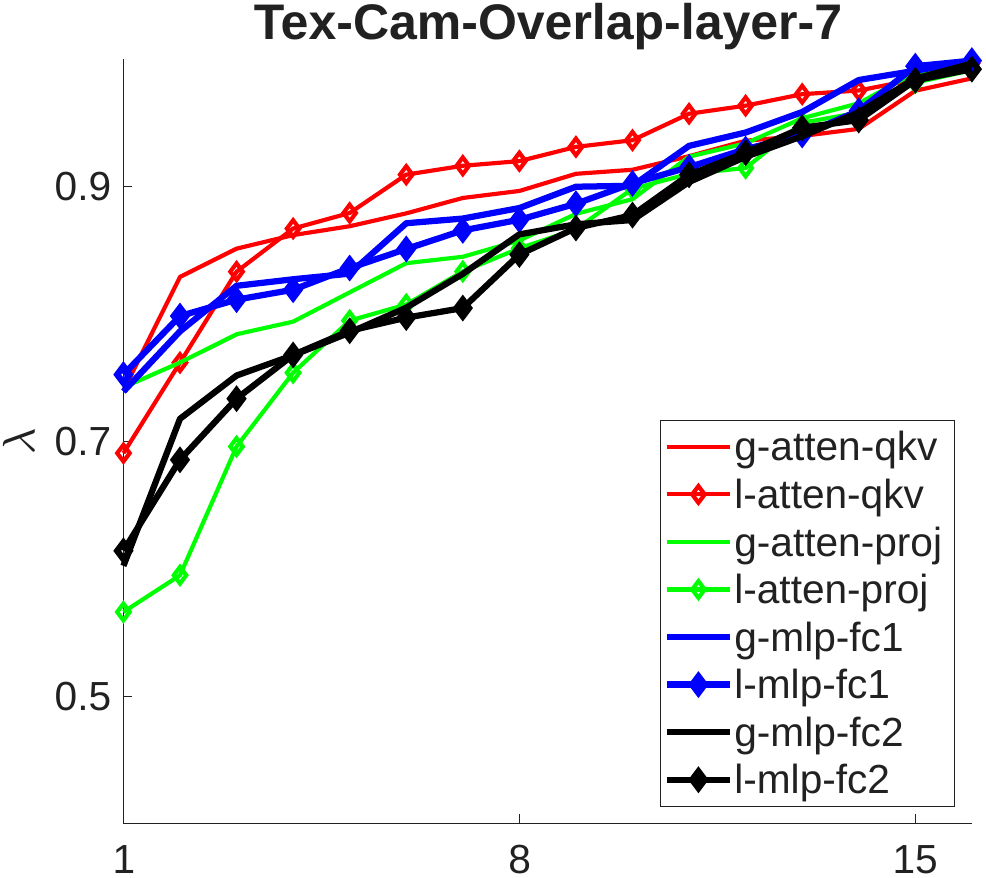}
&
\includegraphics[width=0.21\textwidth]{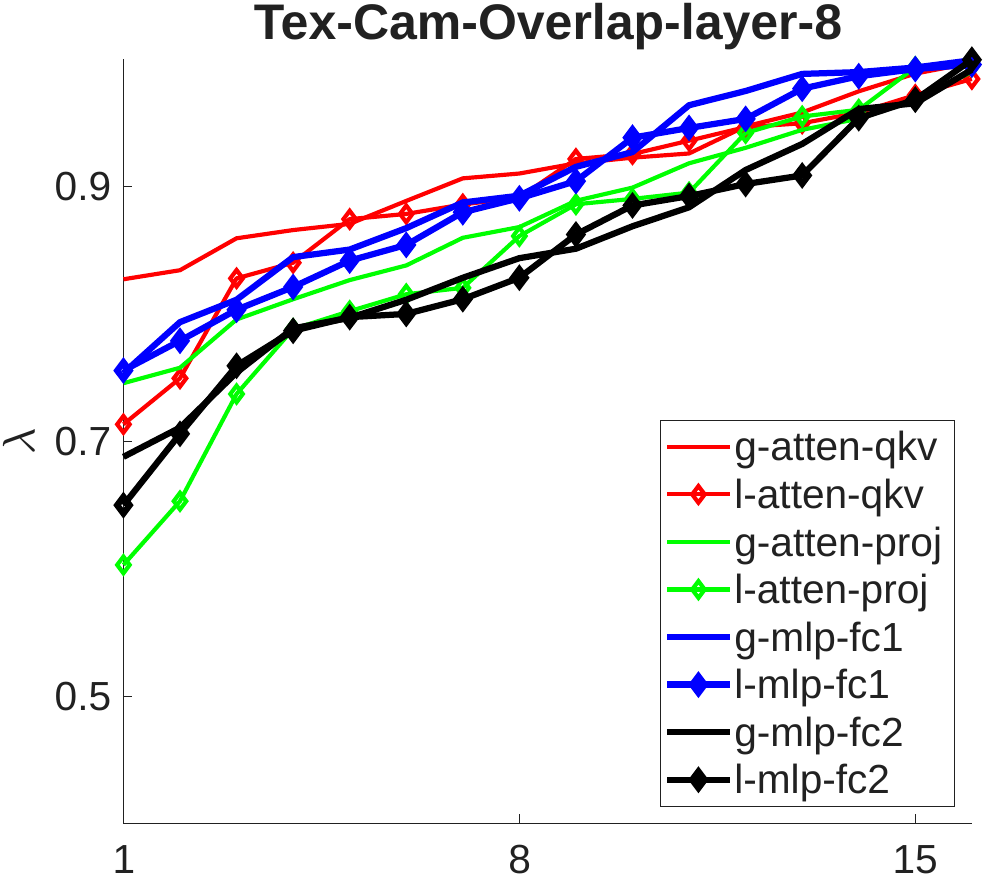}      
\\
\includegraphics[width=0.21\textwidth]{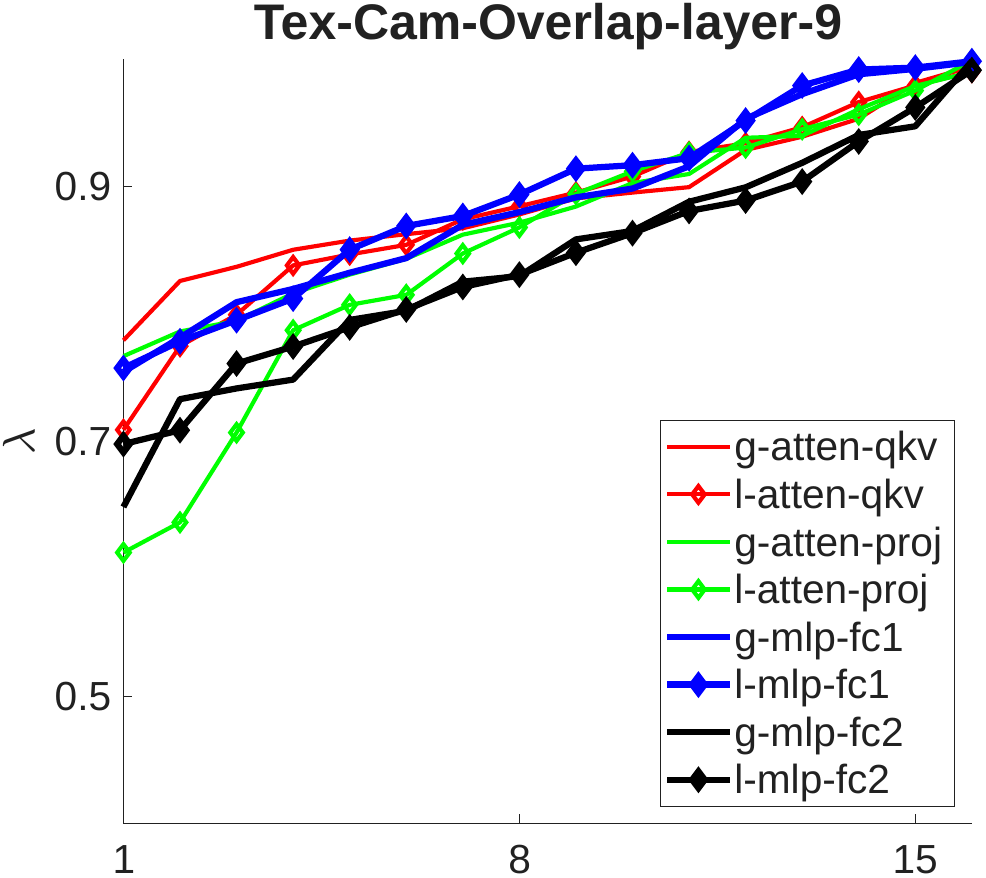}
& 
\includegraphics[width=0.21\textwidth]{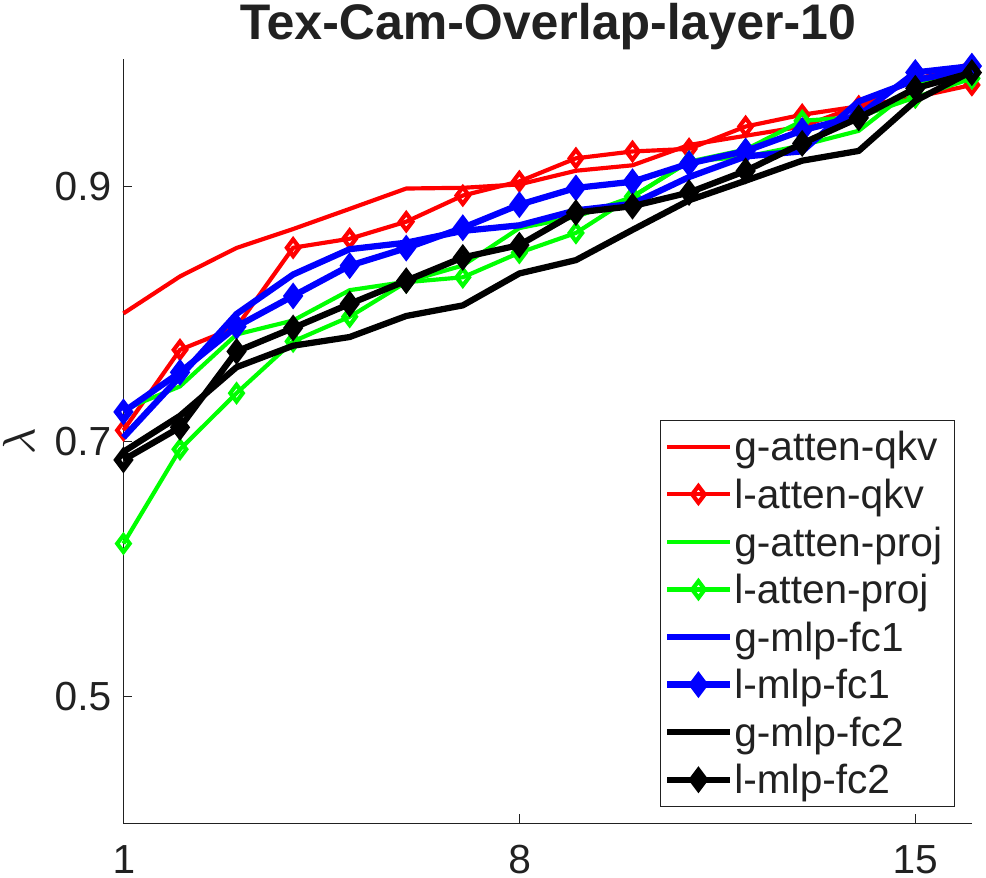}
&
\includegraphics[width=0.21\textwidth]{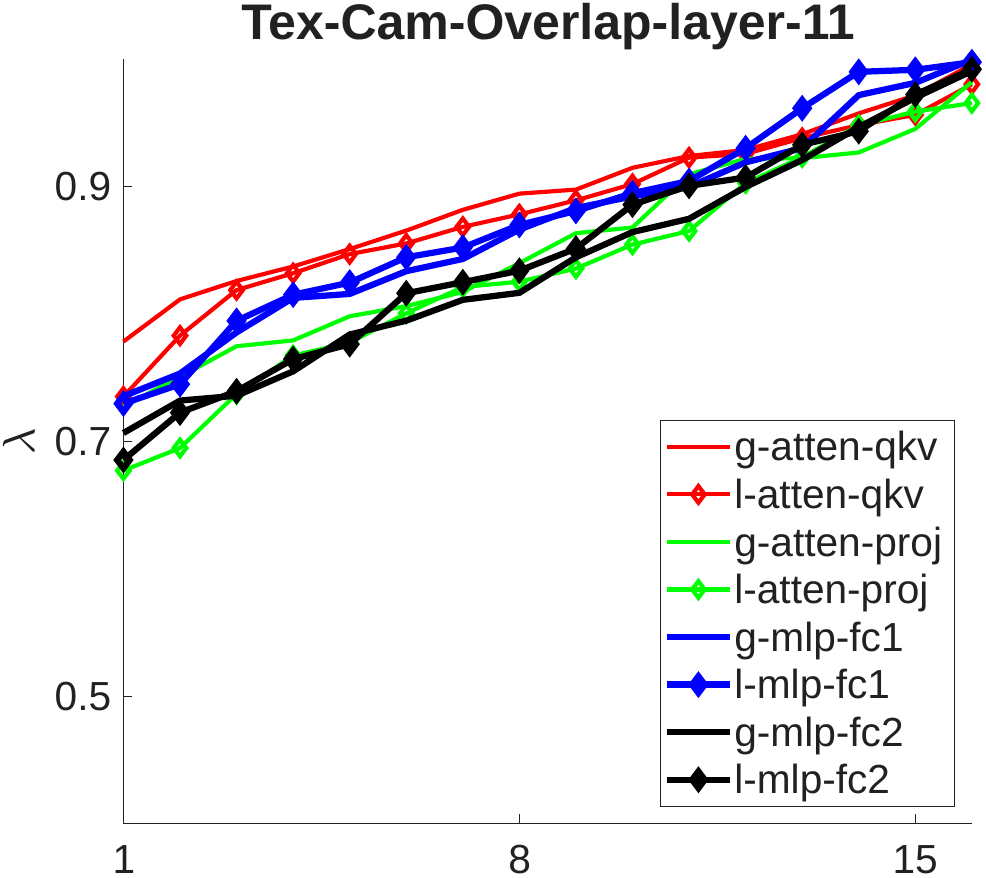}
&
\includegraphics[width=0.21\textwidth]{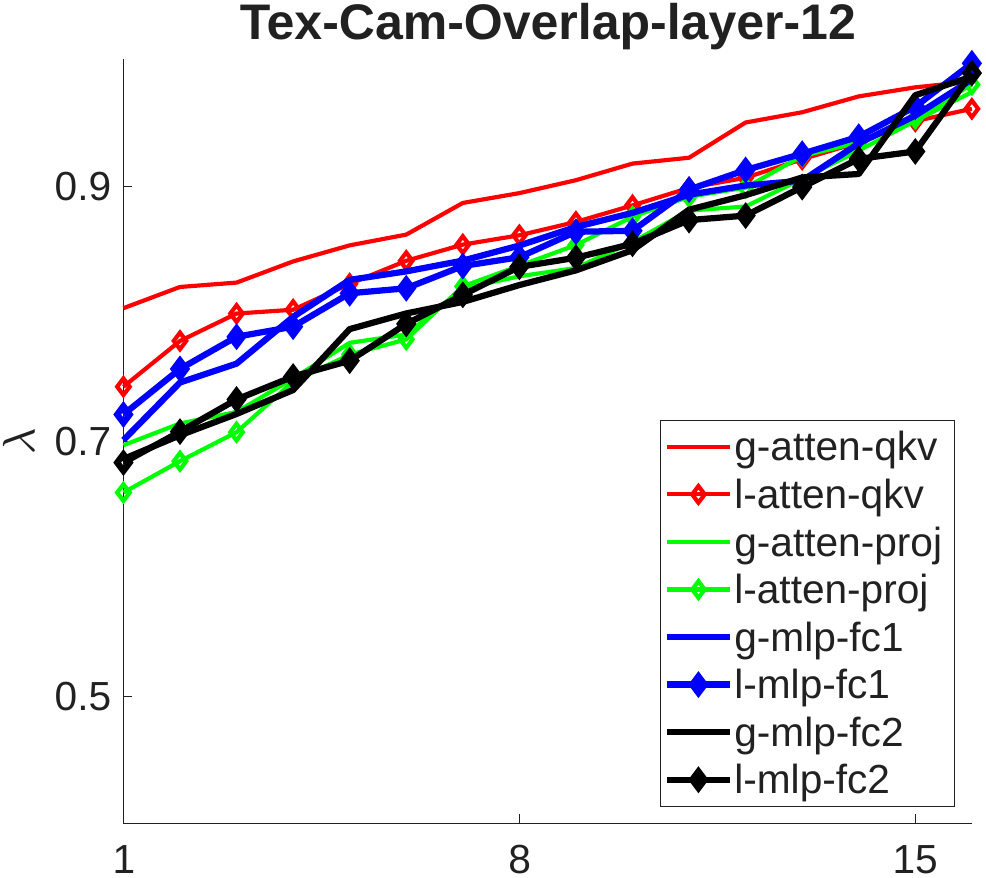}      
\\
\includegraphics[width=0.21\textwidth]{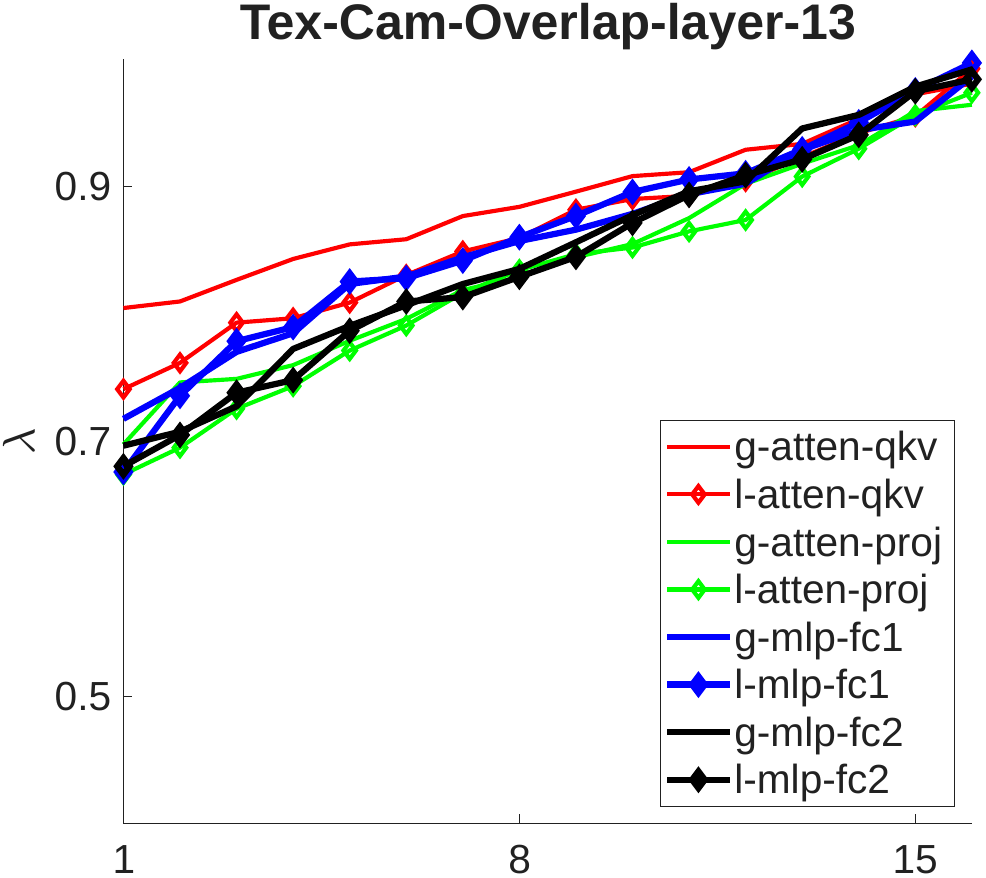}
& 
\includegraphics[width=0.21\textwidth]{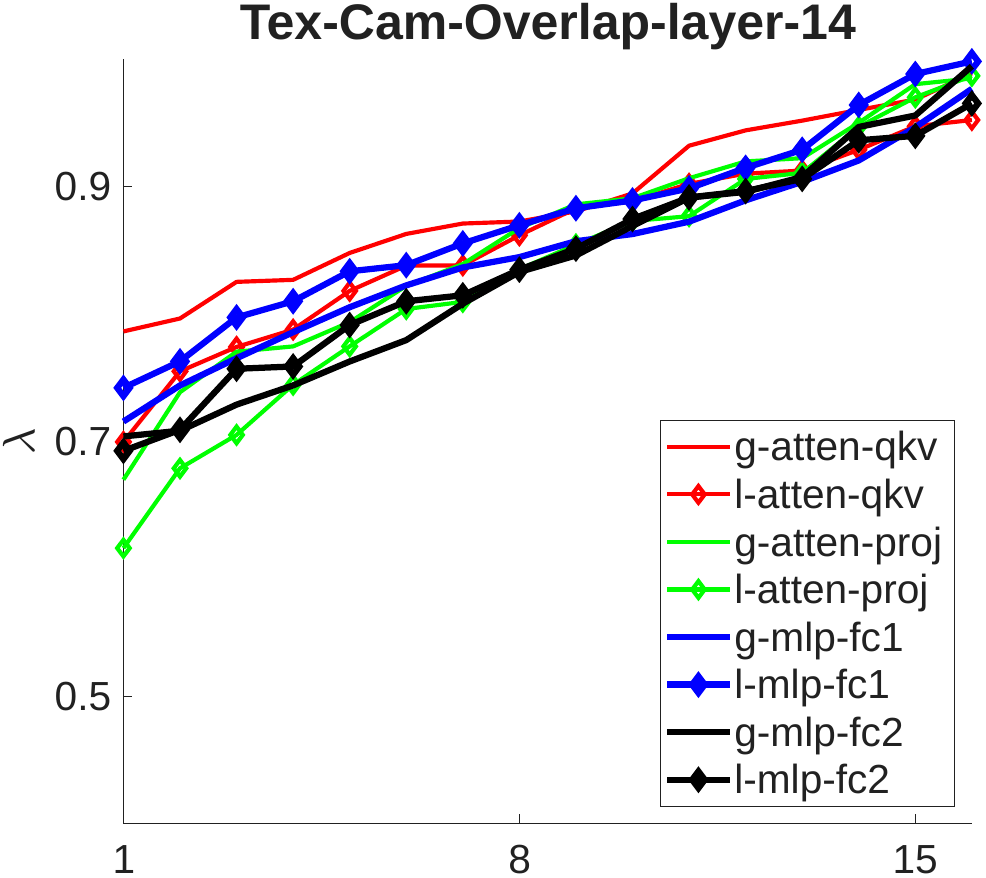}
&
\includegraphics[width=0.21\textwidth]{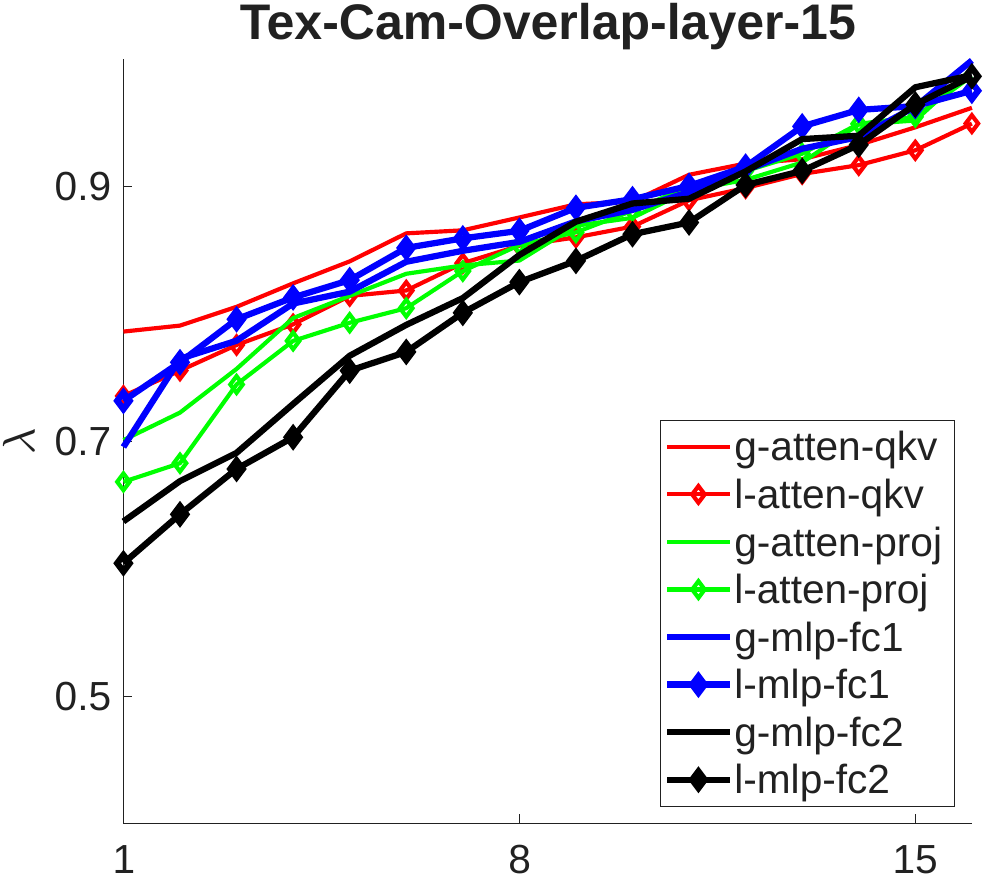}
&
\includegraphics[width=0.21\textwidth]{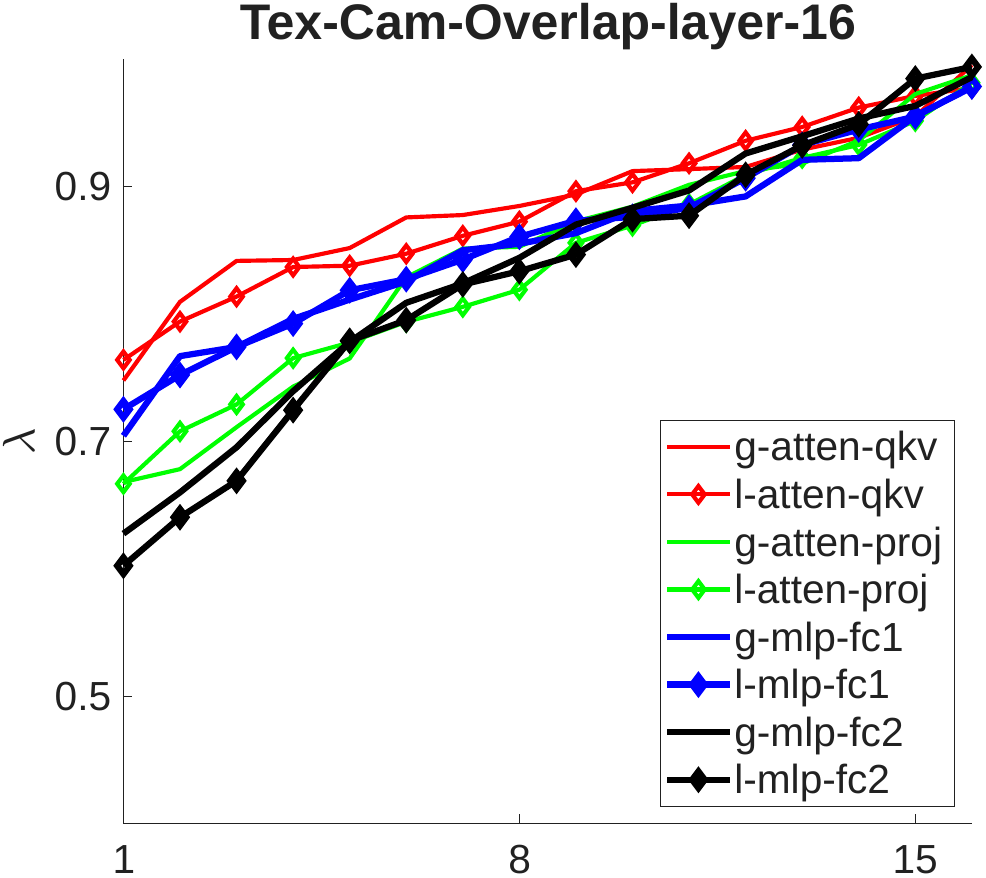}      
\\
\includegraphics[width=0.21\textwidth]{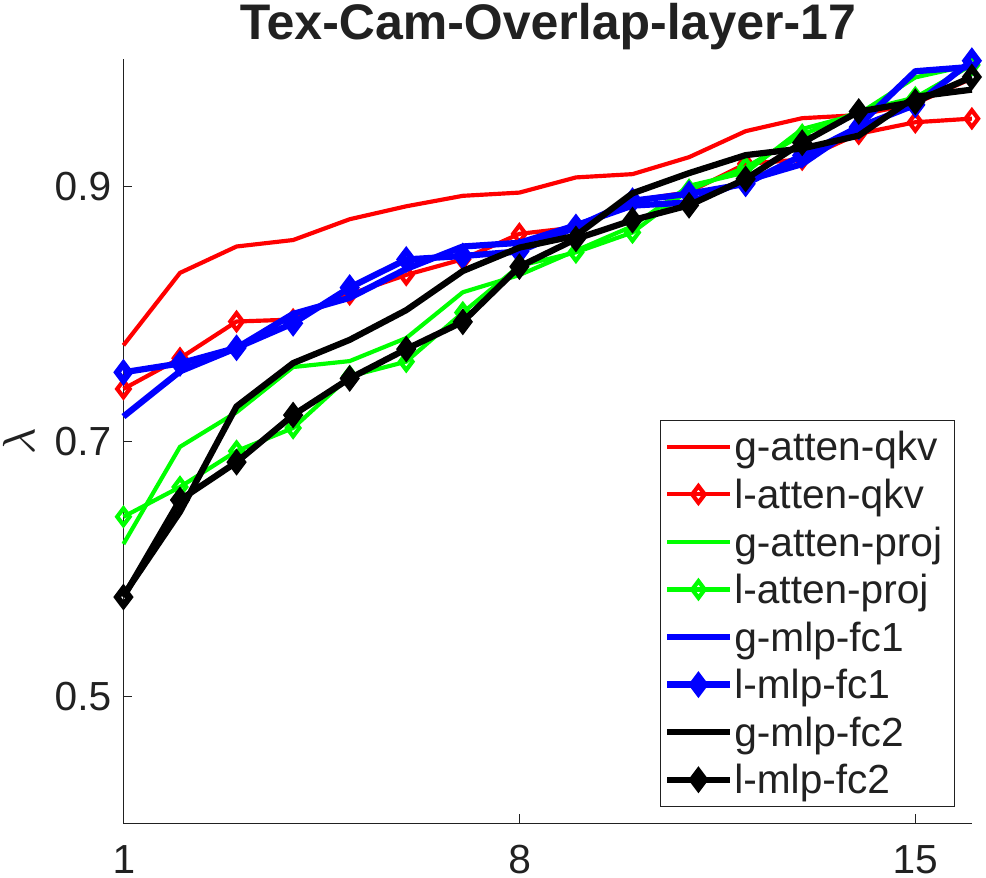}
& 
\includegraphics[width=0.21\textwidth]{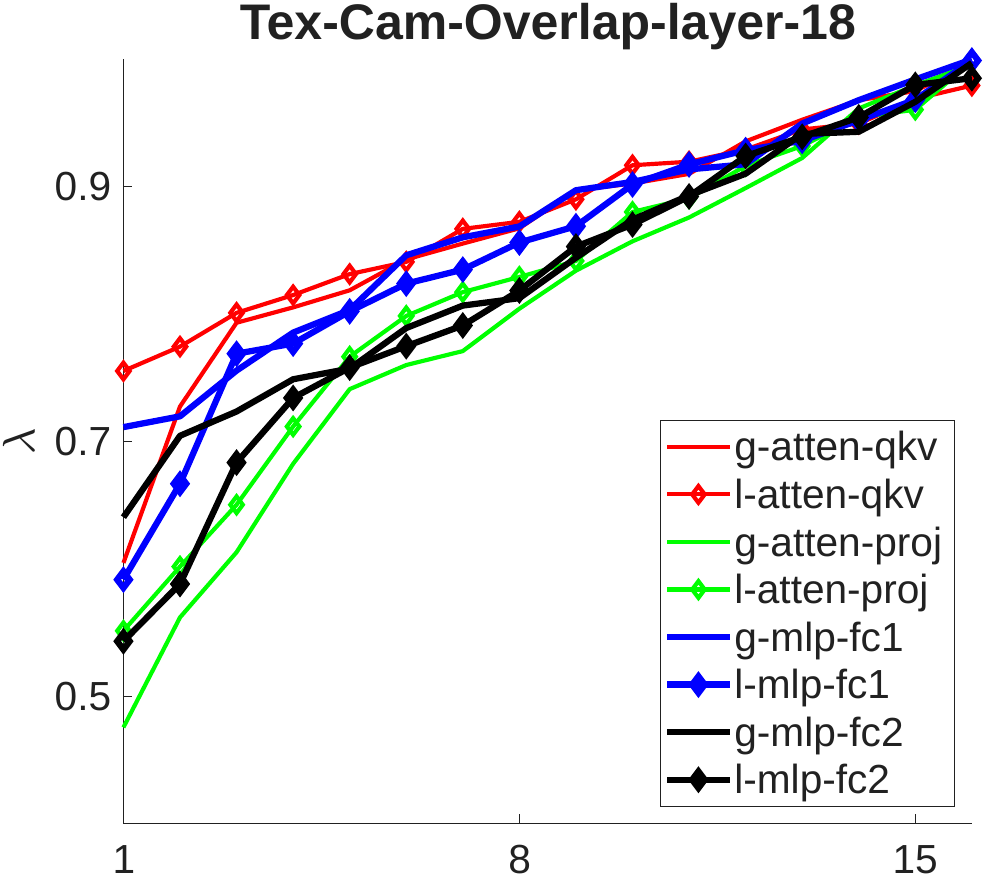}
&
\includegraphics[width=0.21\textwidth]{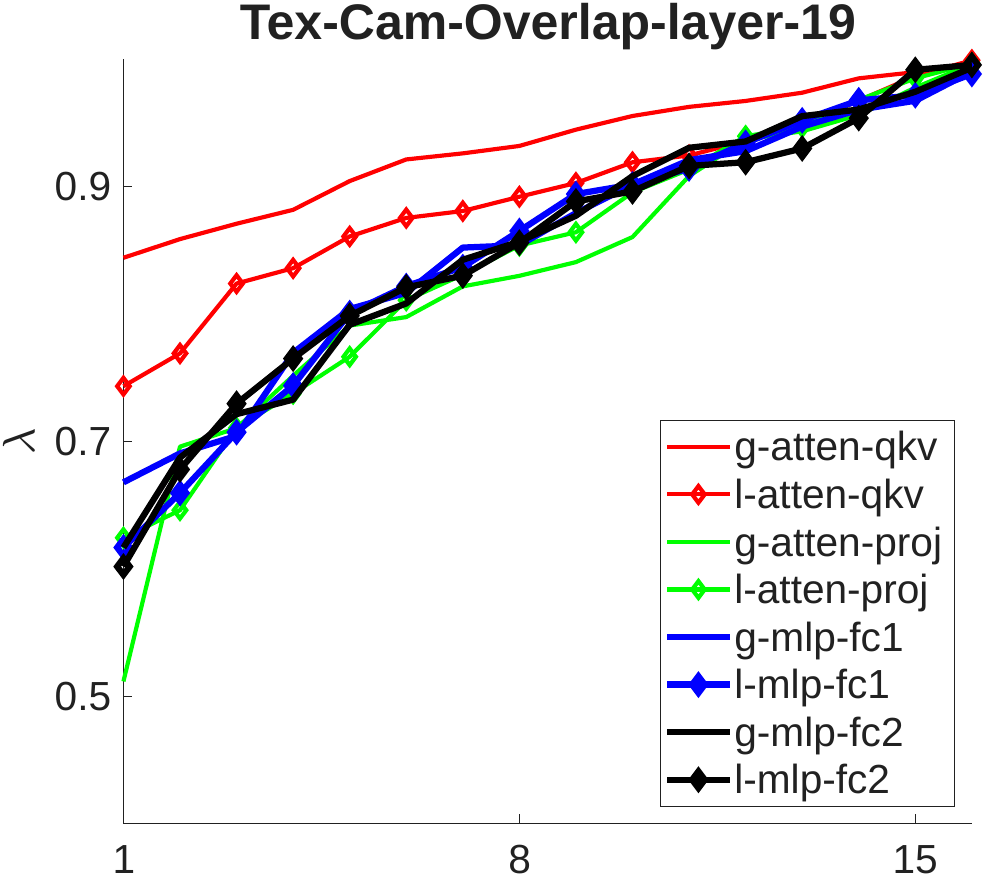}
&
\includegraphics[width=0.21\textwidth]{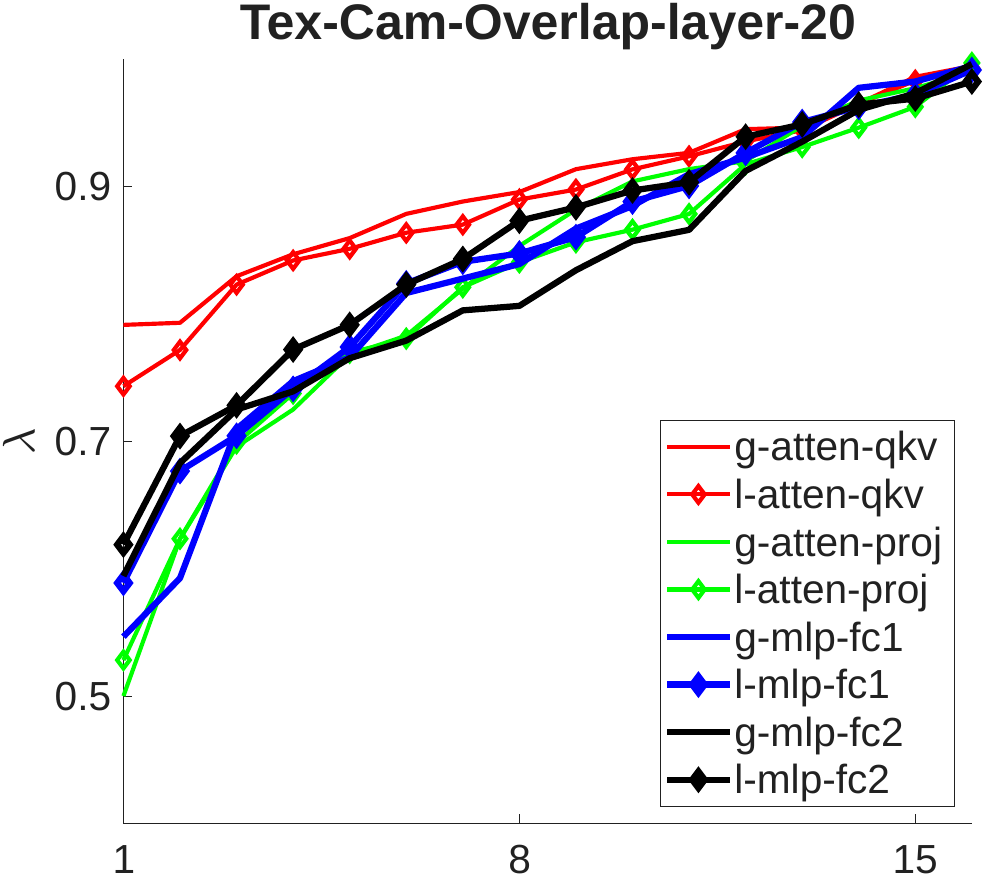}      
\\
\includegraphics[width=0.21\textwidth]{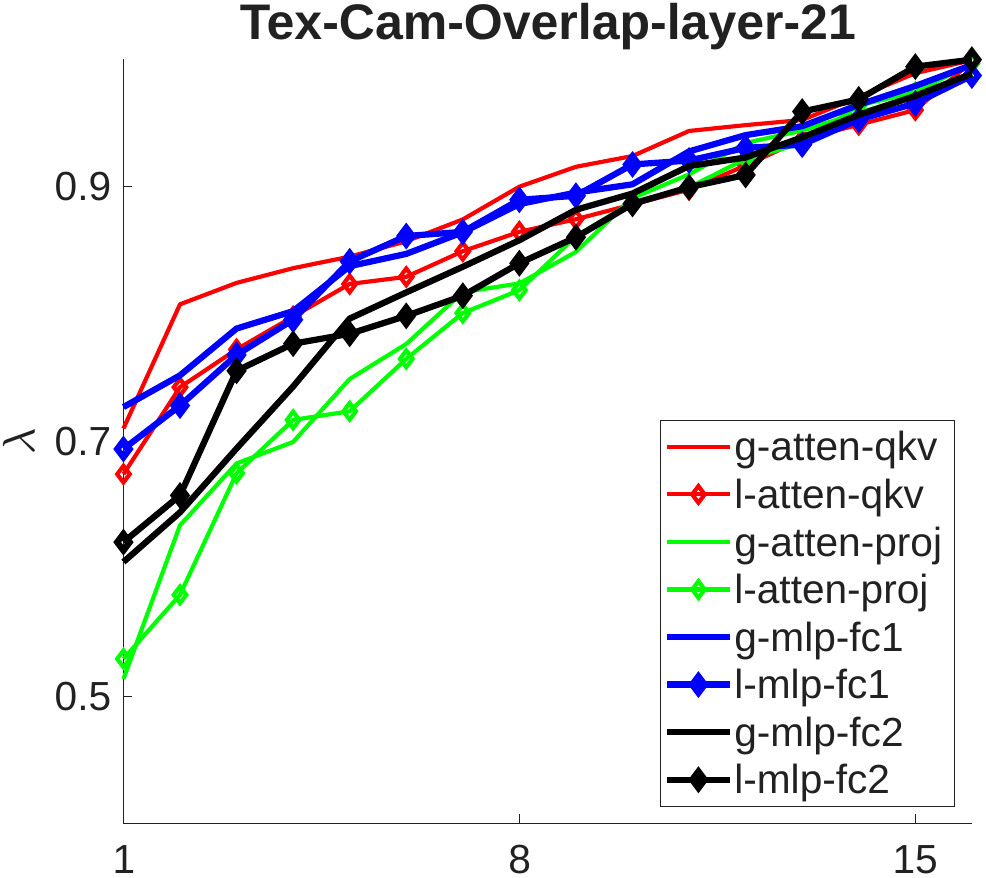}
& 
\includegraphics[width=0.21\textwidth]{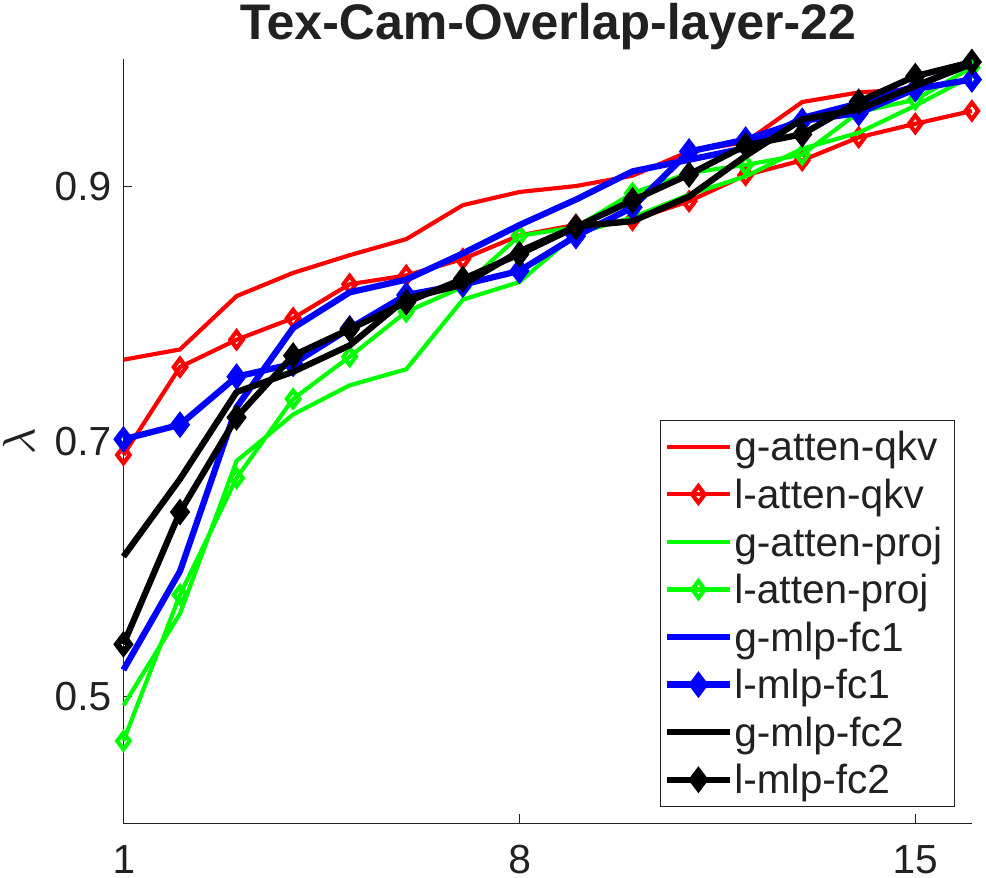}
&
\includegraphics[width=0.21\textwidth]{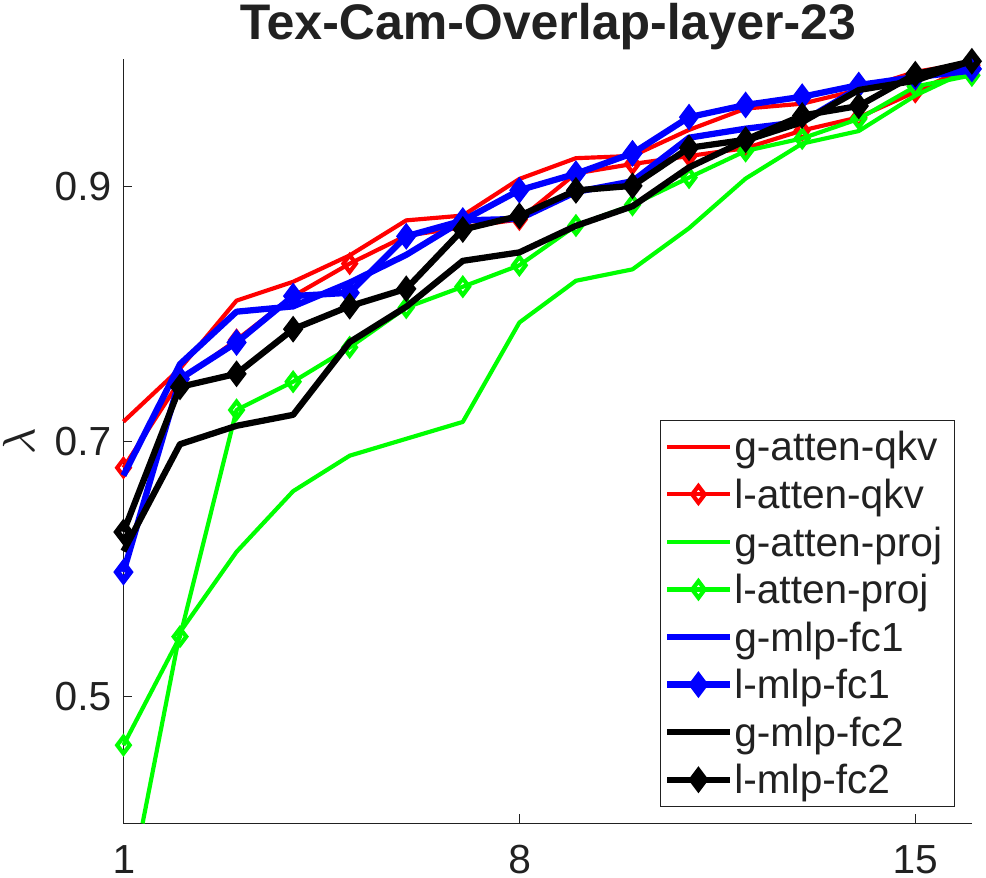}
&
\includegraphics[width=0.21\textwidth]{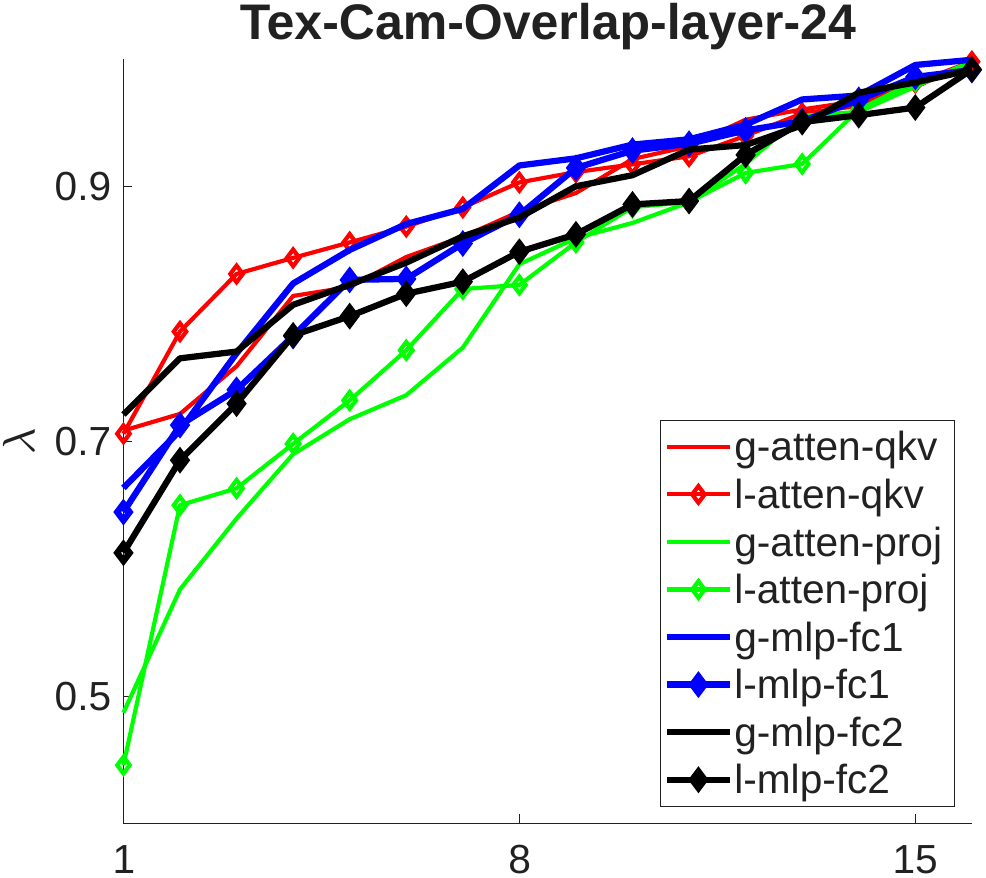}      

\end{tabular}

\captionof{figure}{The overlap ratio between subspaces that correspond to variations in texture and camera motion. }

% \label{Fig:Subspace:Magnitudes}    
\vspace{-3em}
\end{table*}

\clearpage

\subsection{Texture vs Lighting}

\begin{table*}[bp]
\centering
\setlength\tabcolsep{6pt}
\begin{tabular}{cccc}
\includegraphics[width=0.21\textwidth]{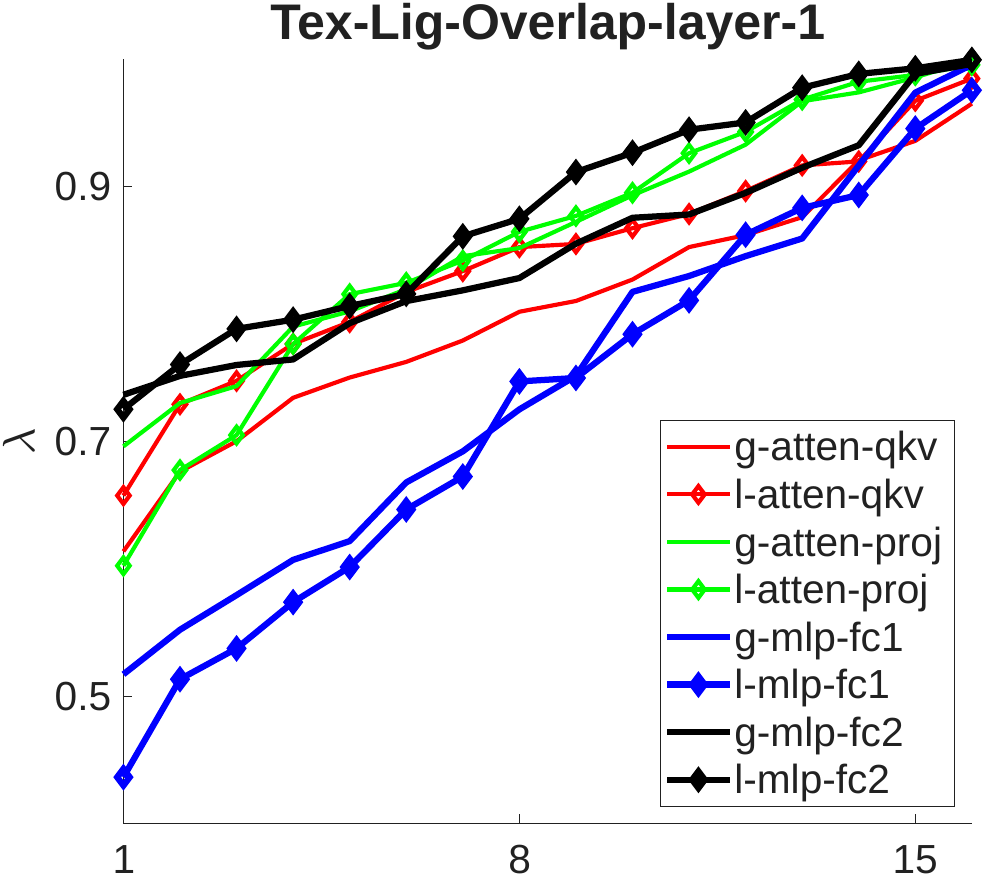}
& 
\includegraphics[width=0.21\textwidth]{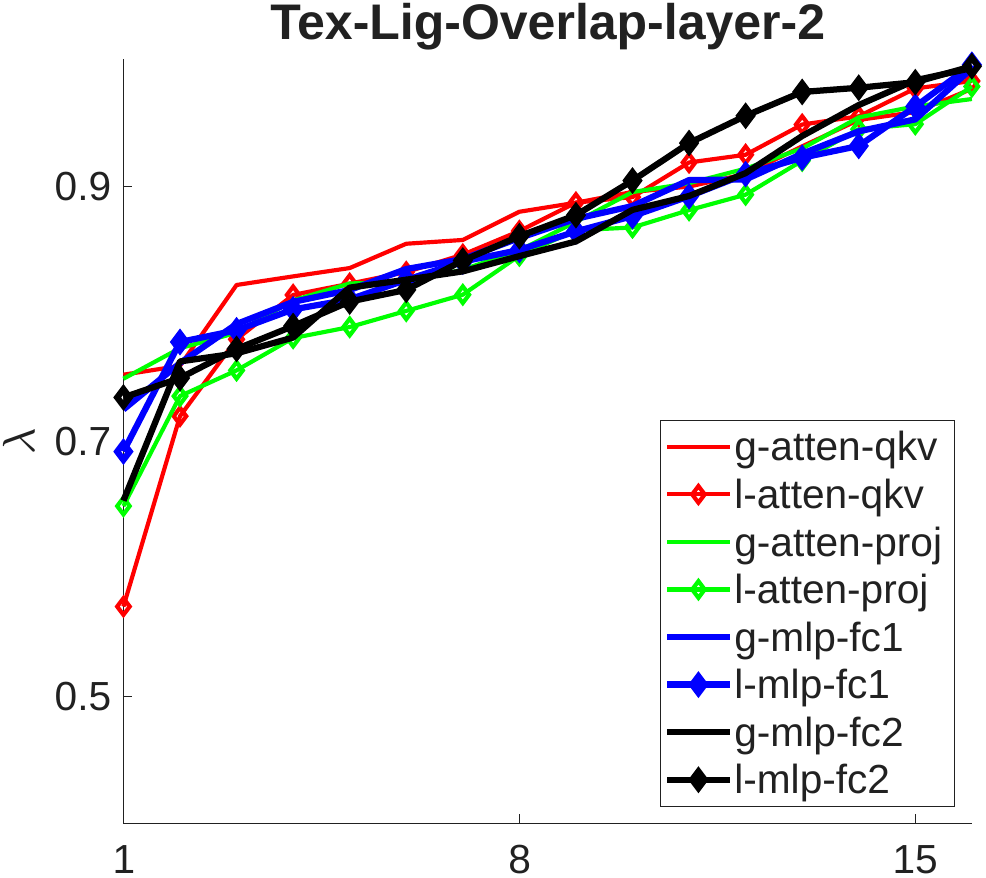}
&
\includegraphics[width=0.21\textwidth]{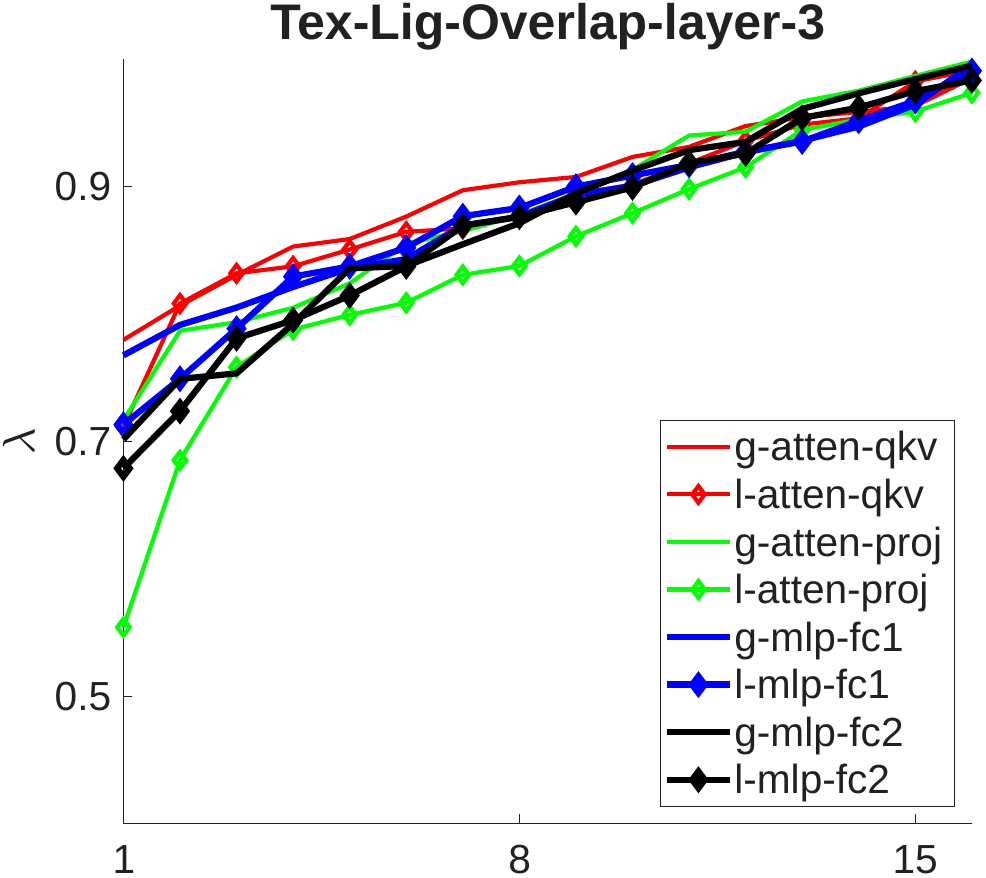}
&
\includegraphics[width=0.21\textwidth]{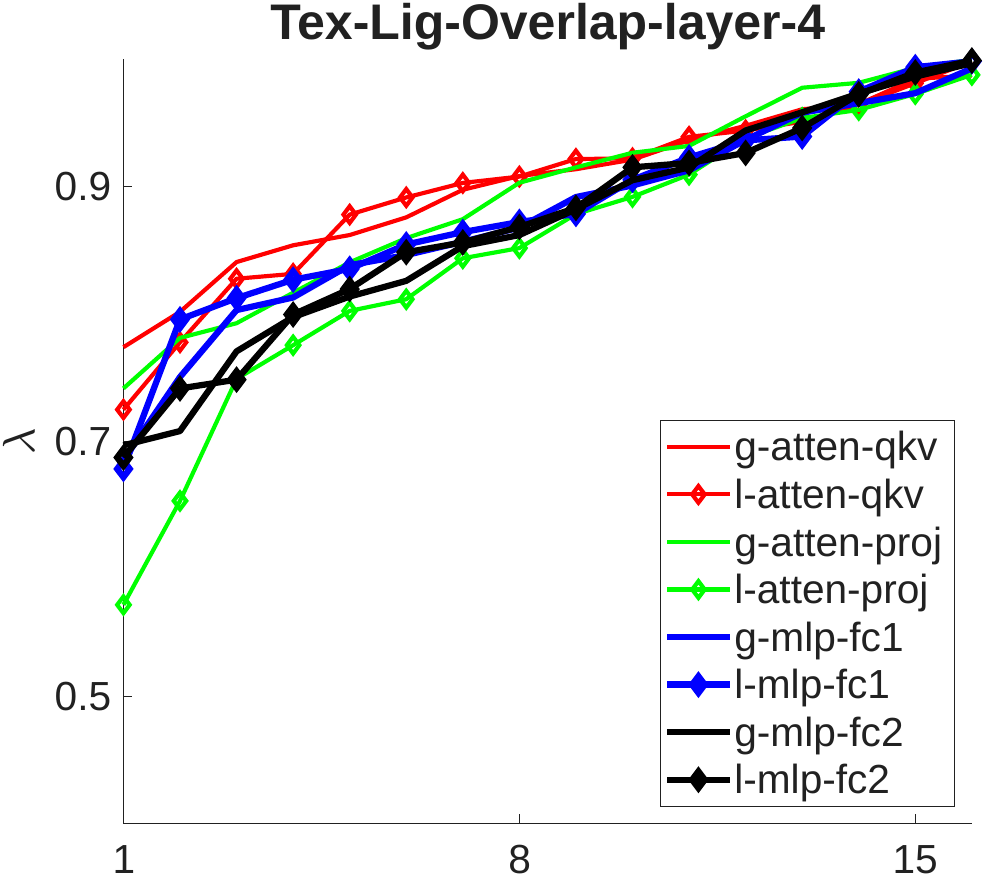}      
\\
\includegraphics[width=0.21\textwidth]{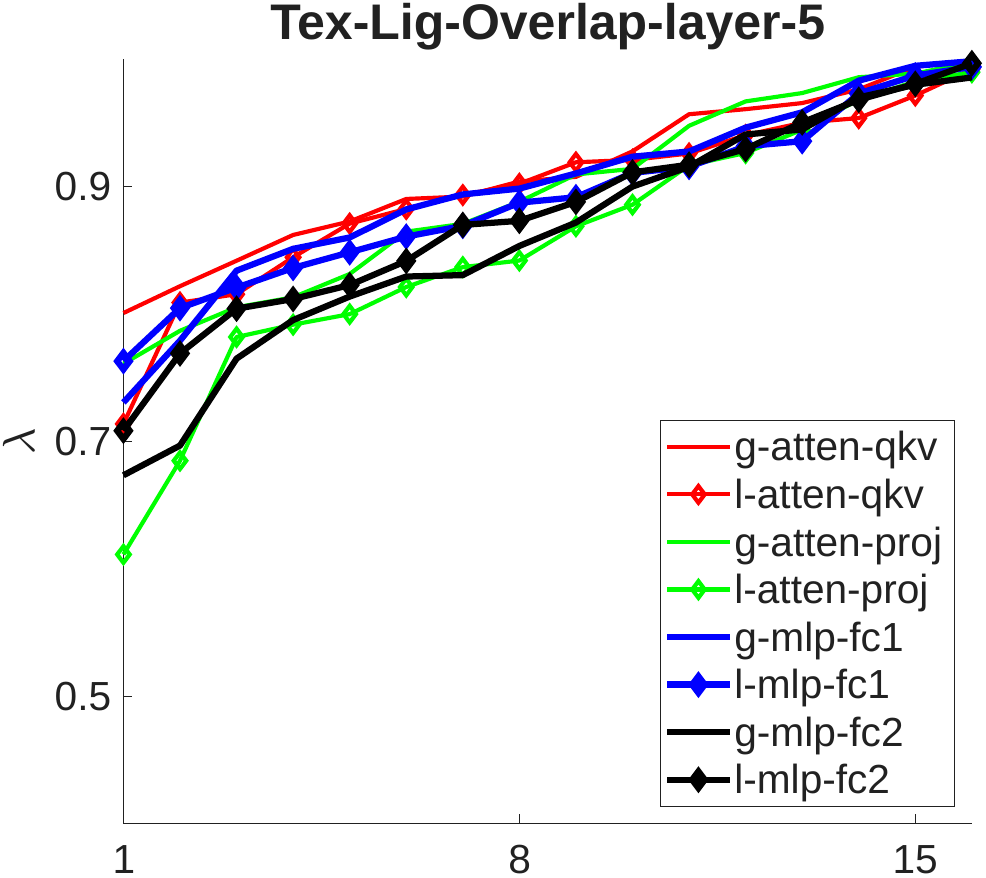}
& 
\includegraphics[width=0.21\textwidth]{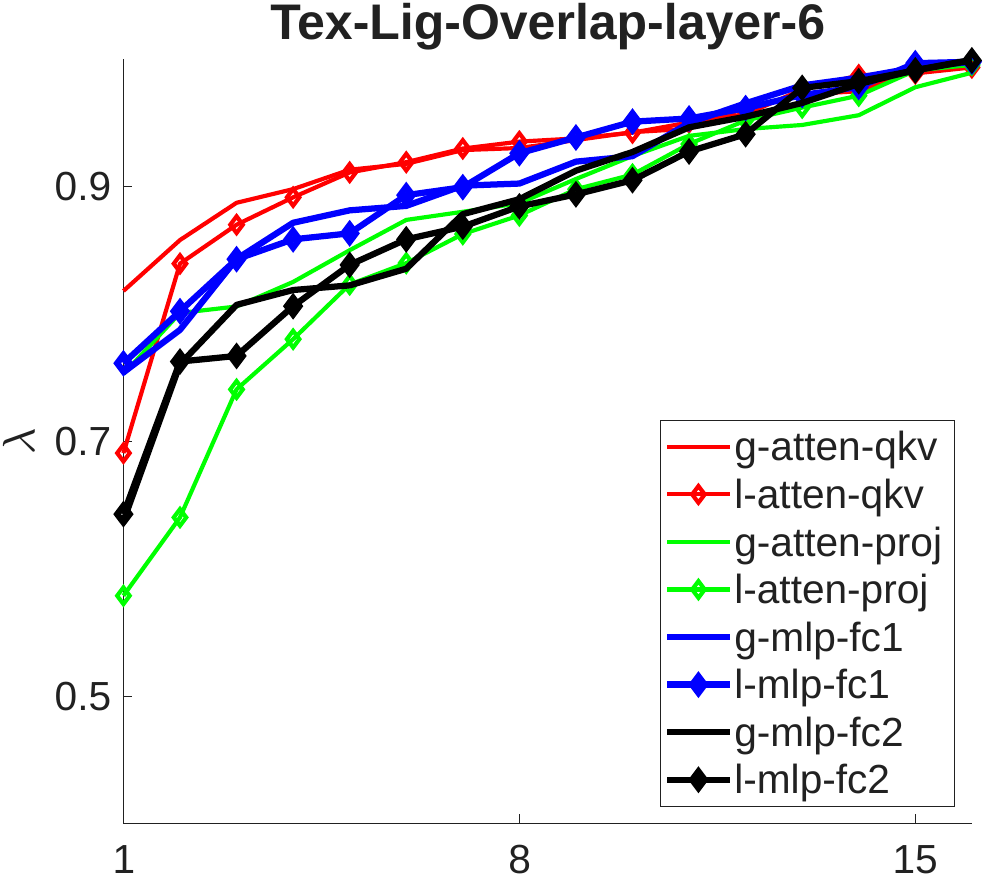}
&
\includegraphics[width=0.21\textwidth]{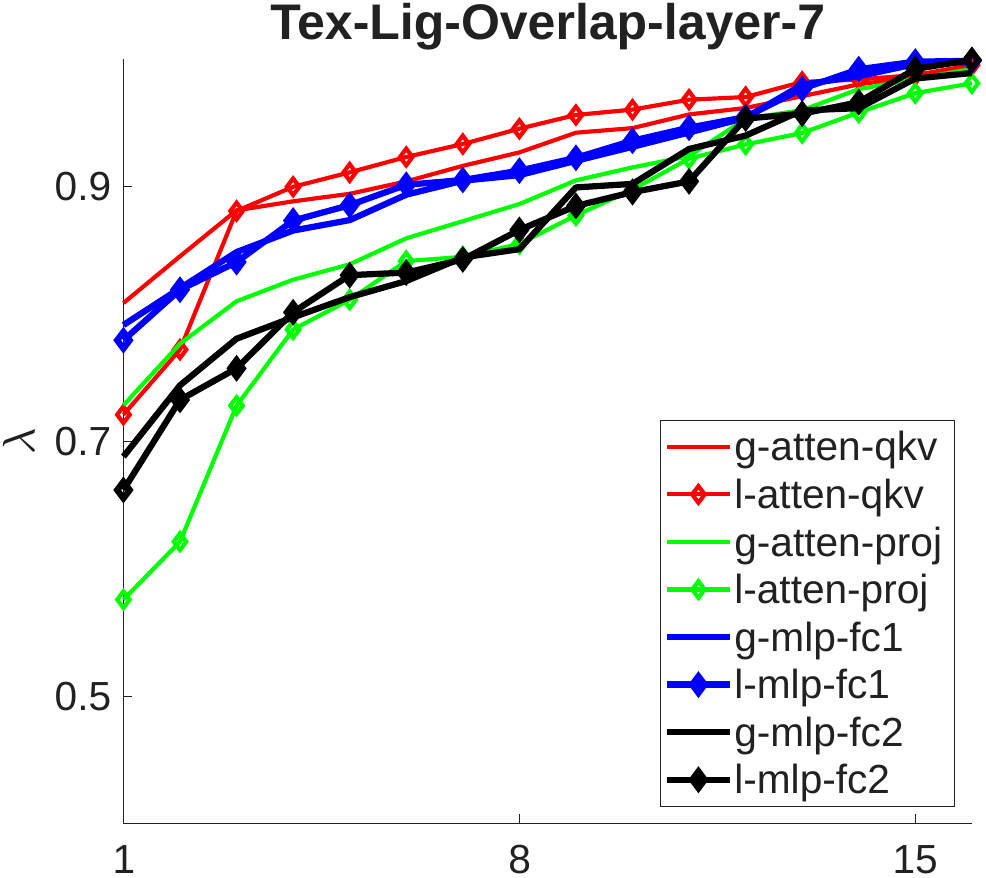}
&
\includegraphics[width=0.21\textwidth]{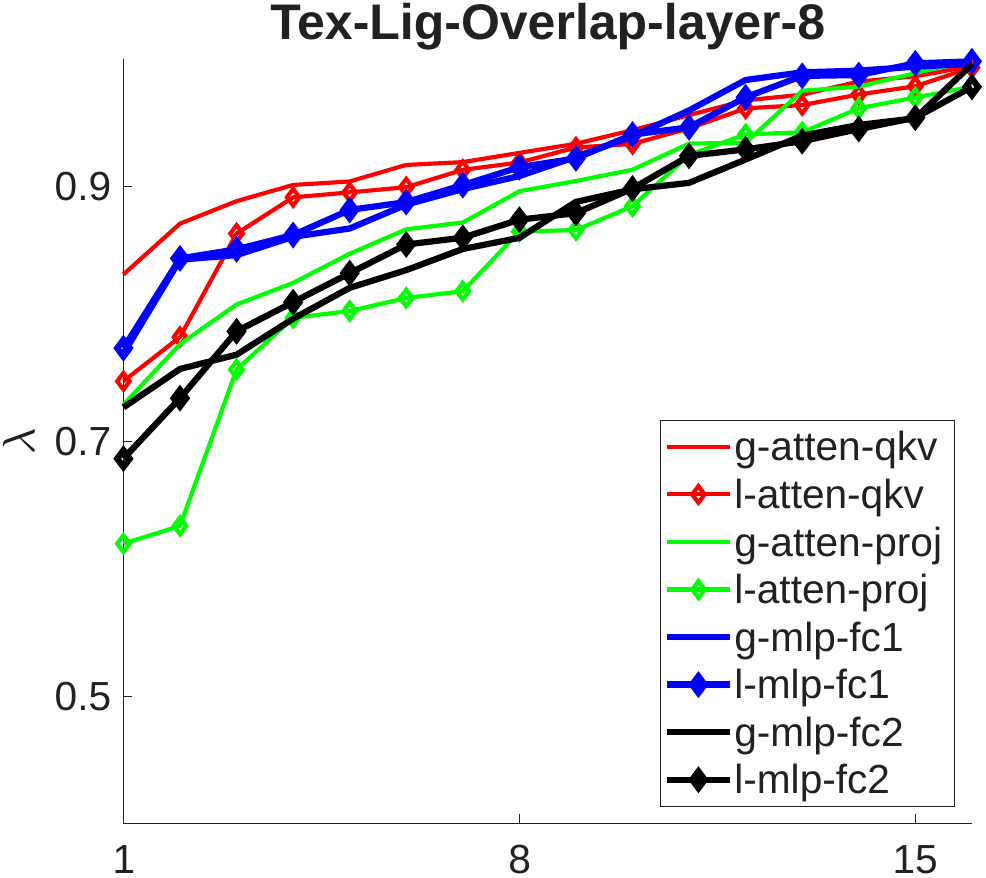}      
\\
\includegraphics[width=0.21\textwidth]{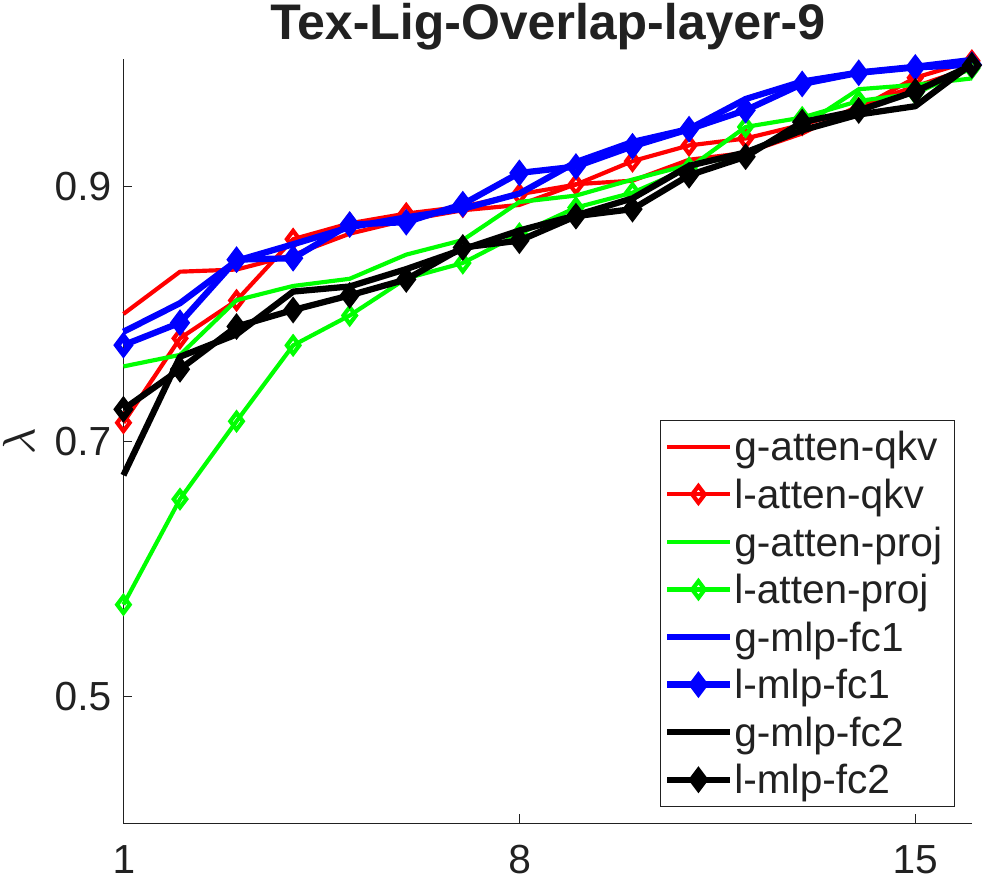}
& 
\includegraphics[width=0.21\textwidth]{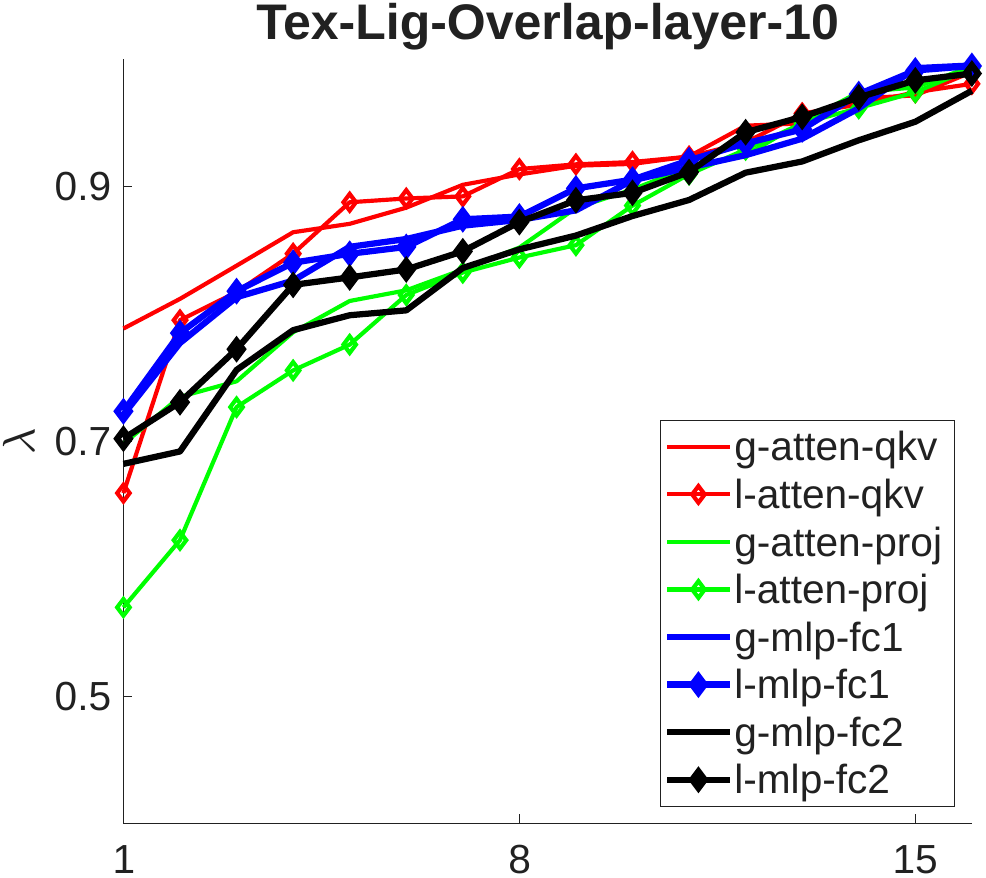}
&
\includegraphics[width=0.21\textwidth]{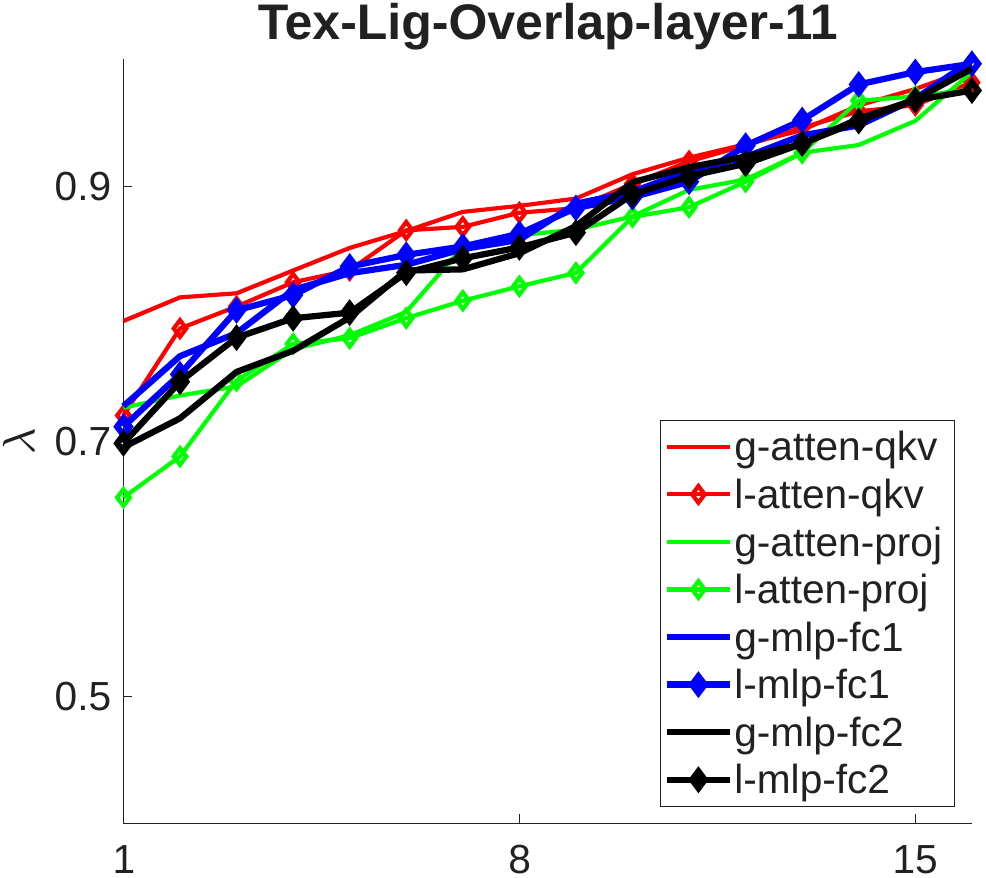}
&
\includegraphics[width=0.21\textwidth]{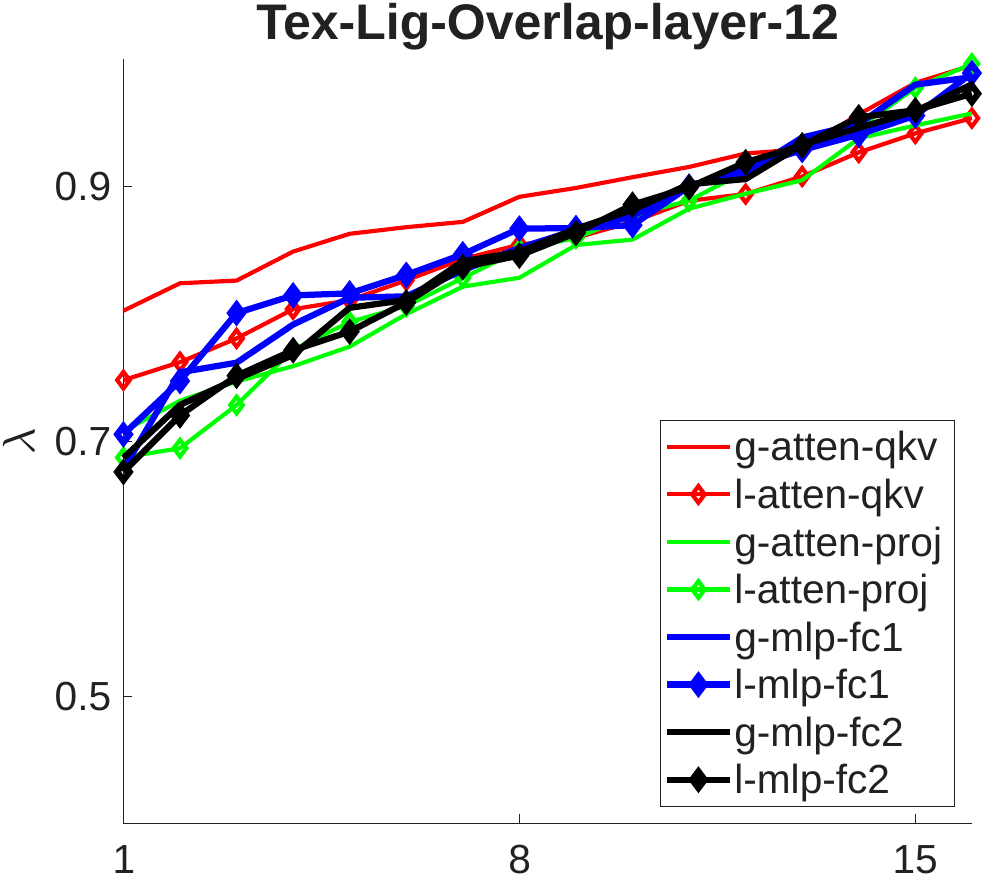}      
\\
\includegraphics[width=0.21\textwidth]{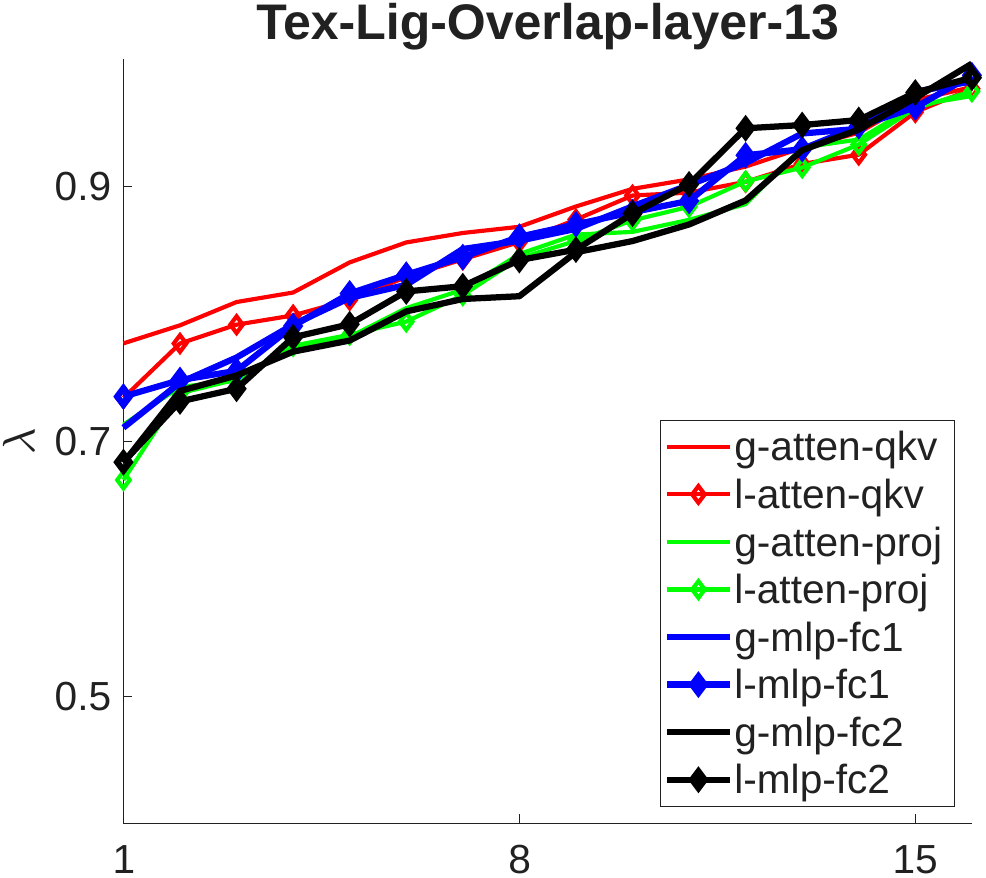}
& 
\includegraphics[width=0.21\textwidth]{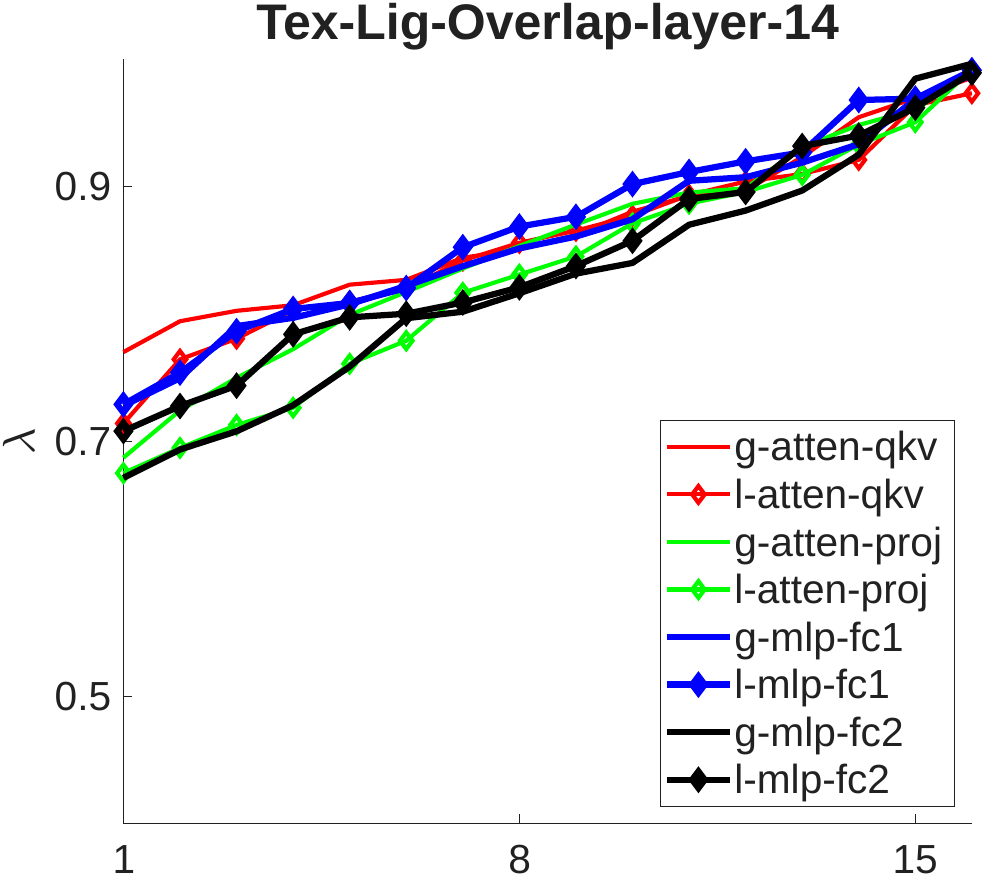}
&
\includegraphics[width=0.21\textwidth]{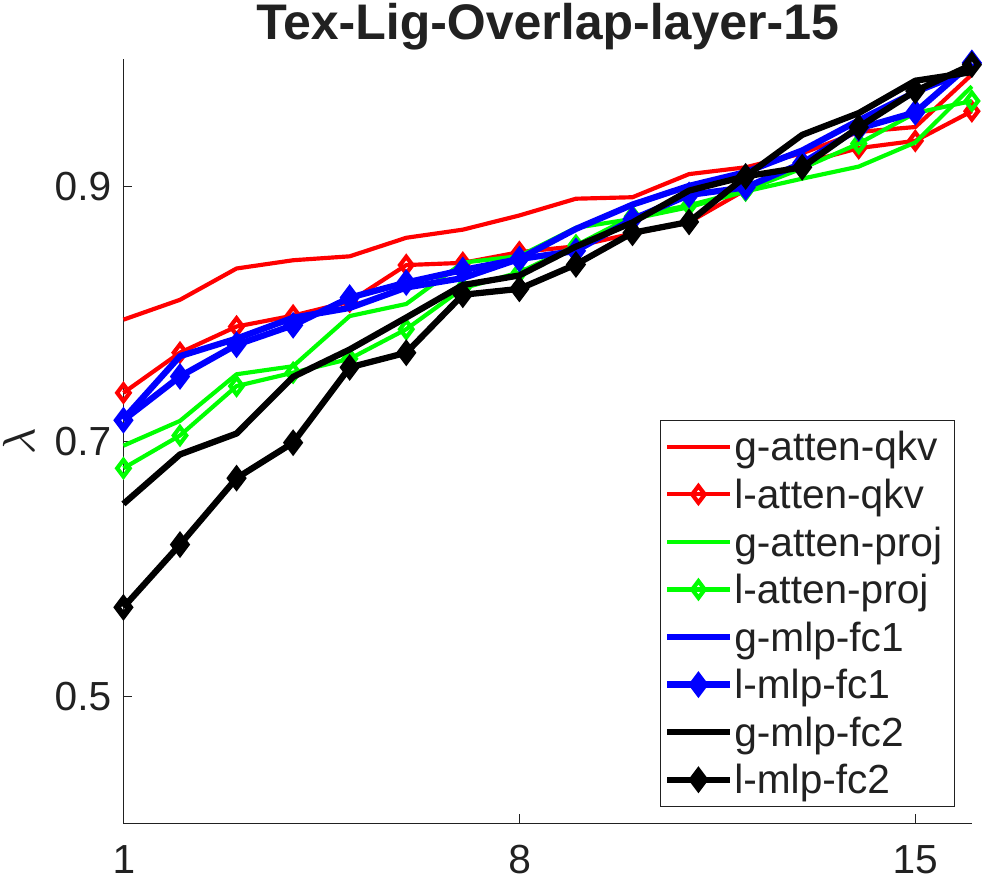}
&
\includegraphics[width=0.21\textwidth]{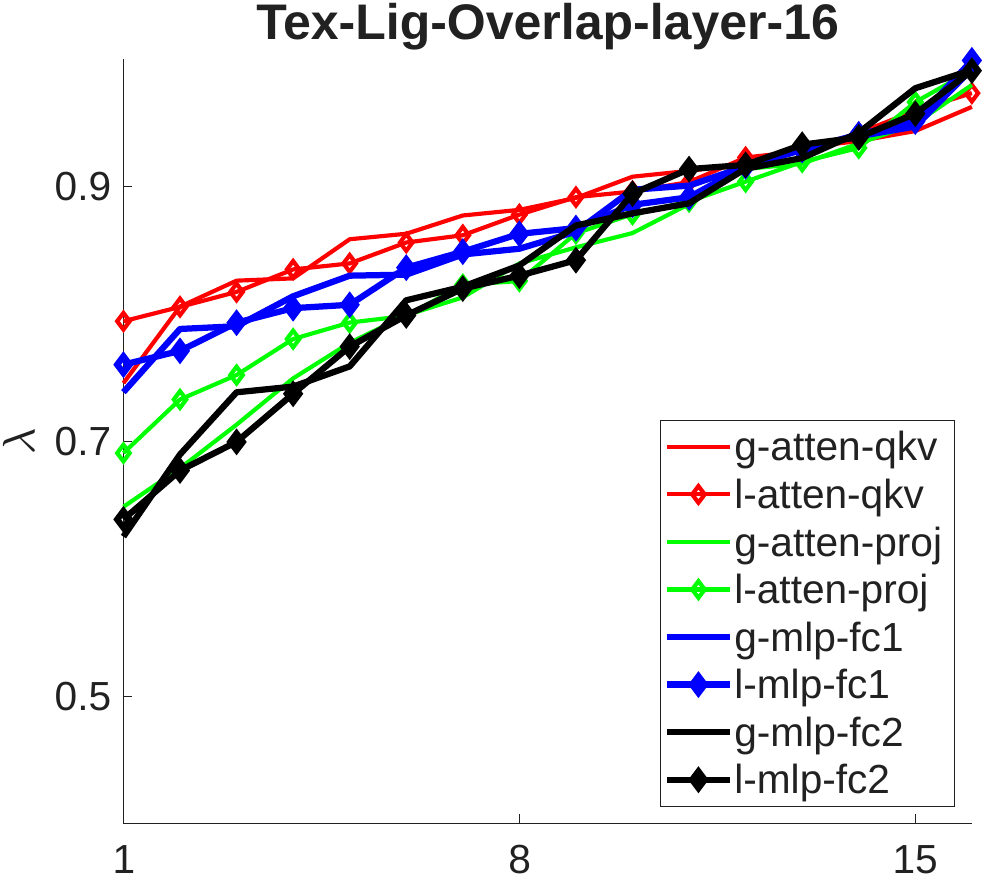}      
\\
\includegraphics[width=0.21\textwidth]{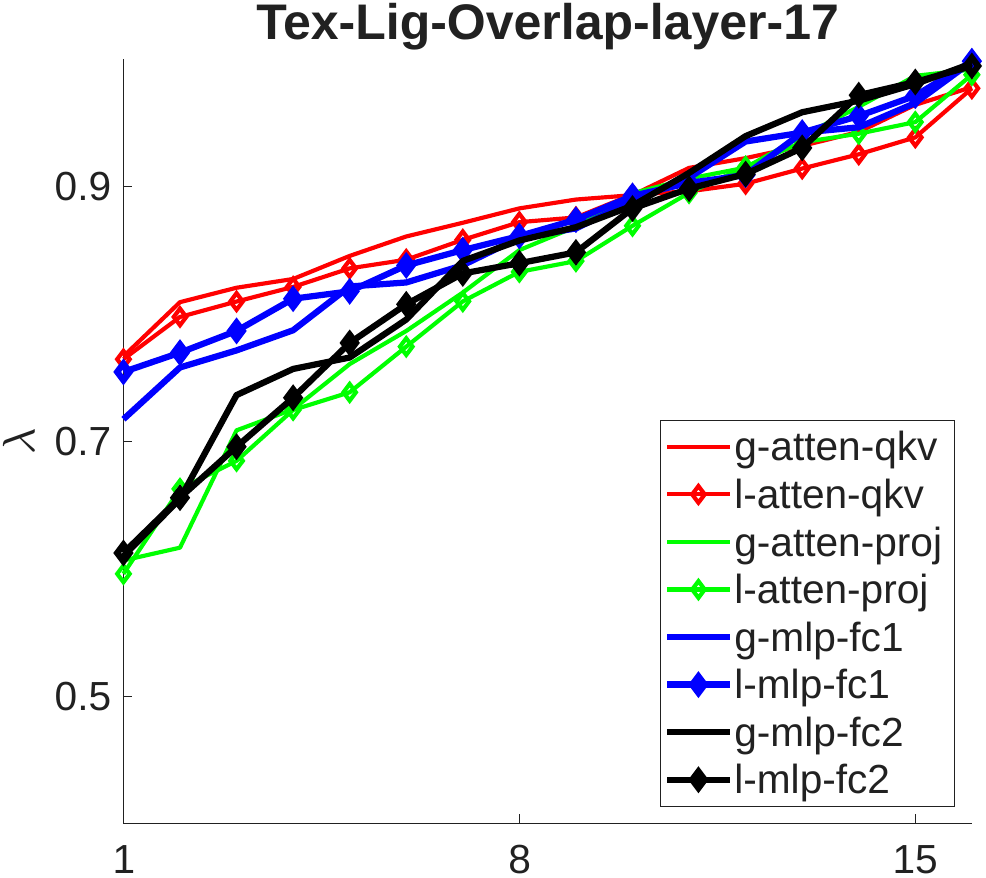}
& 
\includegraphics[width=0.21\textwidth]{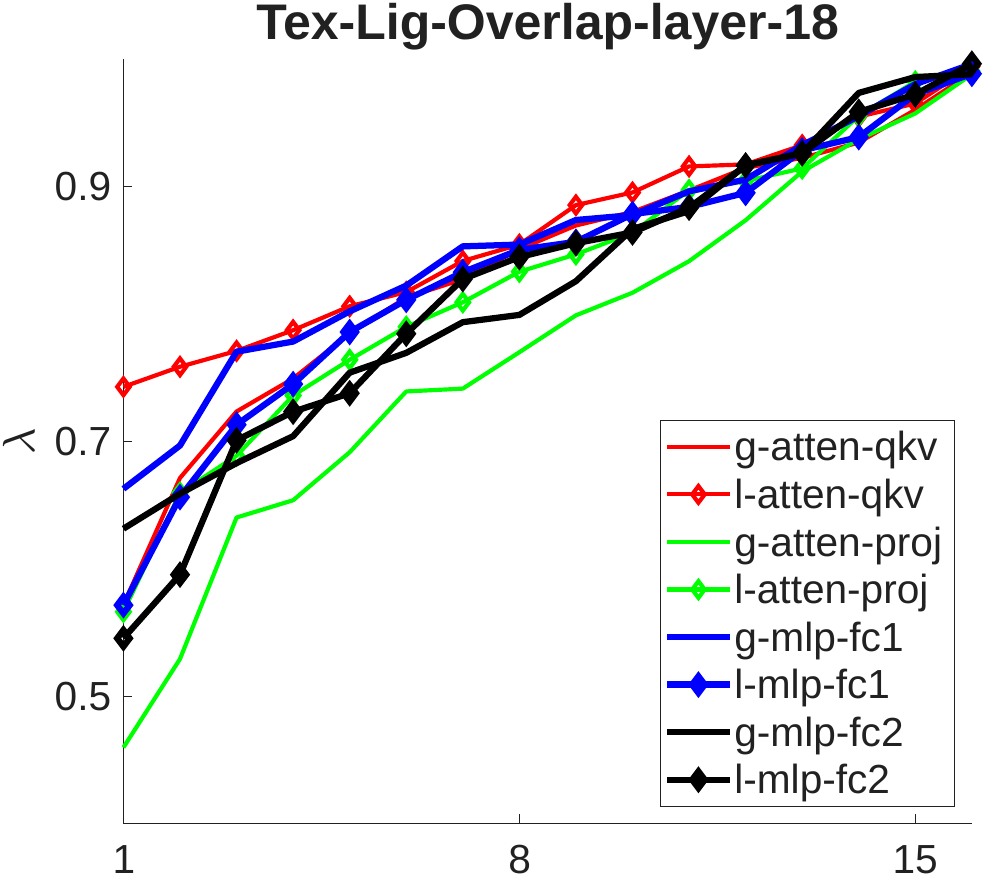}
&
\includegraphics[width=0.21\textwidth]{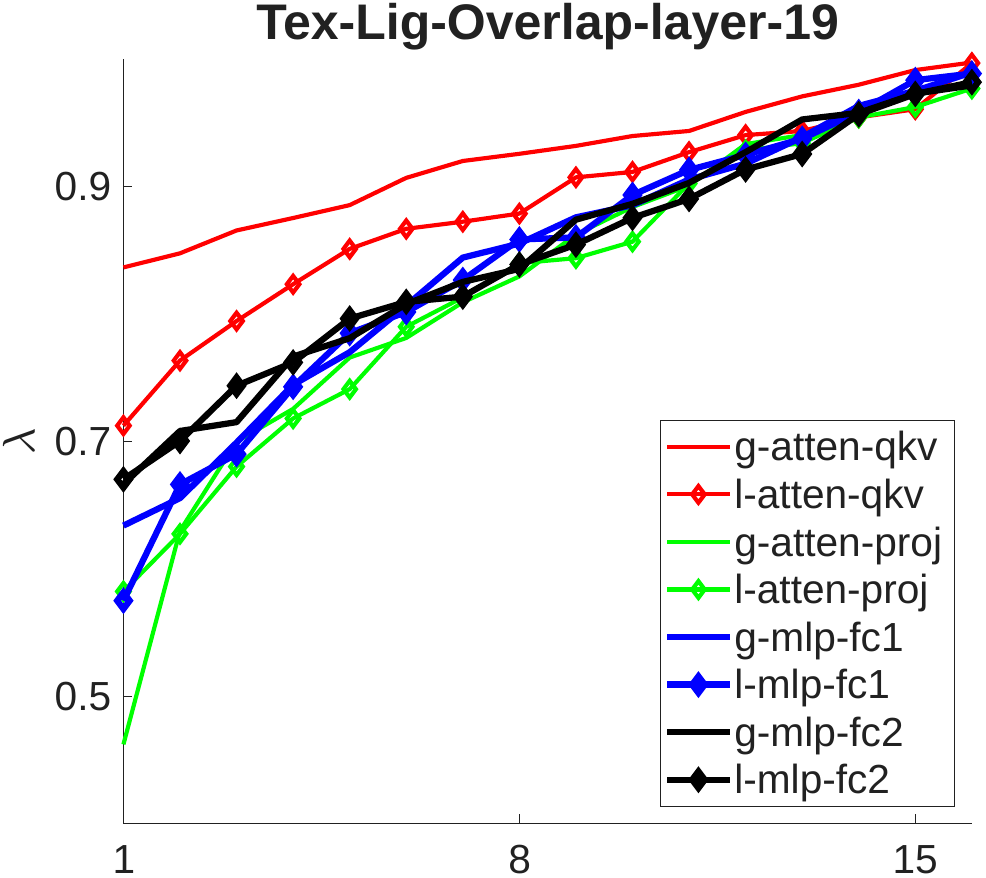}
&
\includegraphics[width=0.21\textwidth]{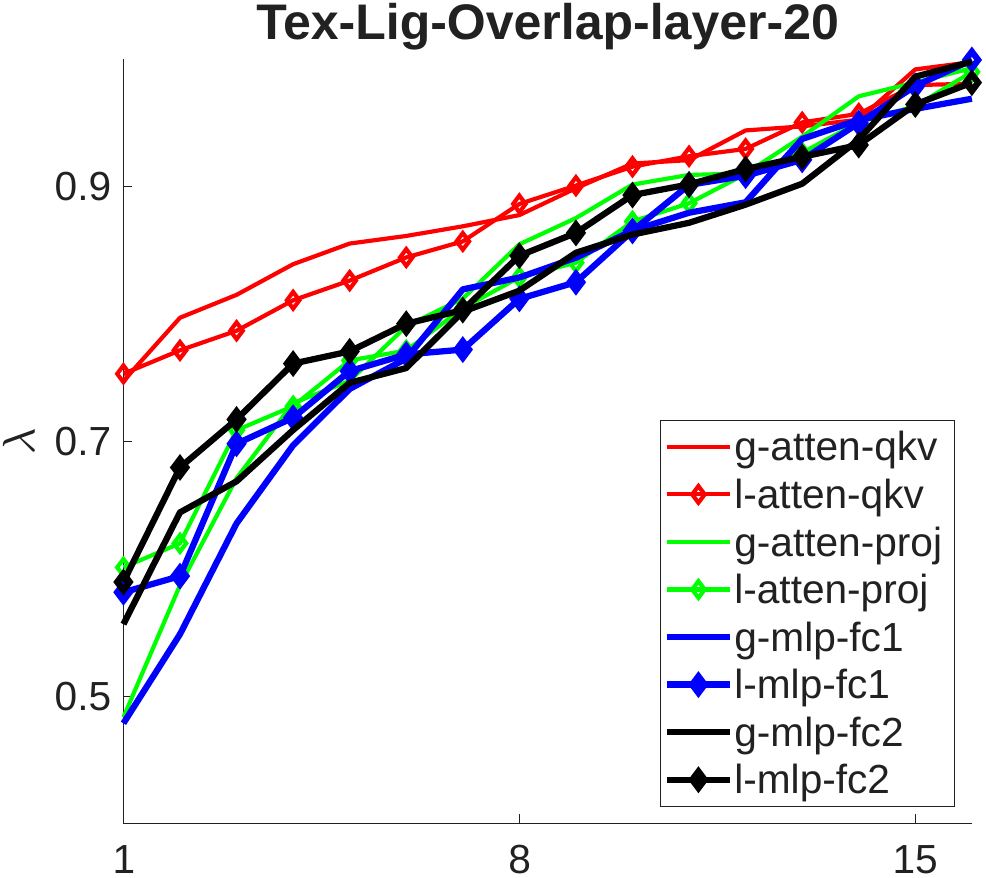}      
\\
\includegraphics[width=0.21\textwidth]{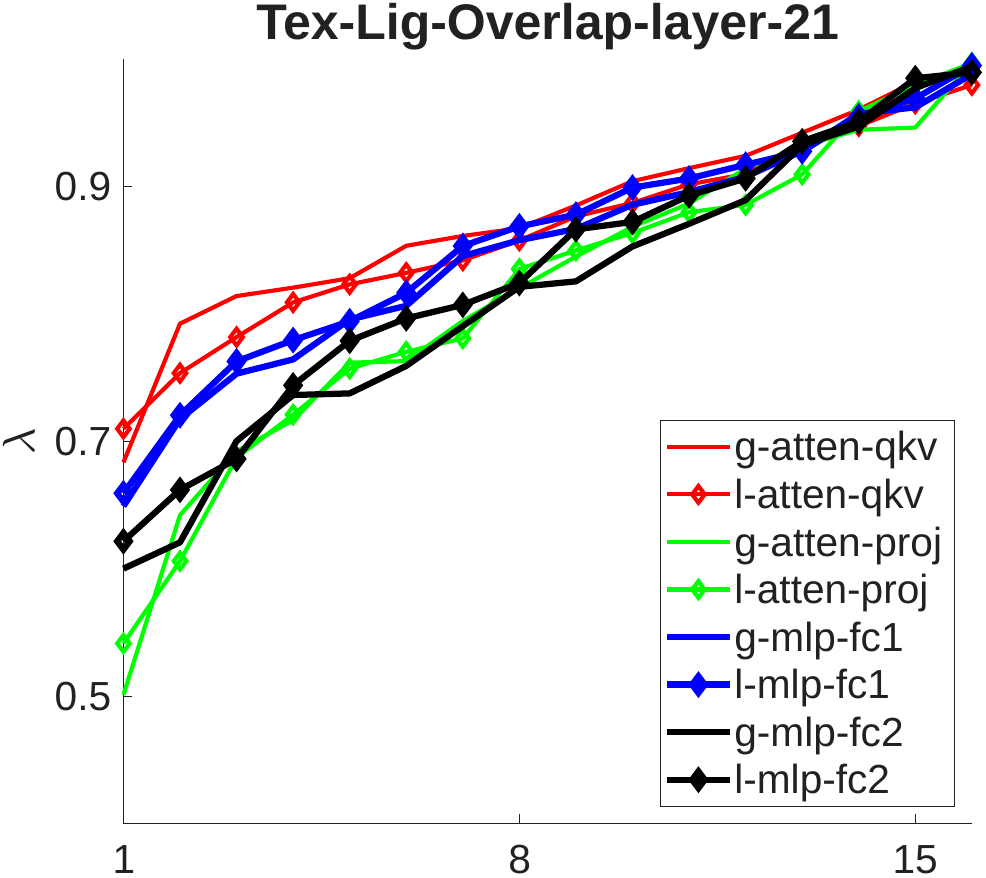}
& 
\includegraphics[width=0.21\textwidth]{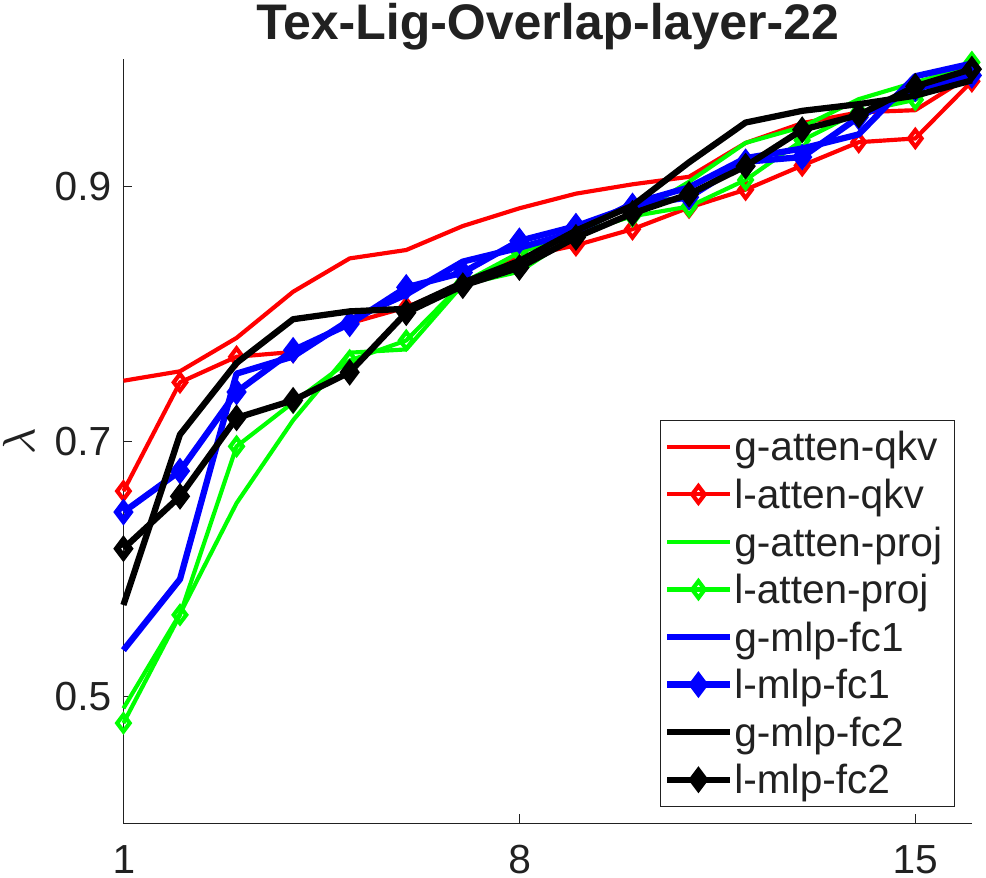}
&
\includegraphics[width=0.21\textwidth]{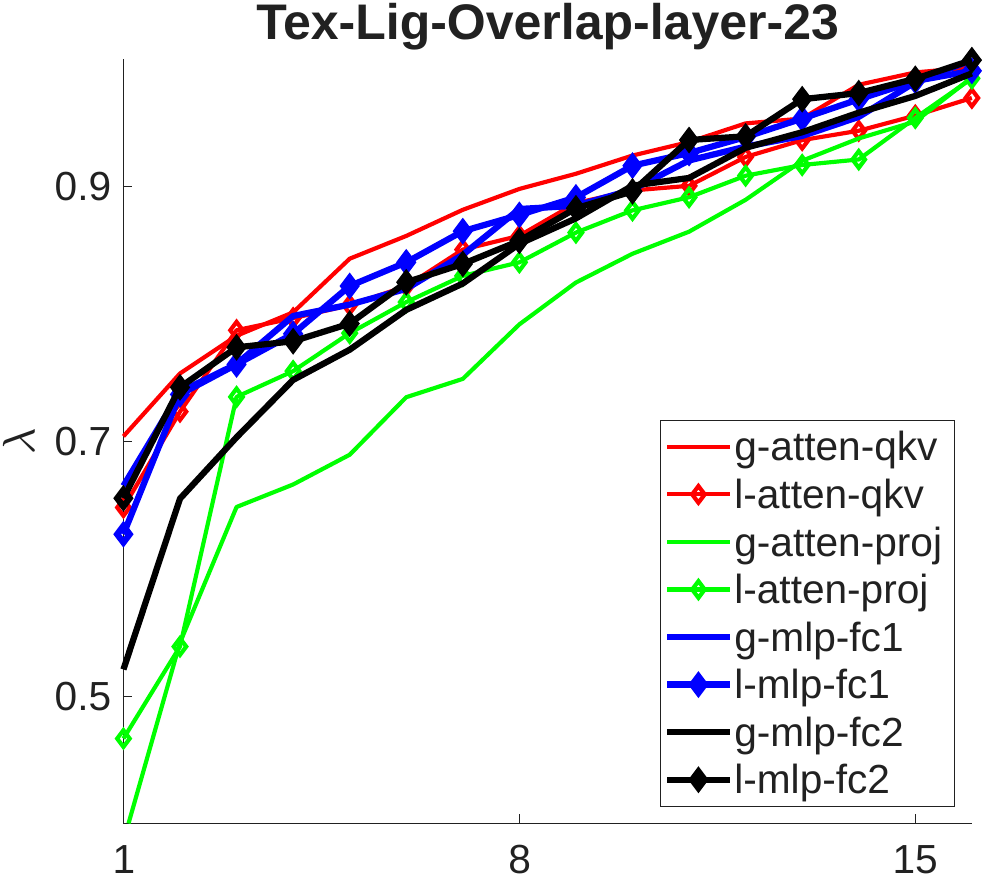}
&
\includegraphics[width=0.21\textwidth]{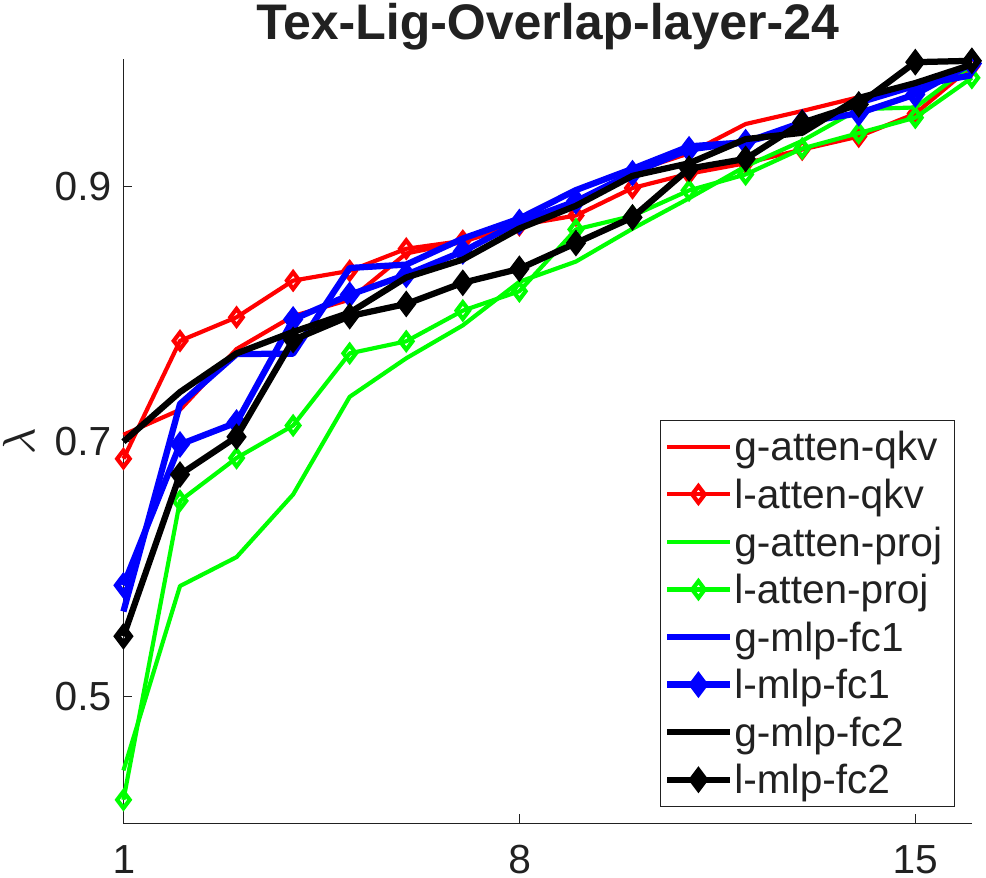}      

\end{tabular}

\captionof{figure}{The overlap ratio between subspaces that correspond to variations in texture and lighting. }

% \label{Fig:Subspace:Magnitudes}    
\vspace{-3em}
\end{table*}

\clearpage

\subsection{Camera vs Lighting}

\begin{table*}[bp]
\centering
\setlength\tabcolsep{6pt}
\begin{tabular}{cccc}
\includegraphics[width=0.21\textwidth]{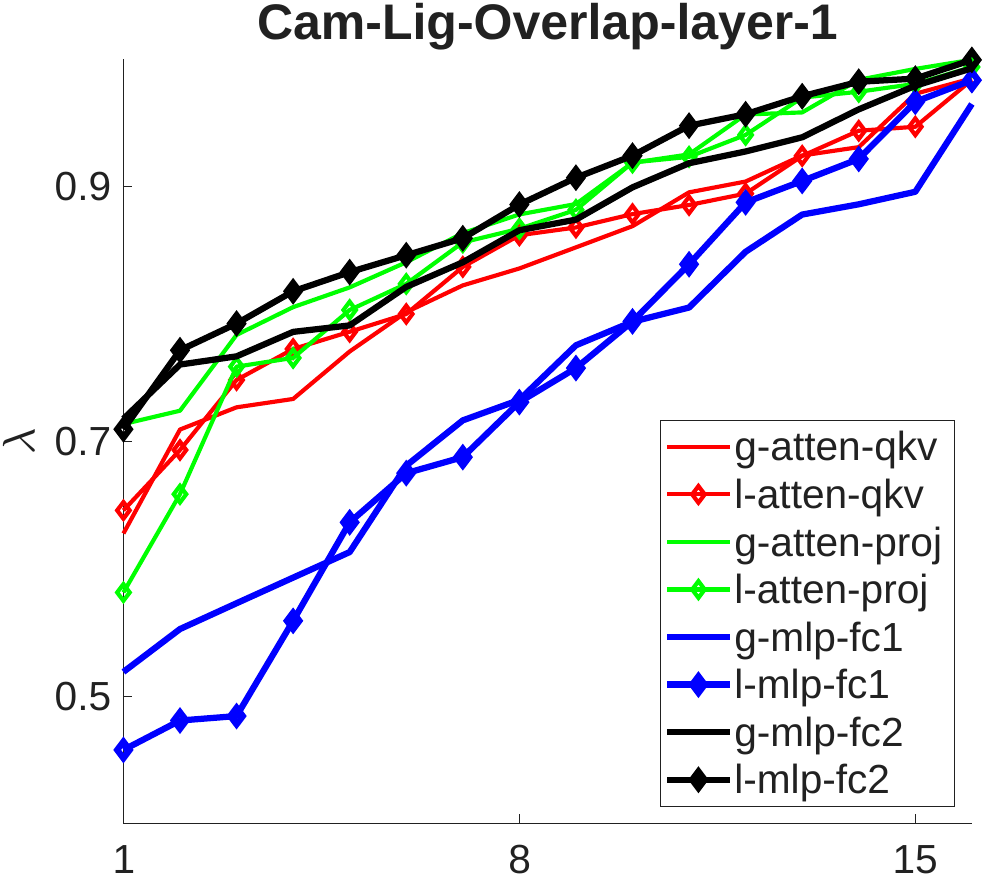}
& 
\includegraphics[width=0.21\textwidth]{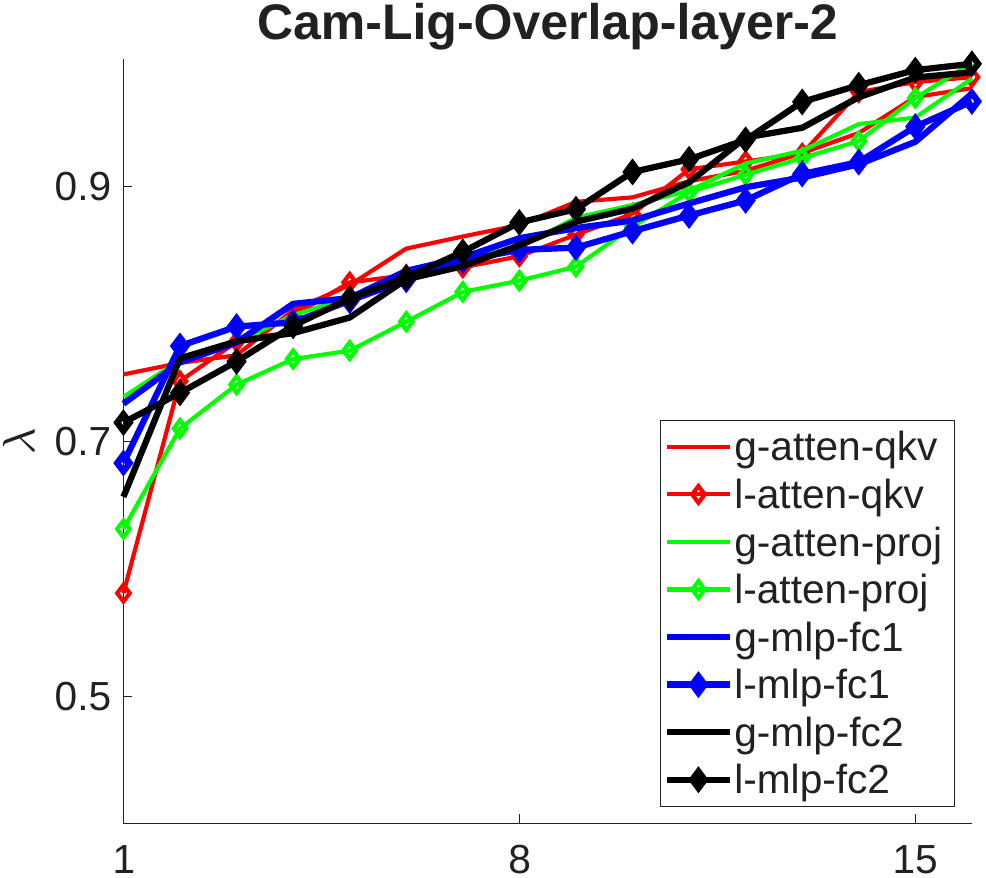}
&
\includegraphics[width=0.21\textwidth]{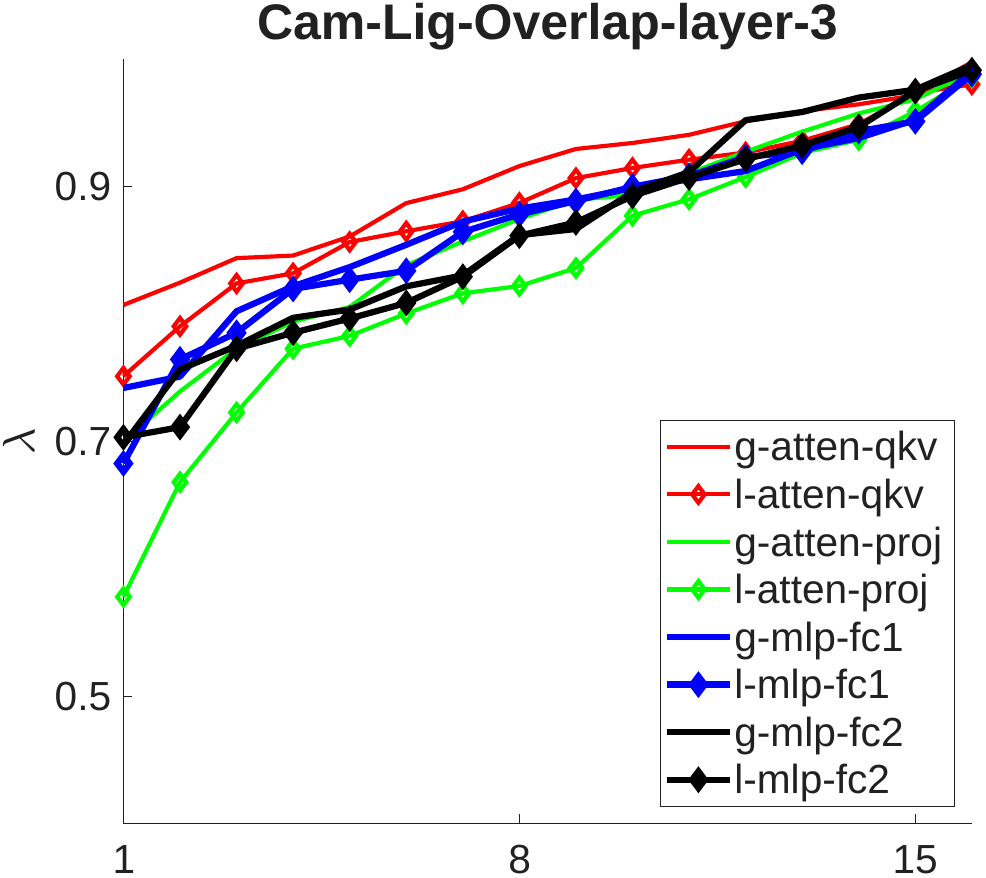}
&
\includegraphics[width=0.21\textwidth]{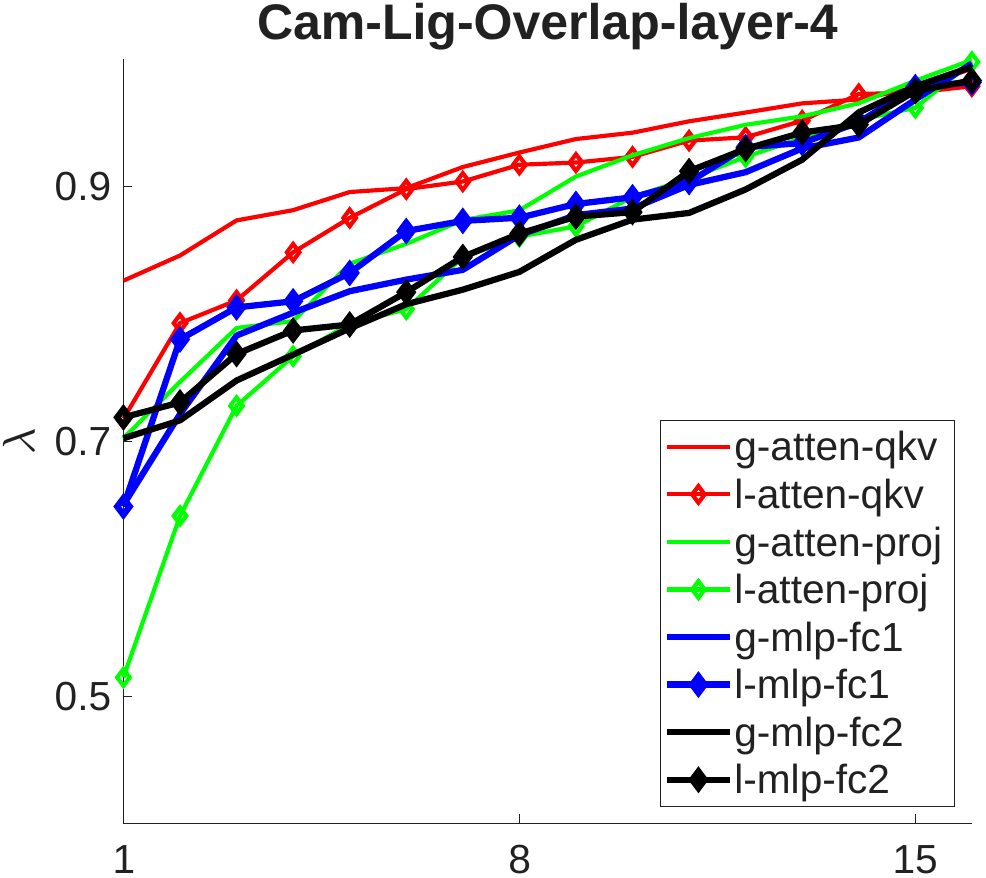}      
\\
\includegraphics[width=0.21\textwidth]{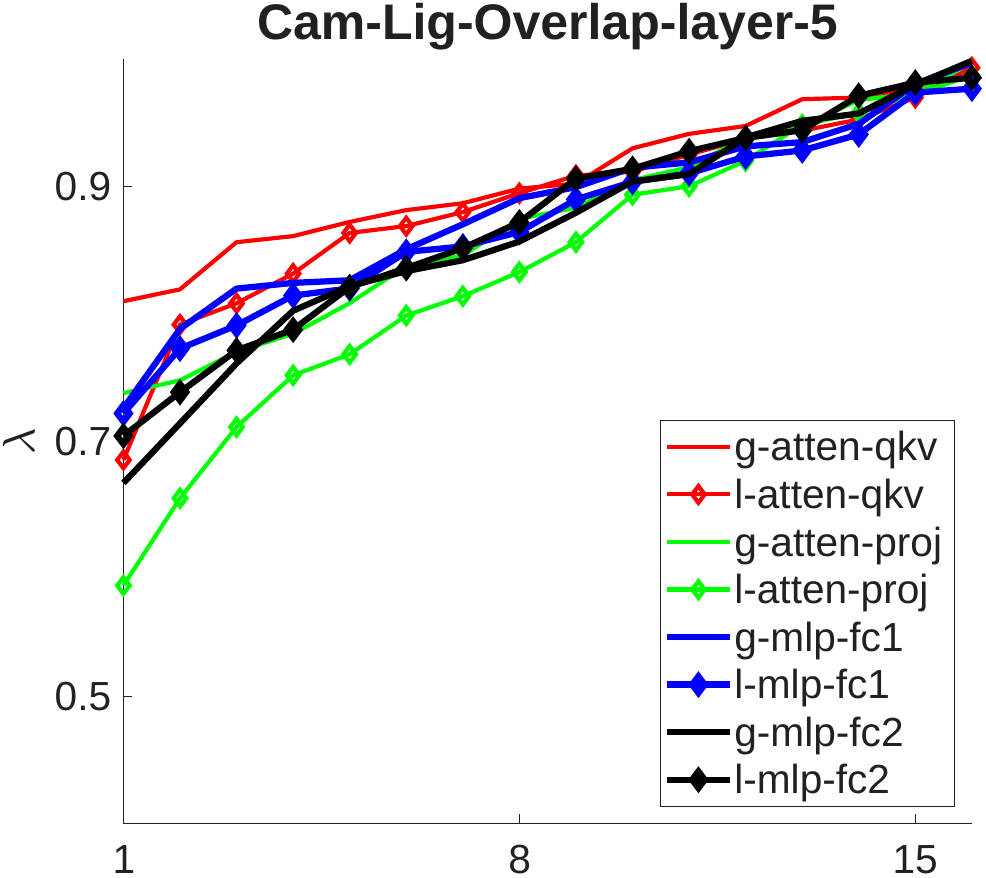}
& 
\includegraphics[width=0.21\textwidth]{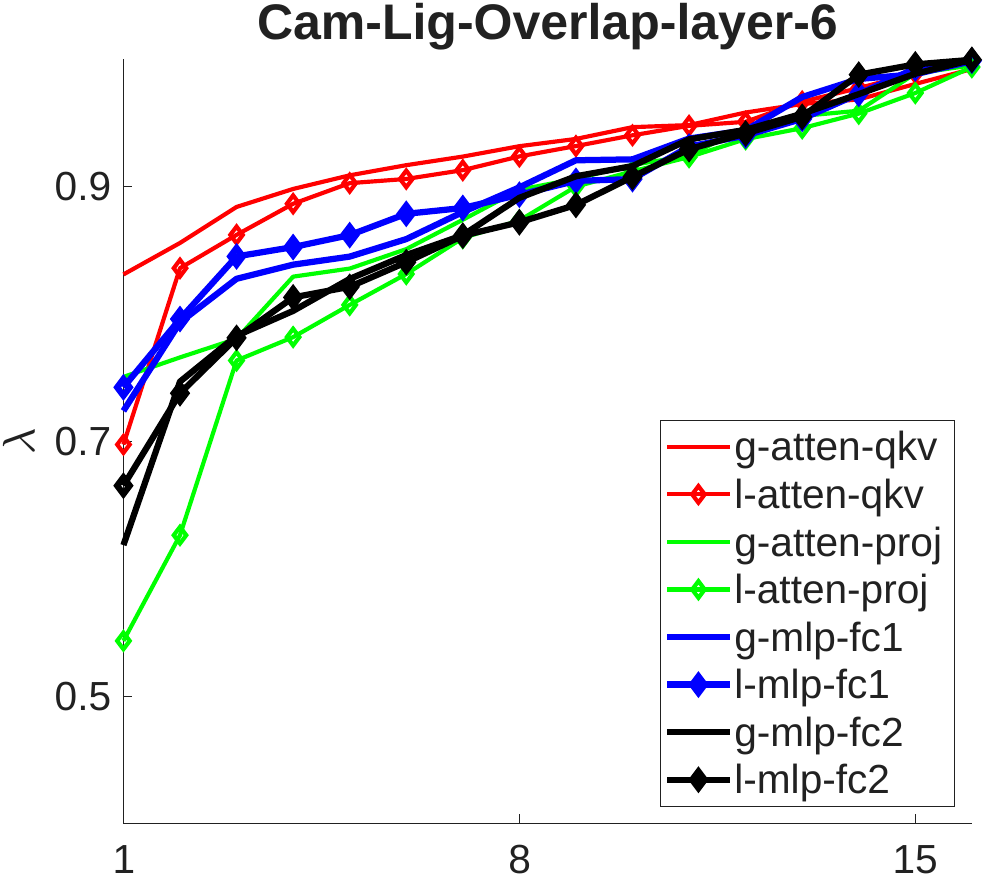}
&
\includegraphics[width=0.21\textwidth]{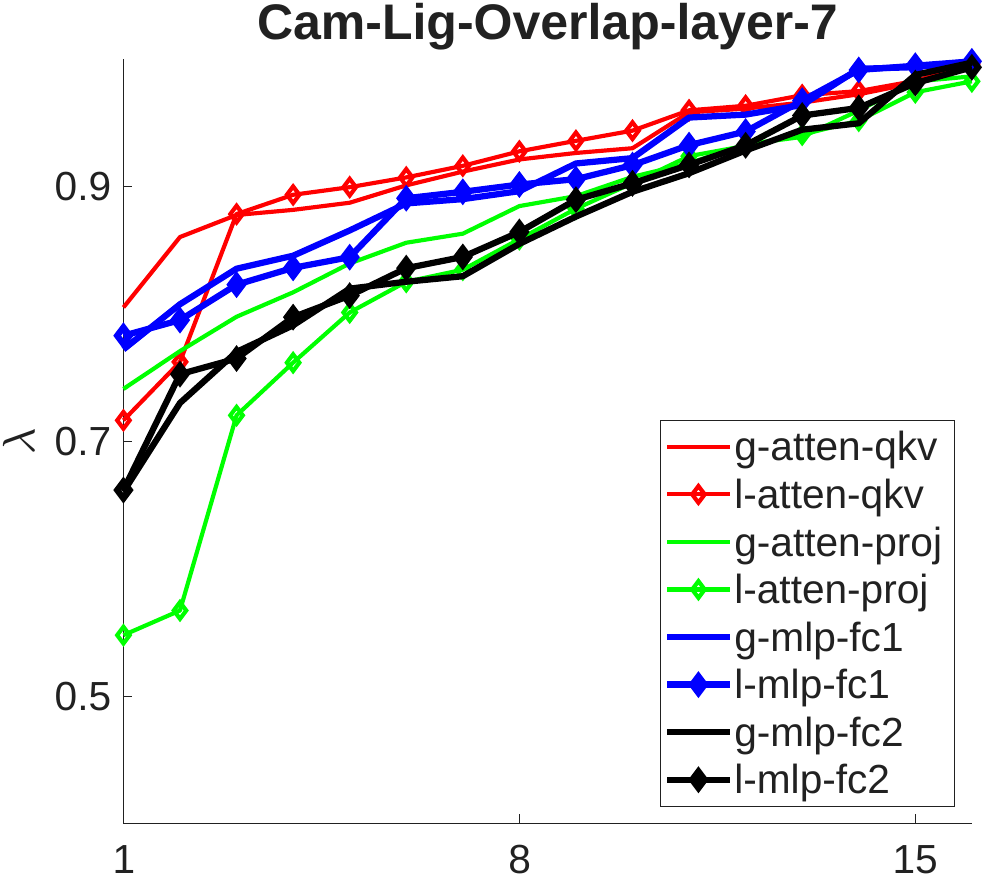}
&
\includegraphics[width=0.21\textwidth]{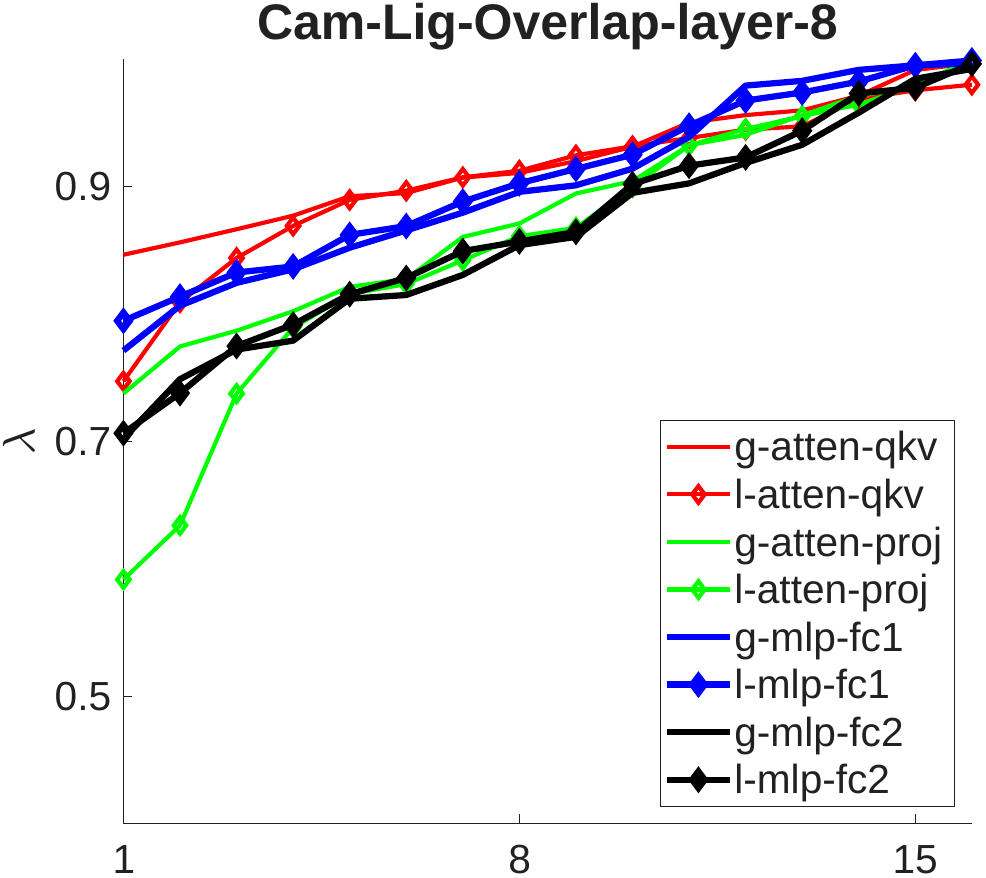}      
\\
\includegraphics[width=0.21\textwidth]{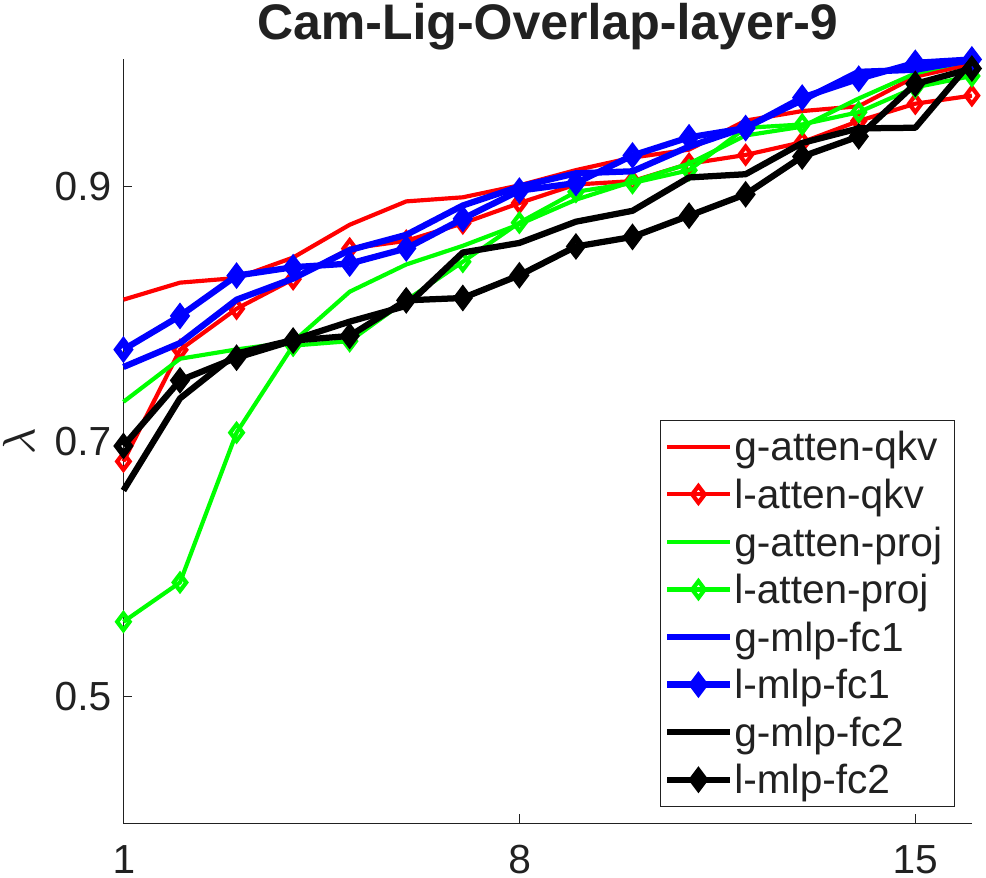}
& 
\includegraphics[width=0.21\textwidth]{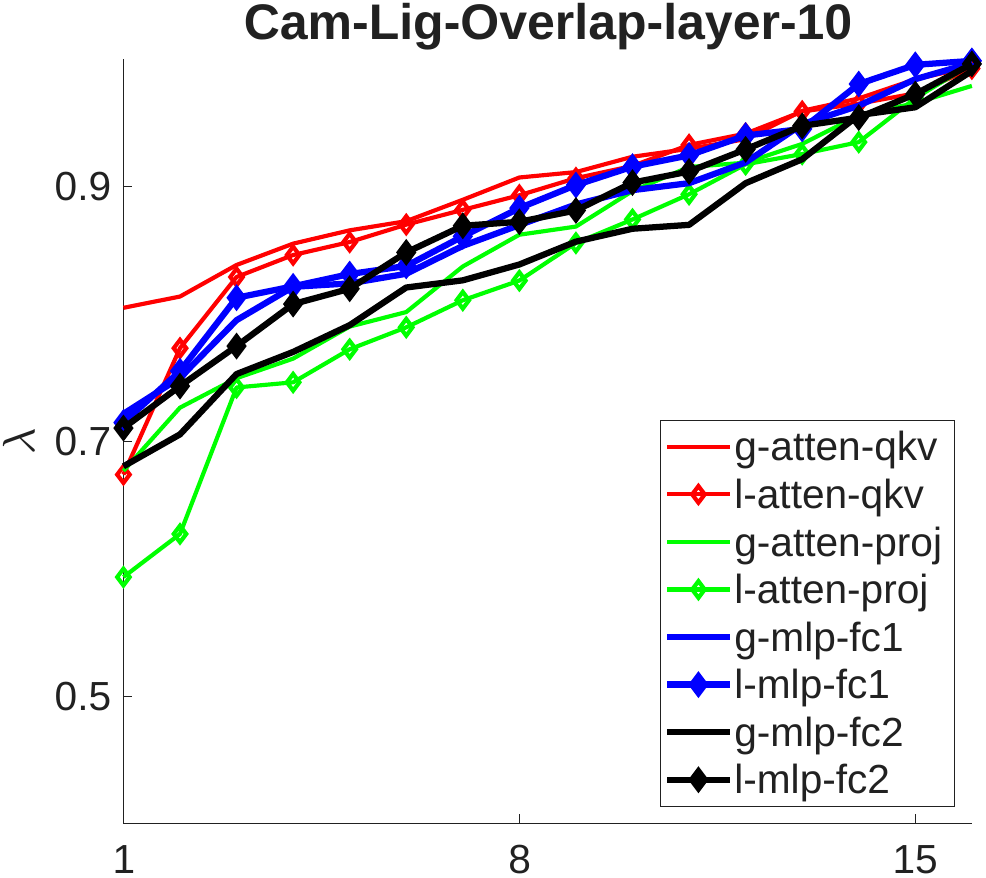}
&
\includegraphics[width=0.21\textwidth]{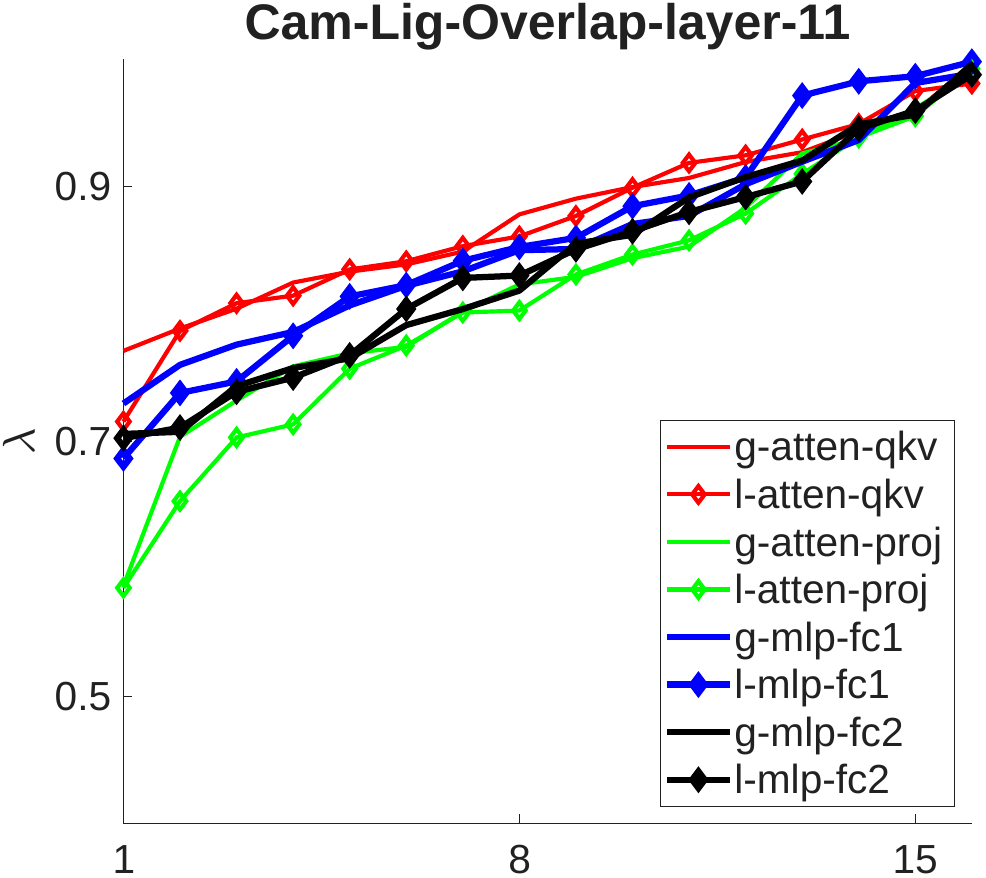}
&
\includegraphics[width=0.21\textwidth]{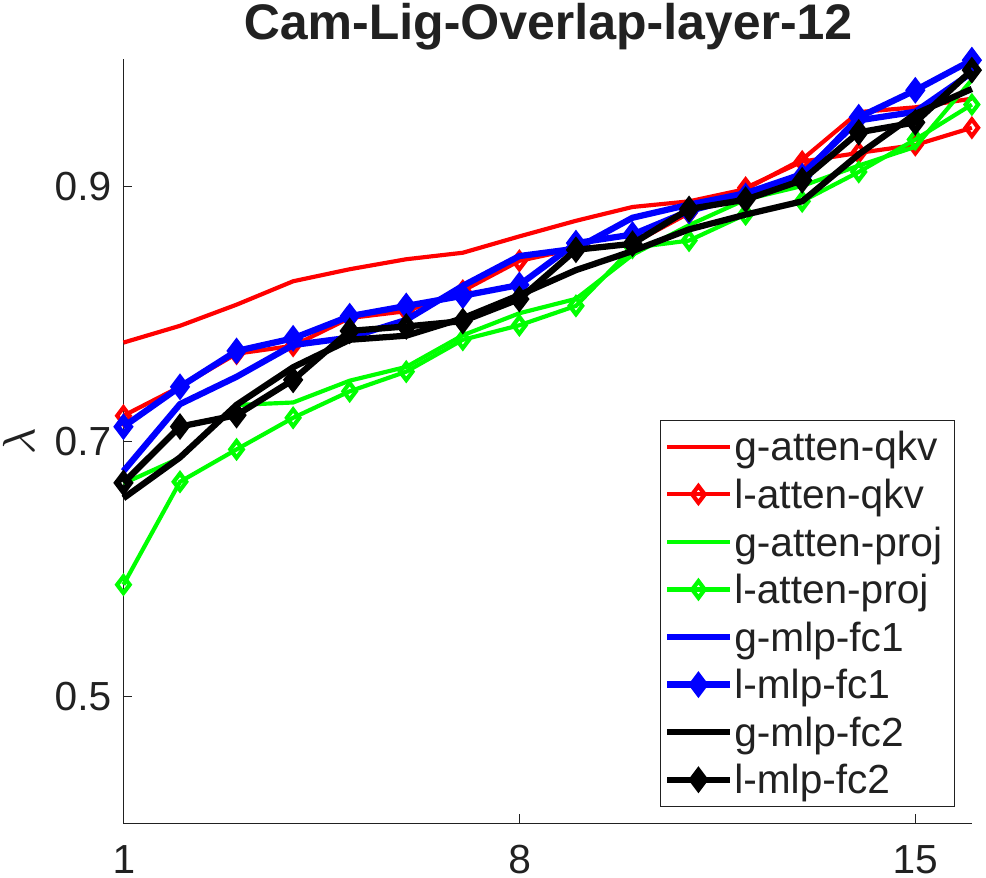}      
\\
\includegraphics[width=0.21\textwidth]{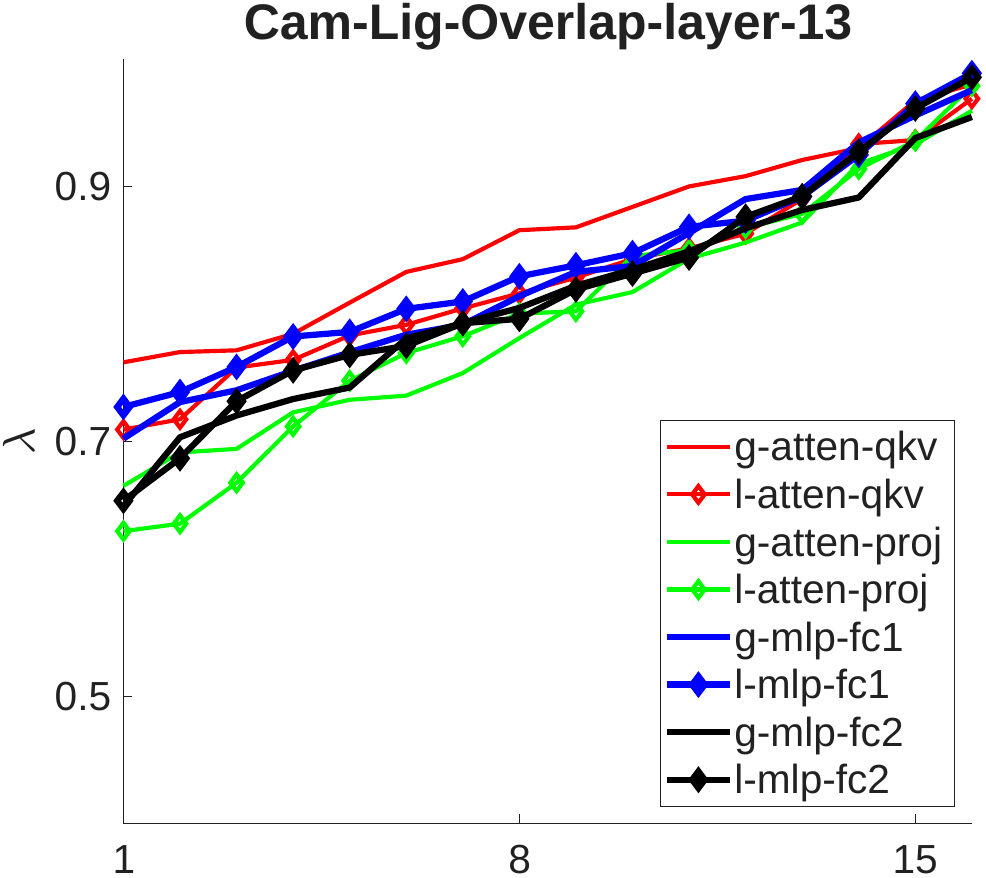}
& 
\includegraphics[width=0.21\textwidth]{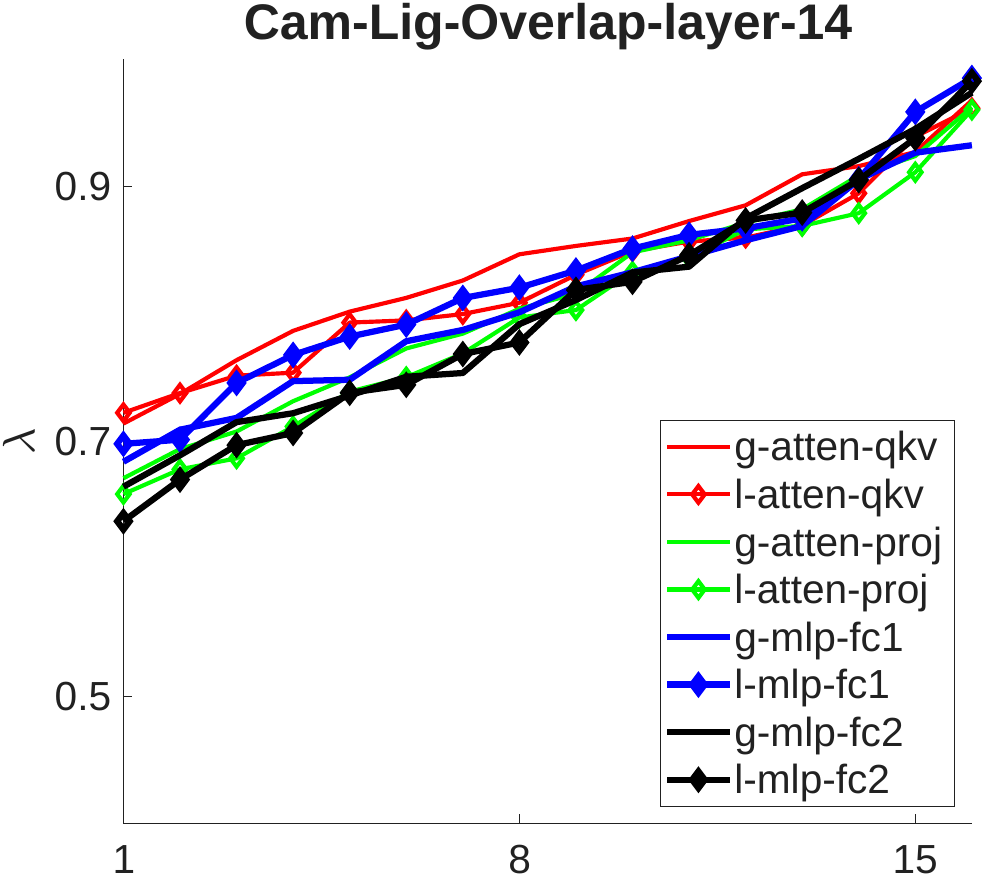}
&
\includegraphics[width=0.21\textwidth]{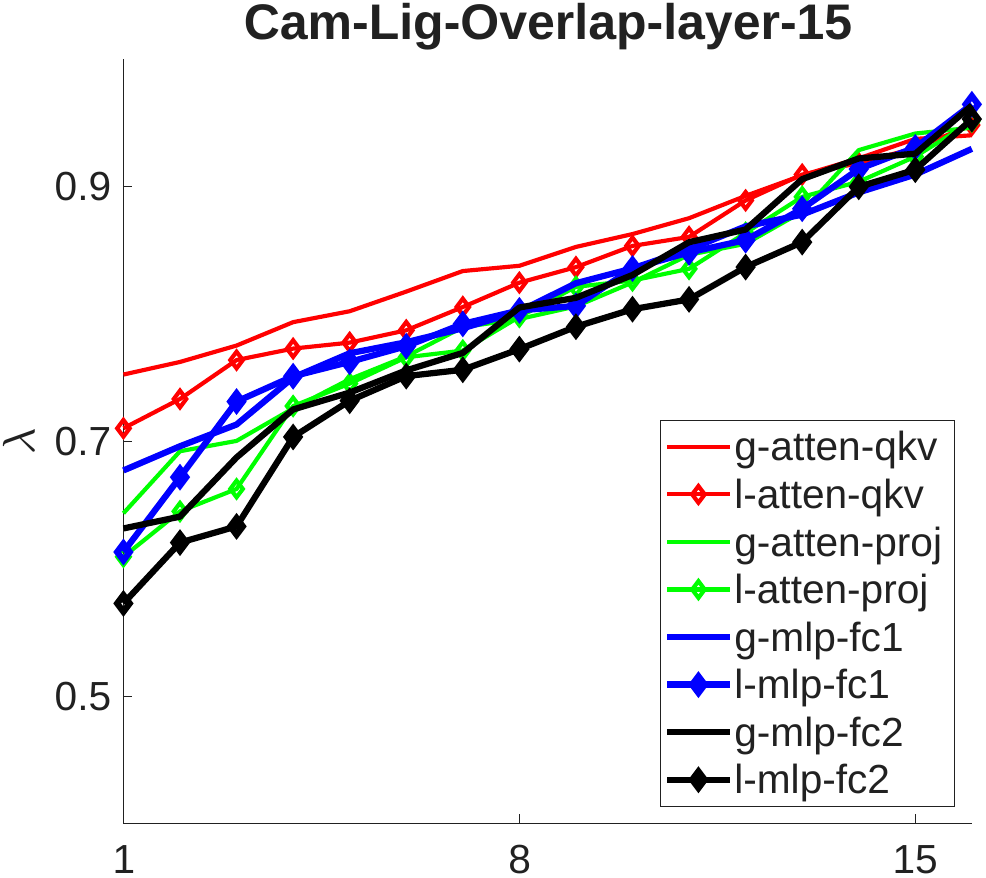}
&
\includegraphics[width=0.21\textwidth]{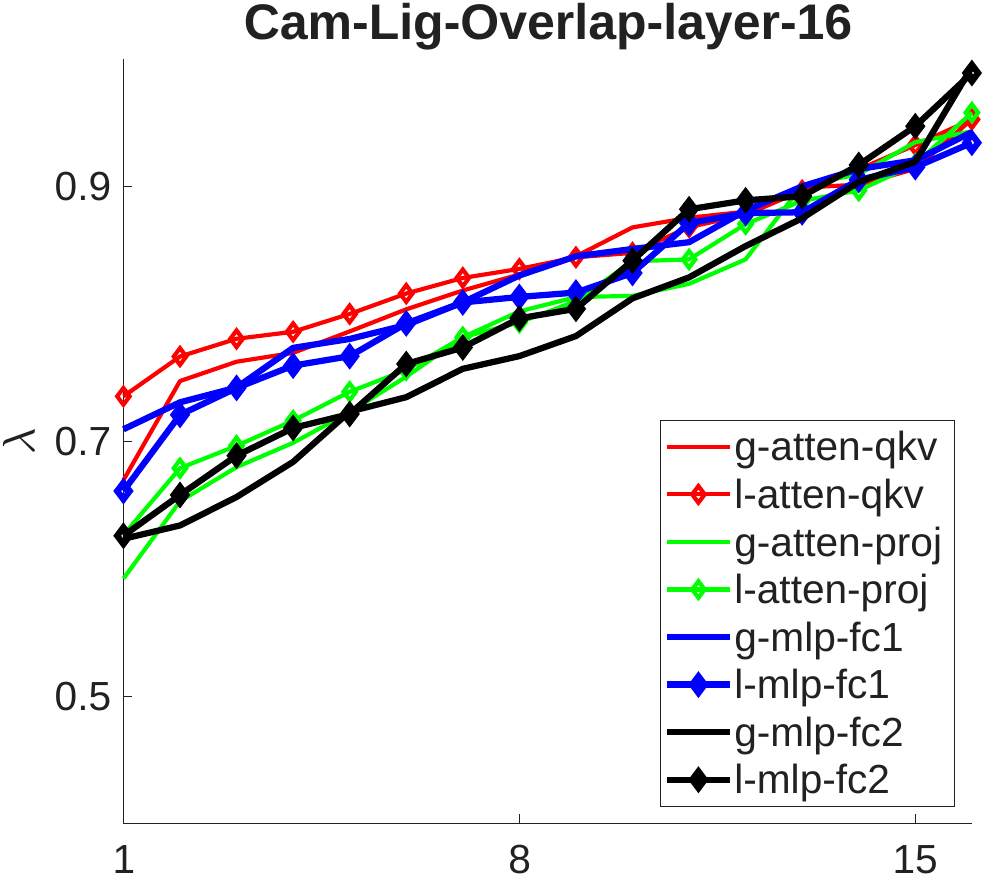}      
\\
\includegraphics[width=0.21\textwidth]{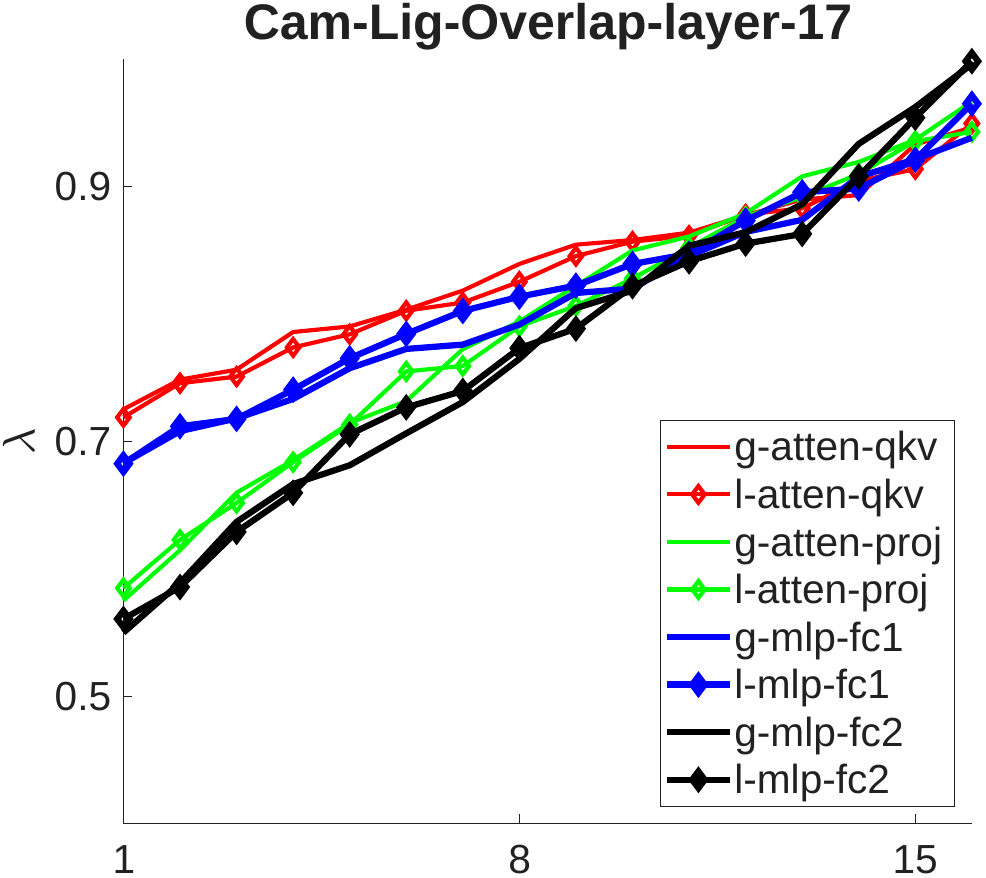}
& 
\includegraphics[width=0.21\textwidth]{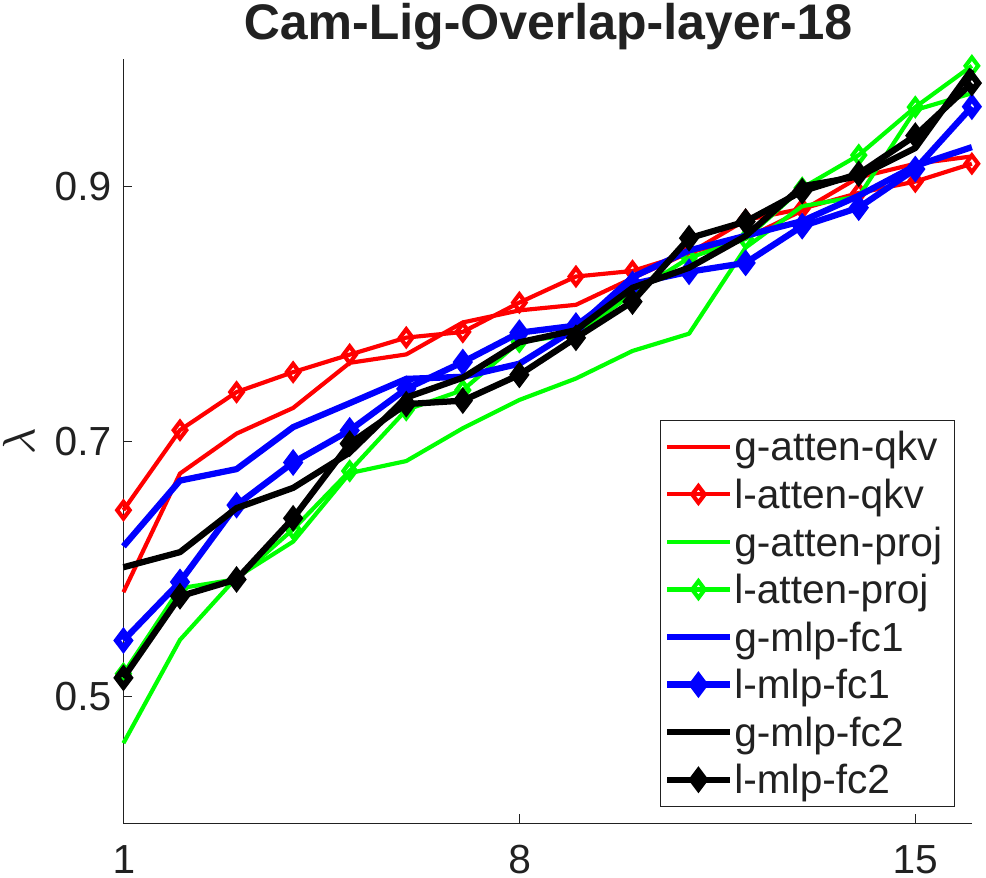}
&
\includegraphics[width=0.21\textwidth]{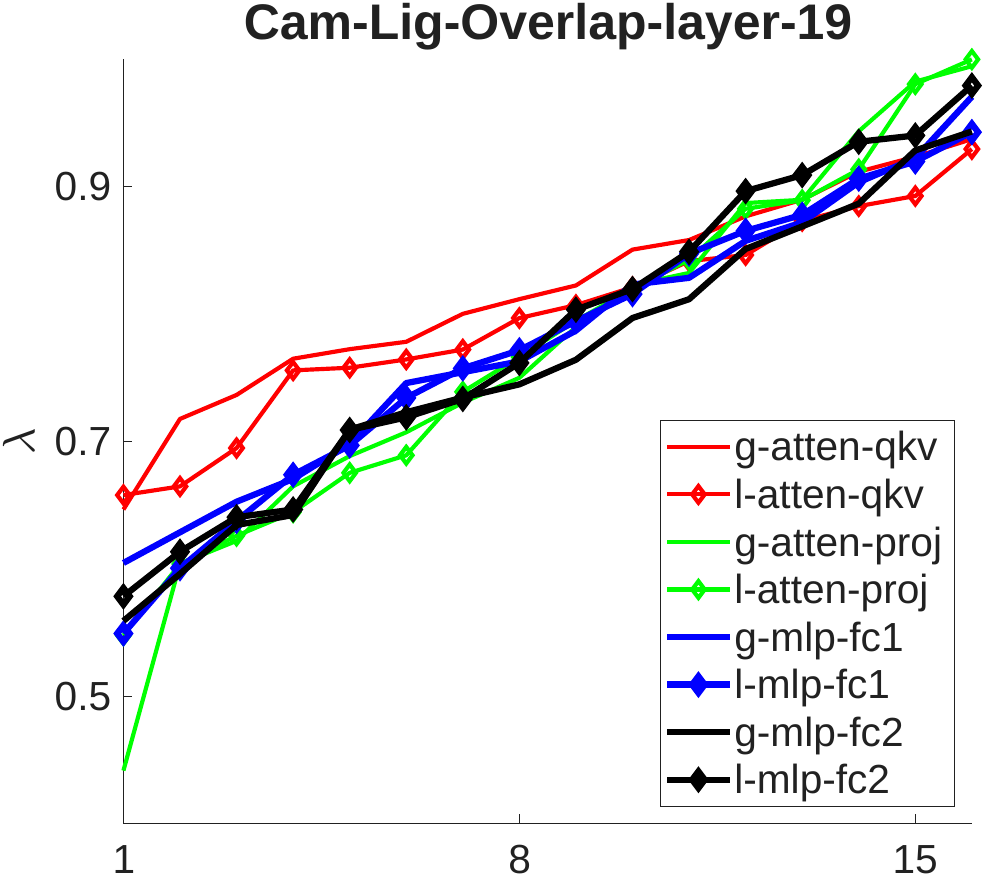}
&
\includegraphics[width=0.21\textwidth]{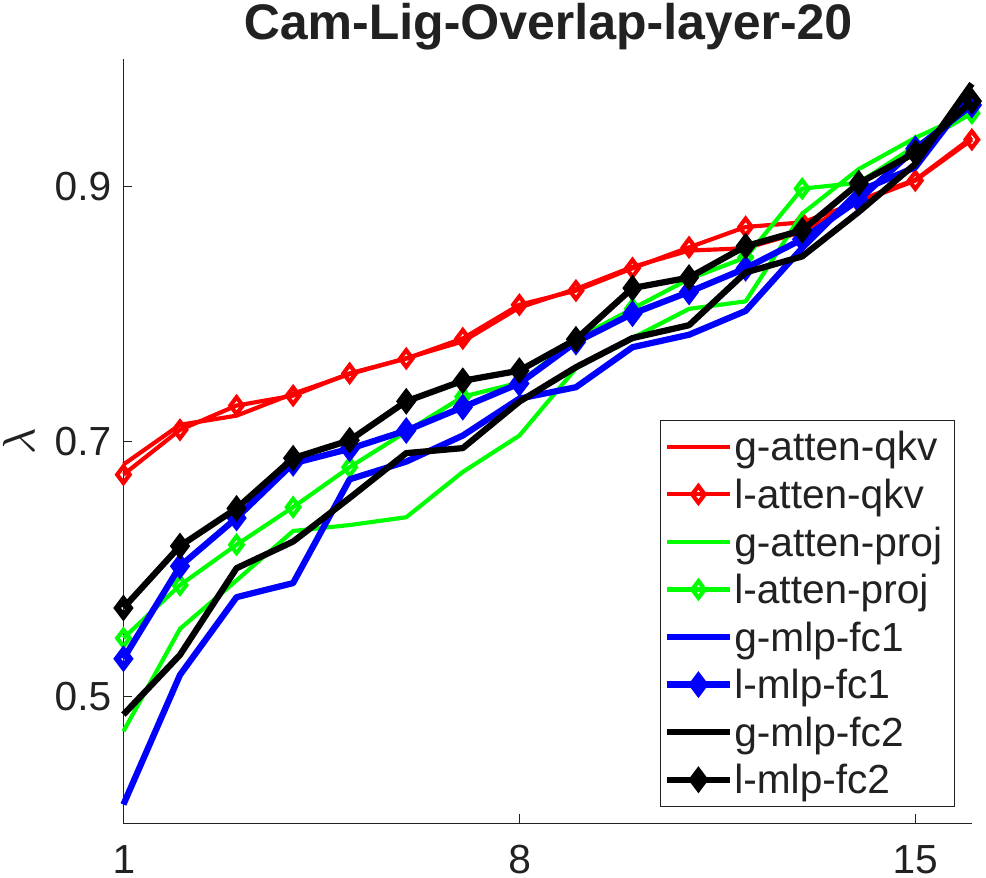}      
\\
\includegraphics[width=0.21\textwidth]{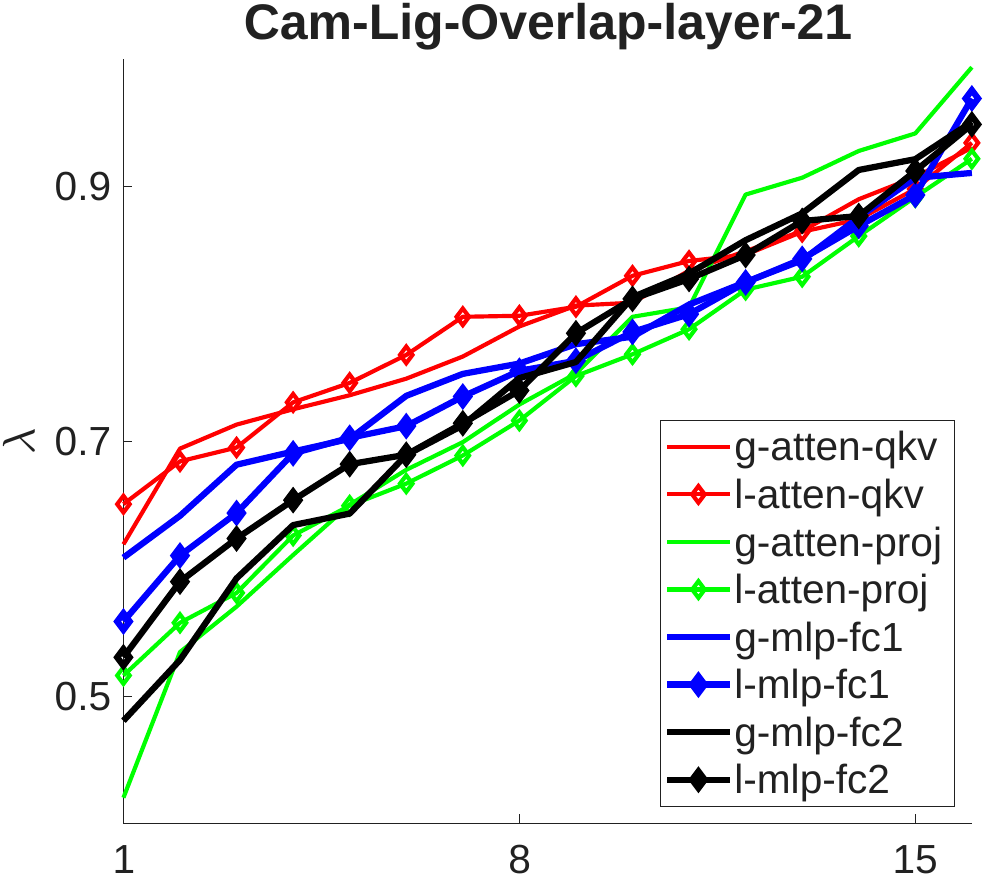}
& 
\includegraphics[width=0.21\textwidth]{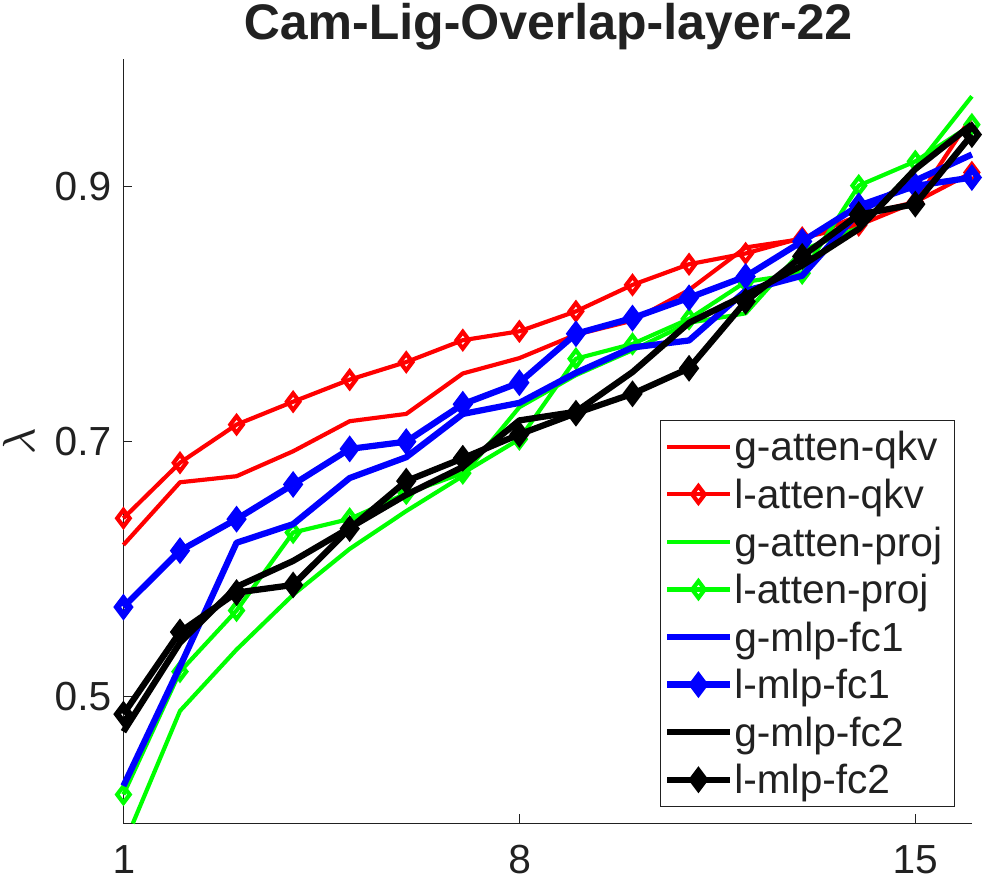}
&
\includegraphics[width=0.21\textwidth]{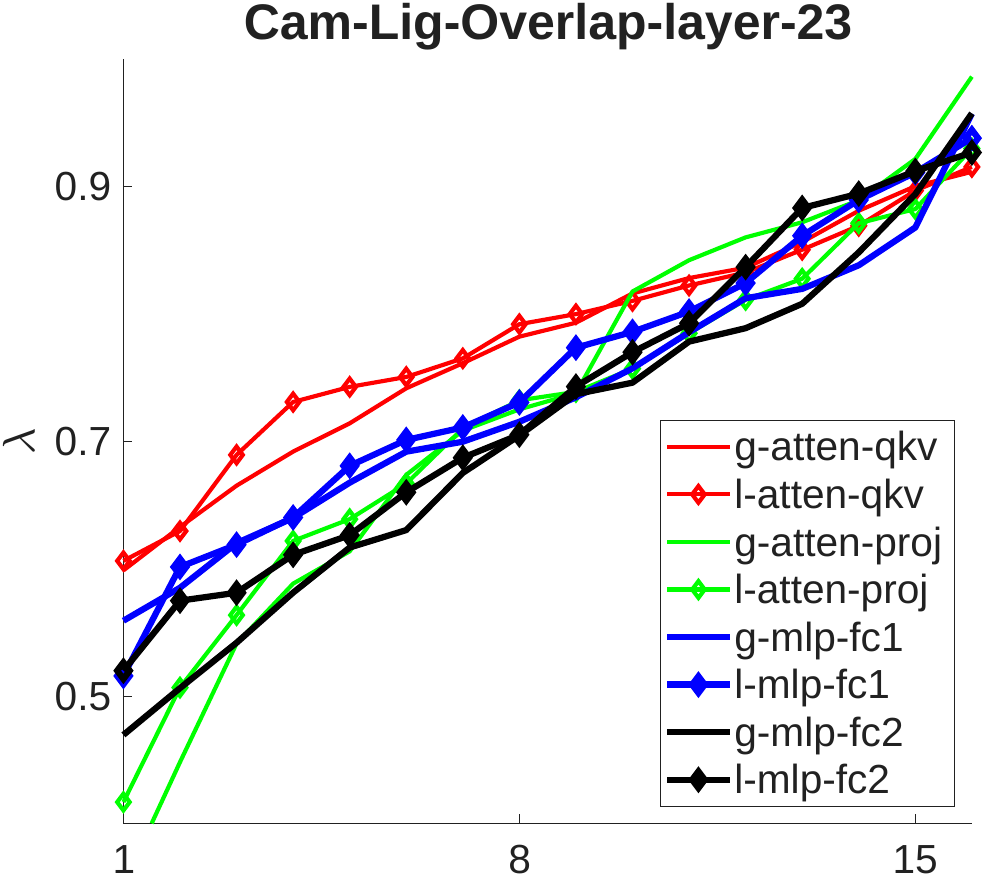}
&
\includegraphics[width=0.21\textwidth]{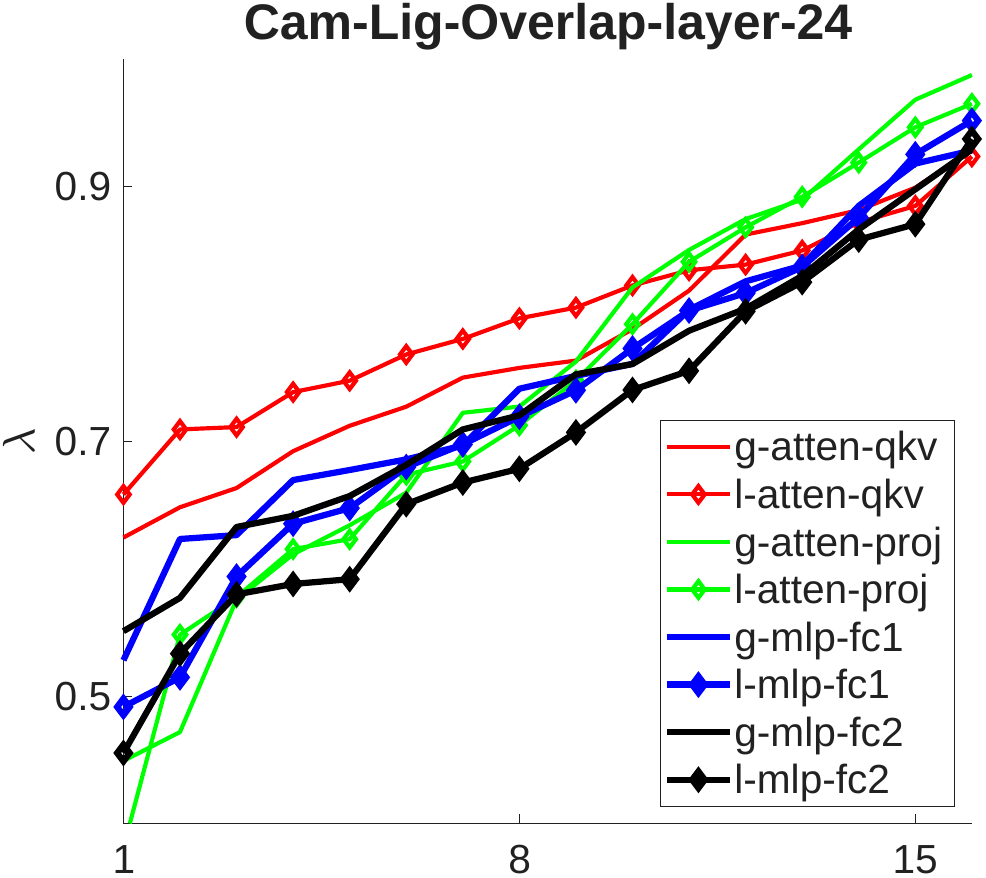}      

\end{tabular}

\captionof{figure}{The overlap ratio between subspaces that correspond to variations in camera motion and lighting. }

% \label{Fig:Subspace:Magnitudes}    
\vspace{-3em}
\end{table*}